\documentclass[lettersize,journal]{IEEEtran}
\usepackage{amsmath,amsfonts}
\usepackage{algorithmic}
\usepackage{algorithm}
\usepackage{array}
\usepackage[caption=false,font=normalsize,labelfont=sf,textfont=sf]{subfig}
\usepackage{textcomp}
\usepackage{stfloats}
\usepackage{url}
\usepackage{verbatim}
\usepackage{graphicx}
\usepackage{cite}
\usepackage{diagbox}
\usepackage{multirow}
\usepackage[dvipsnames]{xcolor}
\def\redm#1{{\textcolor{black}{#1}}}

\usepackage{changepage}

\def\swfig1{0.475\linewidth}
\def\swfigv1{0.102\linewidth}
\def\swflicker{0.161\linewidth}
\def\swjitterr{0.132\linewidth}

\def\swteasa{0.126\linewidth}

\def\swexp{0.121\linewidth}
\def\swablscale{0.3\linewidth}
\def\swablsm{0.2\linewidth}
\def\swffhqvfhq{0.24\linewidth}
\def\swlankmark{0.135\linewidth}

\begin{document}

\title{Analysis and Benchmarking of Extending Blind Face Image Restoration to Videos}

\author{Zhouxia Wang, Jiawei Zhang, Xintao Wang, Tianshui Chen, Ying Shan,~\IEEEmembership{Senior Member,~IEEE,} Wenping Wang,~\IEEEmembership{Fellow,~IEEE,} and Ping Luo~\IEEEmembership{Member,~IEEE,}
\thanks{Zhouxia Wang and Ping Luo are with The University of Hong Kong, Hong Kong, China (Email: wzhoux@connect.hku.hk and pluo@cs.hku.hk). Jiawei Zhang is with SenseTime Research, Shenzhen, China (Email: zhjw1988@gmail.com). Xintao Wang and Ying Shan are with ARC Lab, Tencent PCG, Shenzhen, China (Email: xintao.alpha@gmail.com and yingsshan@tencent.com). Tianshui Chen is with The Guangdong University of Technology, Guangzhou, China (Email: tianshuichen@gmail.com). Wenping Wang is with Texas A\&M University, College Station, US (Email: wenping@cs.hku.hk).\\
This paper is partially supported by the National Key R\&D Program of China No.2022ZD0161000 and the General Research Fund of Hong Kong No.17200622 and 17209324.\\
This paper has supplementary downloadable material available at http://ieeexplore.ieee.org., provided by the author. The material features a video that succinctly yet vividly presents the data collection process, systematic analyses, and a comparison of restored facial videos. Contact wzhoux@connect.hku.hk for further questions about this work.\\
Xintao Wang and Ping Luo are the corresponding authors.}
}

\markboth{IEEE TRANSACTIONS ON IMAGE PROCESSING}%
{Shell \MakeLowercase{\textit{et al.}}: A Sample Article Using IEEEtran.cls for IEEE Journals}


\maketitle

\begin{abstract}

Recent progress in blind face restoration has resulted in producing high-quality restored results for static images. However, efforts to extend these advancements to video scenarios have been minimal, partly because of the absence of benchmarks that allow for a comprehensive and fair comparison. In this work, we first present a fair evaluation benchmark, in which we first introduce a Real-world Low-Quality Face Video benchmark (RFV-LQ), evaluate several leading image-based face restoration algorithms, and conduct a thorough systematical analysis of the benefits and challenges associated with extending blind face image restoration algorithms to degraded face videos. Our analysis identifies several key issues, primarily categorized into two aspects: significant jitters in facial components and noise-shape flickering between frames. To address these issues, we propose a Temporal Consistency Network (TCN) cooperated with alignment smoothing to reduce jitters and flickers in restored videos. TCN is a flexible component that can be seamlessly plugged into the most advanced face image restoration algorithms, ensuring the quality of image-based restoration is maintained as closely as possible. Extensive experiments have been conducted to evaluate the effectiveness and efficiency of our proposed TCN and alignment smoothing operation.
Project page: \url{https://wzhouxiff.github.io/projects/FIR2FVR/FIR2FVR}.

\end{abstract}

\begin{IEEEkeywords}
Blind face restoration, Face video restoration, Real-world benchmark, Low-level vision.
\end{IEEEkeywords}

\begin{figure*}[ht]
\setlength\tabcolsep{1pt}
\scriptsize
\centering
\begin{tabular}{ccccccc}
    \includegraphics[width=\swteasa]{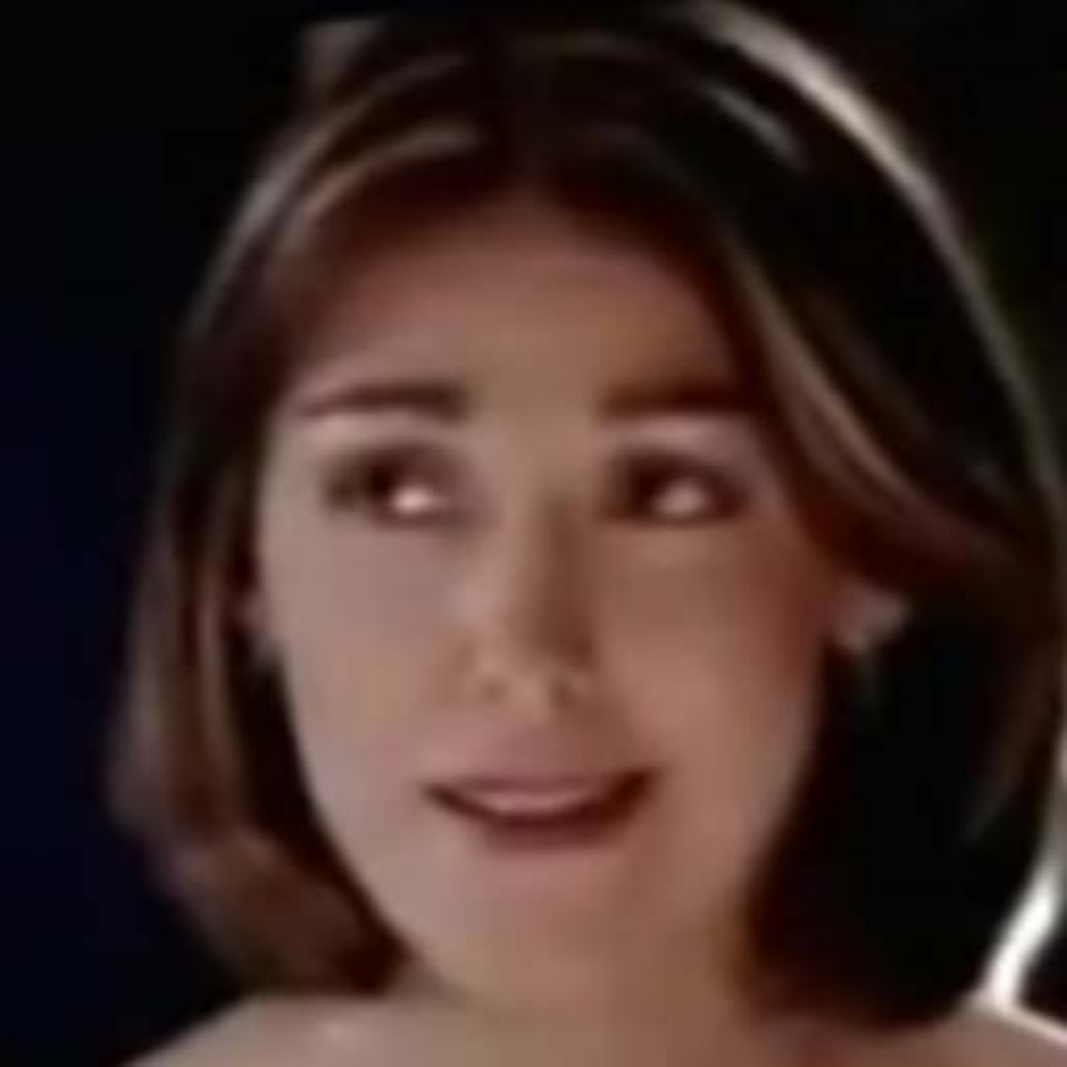} &
    \includegraphics[width=\swteasa]{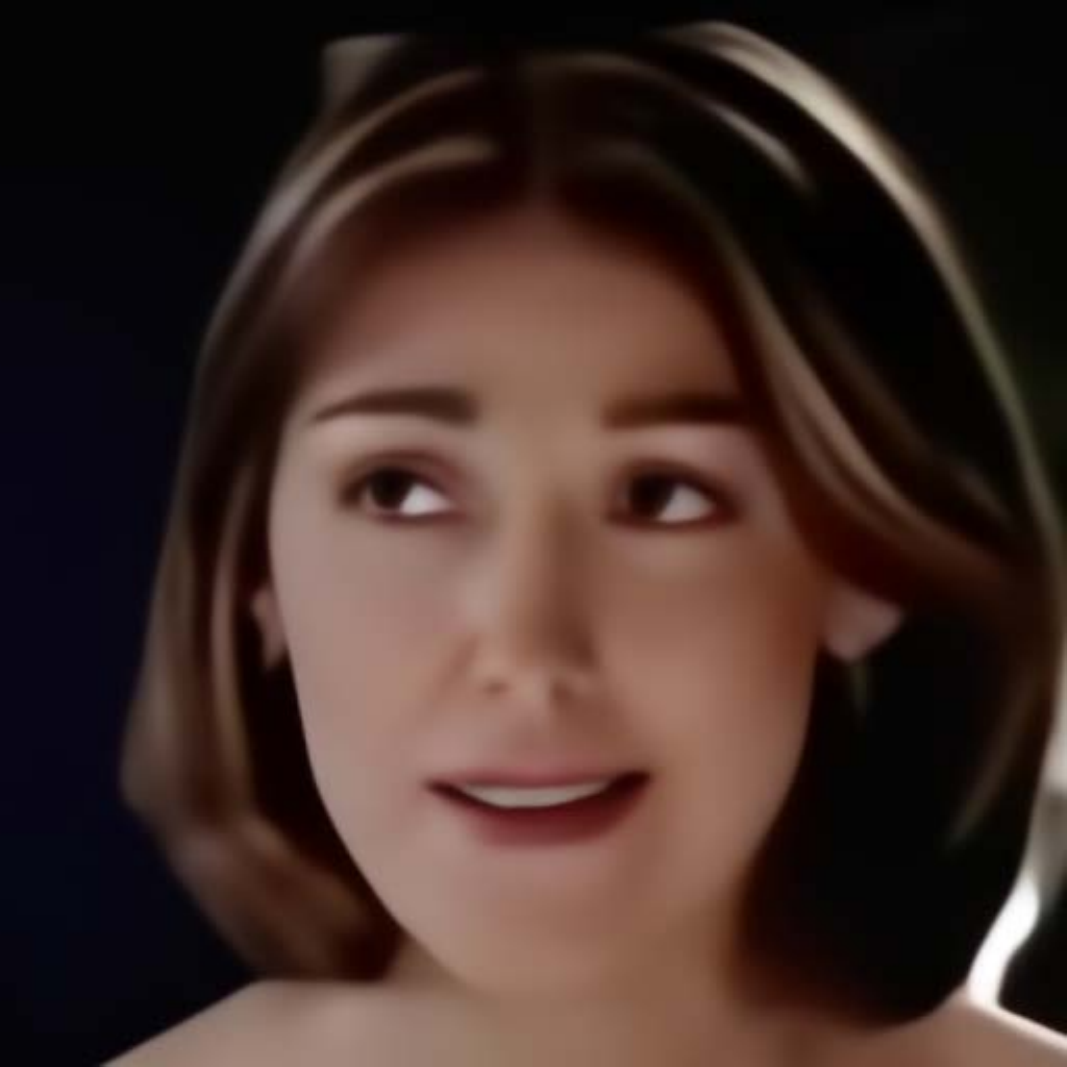} &
    \includegraphics[width=\swteasa]{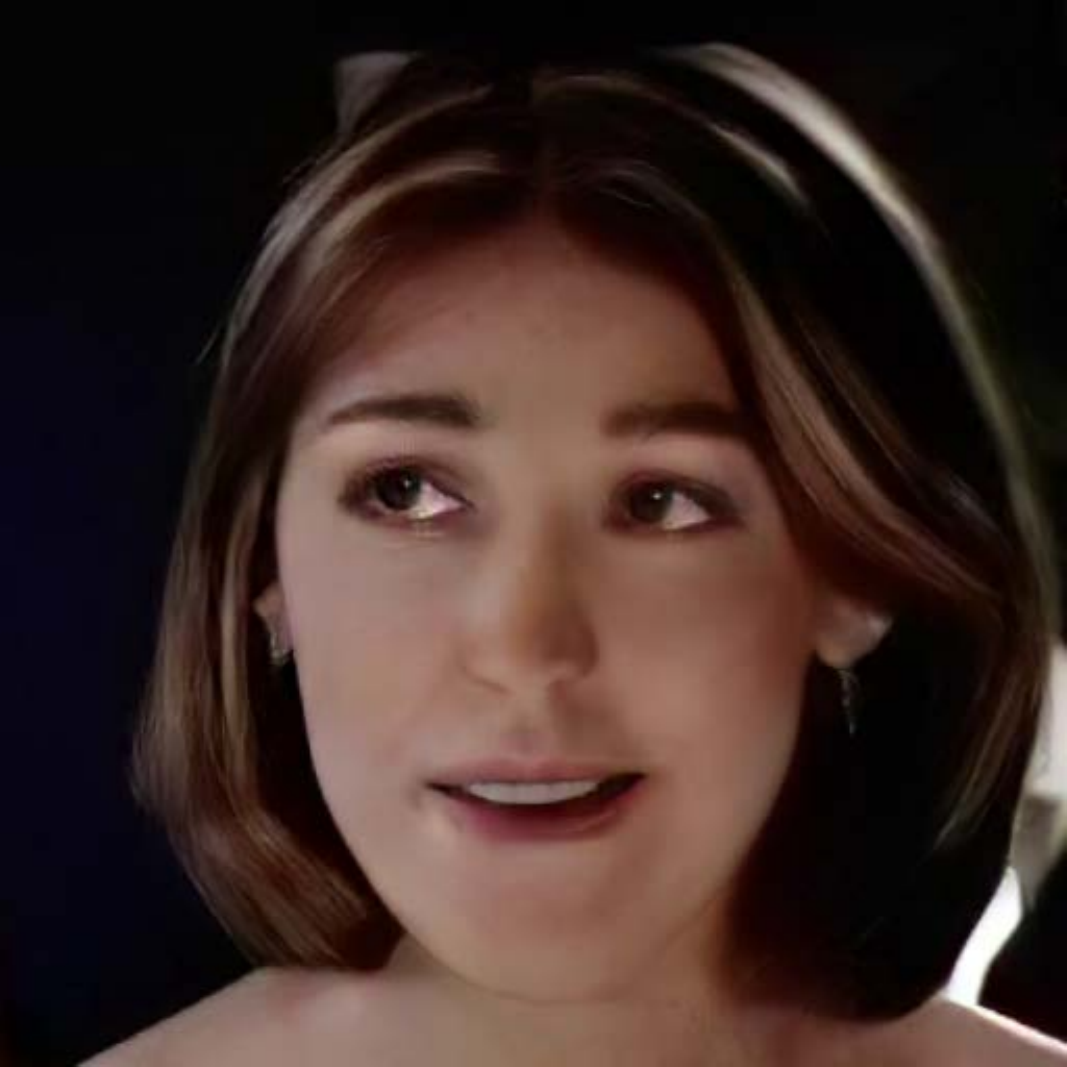} &
    \includegraphics[width=\swteasa]{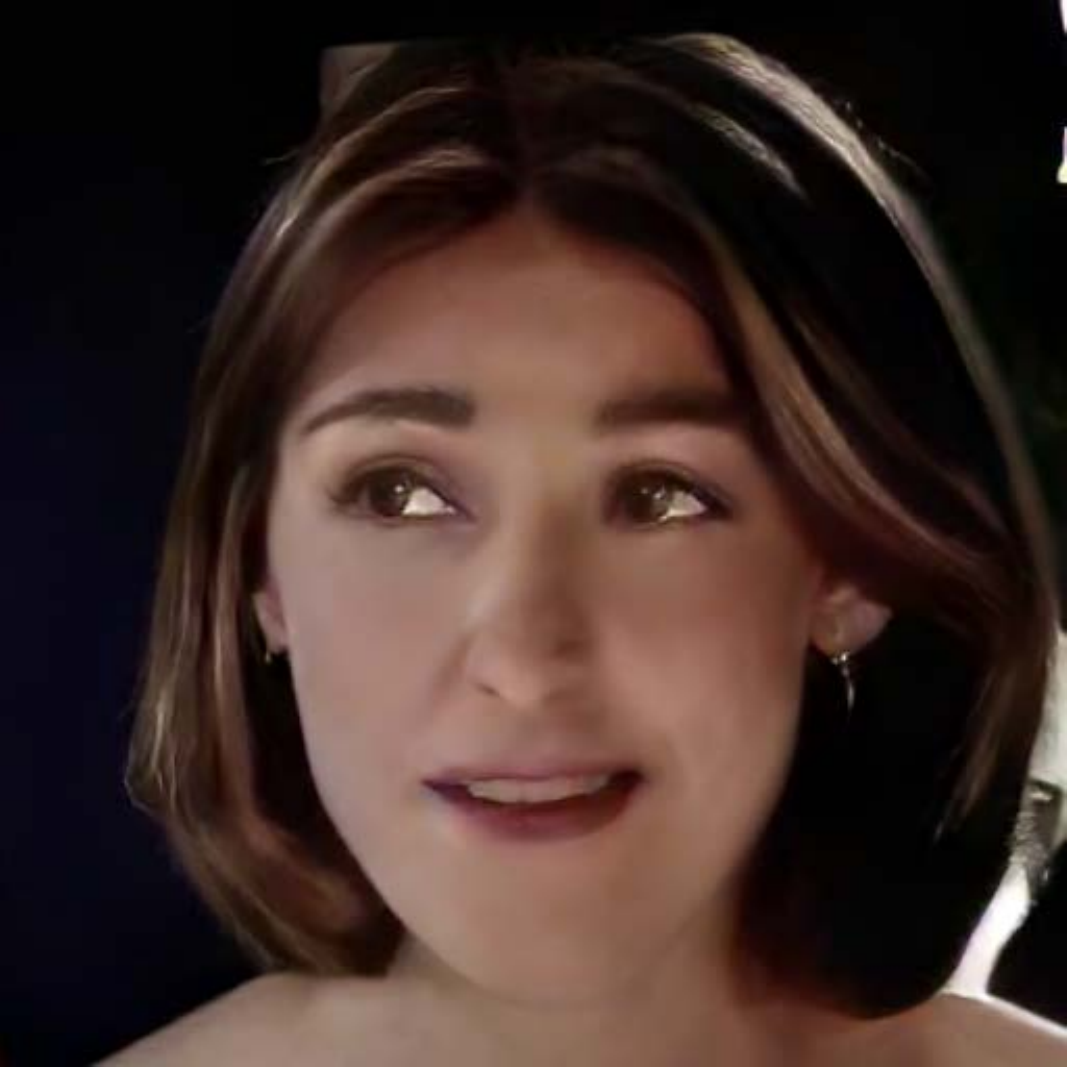} &
    \includegraphics[width=\swteasa]{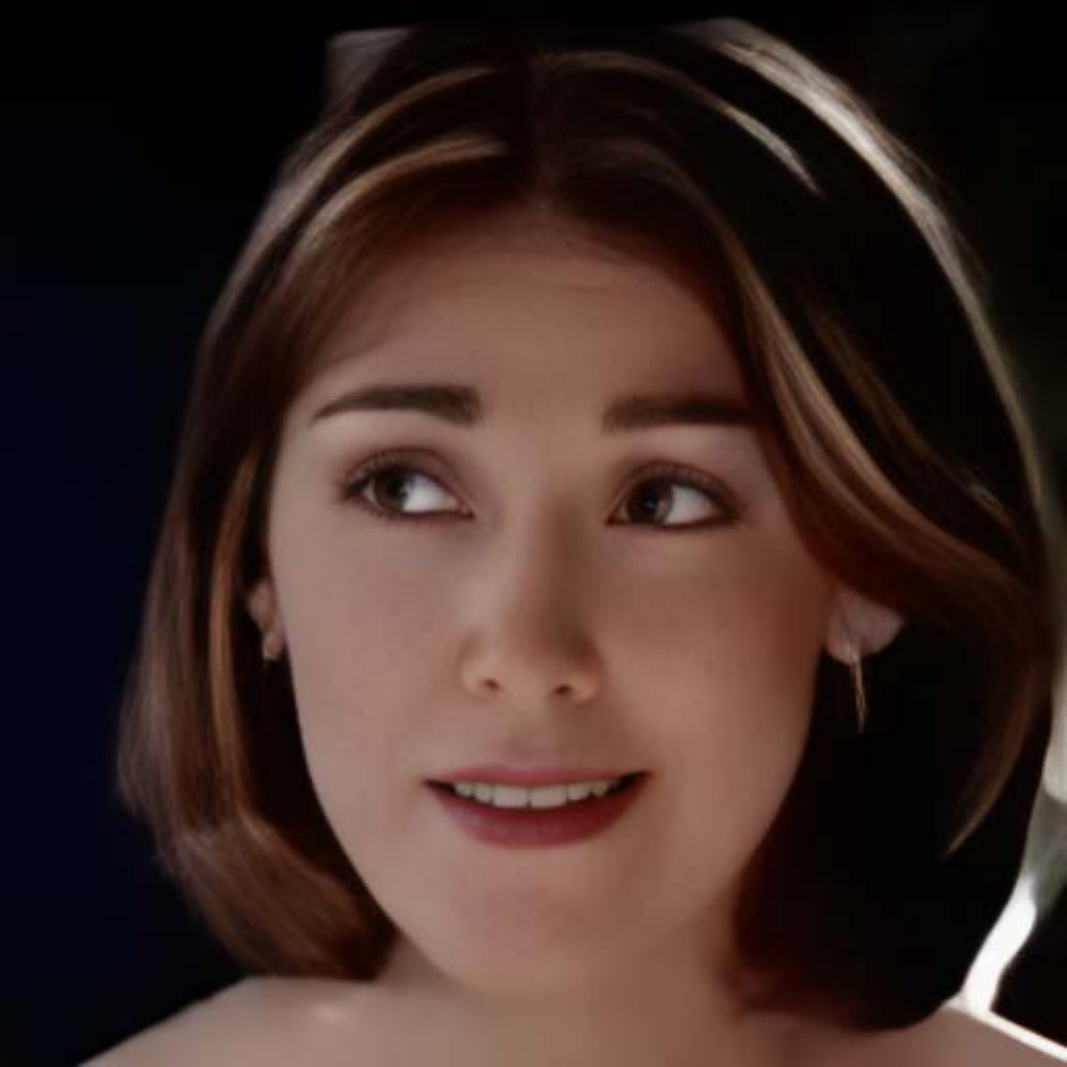} &
    \includegraphics[width=\swteasa]{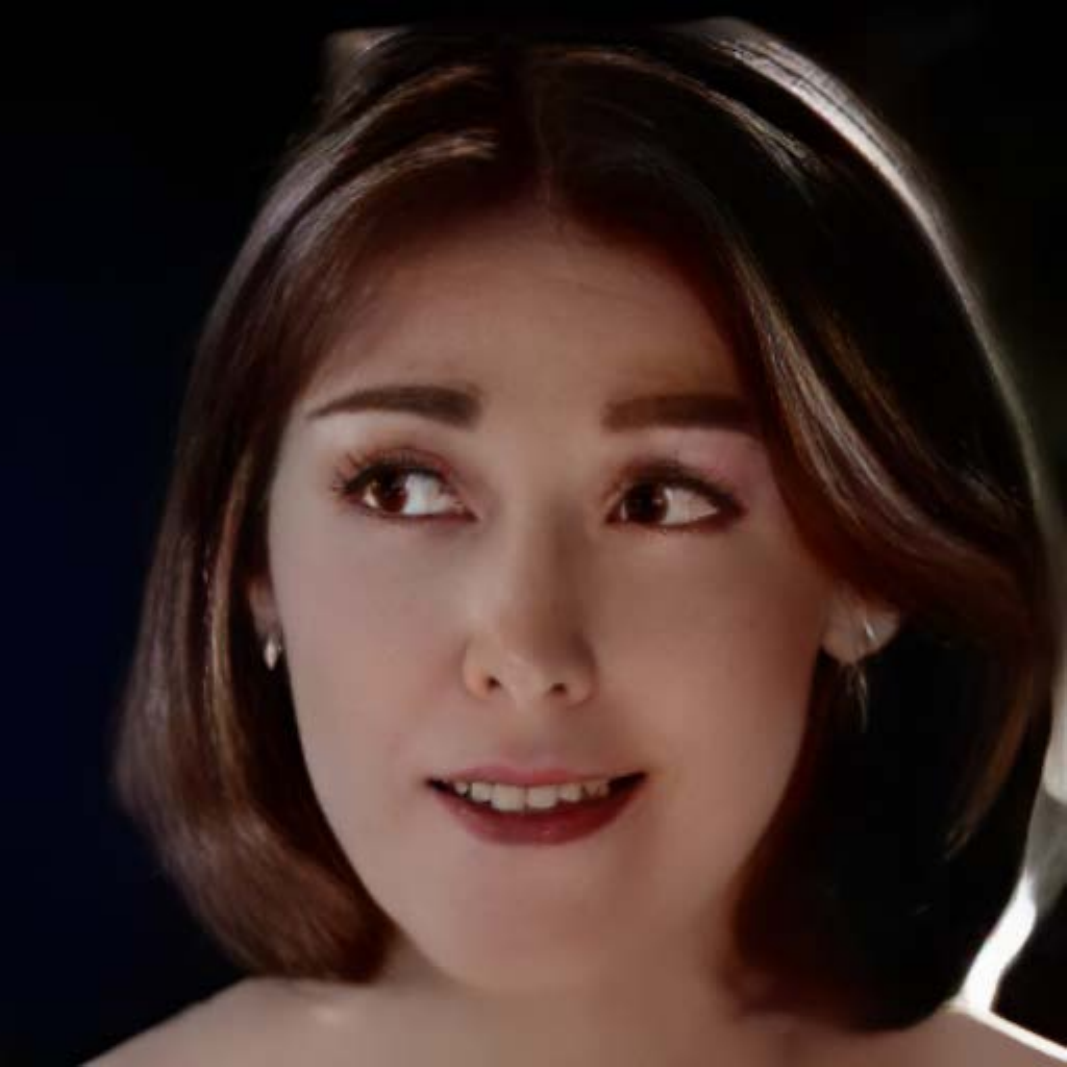} &
    \includegraphics[width=\swteasa]{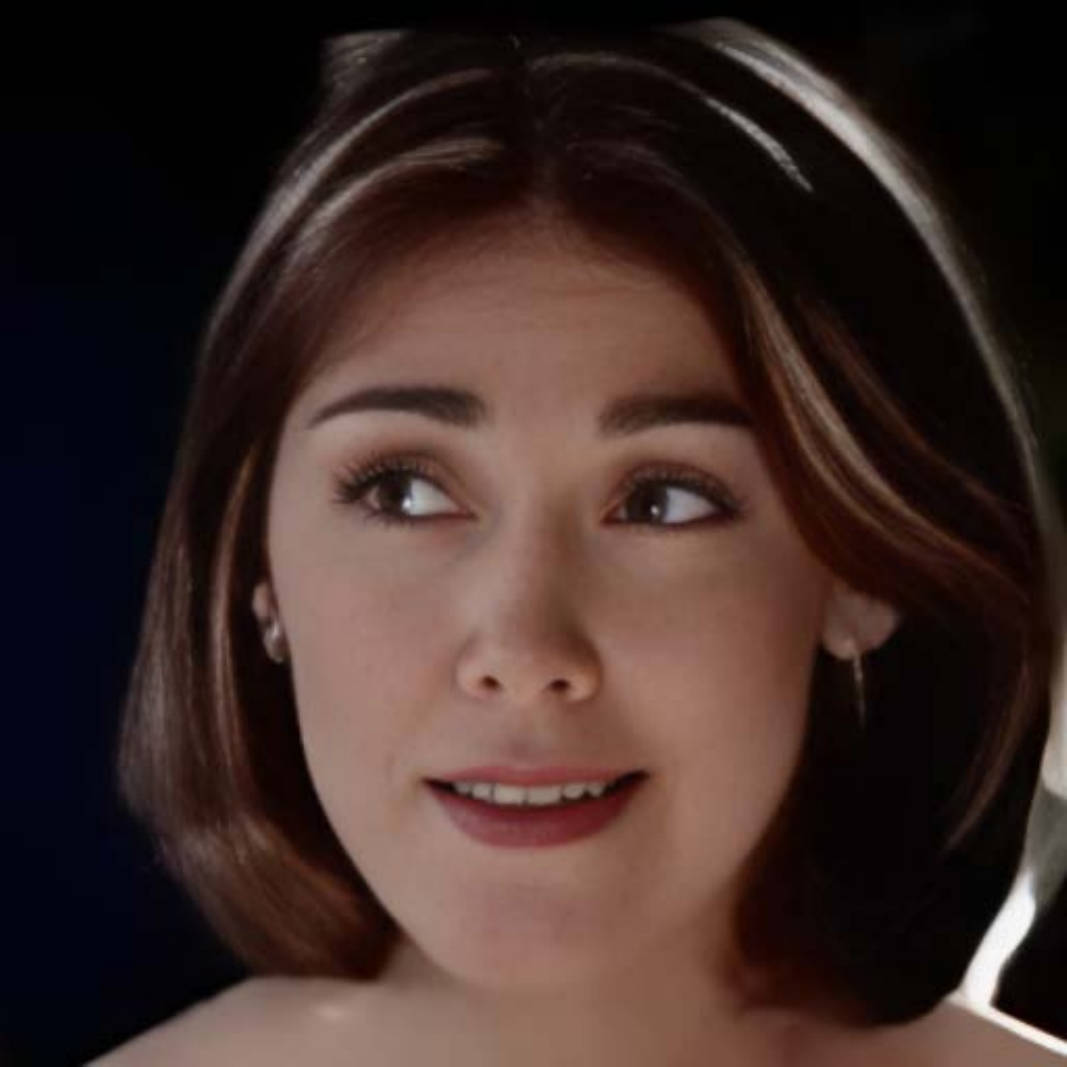} \\
    \includegraphics[width=\swteasa]{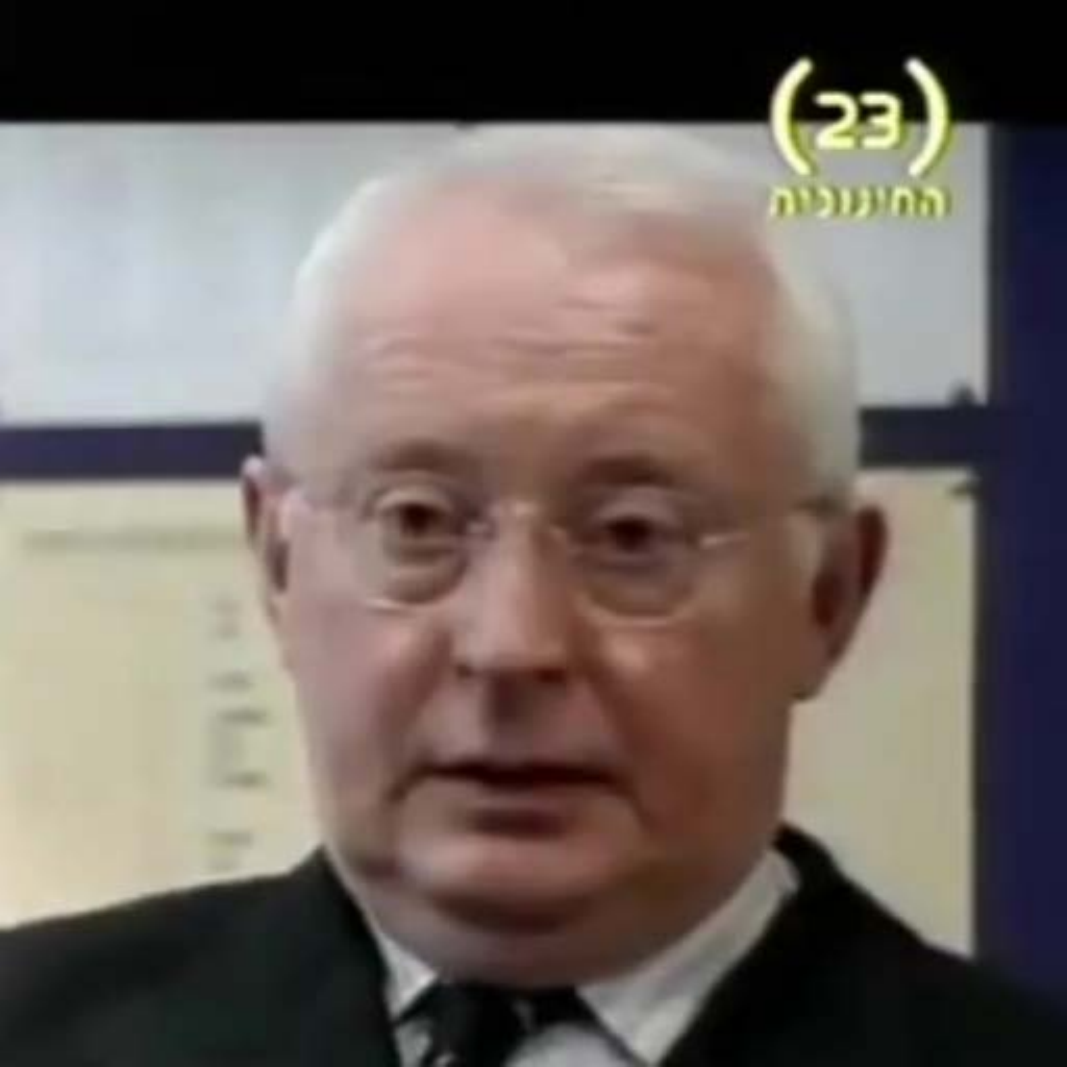} &
    \includegraphics[width=\swteasa]{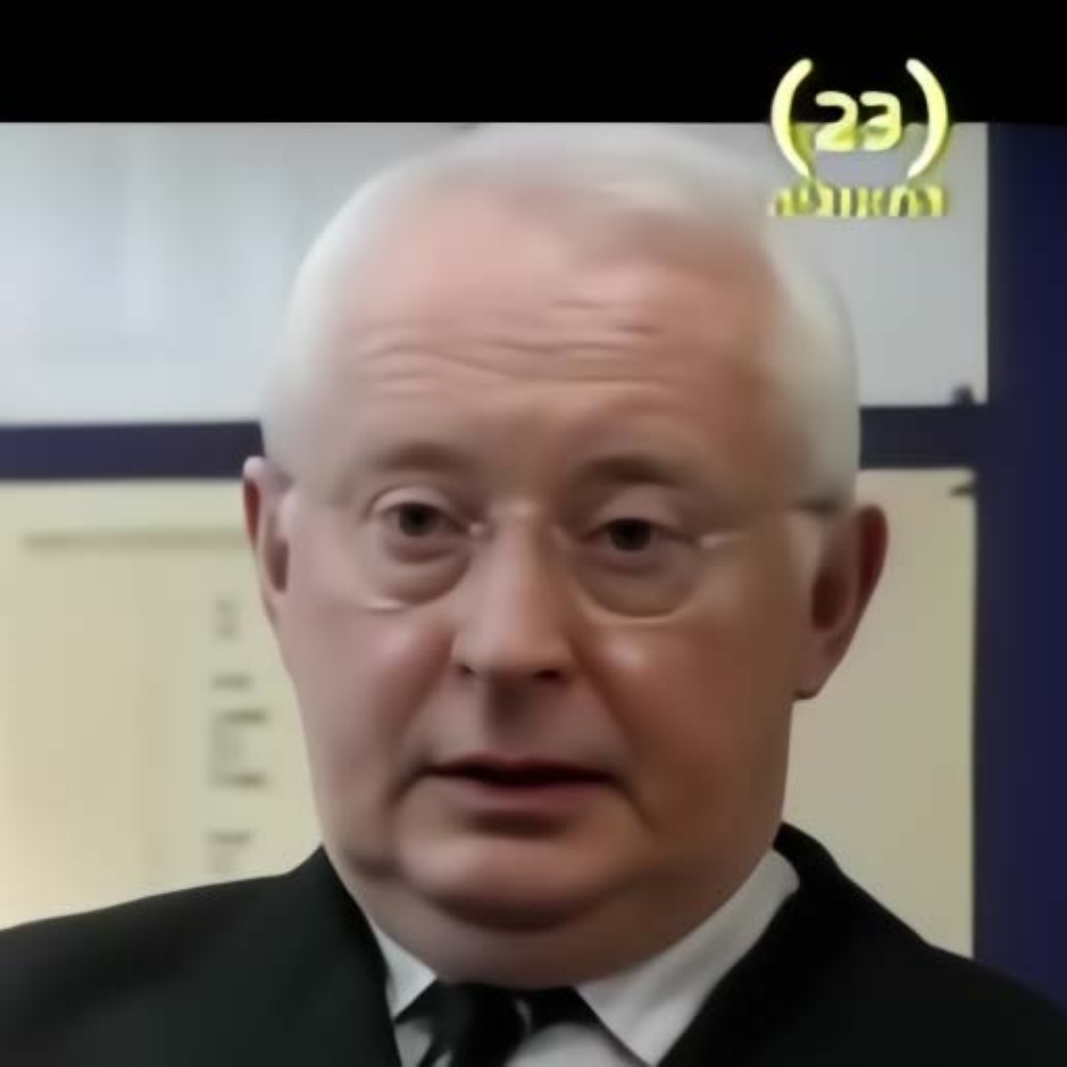} &
    \includegraphics[width=\swteasa]{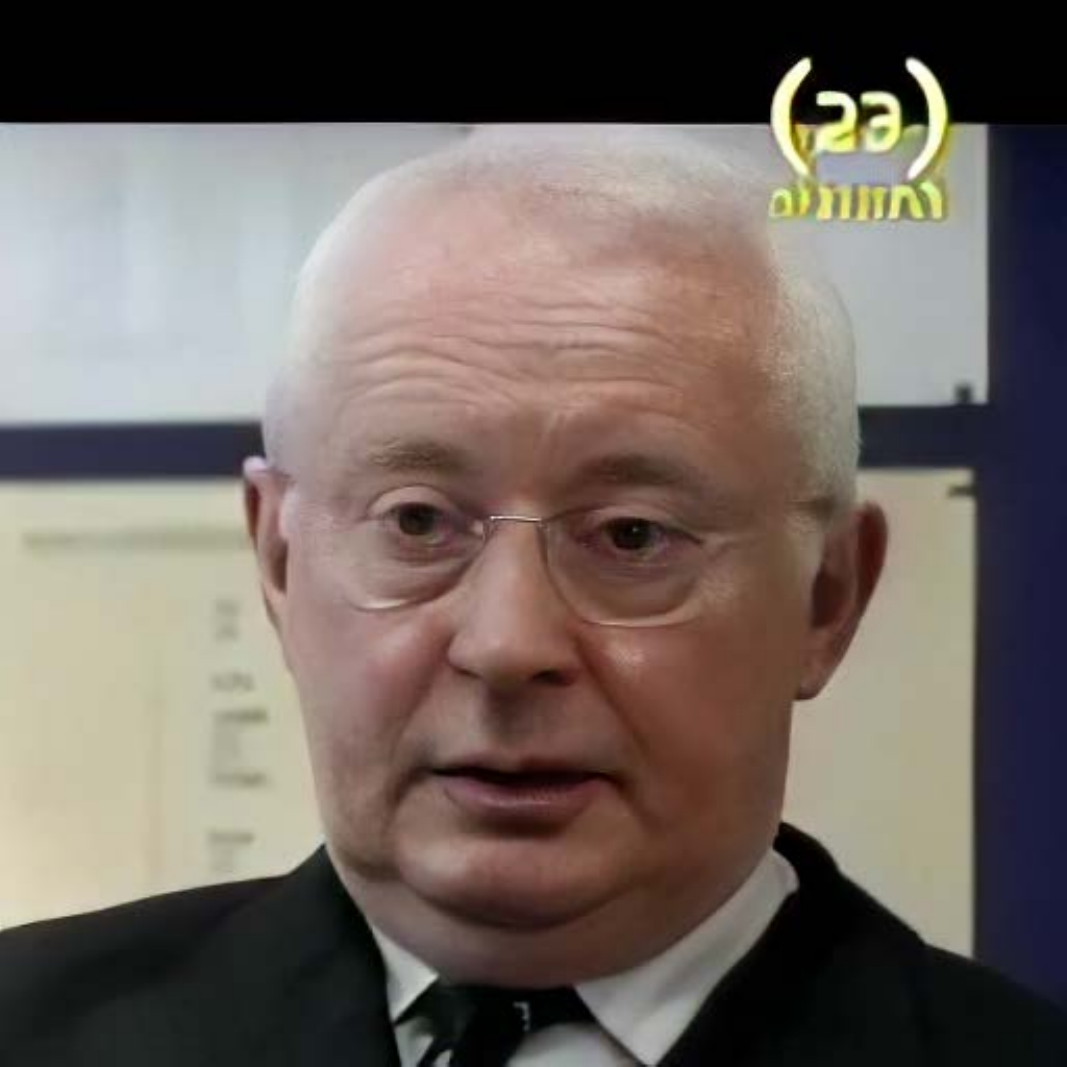} &
    \includegraphics[width=\swteasa]{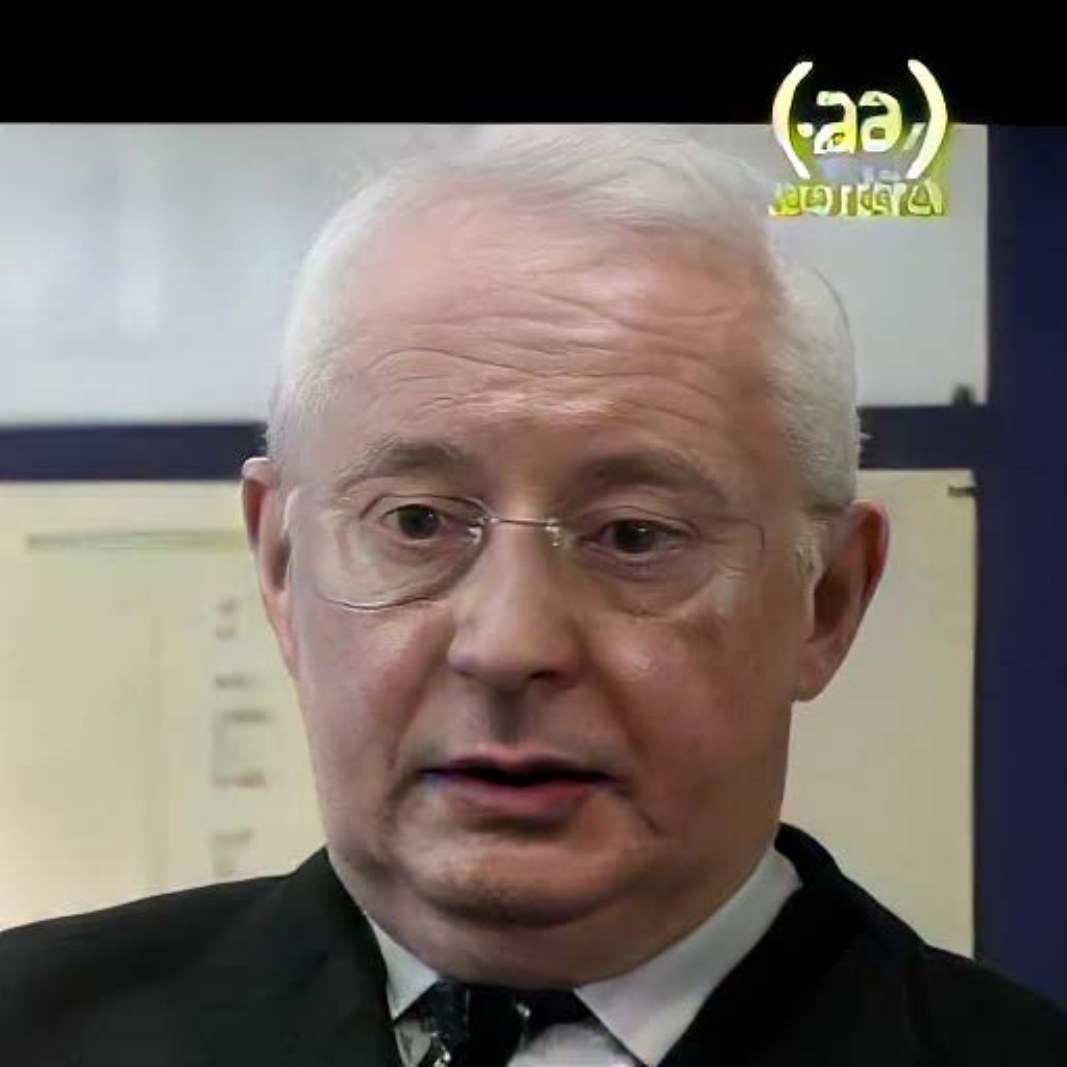} &
    \includegraphics[width=\swteasa]{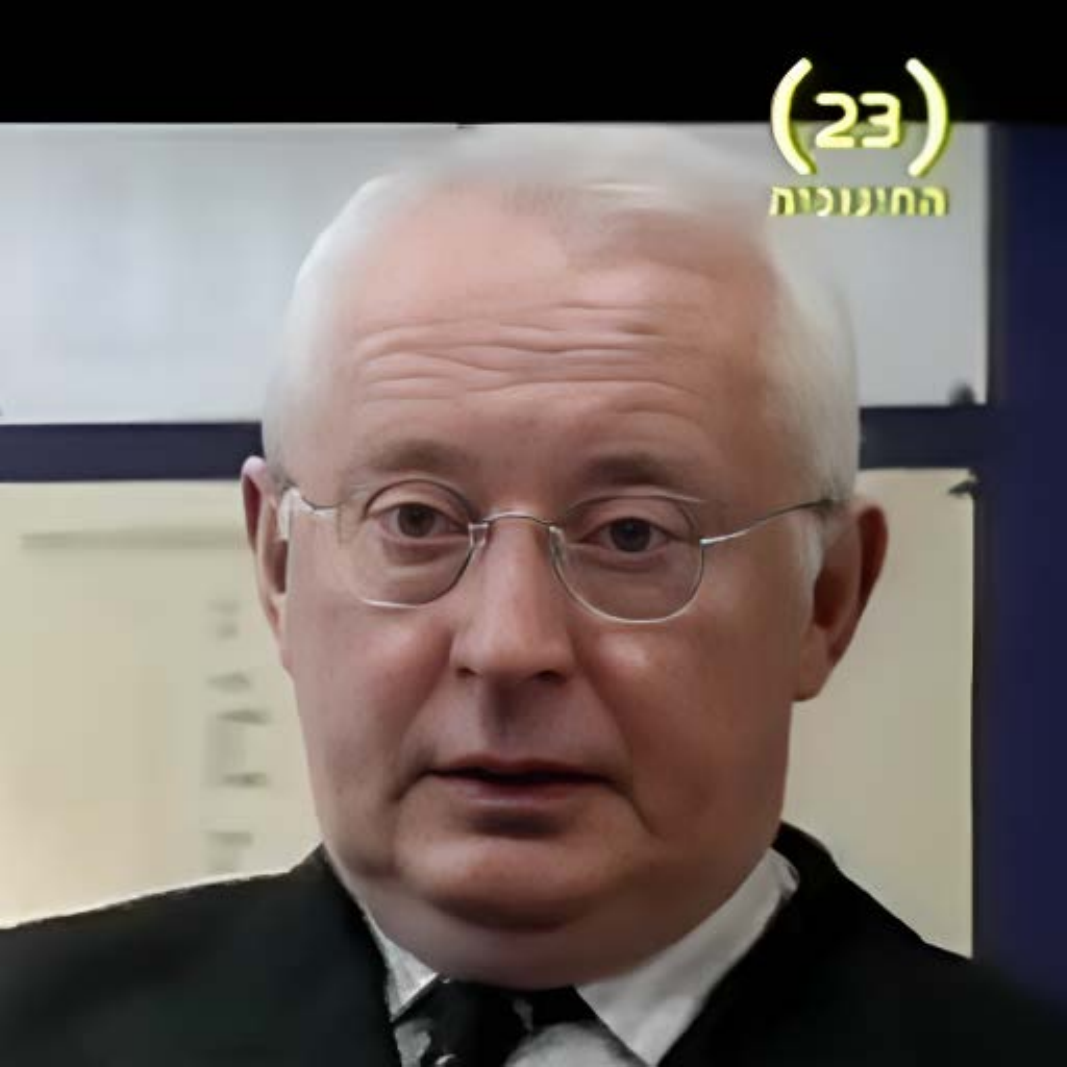} &
    \includegraphics[width=\swteasa]{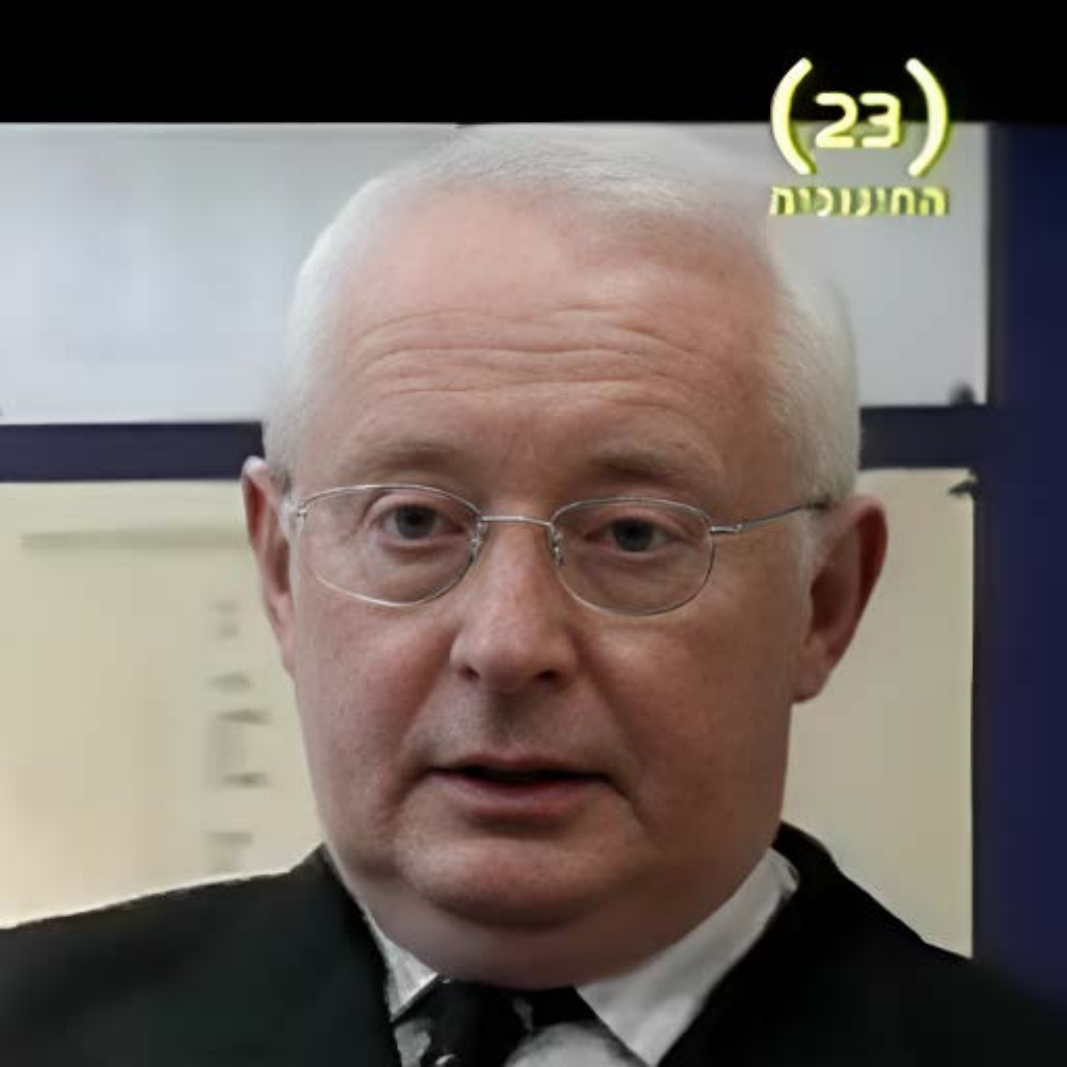} &
    \includegraphics[width=\swteasa]{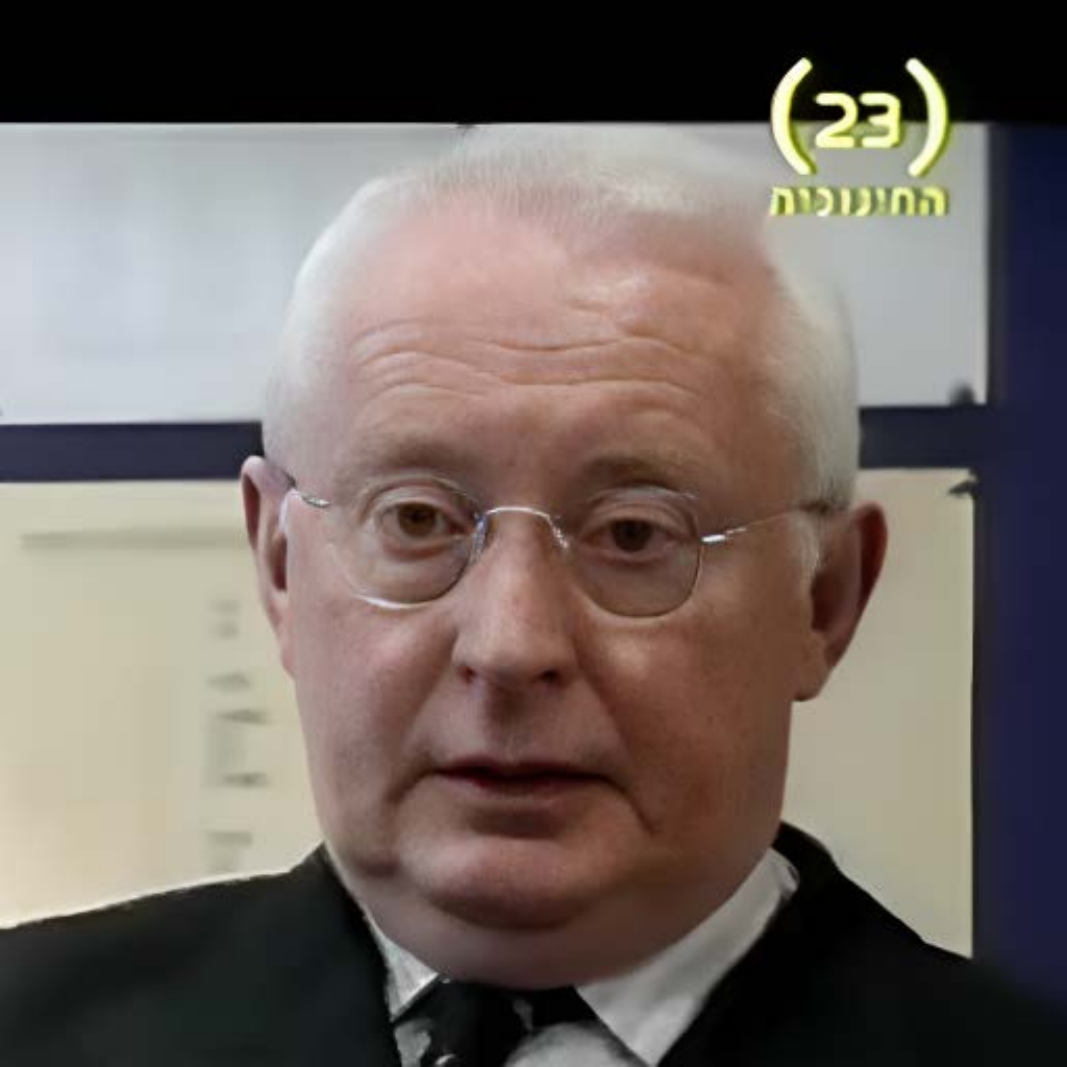} \\
    \includegraphics[width=\swteasa]{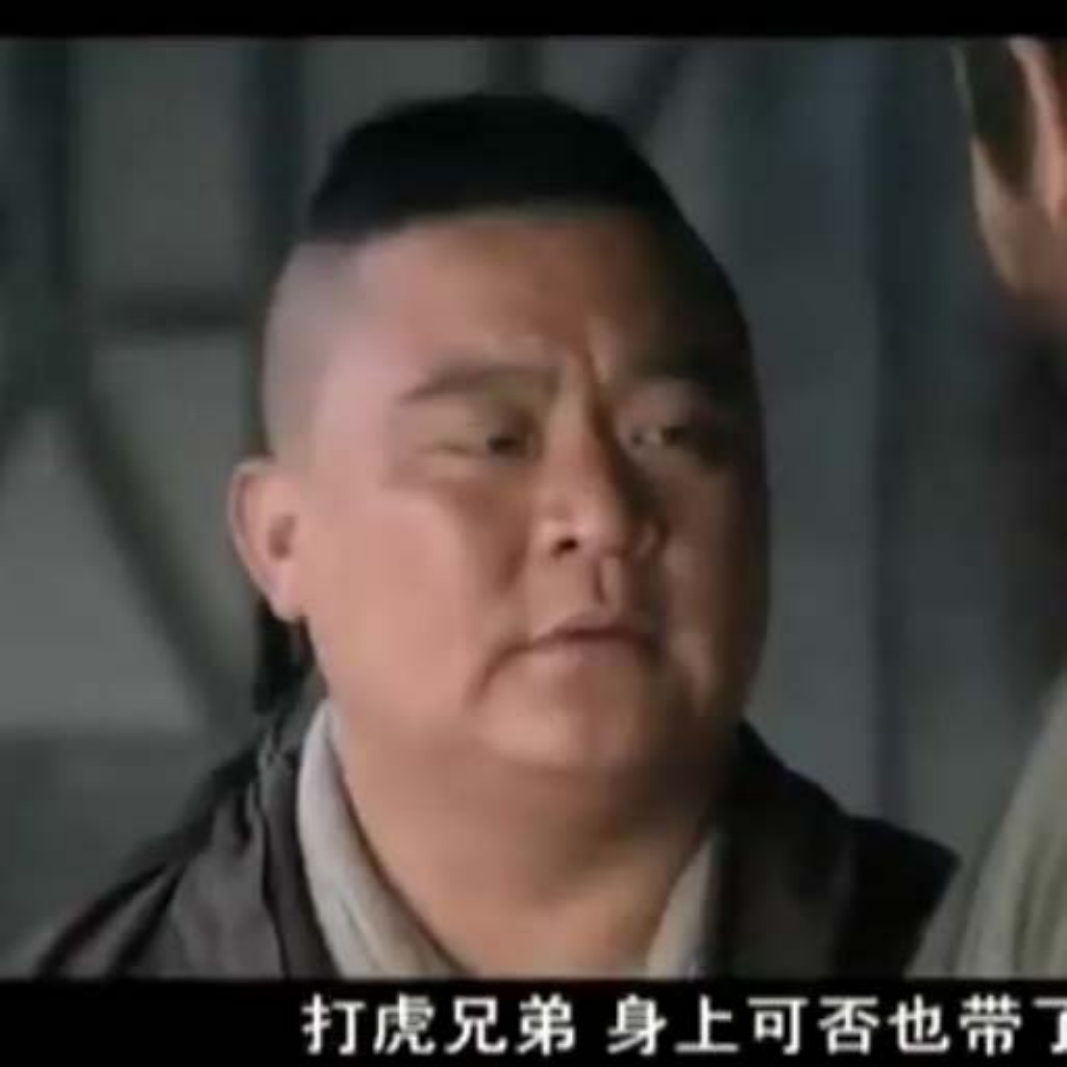} &
    \includegraphics[width=\swteasa]{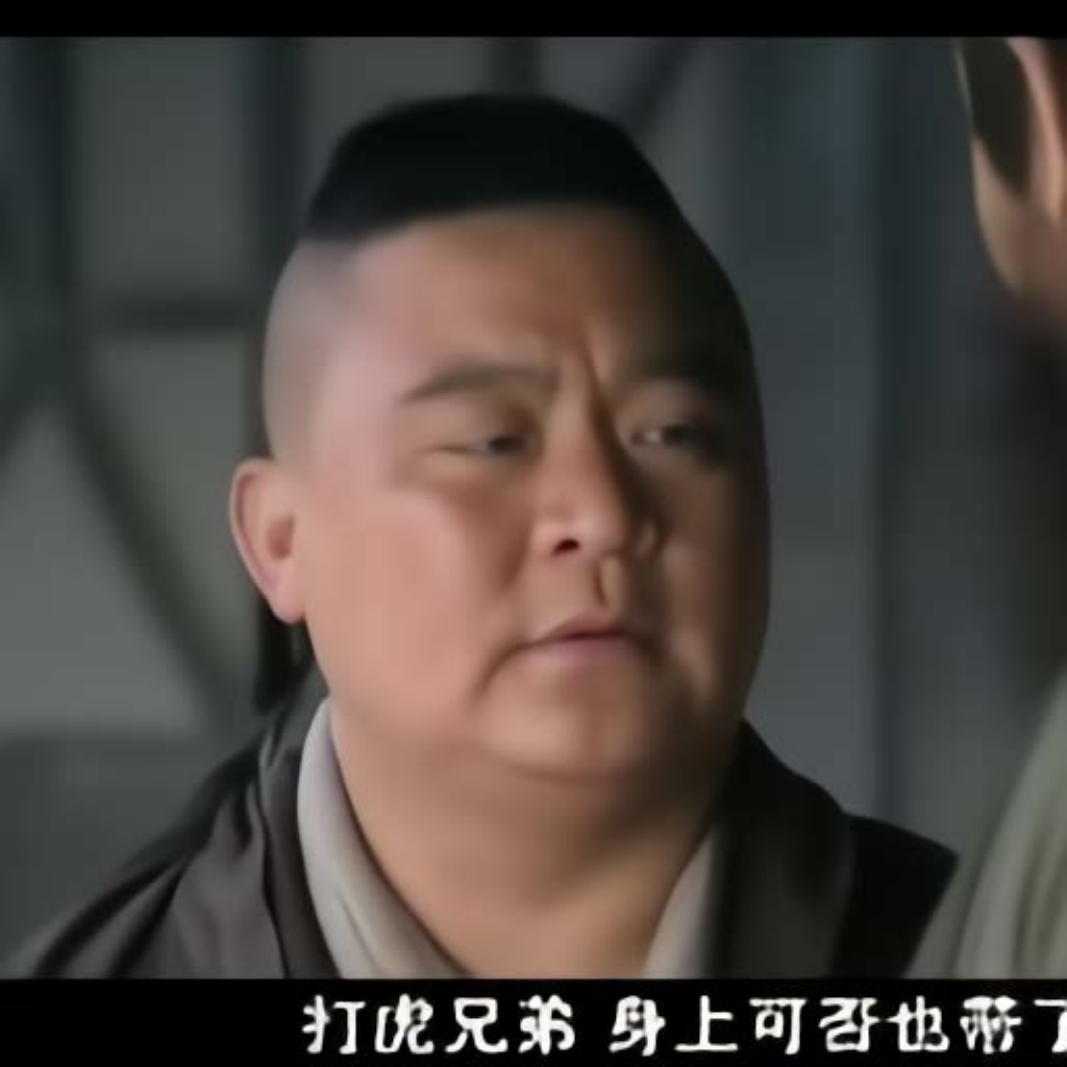} &
    \includegraphics[width=\swteasa]{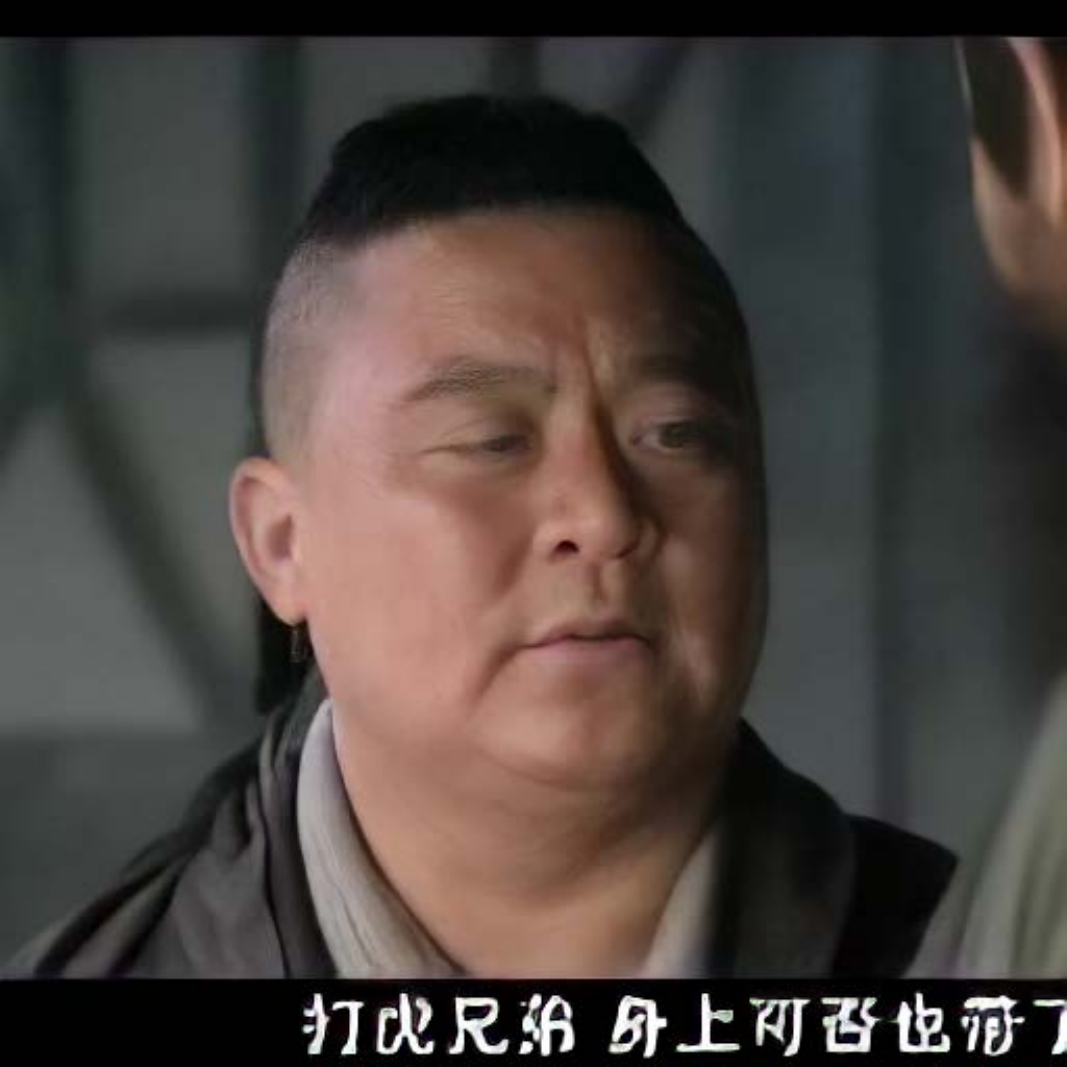} &
    \includegraphics[width=\swteasa]{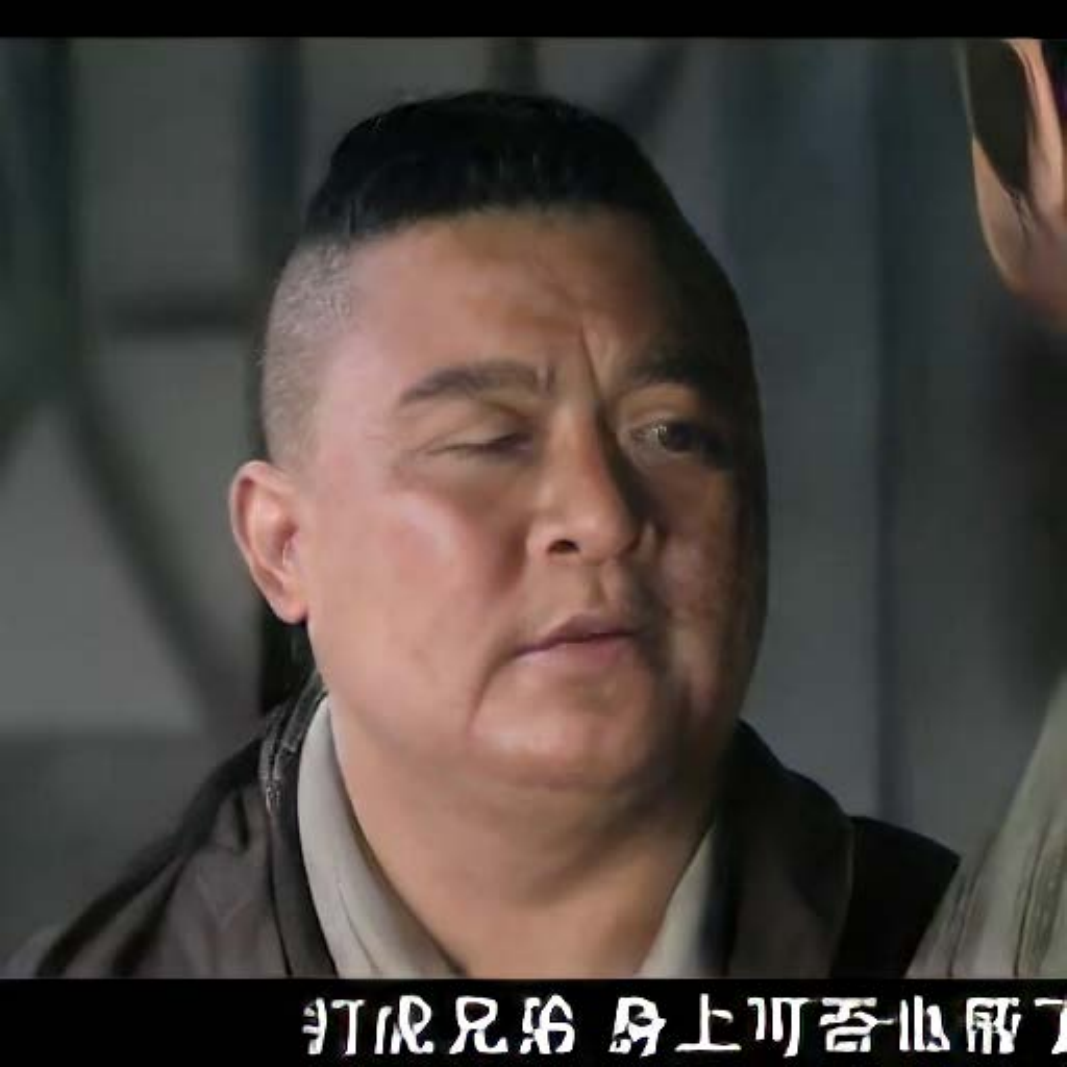} &
    \includegraphics[width=\swteasa]{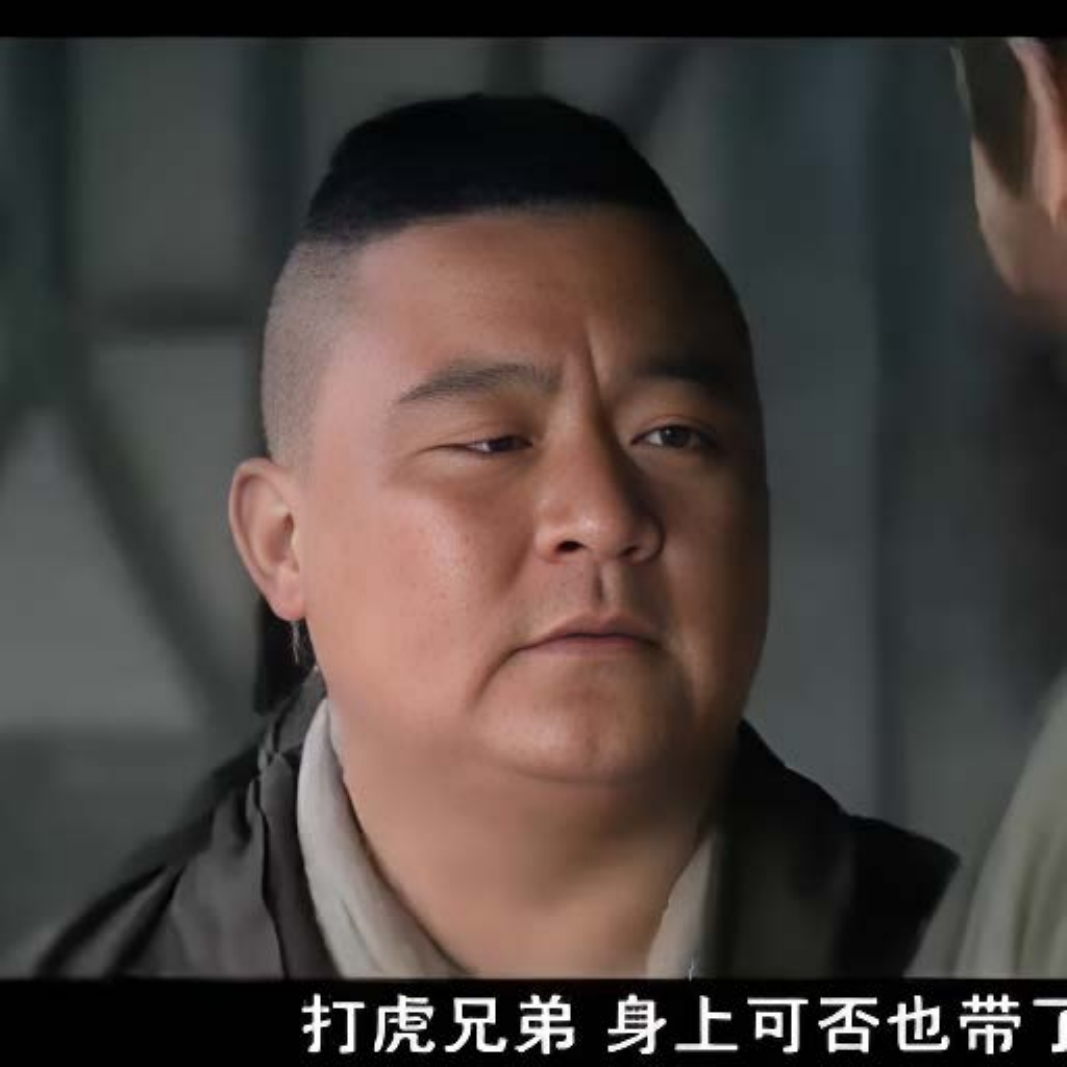} &
    \includegraphics[width=\swteasa]{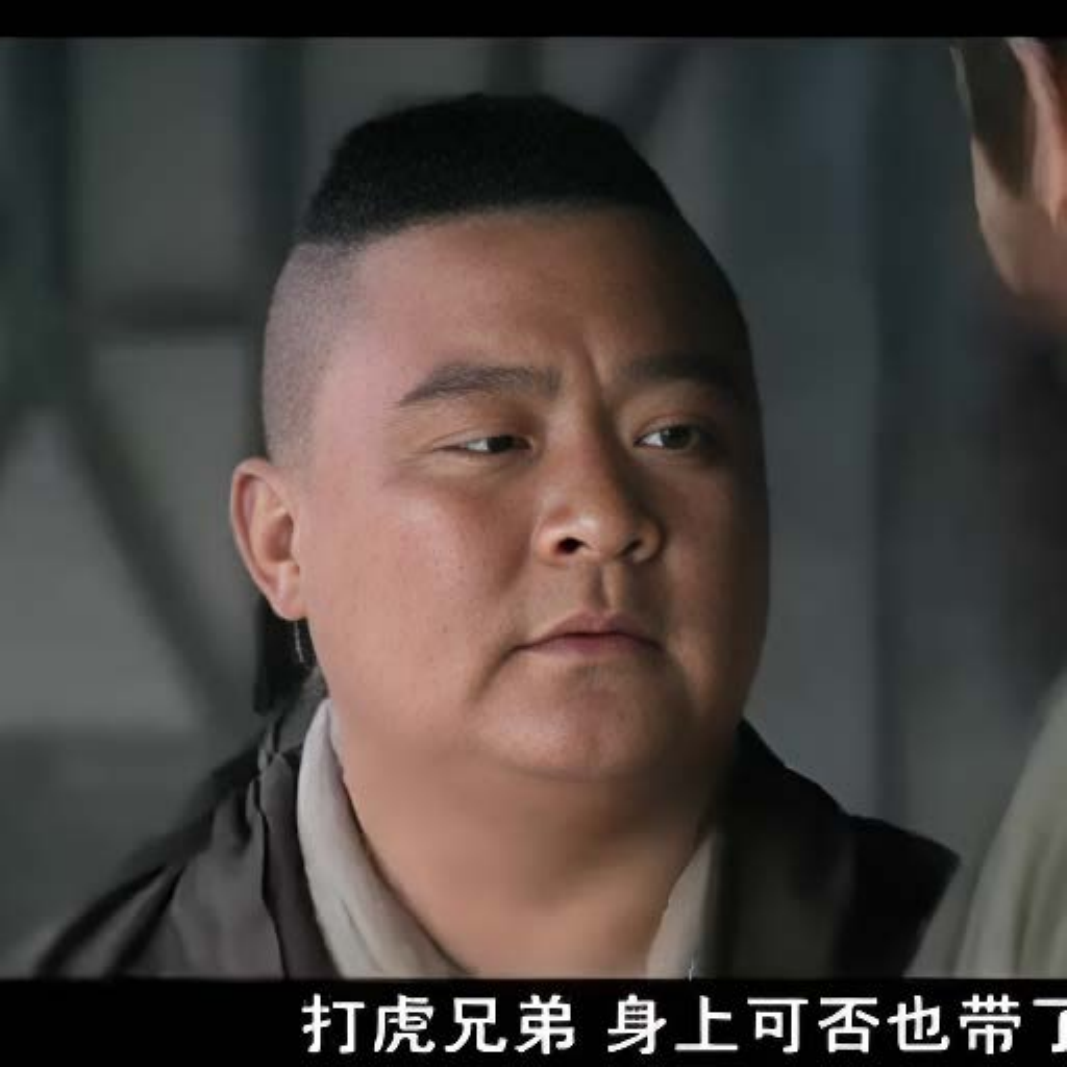} &
    \includegraphics[width=\swteasa]{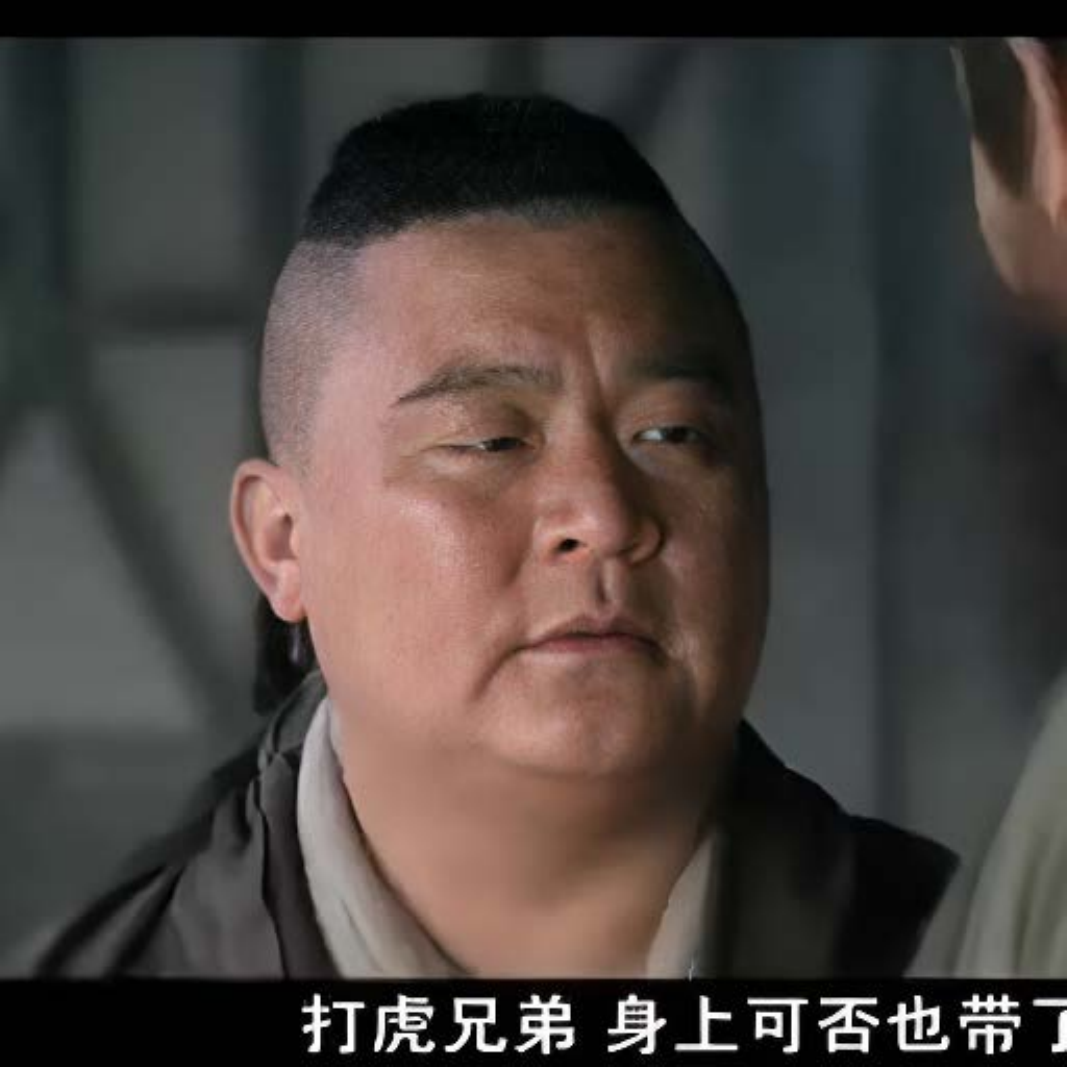} \\
    Input & EDVR~\cite{wang2019edvr} & EDVR-GAN~\cite{wang2019edvr} & BasicVSR-GAN~\cite{chan2021basicvsr} & GFP-GAN~\cite{wang2021towards} & RestoreFormer++~\cite{wang2023restoreformer++} & CodeFormer~\cite{zhou2022towards}
\end{tabular}
\caption{
\textbf{Qualitative comparisons between video-based and image-based methods.} Existing state-of-the-art image-based face restoration algorithms, such as GFP-GAN~\cite{wang2021towards}, RestoreFormer++~\cite{wang2023restoreformer++}, and Codeformer~\cite{zhou2022towards}, achieve higher restoration quality than existing video-based methods, such as EDVR~\cite{wang2019edvr}, EDVR-GAN~\cite{wang2019edvr}, and BasicVSR-GAN~\cite{chan2021basicvsr}, especially in recovering \textbf{teeth with rich details, glasses with complete structure, and eyes with a natural look}. \textbf{Corresponding videos are in the supplementary materials.}}
\label{fig:fvr_fir}
\end{figure*}

\section{Introduction}

Blind face restoration is a challenging task that aims to recover a high-quality face image from its degraded one without knowing the type and degree of degradation. The degradation can be caused by various factors, such as blur, noise, downsampling, compression artifacts, \textit{etc}. Recent methods have achieved impressive results on blind face image restoration~\cite{chen2021progressive,li2020enhanced,li2018learning,li2020blind,wan2020bringing,yang2020hifacegan}, especially by leveraging generative priors~\cite{wang2021towards,yang2021gan} and codebook~\cite{wang2022restoreformer,wang2023restoreformer++,gu2022vqfr,zhou2022towards}. However, blind face video restoration remains relatively under-explored, despite its practical significance for enhancing low-quality videos that contain degraded faces, such as talk shows, TV series, and movies.

In this paper, we first advance the research on blind face video restoration by introducing a \textbf{R}eal-world \textbf{L}ow-\textbf{Q}uality \textbf{F}ace \textbf{V}ideo benchmark (RFV-LQ), which poses significant challenges for existing methods. RFV-LQ consists of low-quality face videos that are carefully collected from various sources, such as old talk shows, TV series, and movies. We conduct extensive experiments to evaluate the performance of current representative video restoration methods on RFV-LQ, such as EDVR~\cite{wang2019edvr} and BasicVSR~\cite{chan2021basicvsr}. \textcolor{black}{Although these methods originally aim for video super-resolution, Xie \textit{et al.}~\cite{xie2022vfhq} retrained them with degraded face videos synthesized from VFHQ~\cite{xie2022vfhq}, a high-quality face video dataset, for blind face video restoration.} As shown in Fig.~\ref{fig:fvr_fir}, these video-based methods fail to recover facial details. Their results are relatively blurry and lack facial details, especially in the areas of nose and teeth. We apply three state-of-the-art blind face image restoration algorithms, GFP-GAN~\cite{wang2021towards}, RestoreFormer++~\cite{wang2023restoreformer++}, and CodeFormer~\cite{zhou2022towards} on RFV-LQ by processing face videos frame-by-frame. We observe that these image-based methods can restore faces from videos with richer facial details, especially in the mouth area (examples are the last three columns in Fig.~\ref{fig:fvr_fir}, and its advantages over video-based methods in quantitative metrics are in Table~\ref{tab:analyze}). 

\textcolor{black}{
Motivated by this observation, we propose to systemically analyze the benefits and challenges of extending current face image restoration algorithms to face video restoration and propose effective solutions based on our analyses. Specifically, we evaluate 3 state-of-the-art blind face image restoration algorithms on RFV-LQ, including GFP-GAN~\cite{wang2021towards}, RestoreFormer++~\cite{wang2023restoreformer++}, and CodeFormer~\cite{zhou2022towards}. Benefiting from their high-quality training data and mature techniques, especially the introduction of high-quality priors, their restored performance is significantly higher compared to video-based methods.
However, there also exists some challenges, such as noticeable jitters in the facial components (Fig.~\ref{fig:jitter}), and significant noise-shape flickers between frames (Fig.~\ref{fig:flicker}).  We analyze these challenges with qualitative visualizations and quantitative metrics and identify that there are three main factors that lead to the jitters and flickers: the employment of GAN~\cite{gan} loss, the ignorance of the temporal information within the videos, and the bias introduced by face alignment required by image-based face restoration algorithms.
}

To address these challenges, we propose a temporal consistency network (TCN) cooperating with alignment smoothing. The alignment smoothing averages the detected landmarks of several neighboring frames to reduce the bias introduced by face misalignment between frames, while TCN models the temporal information in a video to alleviate the jitters and flickers. TCN can be easily integrated with existing face image restoration networks. It consists of Cross-Attended Swin Transformer Layers (CASTL), which use cross-attention blocks to capture the temporal information accorss the current and previous frames. Unlike previous works~\cite{bonneel2015blind,lai2018learning,lei2020blind} that deploy temporal consistency in the post-processing. Our TCN concurrently injects the temporal information into the restored results while recovering from the original degraded face video with well-trained image-based face restoration algorithms. It can enhance the temporal consistency of the restored results and in the meantime has the potential to correct the restored error raised by image-based models.

The contributions of this paper can be summarized as:
\begin{itemize}
    \item \redm{We systematically analyze the potential benefits and challenges of extending existing state-of-the-art image-based face restoration algorithms to blind face video restoration, both qualitatively and quantitatively.}
    \item \redm{We propose a transformer-based Temporal Consistency Network (TCN) combined with alignment smoothing to reduce jitters and flickers in the restored results while maintaining the original quality of face image restoration models. This lightweight plug-and-play component has the potential to correct the restored bias introduced by the image-based methods and offers a processing time that is approximately 3x to 10x faster then previous methods.}
    \item \redm{We introduce a real-world low-quality face video benchmark (RFV-LQ) from diverse real-world degraded raw videos. It can serve as a valuable resource for future research in face video restoration.}
    \item We have conducted extensive experiments to demonstrate the superiority of our proposed method against previous temporal consistency algorithms.
\end{itemize}

The remainder of this paper is structured as follows. In Section~\ref{sec:related_work}, we review the most relevant previous works. In Section~\ref{sec:systemic_analyses}, we introduce a real-world low-quality face video benchmark and provide insightful analyses of the benefits and challenges of applying image-based face restoration algorithms to real-world degraded face videos. Subsequently, in Section~\ref{sec:method}, we present our proposed method in detail, designed to address these challenges. In Section~\ref{sec:exp}, we evaluate the effectiveness and efficiency of our proposed method through extensive experimentation. Finally, we draw our conclusions in Section~\ref{sec:conclusion}.

\begin{table*}[!t]
\caption{
\redm{
\textbf{Quantitative analyses on video-based and image-based blind face restoration methods on our proposed degraded face video benchmark RFV-LQ.}
The quantitative results indicate that the image-based methods perform better in the restored quality, especially in the facial components, but with room for improvement in MSI and Warping Error, which represent the jitters and flicker in the restored results.
The \textcolor{red}{best} and \underline{second-best} scores of each metric are highlighted in red and underlined, respectively.
}
}
\centering
\resizebox{\linewidth}{!}{
\begin{tabular}{c|ccccc|cc}
\hline
Method & FID~$\downarrow$ & FID-Eyes~$\downarrow$ & FID-Nose~$\downarrow$ & FID-Mouth~$\downarrow$ & FID-Hair~$\downarrow$ & MSI~$\downarrow$ & Warping Error~$\downarrow$ \\
\hline
\textbf{Video-Based Methods:} & &&&&&& \\
EDVR~\cite{wang2019edvr} & 105.0535 & 123.2287 & 96.0719 & 145.2019 & 126.7183 & 5.3281 & 0.000739 \\
EDVR-GAN~\cite{xie2022vfhq} & 66.6829 & 55.9353 & 37.1725 & 69.5077 & 56.4627 & 5.7391 & 0.00087 \\
BasicVSR~\cite{chan2021basicvsr} & 101.7159 & 133.0369 & 106.6571 & 155.5566 & 127.0548 & 5.3952 & 0.000747 \\
BasicVSR-GAN~\cite{xie2022vfhq} & \textcolor{red}{65.7563} & 58.8753 & 46.7167 & 78.9885 & 62.7220 & 6.3402 & 0.001108 \\
\hline
\hline
\textbf{Image-Based Methods:} & &&&&&& \\
GFP-GAN~\cite{wang2021towards} & 70.3217 & \underline{50.4729} & 33.9919 & 47.9370 & 59.3655 & 5.7917 & 0.000920 \\
RestoreFormer++~\cite{wang2023restoreformer++} & \underline{66.1725} & 53.9941 & \textcolor{red}{31.7954} & \textcolor{red}{43.5653} & \underline{52.3961} & 6.873 & 0.001058 \\
CodeFormer~\cite{zhou2022towards} & 69.6650 & \textcolor{red}{49.1728} & \underline{32.2229} & \underline{44.6944} & \textcolor{red}{51.6947} & 6.9899 & 0.001035 \\
\hline
\end{tabular}
}
\label{tab:analyze}
\end{table*}

\section{Related Works}
\label{sec:related_work}

\subsection{Blind Face Image Restoration}
\redm{Recent years have witnessed the development of blind face image restoration~\cite{chen2018fsrnet,kim2019progressive,zhu2016deep,shen2018deep,chen2021progressive,yu2018face,dogan2019exemplar,li2020enhanced,li2018learning,li2020blind,wang2022restoreformer,wang2023restoreformer++,gu2022vqfr,zhou2022towards,8733990,10087319,1518942,yasarla2020deblurring}, especially with the powerful neural network~\cite{krizhevsky2012imagenet,simonyan2014very,he2016deep}.
}
The mainstream algorithms that have been proven to attain high-quality faces are prior-based methods.
These priors can be divided into three categories: geometric priors, reference, and generative priors.
Since face is highly structured, geometric priors, such as facial landmarks~\cite{chen2018fsrnet,kim2019progressive,zhu2016deep}, face parsing maps~\cite{shen2018deep,chen2021progressive}, and facial component heatmaps~\cite{yu2018face}, are beneficial to the reconstruction of the facial structure while restoring degraded face images.
Different from the geometric priors that mainly provide auxiliary facial structures, reference tends to acquire high-quality facial details from existing high-quality face images.
The reference can be some high-quality exemplars with the same identity as the degraded face image~\cite{dogan2019exemplar,li2020enhanced,li2018learning}. It also can be high-quality facial features of abundant high-quality face images extracted with an off-line feature extractor~\cite{li2020blind,li2022learning} or learning by a Vector-quantized face reconstruction model~\cite{wang2022restoreformer,wang2023restoreformer++,gu2022vqfr,zhou2022towards}.
Besides, the generative priors encapsulated in the powerful generation model, e.g. StyleGAN2~\cite{karras2020analyzing}, are also helpful for face image restoration.
Since iteratively optimizing the latent for GAN inversion~\cite{menon2020pulse,gu2020image} or directly encoding degraded face to latent space~\cite{wan2020bringing} tend to lose fidelity of the restored faces, most of the related algorithms propose to embed the generative priors into a decoder structure for fusing both the information of the degraded face image and its corresponding priors.
Currently, the algorithms based on generative priors, such as GFP-GAN~\cite{wang2021towards} and GPEN~\cite{yang2021gan}, and reference learning with a Vector-quantized face reconstruction model, such as RestoreFormer~\cite{wang2022restoreformer}, RestoreFormer++~\cite{wang2023restoreformer++}, VQFR~\cite{gu2022vqfr}, and CodeFormer~\cite{zhou2022towards}, perform better than the other existed methods for blind face restoration.
In this paper, we apply these advanced methods to real-world face video restoration and make an insightful analysis of their manifestations.
We expect to explore the potential of face image restoration methods for face video restoration for attaining high-quality face videos with limited training data.

\subsection{Image Processing Models for Videos}
Constrained by the limited memory and computational resource, and the availability of training datasets, there are also some previous works~\cite{bonneel2015blind,lai2018learning,lei2020blind,xu2022temporally,chandran2022temporally} attempt to process videos by leveraging its image processing algorithms counterpart.
They face a common challenge -- temporal consistency, which corresponds to the jitters and flickers that occurred in our works.
These works, whether task-specific~\cite{xu2022temporally,chandran2022temporally} or task-agnostic~\cite{bonneel2015blind,lai2018learning,lei2020blind}, tend to improve temporal consistency as a \textit{post-processing} operation.
Generally, they~\cite{lai2018learning,chandran2022temporally} deploy a consistency network to model the temporal information in the videos that have been pre-processed by its image processing model and design a temporal loss to supervise the learning.
One different work is DVP~\cite{lei2020blind}. It proposes to rebuild temporal consistency with deep video prior.
These methods are implemented with a relatively large network and largely change the results of its image processing algorithm. 
On the contrary, in this work, we attempt to enforce temporal consistency and mitigate the jitters and flickers in the restored results attained by face image algorithms with a lightweight transformer block.
Instead of acting as a post-processing operation, this transformer block can be flexibly plugged into existing face image models and enhance the temporal consistency while restoring each frame.
Besides, since it only modified very few latent features in the face image model, it can preserve the restoration effect of the face image model to a great extent.

\begin{figure}[t]
\setlength\tabcolsep{1pt}
\scriptsize
\centering
\includegraphics[width=0.98\linewidth]{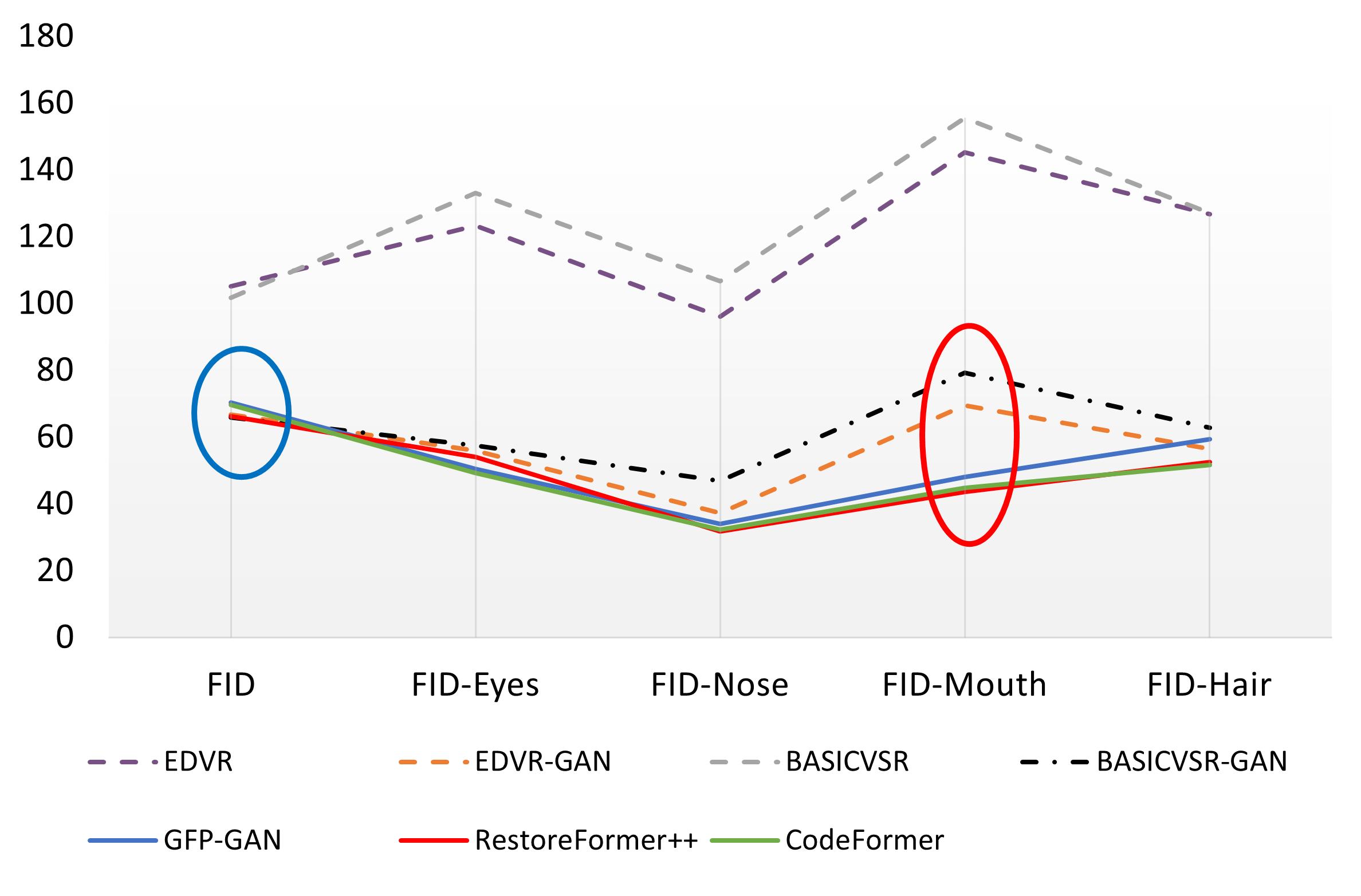}

\caption{\redm{\textbf{FID V.S. Component FIDs}. While the scores of image-based methods (including GFP-GAN~\cite{wang2021towards}, RestoreFormer++~\cite{wang2023restoreformer++}, and CodeFormer~\cite{zhou2022towards}) and video-based methods implemented with GAN loss (such as BasicVSR-GAN~\cite{gu2022vqfr}) are close in terms of FID (\textcolor{blue}{blue circle}), the advantages of image-based methods become significant when measured with Component FIDs, especially in terms of FID-Mouth (\textcolor{red}{red circle}). It is more consistent with the observation in the qualitative results (refer to the high-quality restored teeth of image-based methods in Fig.~\ref{fig:fvr_fir}), which demonstrates the accuracy of Component FIDs compared to FID.
}}
\label{fig:component_fids}
\end{figure}

\section{Systemic Analyses and Benckmarking}
\label{sec:systemic_analyses}
\subsection{Dataset}

In this section, before going through the systemic analyses, we first introduce our collected real-world low-quality face video dataset (RFV-LQ), which is used to evaluate the advantages and challenges of existing methods. RFV-LQ contains degraded face videos that are carefully collected from various raw videos downloaded from YouTube, which cover diverse scenarios, such as old talk videos, TV series, and old movies. We follow the data preprocessing proposed in~\cite{xie2022vfhq} to extract degraded face videos from these downloaded raw videos. Specifically, we first use a face detector RetinaNet~\cite{deng2020retinaface} to detect the faces in the raw videos and then use a tracking method SORT~\cite{bewley2016simple} to group them into different face tracks based on identity. In this step, we remove the faces whose size is smaller than $100 \times 100$ since they are too small to be accurately tracked by the existing tracking methods. Then, we use ArcFace~\cite{deng2019arcface}, a face recognition model, to measure the identity distance of faces in every track and discard the track if their identity distance is larger than a threshold (1.24 in our work). To reduce the identity redundancy among tracks, we randomly select two tracks belonging to the same raw video and identity in RFV-LQ, making the dataset diverse yet lightweight. Finally, we obtain 329 real-world degraded face video tracks in RFV-LQ.

\begin{figure*}[!t]
\setlength\tabcolsep{1pt}
\scriptsize
\centering
\begin{tabular}{ccccccc}
    \scalebox{1.35}{\includegraphics[width=\swjitterr]{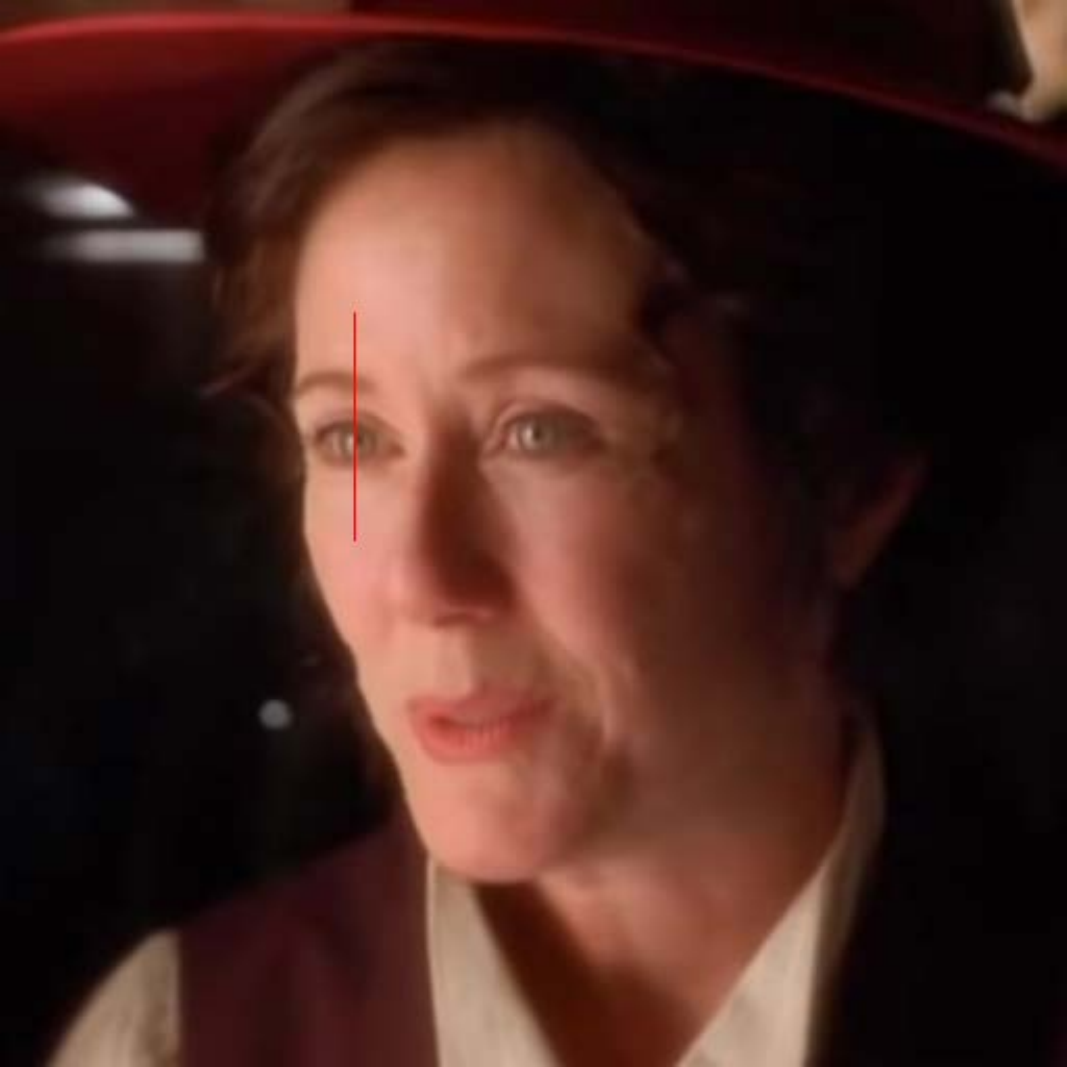}} &
    \includegraphics[width=\swjitterr]{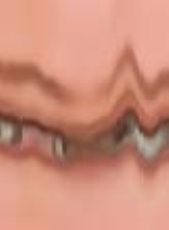} & 
    \includegraphics[width=\swjitterr]{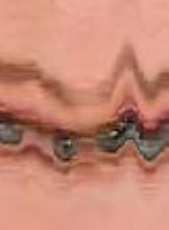} & 
    \includegraphics[width=\swjitterr]{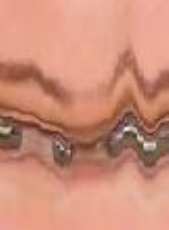} & 
    \includegraphics[width=\swjitterr]{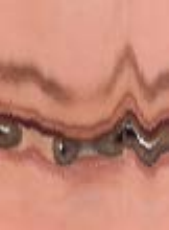} &
    \includegraphics[width=\swjitterr]{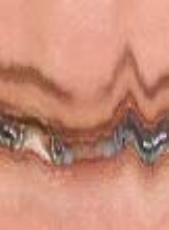} & 
    \includegraphics[width=\swjitterr]{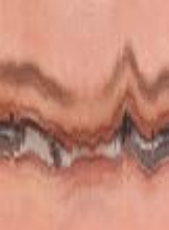}\\
    \scalebox{1.35}{\includegraphics[width=\swjitterr]{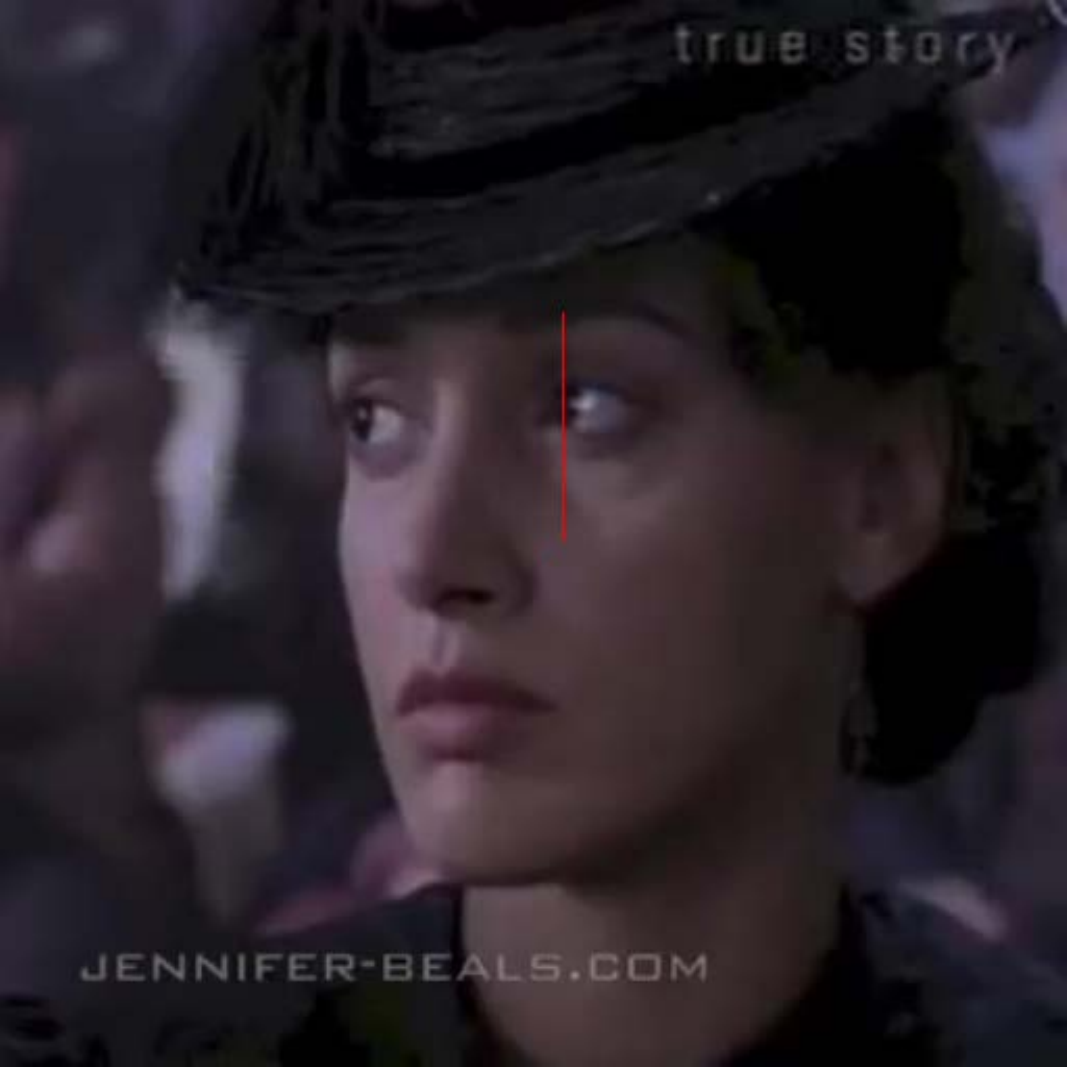}} &
    \includegraphics[width=\swjitterr]{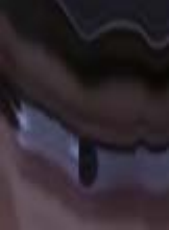} & 
    \includegraphics[width=\swjitterr]{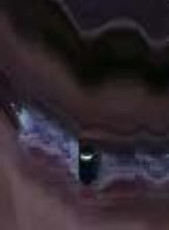} & 
    \includegraphics[width=\swjitterr]{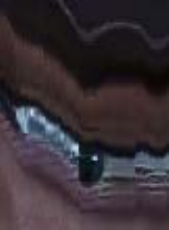} & 
    \includegraphics[width=\swjitterr]{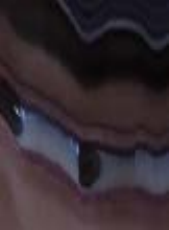} &
    \includegraphics[width=\swjitterr]{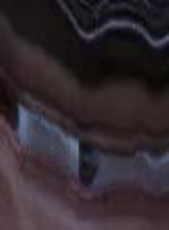} & 
    \includegraphics[width=\swjitterr]{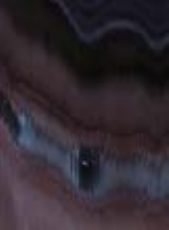}\\
    (a) Image & (b) EDVR~\cite{wang2019edvr} & (c) EDVR-GAN~\cite{wang2019edvr} & (d) BasicVSR-GAN~\cite{chan2021basicvsr} & (e) GFP-GAN~\cite{wang2021towards} & (f) RestoreFormer++~\cite{wang2023restoreformer++} & (g) CodeFormer~\cite{zhou2022towards}
\end{tabular}
\caption{\redm{\textbf{Visualization of jitters in facial components.} (b)$ \sim $(g) show the concatenated vertical slices (the red vertical line in (a))
in every frame along the time by the results of existing video-based and image-based face restoration methods. Compared to the relatively smooth slices in the results of EDVR~\cite{wang2019edvr}, the restored results of the video-based methods with GAN~\cite{gan} and image-based methods exist noticeable jitters.
\textbf{Corresponding videos are in the supplementary materials}.
}}
\label{fig:jitter}
\end{figure*}

\begin{figure*}[ht]
\setlength\tabcolsep{1pt}
\scriptsize
\centering
\begin{tabular}{cccccc}
    \includegraphics[width=\swflicker]{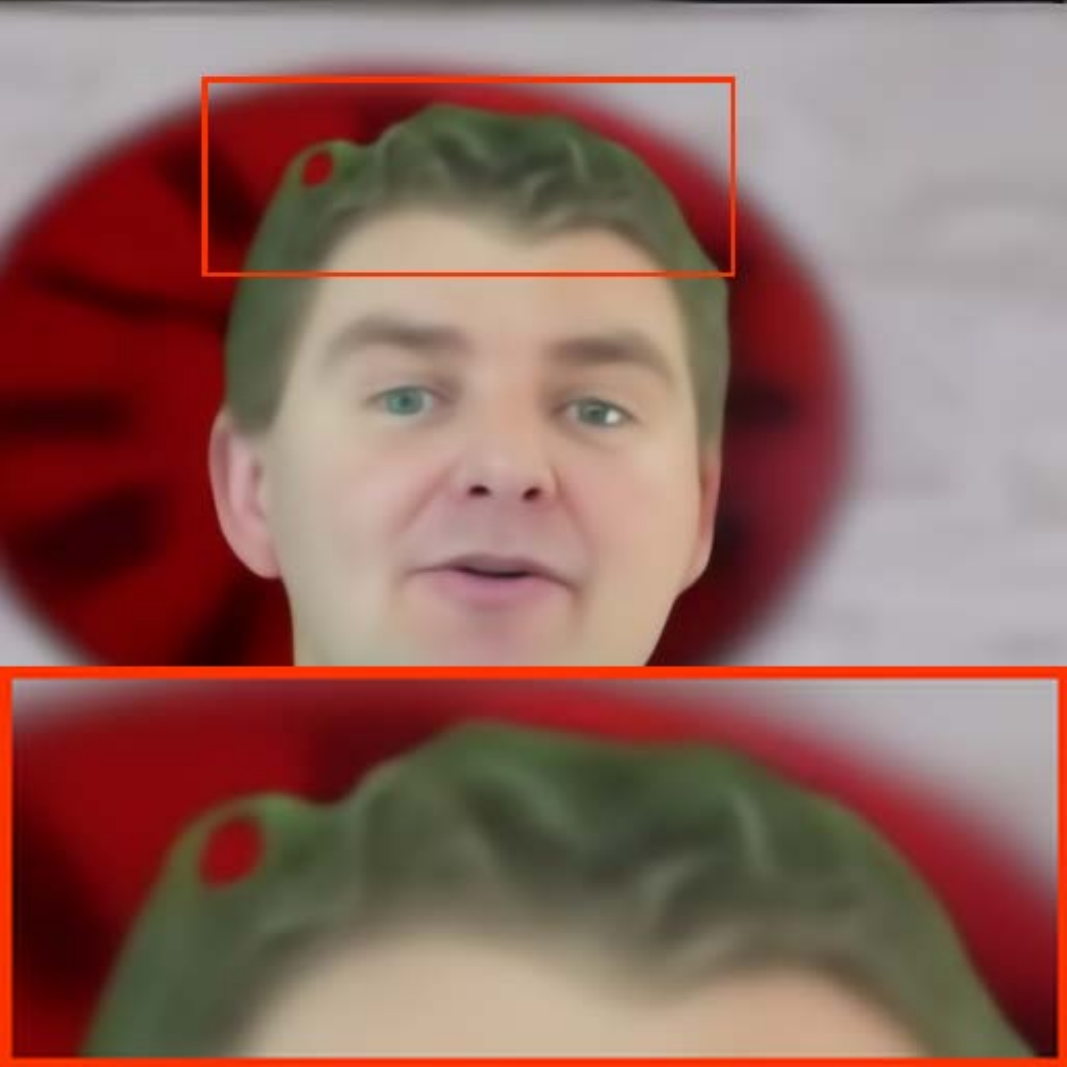} &
    \includegraphics[width=\swflicker]{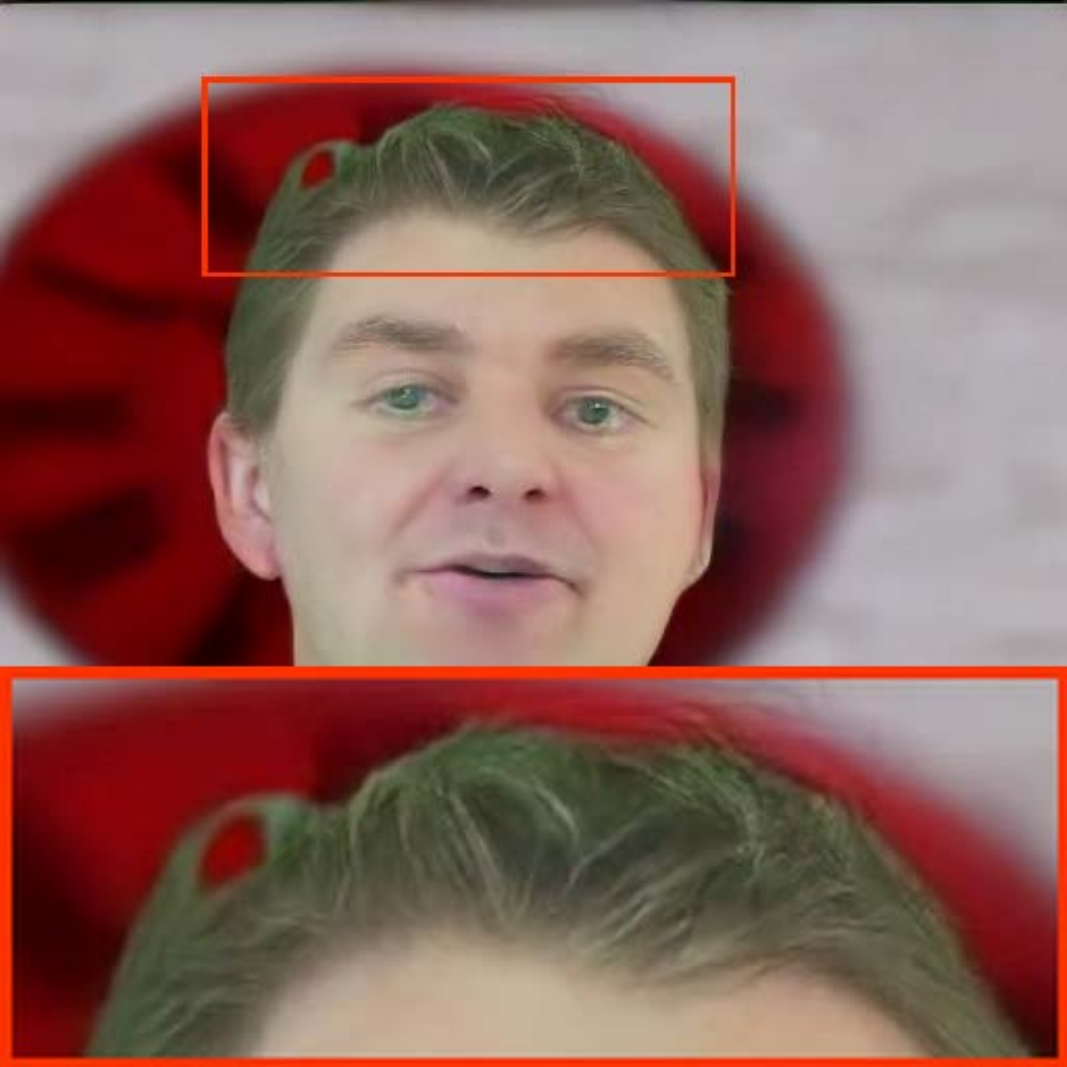} &
    \includegraphics[width=\swflicker]{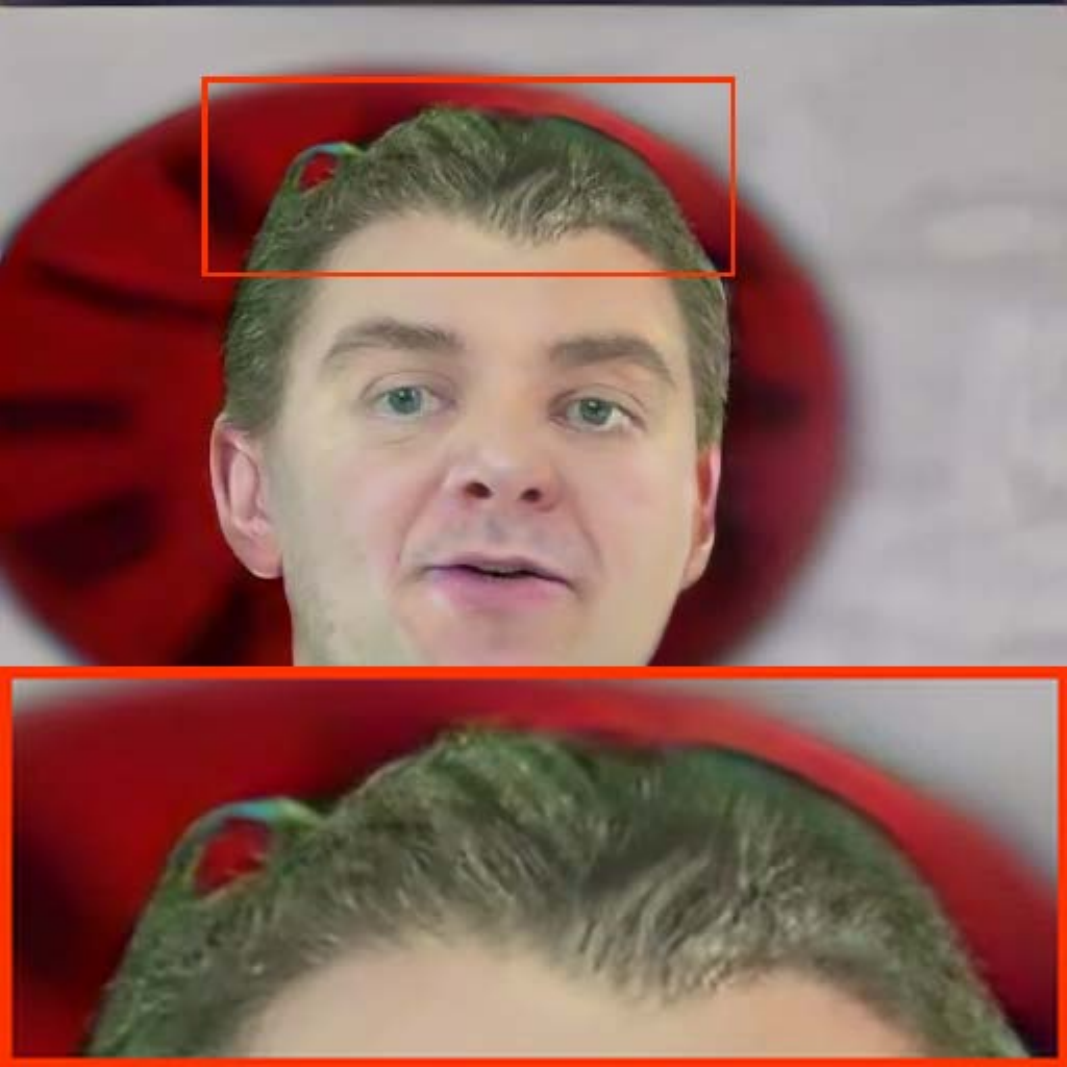} & 
    \includegraphics[width=\swflicker]{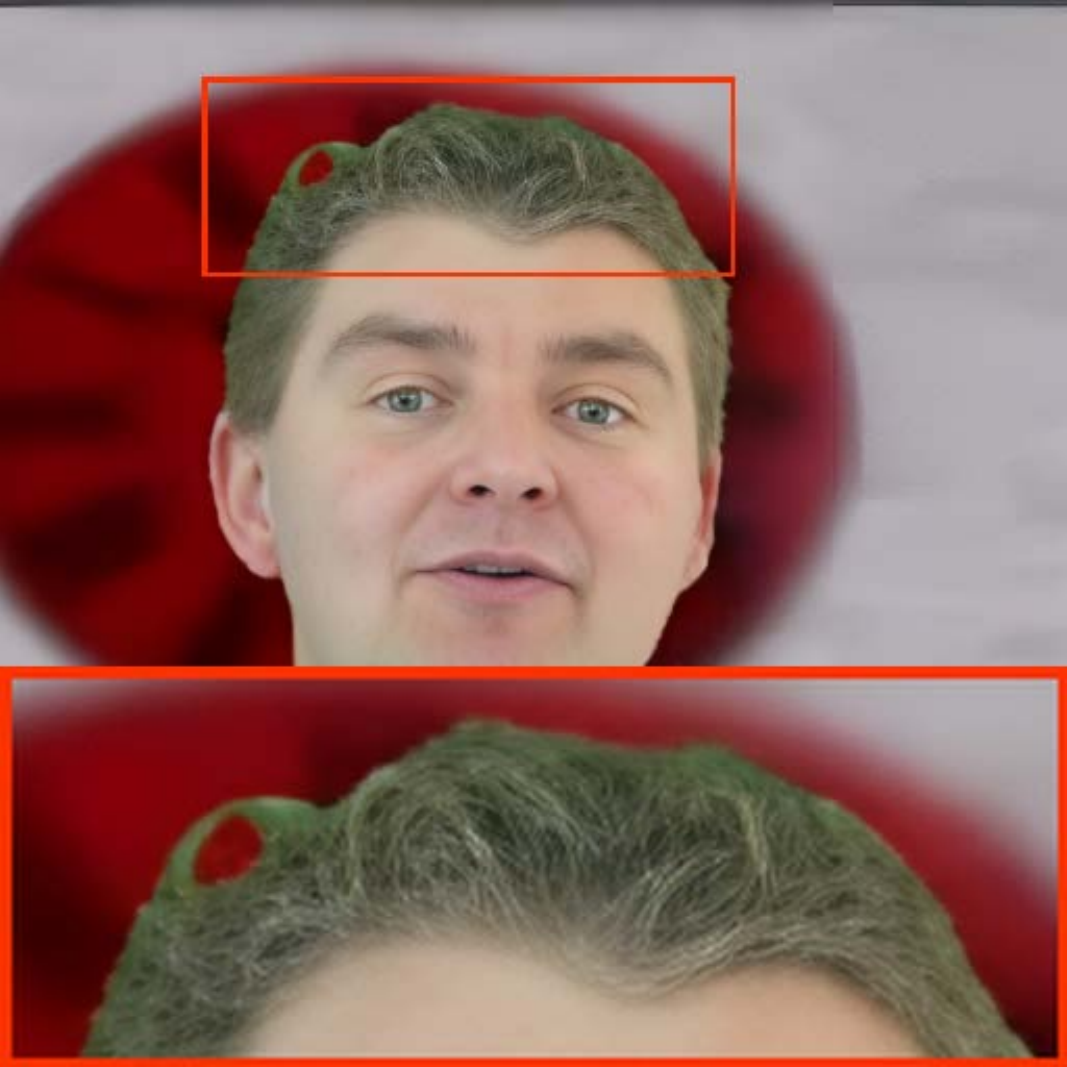} &
    \includegraphics[width=\swflicker]{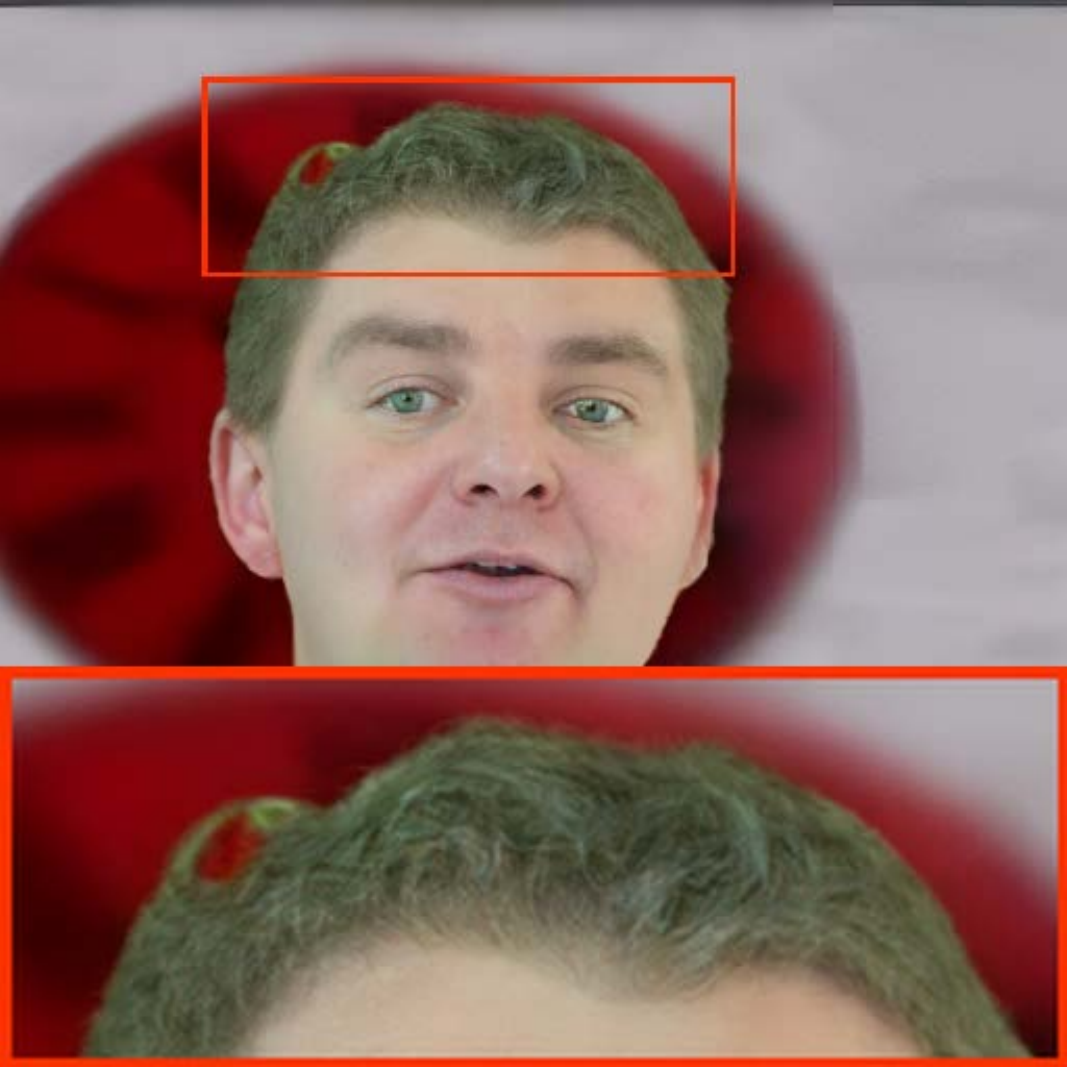} &
    \includegraphics[width=\swflicker]{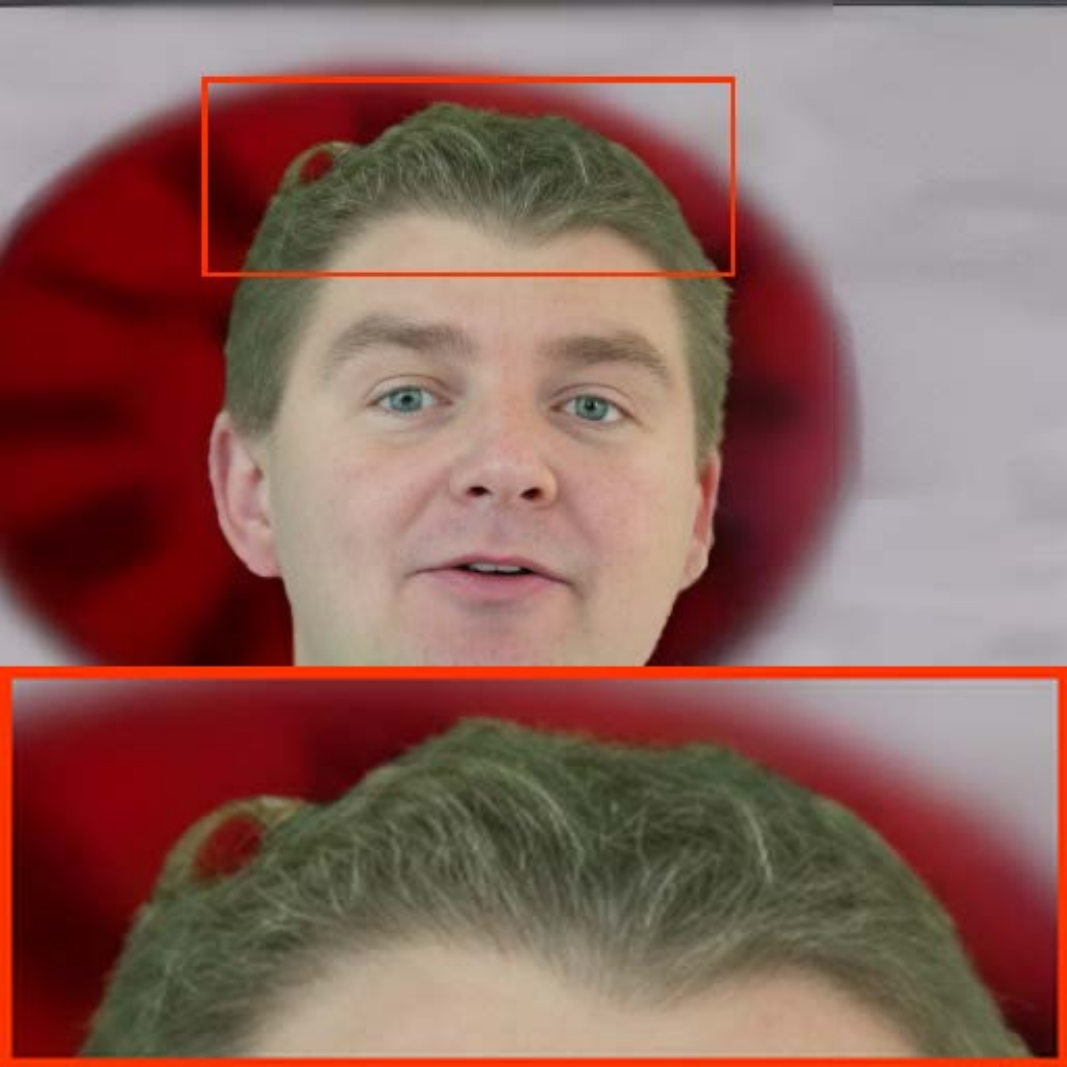}\\
    \includegraphics[width=\swflicker]{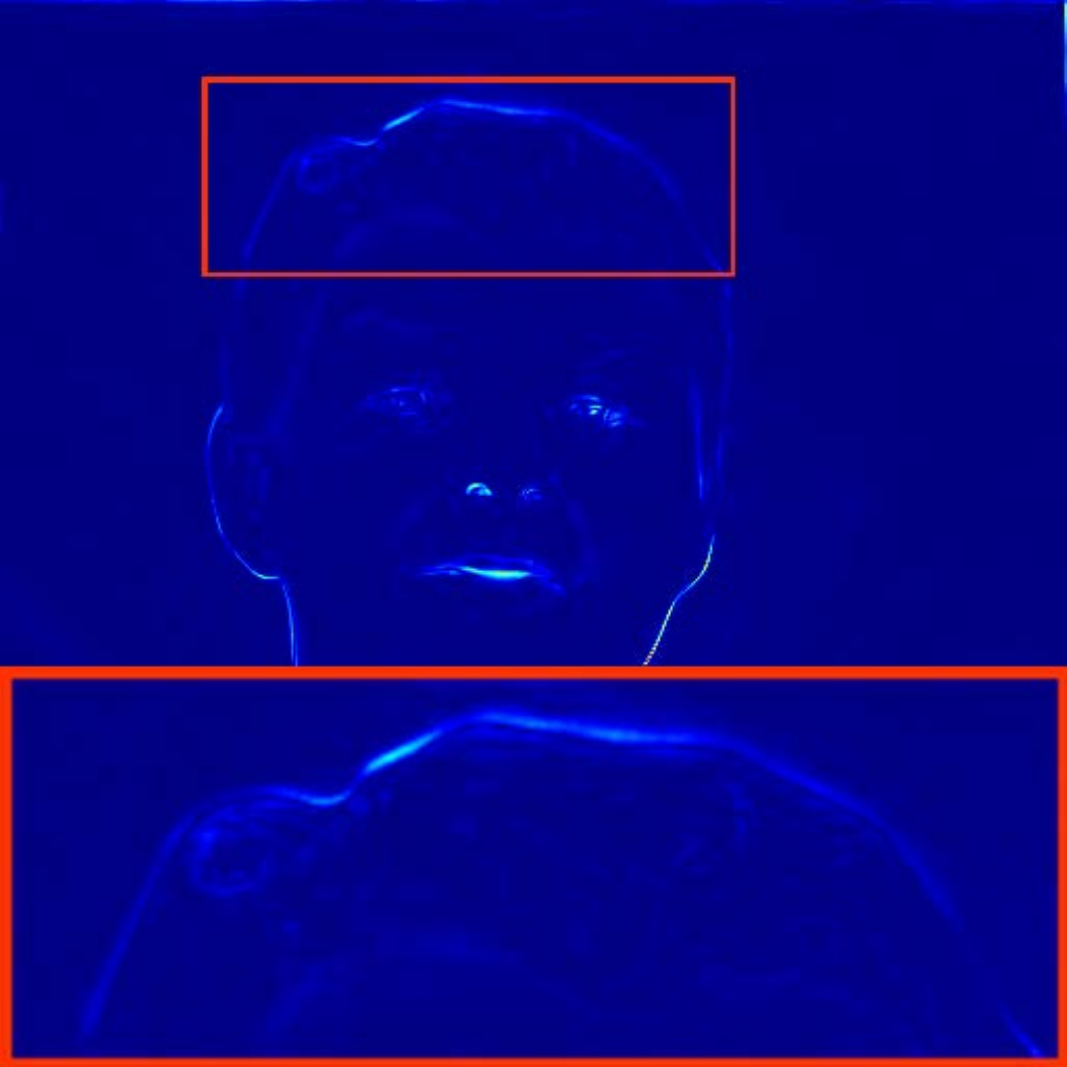} &
    \includegraphics[width=\swflicker]{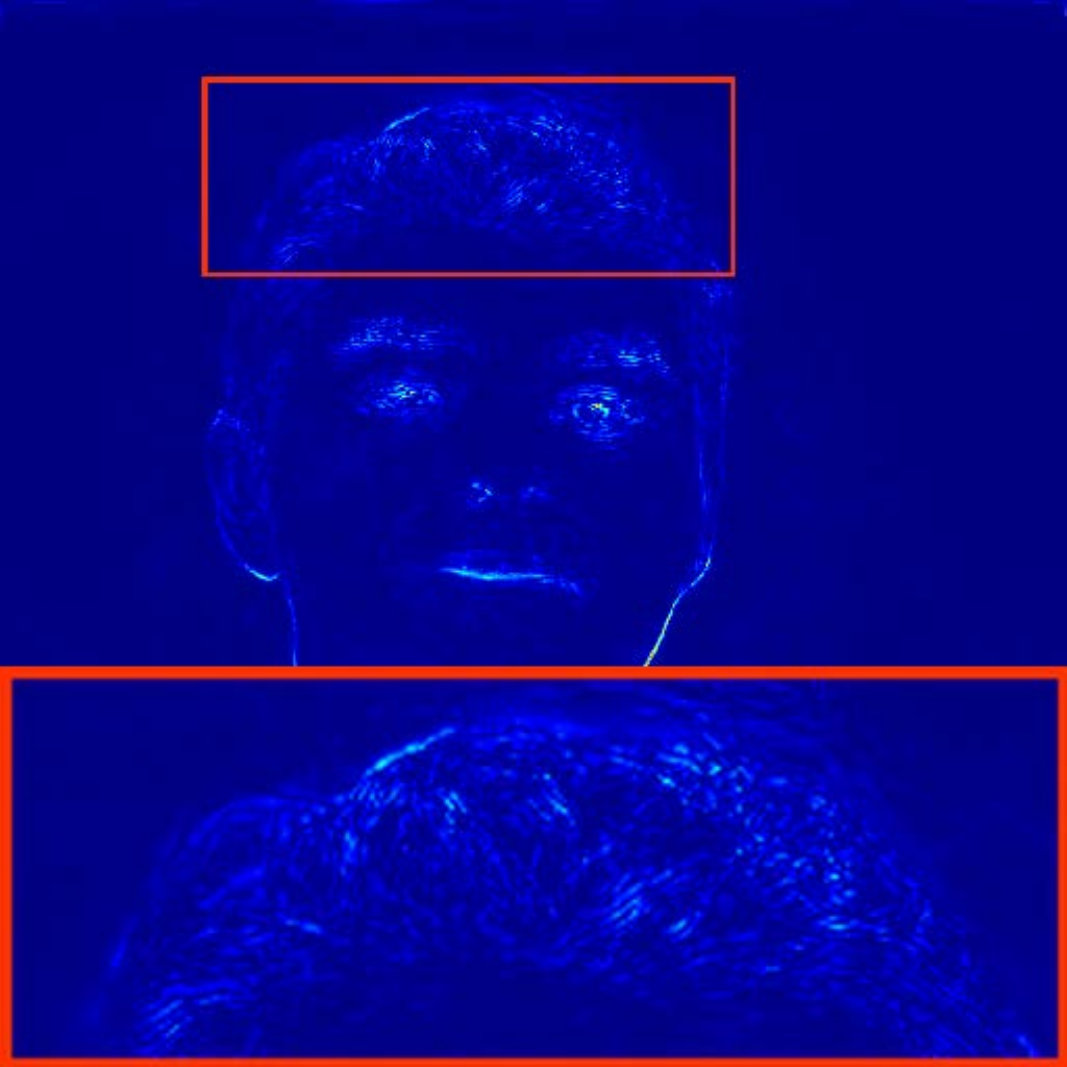} &
    \includegraphics[width=\swflicker]{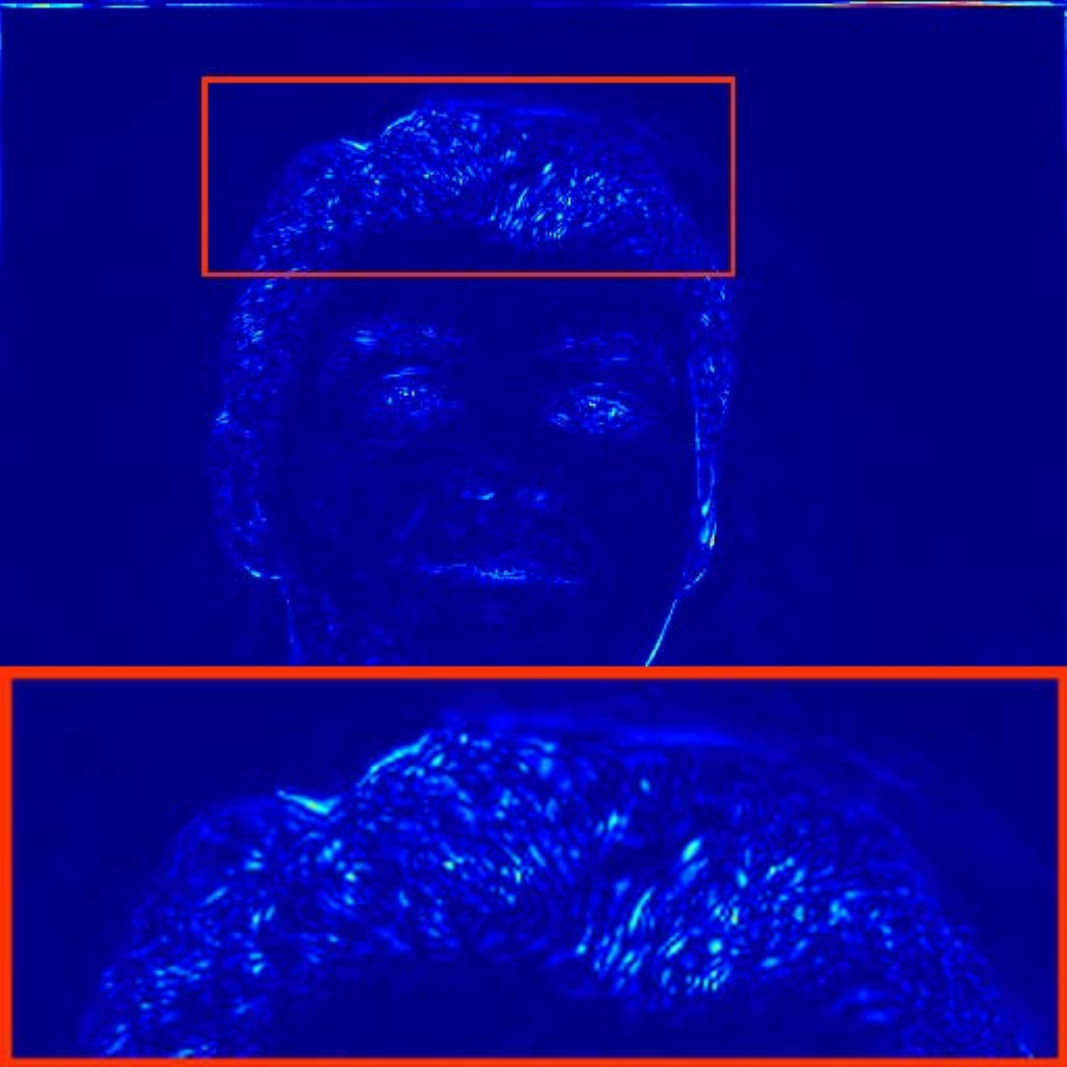} &
    \includegraphics[width=\swflicker]{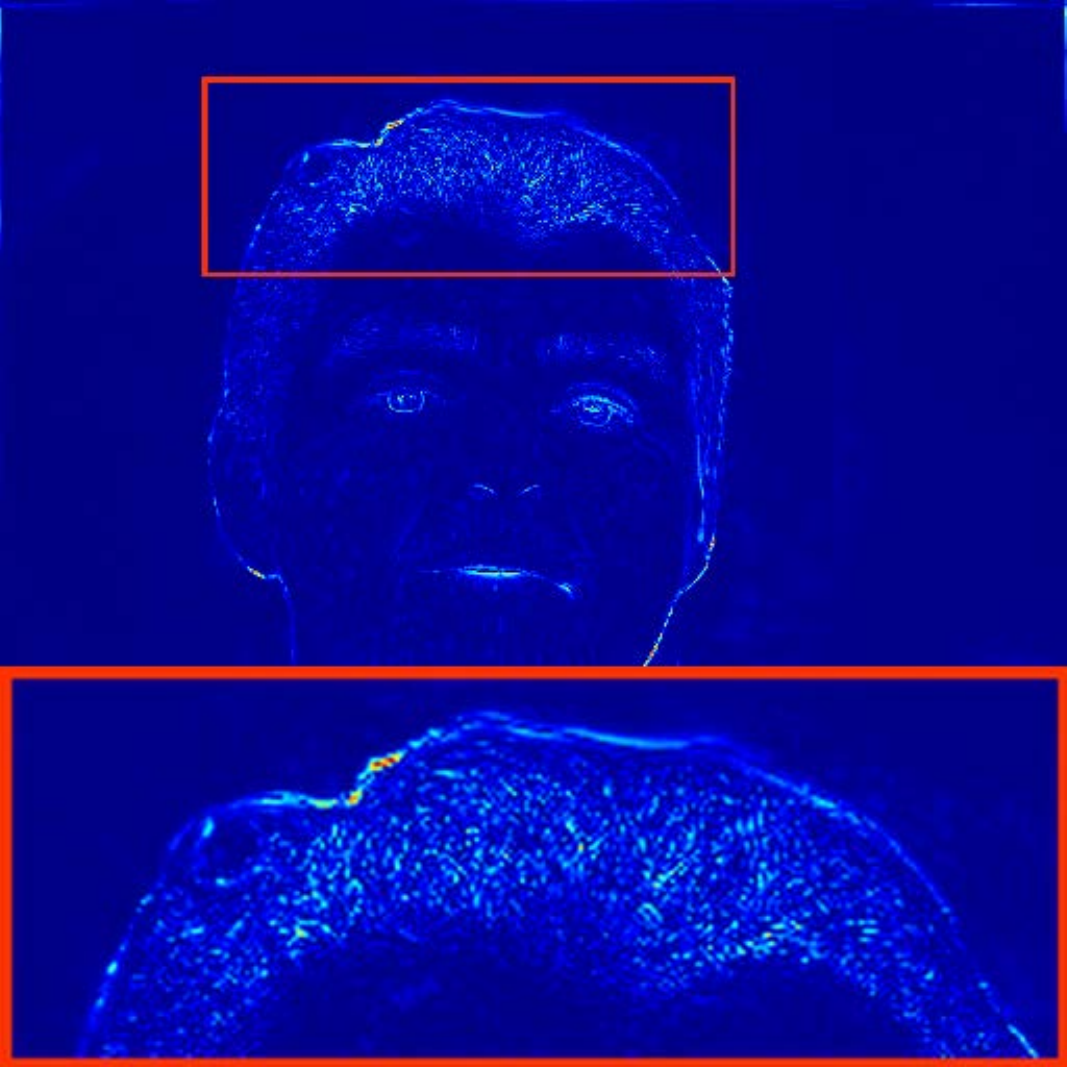} & 
    \includegraphics[width=\swflicker]{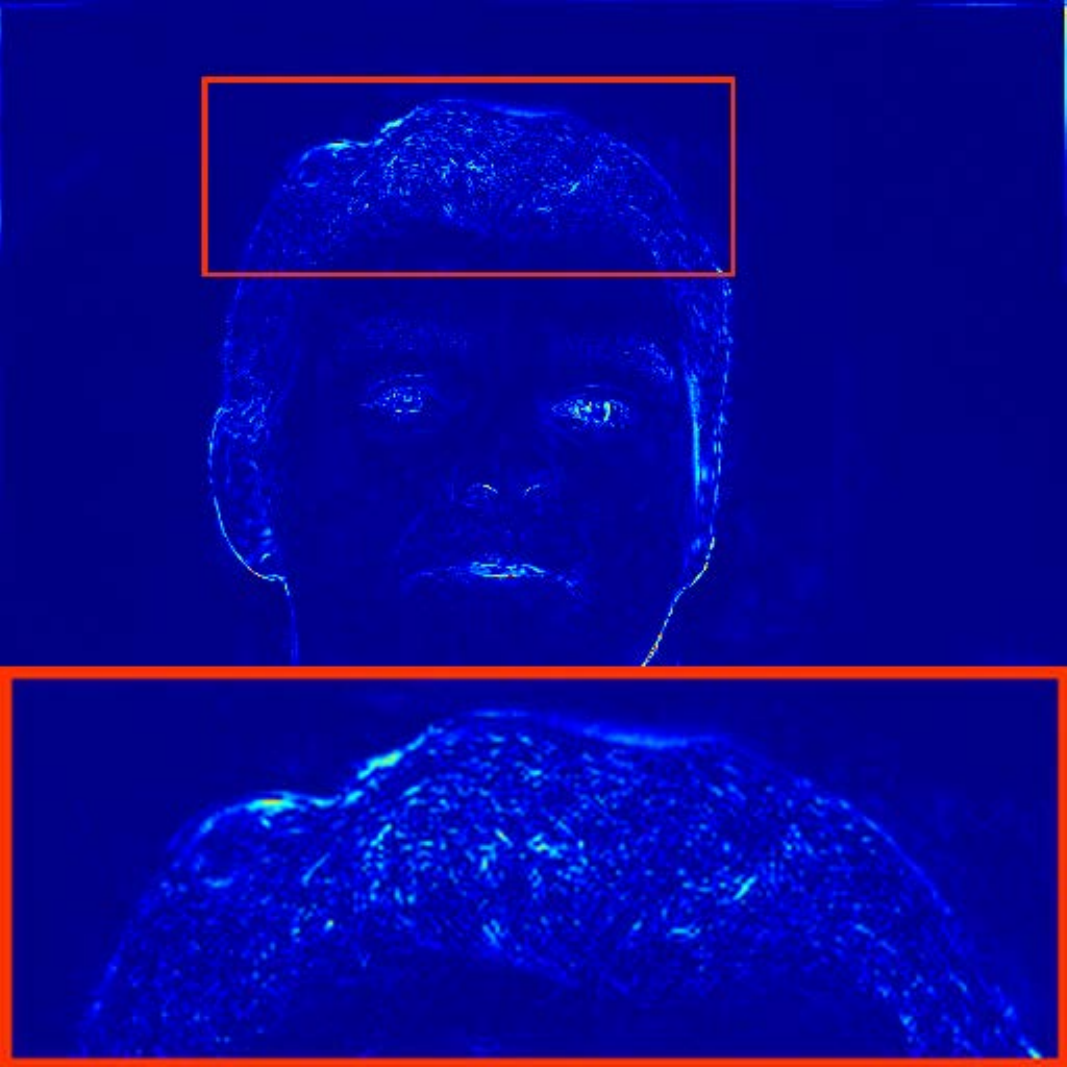} & 
    \includegraphics[width=\swflicker]{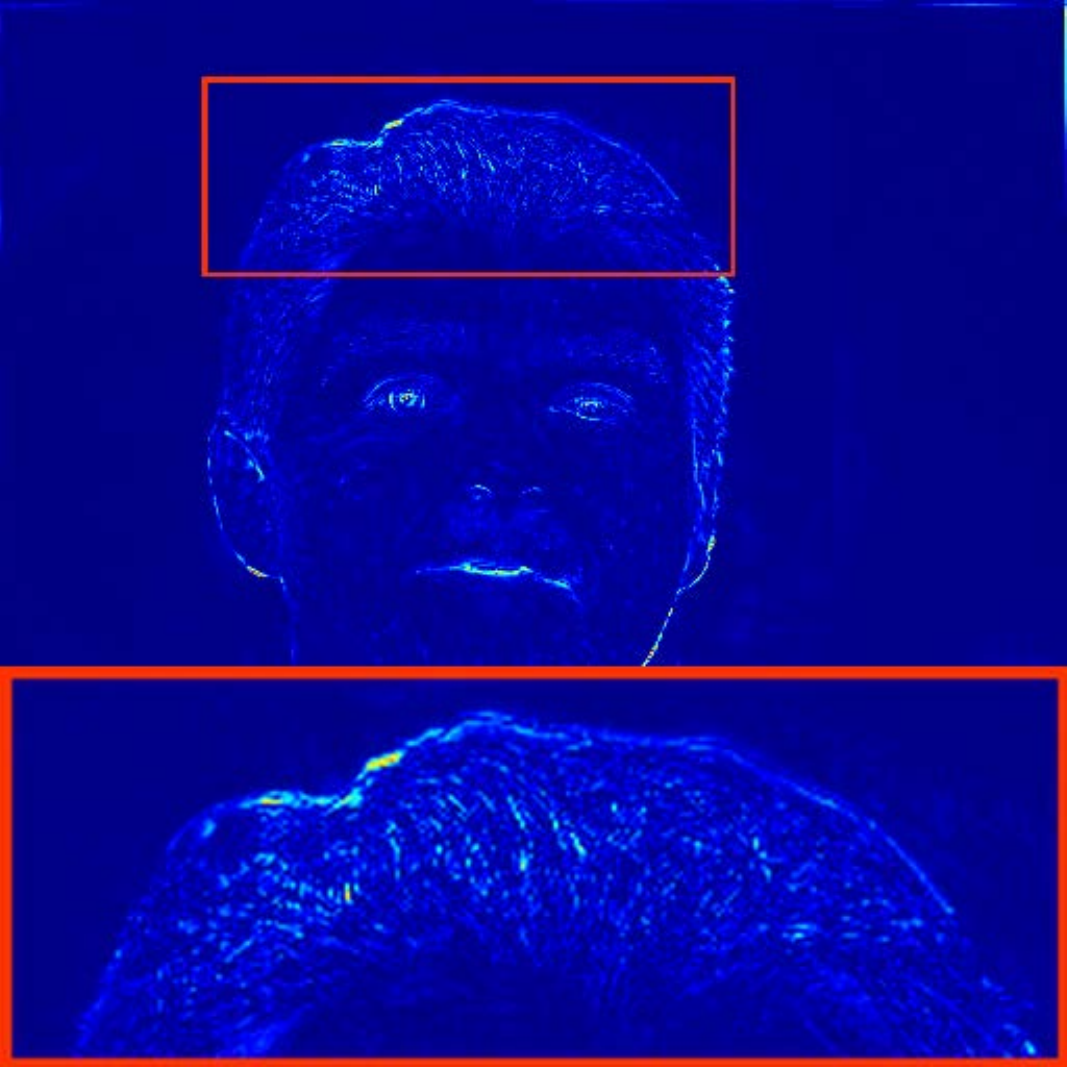}\\
    (a) EDVR~\cite{wang2019edvr} & (b) EDVR-GAN~\cite{wang2019edvr} & (c) BasicVSR-GAN~\cite{chan2021basicvsr} & (d) GFP-GAN~\cite{wang2021towards} & (e) RestoreFormer++~\cite{wang2023restoreformer++} & (f) CodeFormer~\cite{zhou2022towards}
\end{tabular}
\caption{\textbf{Visualization of noise-shape flickers.} The first row shows the restored results of existing video-based and image-based face restoration algorithms, while the second row shows their corresponding warping error map~\cite{lai2018learning} with the previous frame. Although the restored results of video-based methods with GAN loss~\cite{gan} and image-based methods achieve more details (the results of image-based methods look more natural and contain less artifacts), they exist noticeable noise-shape flickers compared to EDVR~\cite{wang2019edvr} without GAN loss. \textbf{Corresponding videos are in the supplementary materials}.
}
\label{fig:flicker}
\end{figure*}

\subsection{Systemic Analyses}
\label{subsec:analyses}

In this section, we systematically analyze the benefits and challenges of extending existing blind face image restoration algorithms to blind face video restoration. We evaluate 3 representative state-of-the-art face image restoration algorithms on RFV-LQ, including GFP-GAN~\cite{wang2021towards}, RestoreFormer++~\cite{wang2023restoreformer++}, and CodeFormer~\cite{zhou2022towards}. These works implement blind face restoration with the help of extra useful priors. Their frameworks can be generalized into a unified framework, denoted as FIR in the left of Fig.~\ref{fig:TCN}. It consists of three components: an encoder, extra-provided priors, and a decoder. The encoder first maps a degraded face image into latent features. Then, high-quality priors closely related to the latent features are obtained from an extra generative model or codebook. (The priors of GFP-GAN~\cite{wang2021towards} are from a well-trained generative model, e.g., StyleGAN2~\cite{karras2020analyzing}, while the priors of RestoreFormer++~\cite{wang2023restoreformer++} and CodeFormer~\cite{zhou2022towards} are from a codebook learned by leveraging the idea of Vector Quantization~\cite{oord2017neural}.) Finally, a decoder reconstructs a high-quality face image from the degraded latent features and priors. 
\redm{
To provide a clearer view of the framework, we omit the connection lines between the encoder (or priors) and decoder in Fig.~\ref{fig:TCN}. However, all these connections are retained while conducting evaluations and extensions, as they play an important role in balancing the fidelity and realism of the restored results.}
\textcolor{black}{To better analyze the performance of these image-based restoration methods, we also evaluate two current representative video-based restoration methods with two settings: EDVR~\cite{wang2019edvr}, EDVR-GAN~\cite{chan2021basicvsr,xie2022vfhq}, BasicVSR~\cite{chan2021basicvsr}, and BasicVSR-GAN~\cite{chan2021basicvsr,xie2022vfhq}. Unlike the image-based algorithms that process the video frame by frame individually, the video-based methods process a sequence of frames at a time, leveraging the temporal information within the video to enhance restoration. Note that EDVR~\cite{wang2019edvr} and BasicVSR~\cite{chan2021basicvsr} are originally designed for video super-resolution, and Xie \textit{et al.}~\cite{xie2022vfhq} retrained them from face video restoration. We employ the trained model provided by Xie \textit{et al.}~\cite{xie2022vfhq} for our evaluation.}

\noindent\textbf{Benefits.} 
As shown in Fig.~\ref{fig:fvr_fir}, the advantages of the image-based face restoration methods lie in their high-quality restoration. Compared to video-based methods, image-based methods produce more realistic and detailed faces, such as teeth with richer details, glasses with a more complete structure, and eyes with a more natural look. To evaluate the restoration quality more comprehensively, we measure it with FID~\cite{fid}, which is represented as the distance of the distribution between the restored face images and real-world high-quality face images. However, we find that BasicVSR-GAN~\cite{chan2021basicvsr} has the lowest FID score (the lower, the better), which contradicts our visual observation. This is mainly because FID is calculated based on the whole face image, including the diverse background, which may result in an inaccurate assessment. On the contrary, human perception mainly concentrates on some key regions of a face image, such as eyes, nose, mouth, hair, etc. Therefore, we propose \textbf{Component FIDs}, which consist of FID-Eyes, FID-Nose, FID-Mouth, and FID-Hair. They calculate the distribution distance between the restored face images and real-world high-quality face images based on the areas of specific facial components.
\redm{The results are in Table~\ref{tab:analyze} and Fig.~\ref{fig:component_fids}. They show that although the difference between image-based methods and video-based methods implemented with GAN loss in terms of FID is extremely small, the advantages of image-based methods become significant when measured with Component FIDs, especially in terms of FID-Mouth. The performance of component-based FID is more consistent with the observation in the qualitative results (such as the high-quality restored teeth attained with the image-based methods in Fig.~\ref{fig:fvr_fir}), which demonstrates that Component FIDs can measure the quality of the restored face results more accurately.}

\begin{figure}[t]
\setlength\tabcolsep{1pt}
\scriptsize
\centering
\begin{tabular}{cccc}
    \includegraphics[width=\swffhqvfhq]{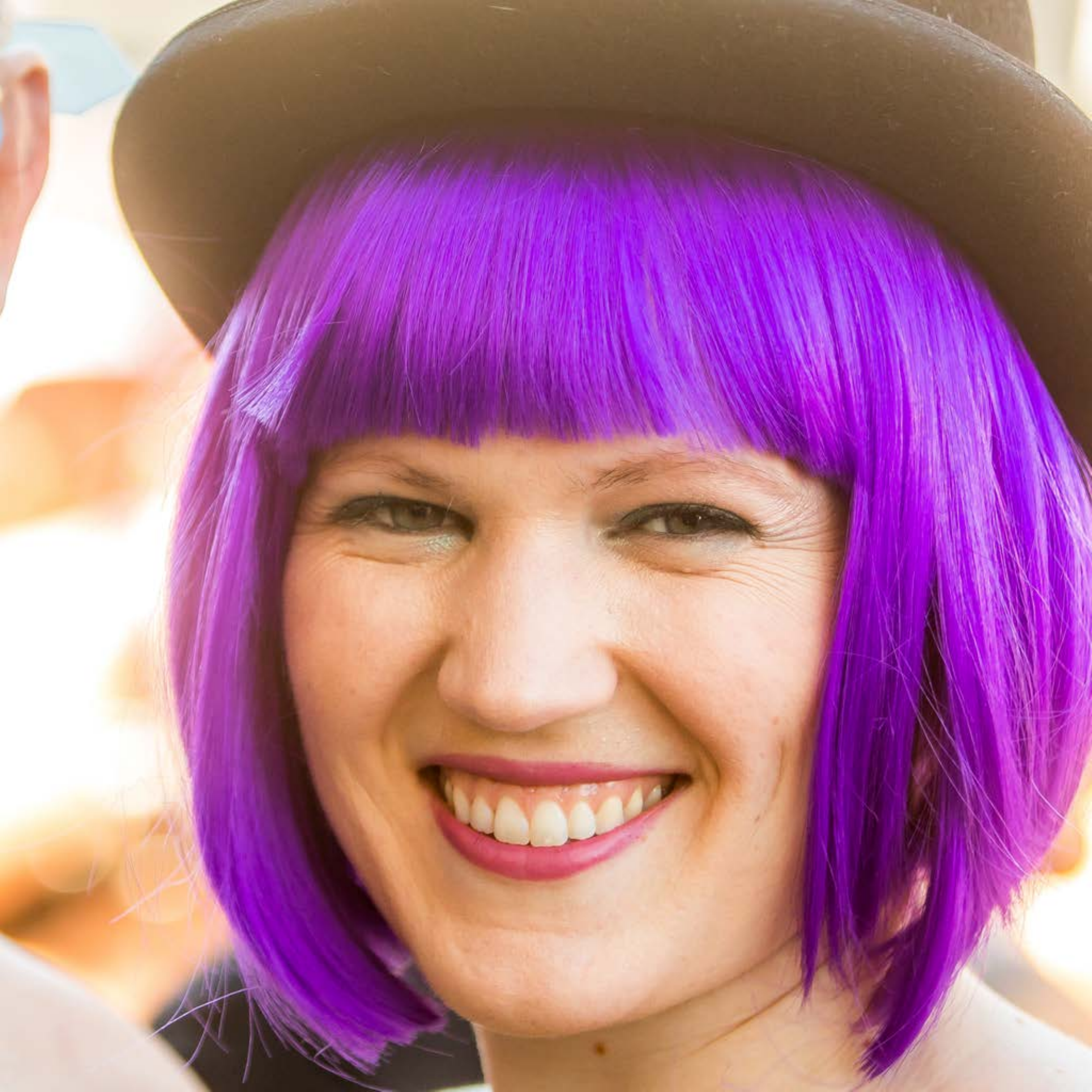} & 
    \includegraphics[width=\swffhqvfhq]{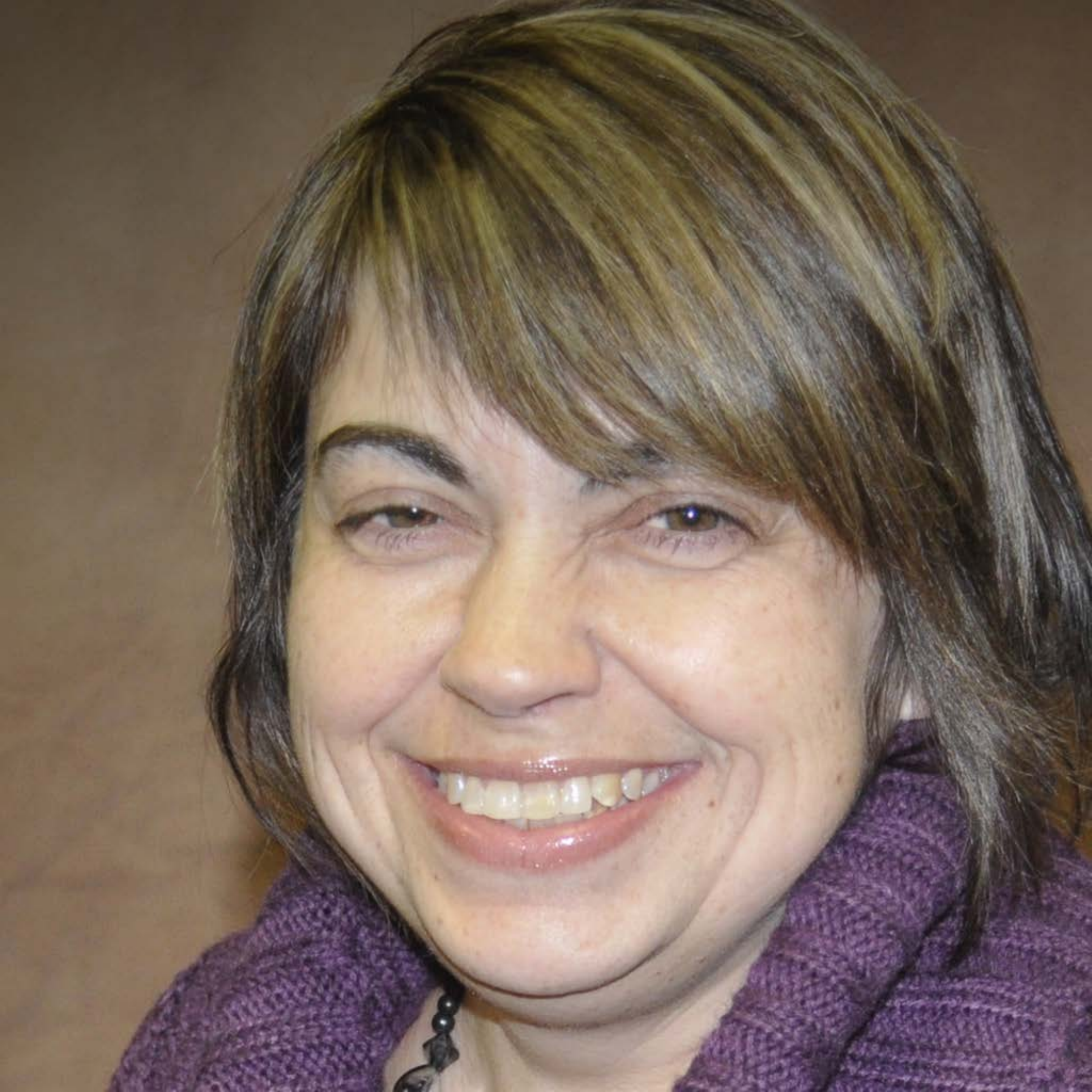} &
    \includegraphics[width=\swffhqvfhq]{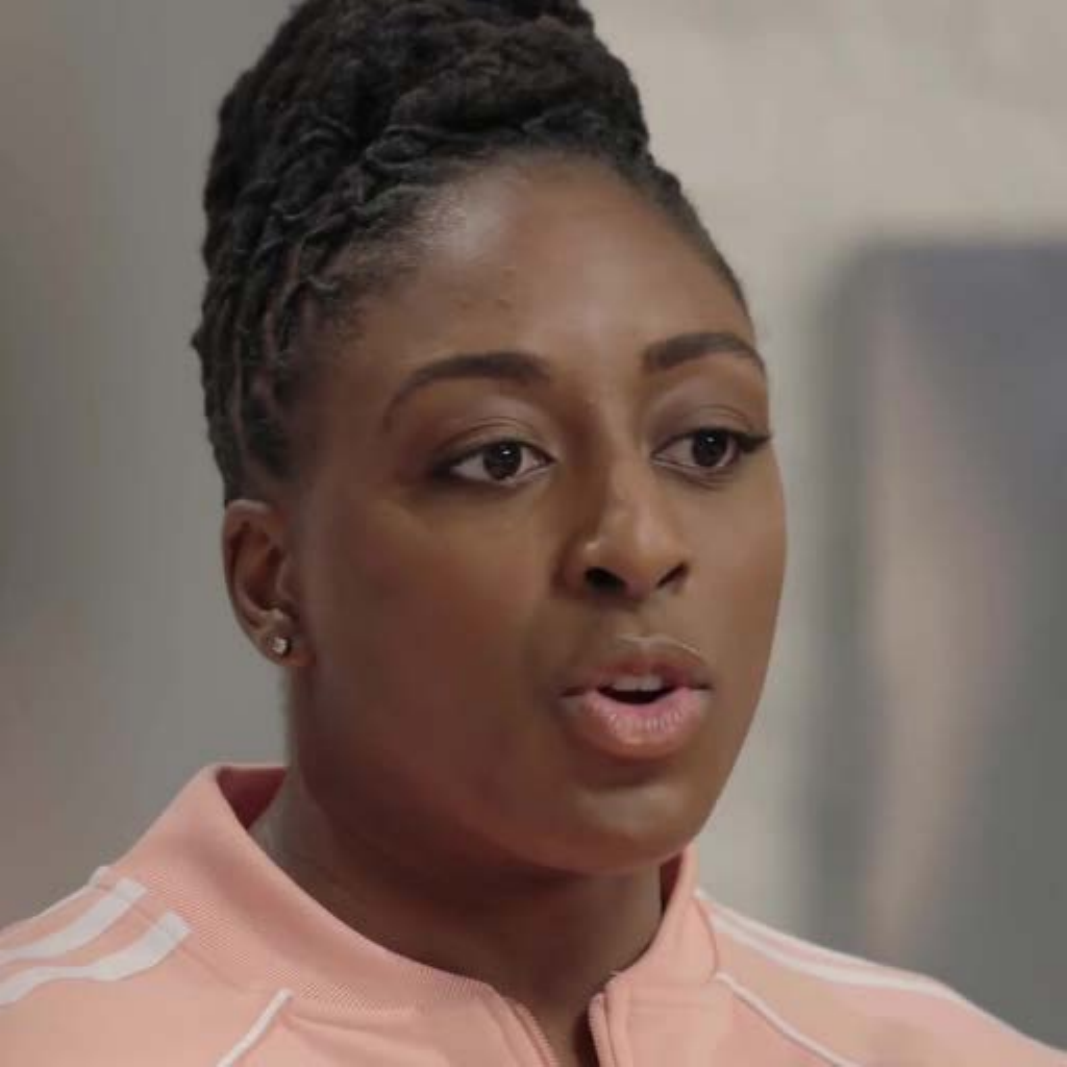} &
    \includegraphics[width=\swffhqvfhq]{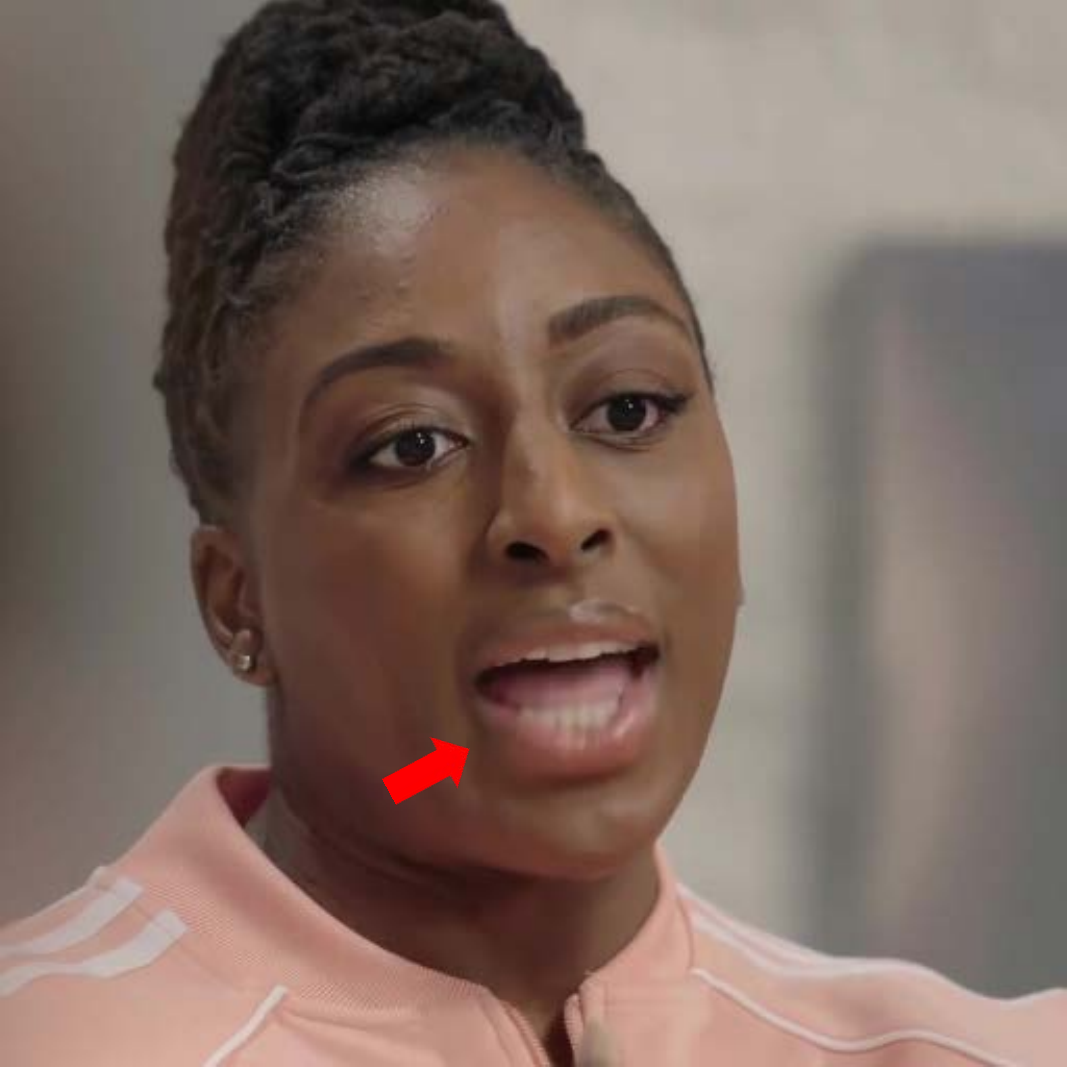} \\
    \multicolumn{2}{c}{(a) FFHQ~\cite{ffhq}} & \multicolumn{2}{c}{(b) VFHQ~\cite{xie2022vfhq}}
\end{tabular}
\caption{\textbf{High-Quality Face Images in FFHQ~\cite{ffhq} and VFHQ~\cite{xie2022vfhq}.} Both FFHQ~\cite{ffhq} and VFHQ~\cite{xie2022vfhq} provide high-quality face images beneficial for research in face image/video restoration. However, due to the inherent motion blur in videos, face images extracted from VFHQ~\cite{xie2022vfhq} display noticeable blur (indicated by the \textcolor{red}{red arrow} in the final face image). This can adversely affect blind face restoration when using VFHQ~\cite{xie2022vfhq} as the training dataset.}
\label{fig:ffhq_and_vfhq}
\end{figure}

\textcolor{black}{
We investigate the factors contributing to the benefit of image-based algorithms for processing degraded face videos and attribute them to \textit{(1) the superior training data} and \textit{(2) mature techniques of image-based face restoration algorithms}. While most image-based methods are trained on high-quality face images from FFHQ~\cite{ffhq}, video-based techniques utilize VFHQ~\cite{xie2022vfhq}, which comprises high-quality face videos with inherent motion blur. Examples from FFHQ~\cite{ffhq} and VFHQ~\cite{xie2022vfhq} are displayed in Fig.~\ref{fig:ffhq_and_vfhq}. Unlike individual face images from FFHQ~\cite{ffhq} provide high-quality facial details, face videos from VFHQ~\cite{xie2022vfhq} naturally exhibit motion blur, as indicated by the \textcolor{red}{red arrow} in the final face image. During training, video-based methods use VFHQ~\cite{xie2022vfhq} videos as Ground Truth, with their synthesized degraded videos serving as input. The inclusion of motion blur in the Ground Truth diminishes the restoration quality of video-based techniques. In addition, image-based methods own mature and advanced techniques~\cite{wang2022survey}, especially those based on generative priors~\cite{yang2021gan,wang2021towards} and codebook~\cite{gu2022vqfr,wang2022restoreformer,wang2023restoreformer++,zhou2022towards}, greatly enhancing the performance of blind face restoration. However, there are few video-based methods specifically designed for blind face video restoration. EDVR~\cite{wang2019edvr} and BasicVSR~\cite{chan2021basicvsr} are originally designed for video super-resolution, and Xie \textit{et al.}~\cite{xie2022vfhq} retrained them from face video restoration. We adopt the trained model provided by Xie \textit{et al.}~\cite{xie2022vfhq} for evaluation.
}

\noindent\textbf{Challenges.} 
Despite the benefit discussed above, image-based algorithms face two challenges in their restored results: \textit{(1) obvious jitters in facial components} and \textit{(2) noticeable noise-shape flickers between frames}. To intuitively illustrate these challenges, we propose employing appropriate qualitative visualizations and quantitative metrics to measure the jitters and flickers.

First, we visualize the jitter by extracting the pixels along a vertical slice near the facial components (such as eyes, nose, and mouth) in every frame and concatenate them over time. Fig.~\ref{fig:jitter} shows some examples, where (b)$ \sim $(g) are the concatenated results of the \textcolor{red}{red line} in (a) for different restoration methods. The smoother the concatenated results, the less the jitters. Compared to the relatively smooth slices in the results of EDVR~\cite{wang2019edvr}, the restored results of the video-based methods with GAN~\cite{gan} and image-based methods exist noticeable jitters, especially the image-based methods, RestoreFormer++~\cite{wang2023restoreformer++} and CodeFormer~\cite{zhou2022towards}. 

To quantify the jitters, we adopt \textbf{MSI} (Motion Stability Index)~\cite{ling2022stableface},
which measures the acceleration variance of facial key points across sequential frames. 
Specifically, we first adopt Awing~\cite{wang2019adaptive} to estimate 98 facial landmarks of each frame in a video, but only 65 of them that represent the facial components (such as eyebrows, eyes, noses and mouth) are used.
We denote these landmarks as $\mathbf{Z}_t \in \mathbb{R}^{N\times2}$, where $t \in [0, T-1]$, and $T$ is the number of frames in a video. $N=65$ is the number of landmarks in each frame. Each landmark is represented as a two-dimensional vector which indicates the coordinates in the height and width directions of an image, respectively.
We regard the interval between two adjacent frames as a unit, then the velocity and acceleration of landmark $i$ at timestep $t$ can be expressed as:
\begin{equation}
    \begin{aligned}
        \mathbf{v}_t^i &= \mathbf{Z}_{t+1}^{i} - \mathbf{Z}_{t}^{i}, t \in [0, T-2]; \\
        \mathbf{a}_t^i &= \mathbf{v}_{t+1}^{i} - \mathbf{v}_{t}^{i}, t \in [0, T-3]. \\
    \end{aligned}
\end{equation}
After that, we attain the variance of $\mathbf{a}^i$ to express the sharpness of jitters in landmark $i$. That is
\begin{equation}
    \sigma(\mathbf{a}^i) = \frac{1}{T-2}\sum_{t\in[0,T-2)}|\mathbf{a}_t^i-\mathbf{\bar{a}}^i|,
\end{equation}
where $\mathbf{\bar{a}}^i$ is the mean of $\{\mathbf{a}_0^i,\mathbf{a}_1^i,\dots,\mathbf{a}_{T-3}^i\}$, and $|\cdot|$ means the $L_1$ norm of vector ($\mathbf{a}_t^i-\bar{\mathbf{a}}^i$).
Finally, the \textbf{MSI} score of a video is described as the average of the reciprocal of the variance of each landmark:
\begin{equation}
    \mathbf{MSI} = \frac{1}{N}\sum_{i=0}^{N}\frac{1}{\sigma(\mathbf{a}^i)+\epsilon_0},
\end{equation}
where $\epsilon_0=10^{-5}$ is for avoiding dividing by zero and $N$ is the number of frames in the video. 
Therefore, the smaller \textbf{MSI} score is, the more stable the motion in the restored video is.
The results in Table~\ref{tab:analyze} show that video-based methods with GAN~\cite{gan} and image-based methods achieve larger scores which indicates they contain more jitters in facial components.

As for the noise-shape flickers, we visualize and measure them with \textbf{Warping Error}~\cite{lai2018learning}. 
Specifically, we estimate the optical flow between two consecutive frames $\mathbf{I}_t^r$ and $\mathbf{I}_{t+1}^r$ using SpyNet~\cite{ranjan2017optical} and warp $\mathbf{I}_{t+1}^r$ to $\mathbf{\hat{I}}_{t+1}^r$ based on the predicted flow. 
Then we take the pixel-wise square differences between $\mathbf{I}_t^r$ and $\mathbf{\hat{I}}_{t+1}^r$ as the warping error map $\mathbf{M}_{warp}$, which is formulated as:
\begin{equation}
    \mathbf{M}_{warp}(\mathbf{I}_t, \mathbf{I}_{t+1}) = ||\mathbf{I}_t - \mathbf{\hat{I}}_{t+1}||_2,
\end{equation}
where $||\cdot||_2$ is the L2-norm.
By further averaging the warping error of all the pixels in a video, we get the warping error score $\mathbf{E}_{warp}$:
\begin{equation}
    \mathbf{E}_{warp} = \frac{1}{H\times W\times T}\sum_{j=0}^{H\times W}\sum_{t=0}^{T-2}\mathbf{M}_{warp}^j(\mathbf{I}_t, \mathbf{I}_{t+1}),
\end{equation}
where $j$ means the $j_{th}$ pixel in frame $\mathbf{I}_t$ or $\mathbf{\hat{I}}_{t+1}$. $H$ and $W$ are respectively the height and width of a frame.

Results in terms of Warping Error are in Fig.~\ref{fig:flicker} and Table~\ref{tab:analyze}, where the noise-shape flickers are highlighted in the warping error map and lead to a high Warping Error score. These results also indicate that video-based methods with GAN~\cite{gan} and image-based methods contain obvious highlighted areas and achieve a high Warping Error score compared to the video-based methods without GAN~\cite{gan} (EDVR~\cite{wang2019edvr} and BasicVSR~\cite{chan2021basicvsr}).

\begin{figure*}[t]
\setlength\tabcolsep{1pt}
\scriptsize
\centering
\begin{tabular}{ccccccc}
    \includegraphics[width=\swlankmark]{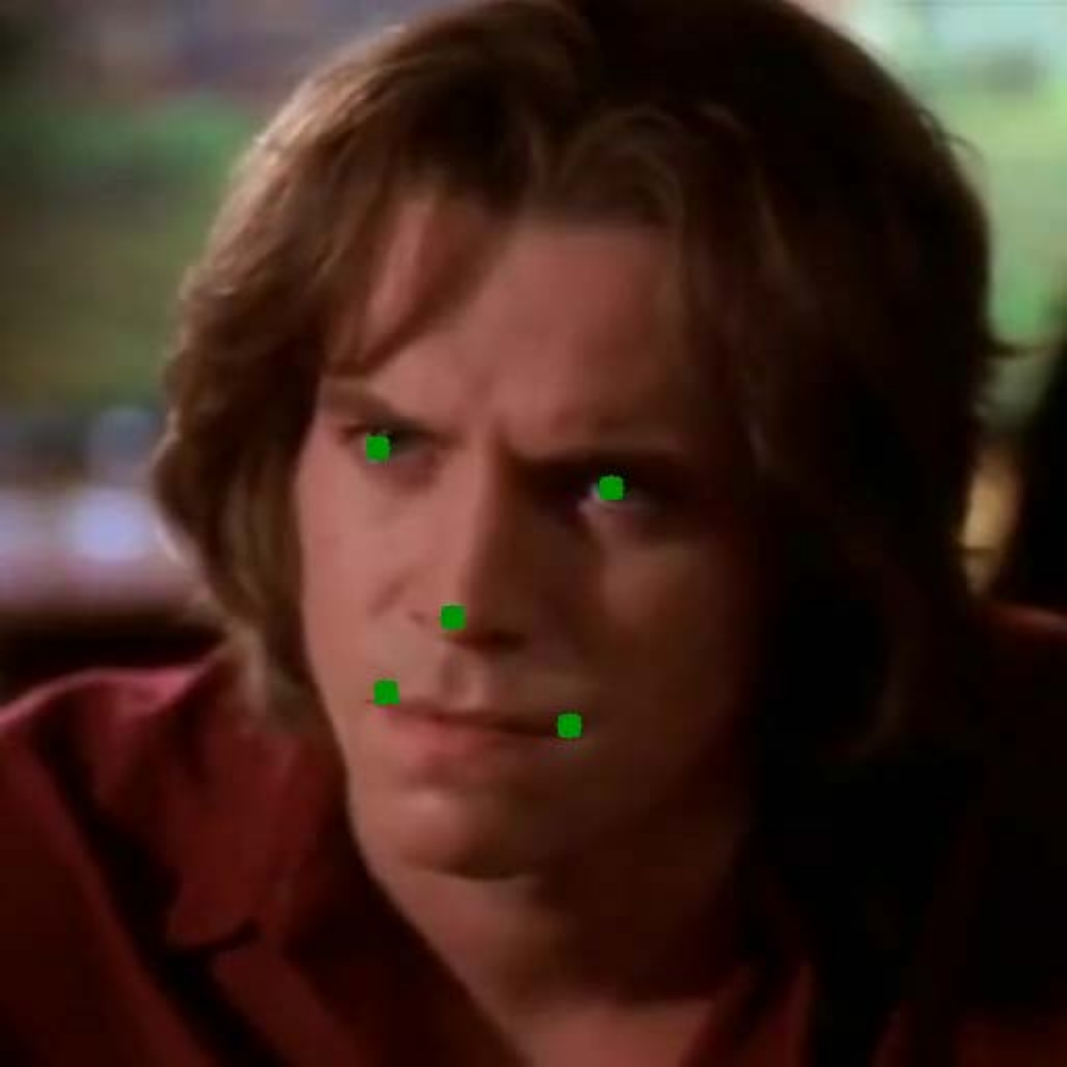} &
    \includegraphics[width=\swlankmark]{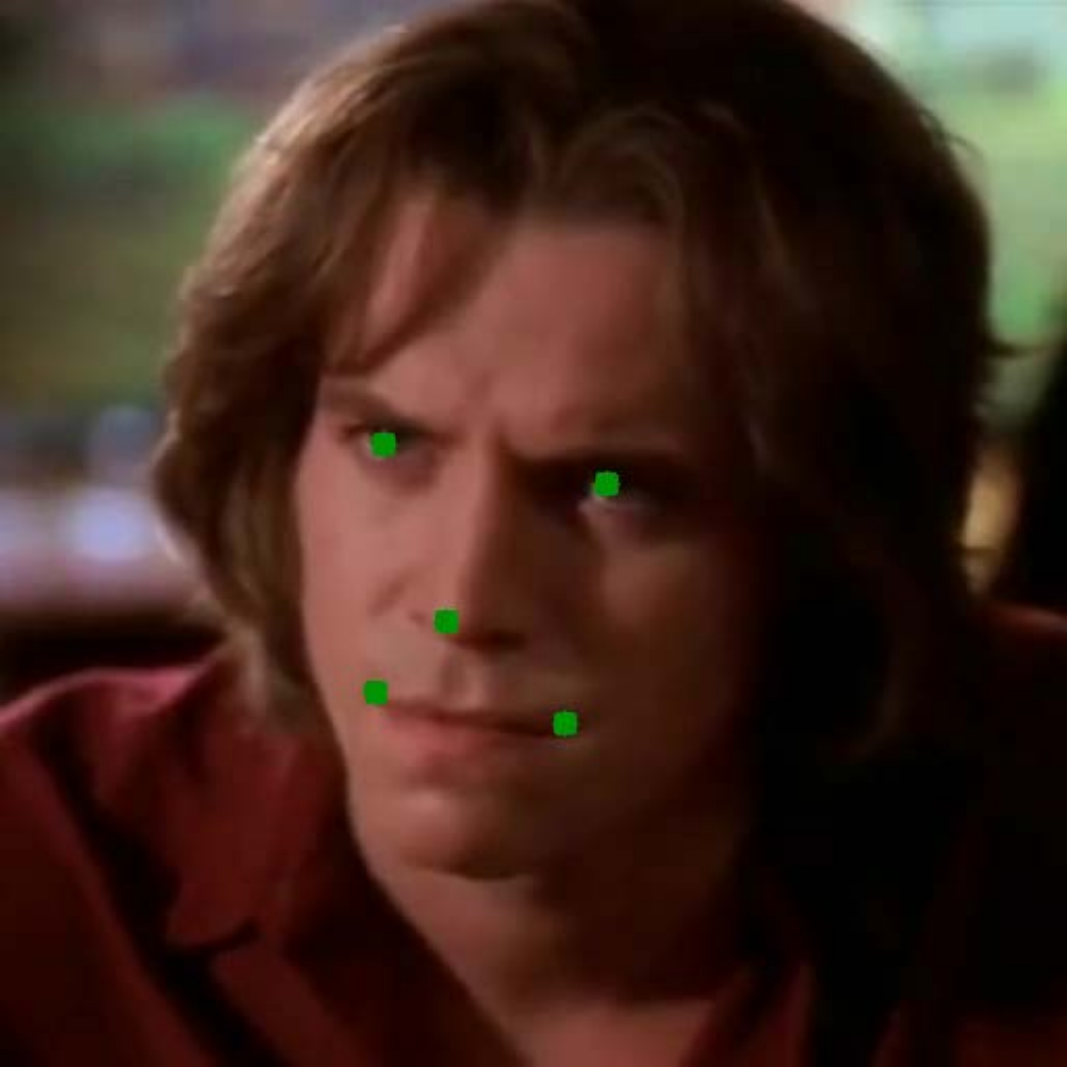} &
    \includegraphics[width=\swlankmark]{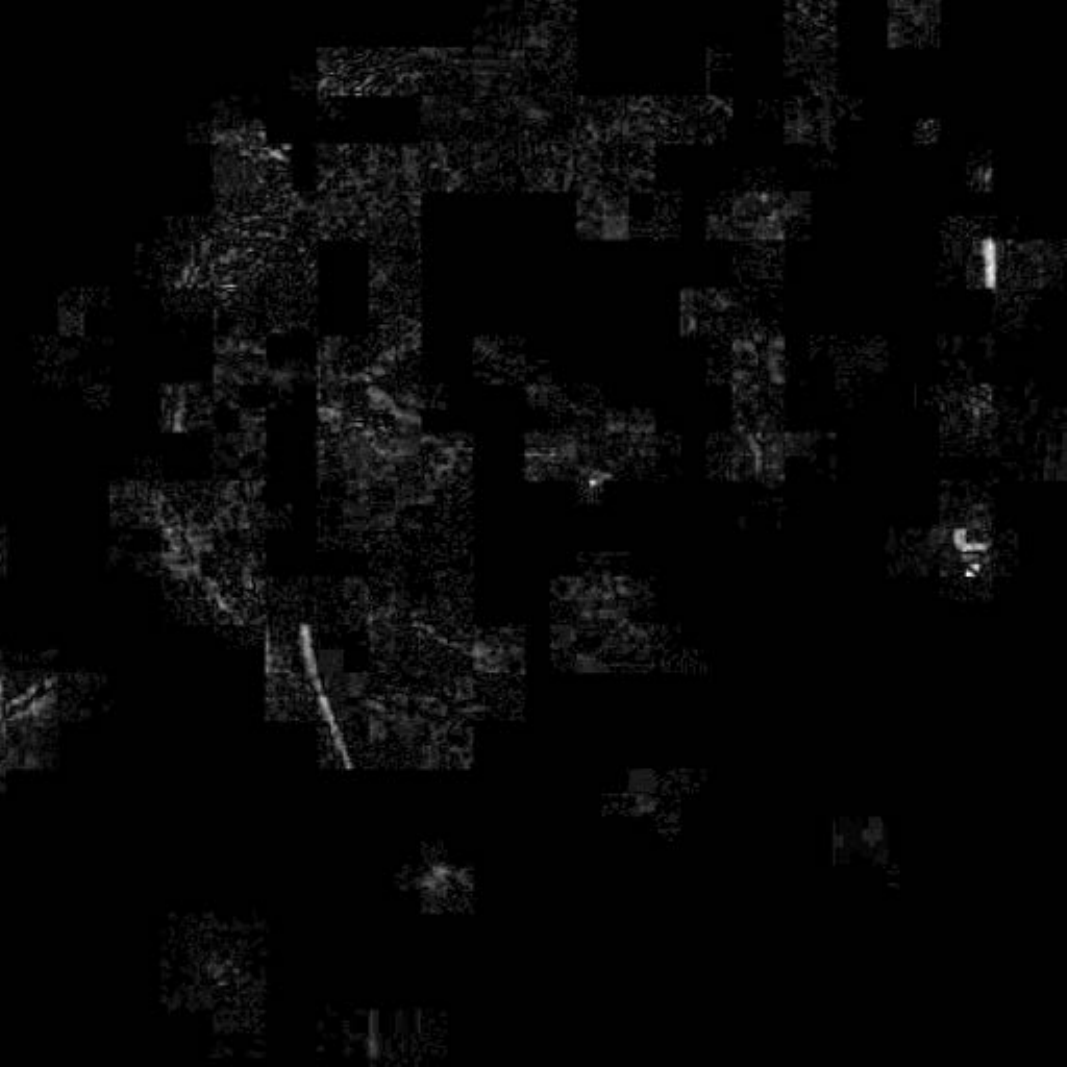} &
    \includegraphics[width=\swlankmark]{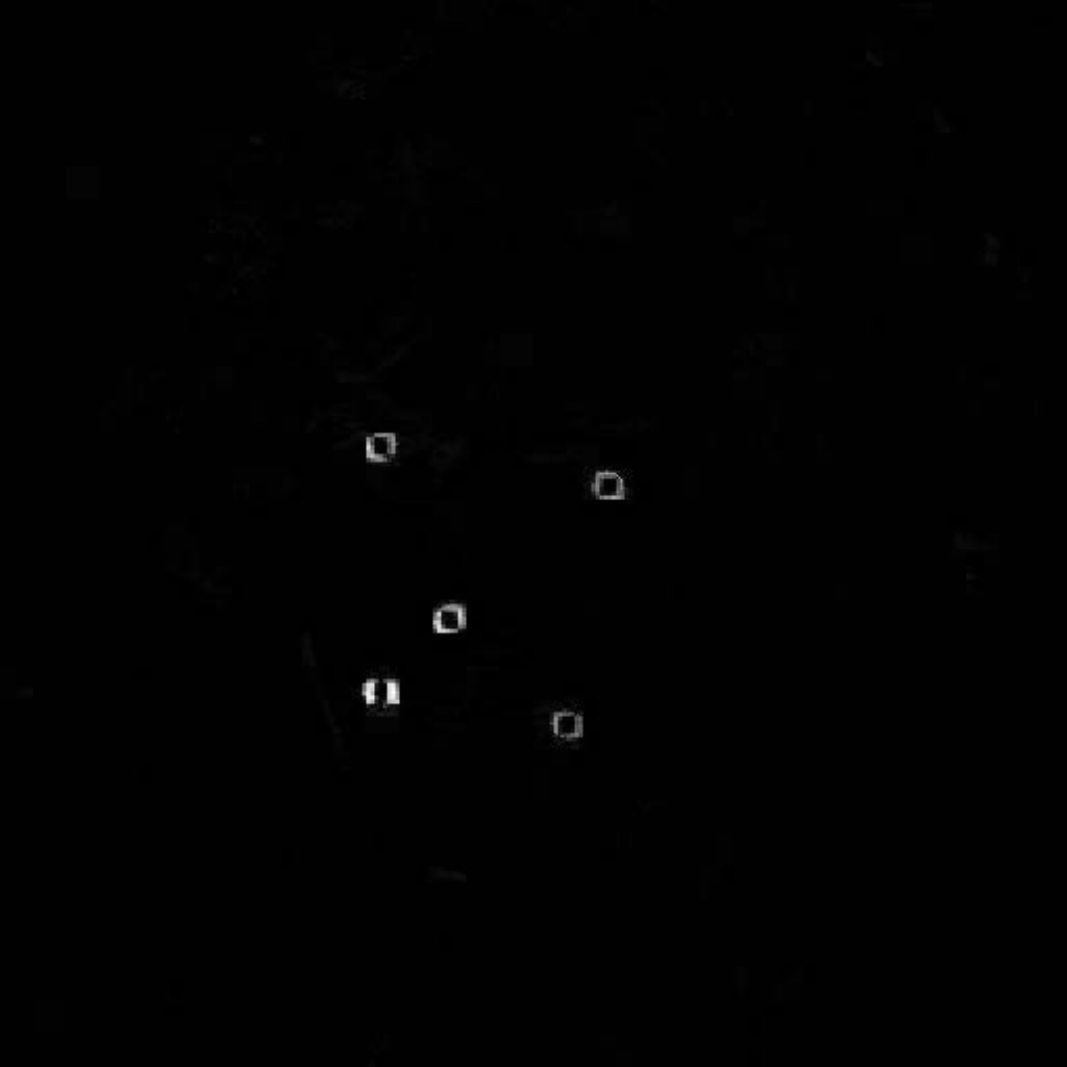} &
    \includegraphics[width=\swlankmark]{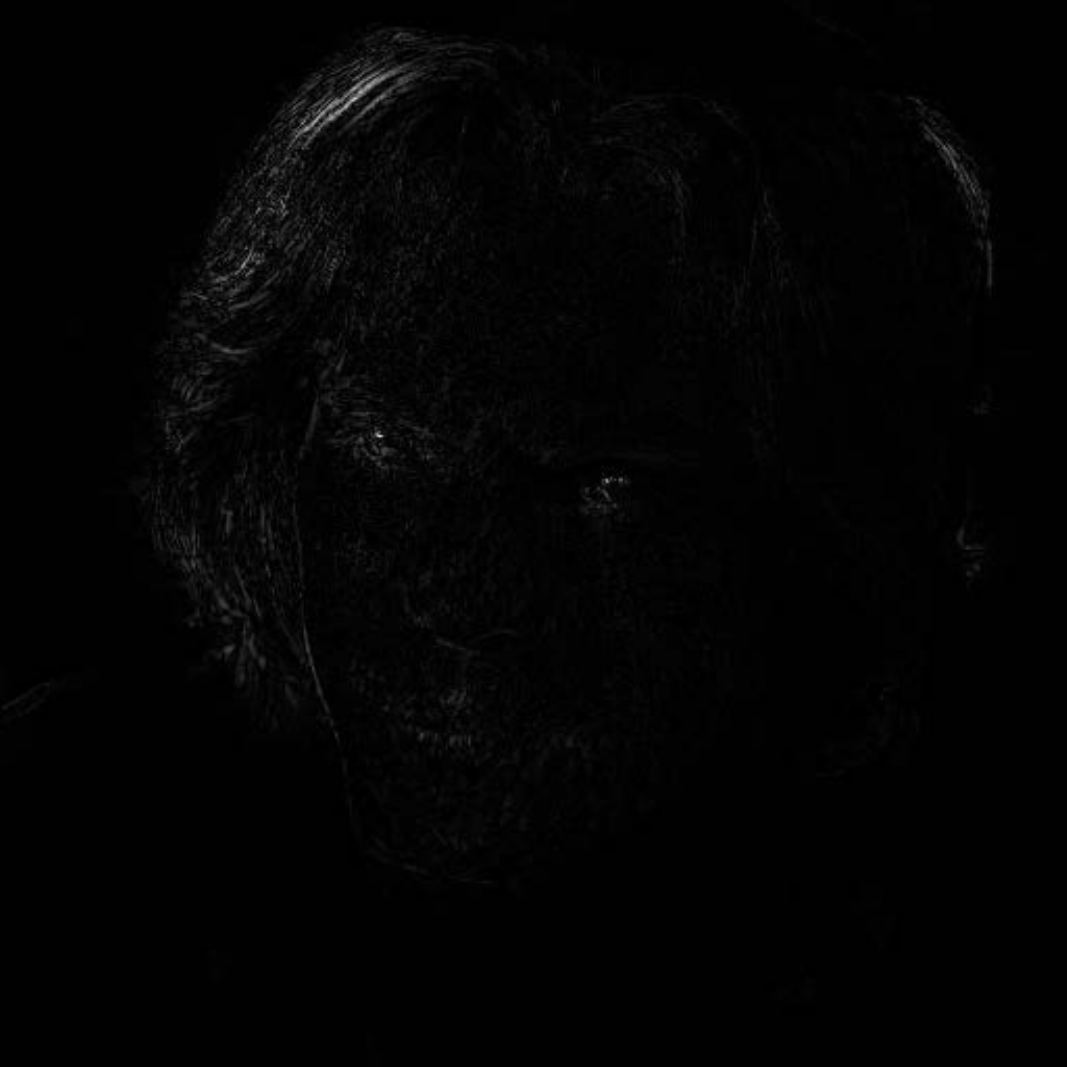} &
    \includegraphics[width=\swlankmark]{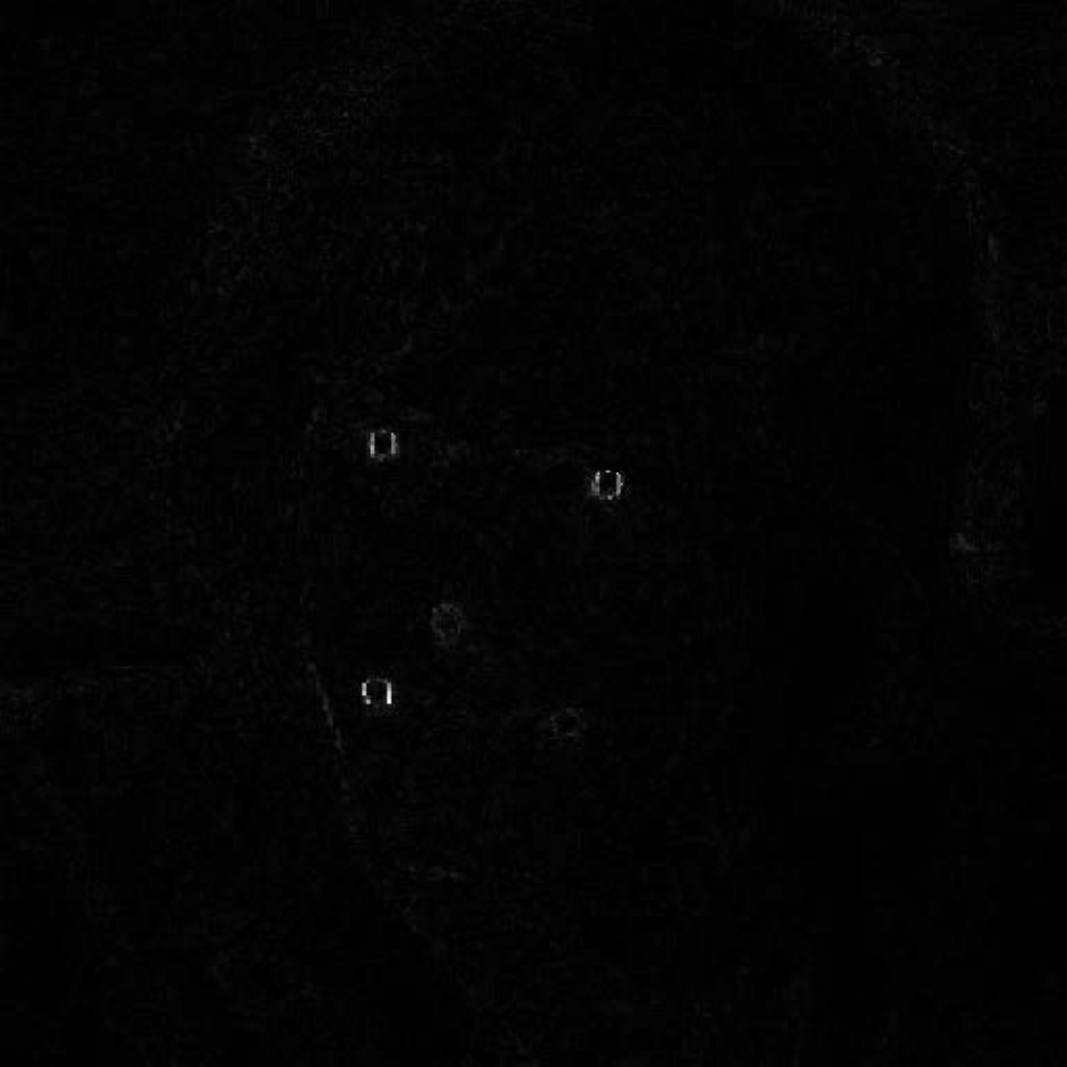} &
    \includegraphics[width=\swlankmark]{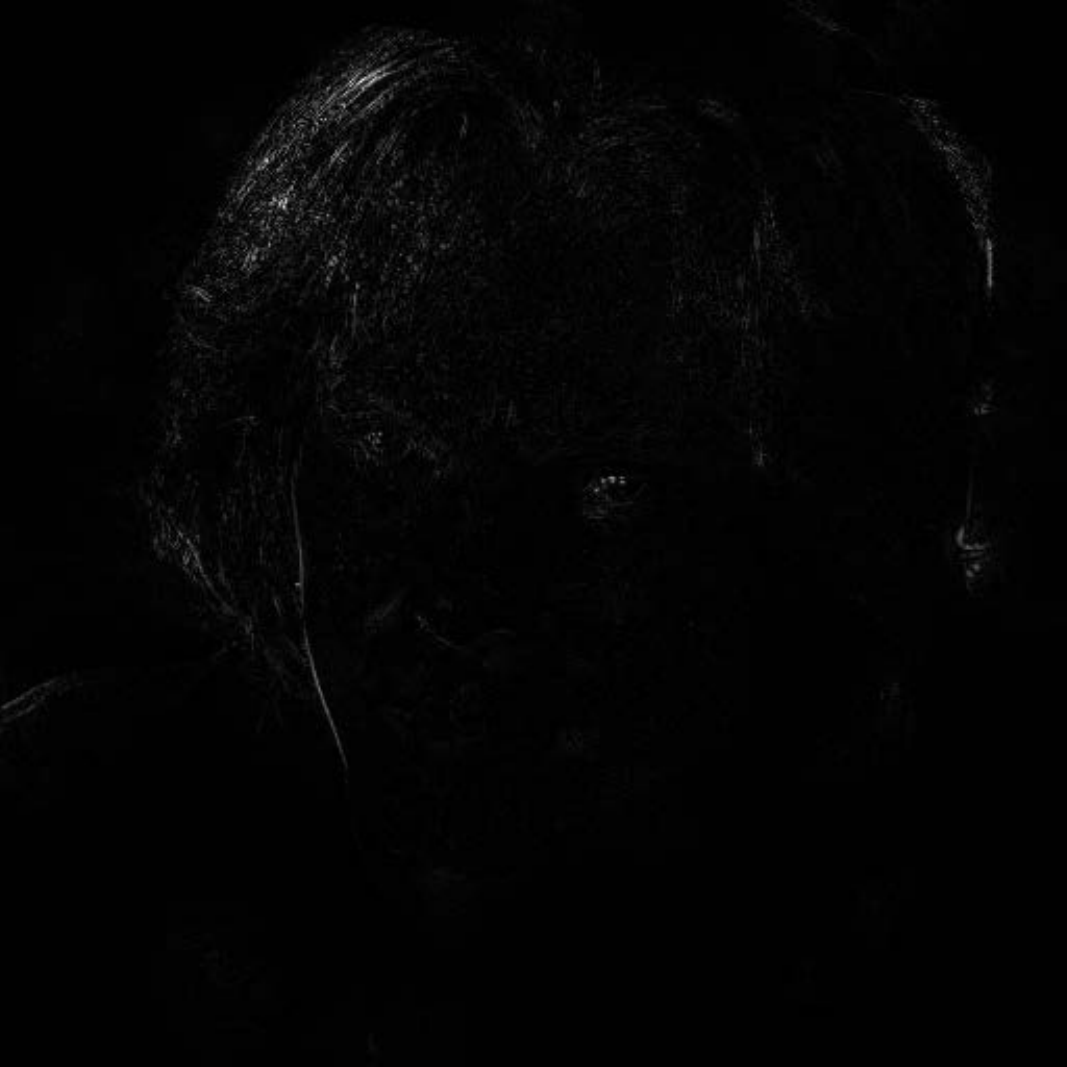} \\
    & & (c) & (d) & (e) & (f) & (g) \\
    (a) F0 & (b) F1 & \tiny $|F0-F1|$ & \tiny $|F0-F1|_{landmark}^{k=0}$ & \tiny $|F0-F1|_{restored}^{k=0}$ & \tiny $|F0-F1|_{landmark}^{k=3}$ & \tiny $|F0-F1|_{restored}^{k=3}$
\end{tabular}
\caption{
\textbf{This illustration demonstrates the negative impact of face alignment bias on face video restoration.} Frames (a) and (b) are two consecutive frames from a video, with landmarks represented as \textcolor{green}{green} points. There is almost no movement between these two faces and their difference value shown in (c) is primarily caused by video compression degradation. However, the difference in landmarks between these two frames, as shown in (d), is significant, attributable to the face alignment bias of existing face detection algorithms such as Awing~\cite{wang2019adaptive}. This bias can lead to noticeable differences in the restored results in (e) (achieved with RestoreFormer++~\cite{wang2023restoreformer++}), particularly in the hair area and facial components, resulting in jitters and flickers in the restored face video. By smoothing the alignment with $k=3$, the difference in the landmarks and restored result in (f) and (g) of the two frames is reduced, indicating a decrease in jitters and flickers in the restored face video. \textbf{For a clearer view, please zoom in and the corresponding videos are in the supplementary materials.}}
\label{fig:landmark_analysis}
\end{figure*}

With an insight analysis of the qualitative and quantitative results of jitters and noise-shape flickers, we conclude that three main factors lead to the issues: 1) \textit{the employment of GAN loss~\cite{gan}}, 2) \textit{the ignorance of temporal information}, and 3) \textit{the bias introduced by face alignment}. Previous discussion shows that EDVR-GAN~\cite{wang2019edvr} and BasicVSR-GAN~\cite{chan2021basicvsr} exhibit severe jitters and flickers than EDVR~\cite{wang2019edvr} and BasicVSR~\cite{chan2021basicvsr}, suggesting that GAN loss is a factor that causes the jitters and noise-shape flickers. Image-based algorithms are also trained with GAN loss, but their jitters and flickers are severer, especially RestoreFormer++~\cite{wang2023restoreformer++} and CodeFormer~\cite{zhou2022towards}, which have higher MSI and Warping Error scores as shown in Table~\ref{tab:analyze}. That is mainly due to the ignorance of temporal information when processing degraded videos with image-based algorithms. Unlike video-based methods that restore a face image with the help of a sequence of neighboring frames, the image-based algorithms process the degraded video frame by frame individually, missing the temporal information needed to maintain the temporal consistency of the restored video and leading to jitters and flickers. 

Moreover, the face alignment required by image-based algorithms can also contribute to jitters and flickers in the restored face video. As illustrated in Fig.~\ref{fig:landmark_analysis}, frames (a) and (b) are two consecutive frames from a video, with landmarks depicted as \textcolor{green}{green} points. Despite minimal movement between these two faces in the frames (the highlighted areas in (c) are primarily due to video compression degradation, as these two frames are extracted from a real-world face video), there is a significant bias between the landmarks of these two frames, as demonstrated in (d). This bias further results in a noticeable difference between the restored results (achieved with RestoreFromer++~\cite{wang2023restoreformer++}) of these two degraded face images, particularly in the hair area and facial components, as shown in (e). This difference manifests as jitters and flickers in a restored face video. Our experiment's results in (f) and (g) indicate that by smoothing the landmarks with $k=3$ (as referred to in Eq.~\ref{eq:smoothing}), the bias of the landmarks and restored results is reduced, suggesting mitigation of jitters and flickers in the restored face video.

\section{Methodology}\label{sec:method}

To adapt image-based face restoration algorithms for degraded face video applications, it is necessary to mitigate the jitters and noise-shape flickers observed in the restored results of image-based methods. We explore potential solutions based on the three previously analyzed contributing factors: the employment of GAN loss~\cite{gan}, the ignorance of temporal information, and the bias introduced by face alignment. Given the significant role GAN loss plays in achieving high-frequency details in blind face restoration, its removal to reduce jitters and noise-shape flickers would substantially compromise restoration quality (as demonstrated by the results from EDVR~\cite{wang2019edvr} and BasicVSR~\cite{chan2021basicvsr}). Therefore, our proposed solutions primarily focus on addressing the latter two factors. Ultimately, we introduce a temporal consistency network cooperating with an alignment smoothing strategy to alleviate jitters and noise-shape flickers in the restored face videos, while ensuring their overall restoration quality.

In this section, we first present the alignment smoothing strategy that we apply to face alignment, which is a preprocessing step for face image restoration methods. Then, we propose a temporal consistency network (TCN) with Cross-Attended Swin Transformer Layers (CASTL), which is incorporated into existing face image restoration methods to exploit the temporal information across the neighboring frames.

\subsection{Alignment smoothing}

To restore faces from the $t$-th degraded frame $\mathbf{I}_t^d$ in a video using the current state-of-the-art face image restoration methods~\cite{wang2021towards,wang2022restoreformer,wang2023restoreformer++,zhou2022towards}, $\mathbf{I}_t^d$ should be aligned to a reference face image $\mathbf{I}_{ref}$ from FFHQ~\cite{ffhq} before the restoration process. The alignment is based on landmarks detected by a face detector Awing~\cite{wang2019adaptive}. We denote the detected landmarks from $\mathbf{I}_t^d$ as $\mathbf{Z}_t^m$, where $m \in [0,1,\dots,M]$ and $M$ is the number of landmarks. These landmarks between the neighboring frames are unstable due to the video degradation, which causes large jitters in facial components and noticeable noise-shape flickers between frames in the aligned frames. To reduce the jitters and flickers in the restored video, we smooth the landmarks by averaging the landmarks of $2k+1$ neighboring frames, which can be formulated as:
\begin{equation}
\label{eq:smoothing}
    \mathbf{\hat{Z}}_t^m = \frac{1}{2k+1}\sum_{i=(t-k)}^{(t+k)}{\mathbf{Z}_i^m}
\end{equation}
Then $\mathbf{I}_t^d$ is aligned to $\mathbf{I}_{ref}$ based on the smoothed landmarks $\mathbf{\hat{Z}}_t^m$ and the aligned face image $\mathbf{\bar{I}}_t^d$ is the input of the restoration methods.

\subsection{Temporal Consistency Network}

\begin{figure}[t]
\centering
    \includegraphics[width=1.0\linewidth]{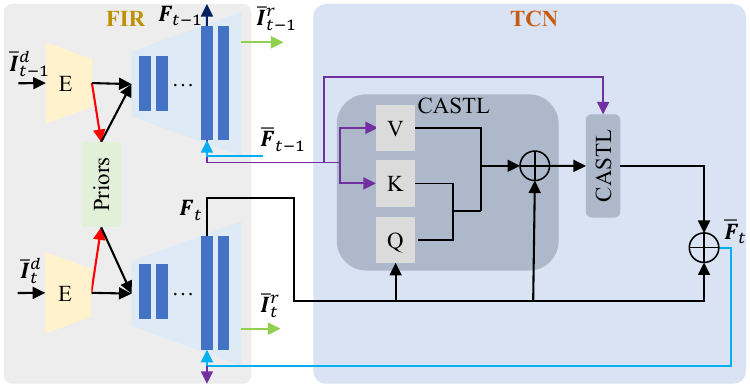}
    \caption{\textbf{Framework of TCN}. Temporal Consistency Network (TCN) is integrated into existing face image restoration (FIR) methods to utilize temporal information for face video restoration. FIR~\cite{wang2021towards,wang2022restoreformer,wang2023restoreformer++,zhou2022towards} extracts hidden feature $\mathbf{F}_{t}$ from the aligned degraded frame $\mathbf{\bar{I}}_{t}^d$. TCN contains two CASTLs. The first CASTL takes $\mathbf{F}_{t}$ as a query while $\mathbf{\bar{F}_{t-1}}$, which is the output of TCN from the previous frame, acts as key and value. Then, the output of TCN \redm{$\mathbf{\bar{F}_{t}}$}, which contains temporal information, replaces $\mathbf{F}_{t}$ as the input of the rest block of the decoder to reconstruct the restored face image $\mathbf{\bar{I}}_t^r$.
    }
    \label{fig:TCN}
\end{figure}

Directly applying existing face image restorations ignores the temporal information, which is crucial to reduce jitters in facial components and noticeable noise-shape flickers between frames. To preserve the restoration quality of the image-based methods while mitigating the jitters and flickers in their restored videos, we incorporate our proposed temporal consistency network (TCN) into existing image-based face restoration methods~\cite{wang2021towards,wang2022restoreformer,wang2023restoreformer++,zhou2022towards} to model the temporal information. As shown in Fig.~\ref{fig:TCN}, our proposed TCN is plugged into FIR, a universal framework of \textbf{F}ace \textbf{I}mage \textbf{R}estoration algorithms. As described in Subsec.~\ref{subsec:analyses}, FIR consists of three components: an encoder, extra-provided priors, and a decoder. 
\redm{
Note that the TCN is inserted into the decoder where the resolution of the hidden feature is upsampled to $256\times 256$ and the detailed analyses of this setting are in Sec.~\ref{sec:ablation_studies}.
}

Specifically, the decoder of a face image restoration method extracts a hidden representation $\mathbf{F}_{t}$ from the degraded frame $\mathbf{\bar{I}}_{t}^d$. We model the temporal information between the current and previous frames with two Swin Transformer Layers based Cross-Attention (CASTL). The first CASTL uses $\mathbf{F}_{t}$ as a query and $\mathbf{\bar{F}_{t-1}}$, which is the output of TCN from the previous frame, as key and value. The output of TCN $\mathbf{\bar{F}}_{t}$ in the current frame is obtained by adding the output of the second CASTL to $\mathbf{F}_{t}$. Finally, $\mathbf{\bar{F}}_{t}$ replaces $\mathbf{F}_{t}$ as the input of the rest block of the decoder to reconstruct the restored face image $\mathbf{\bar{I}}_t^r$.

\subsection{Network training}
\noindent\textbf{Learning Losses.}
A pixel-wise L1 loss $\mathcal{L}_1$, a perceptual loss $\mathcal{L}_p$, and an adversarial loss $\mathcal{L}_a$ are used to train the extended face image restoration methods~\cite{wang2021towards,wang2022restoreformer,wang2023restoreformer++,zhou2022towards} incorporating with the proposed TCN. Besides, we use an additional temporal loss~\cite{lai2018learning} to smooth the jitters in the facial components and alleviate the jitters and flickers. The temporal loss is formulated as follows:
\begin{equation}
    \mathcal{L}_t = \sum_{i=1}^{H \times W}||\mathbf{\bar{I}}_t^r - \mathbf{\hat{\bar{I}}}_{t-1}^r||,
\end{equation}
where $\mathbf{\hat{\bar{I}}}_{t-1}^r$ is attained by warping previous restored frame $\mathbf{\bar{I}}_{t-1}^r$ toward $\mathbf{\bar{I}}_t^r$ with the optical flow estimated between ground true $\mathbf{\bar{I}}_{t-1}^h$ and $\mathbf{\bar{I}}_{t}^h$.
Therefore, the overall loss is:
\begin{equation}
    \mathcal{L} = \lambda_{1}\mathcal{L}_1 + \lambda_{p}\mathcal{L}_p + \lambda_{a}\mathcal{L}_a + \lambda_{t}\mathcal{L}_t,
\end{equation}
where $\lambda_{\cdot}$ are the weighting factors for different losses.

\redm{
\noindent\textbf{Training details.}
We train the proposed TCN on videos from VFHQ~\cite{xie2022vfhq}, a widely-used high-quality face video dataset for blind face video restoration. VFHQ provides the necessary temporal information for TCN learning. However, as discussed in Sec.~\ref{subsec:analyses} and shown in Fig.~\ref{fig:ffhq_and_vfhq}, VFHQ videos inherently suffer from noticeable motion blur and do not match the quality of face images in FFHQ~\cite{ffhq}. To avoid a significant drop in the restored quality of image-based methods, we implement two strategies: 1) We ensure the training face videos are of relatively high quality by filtering out low-quality videos in VFHQ, specifically those with a face size smaller than 800x800 pixels in the original collected video. This filtered dataset is referred to as VFHQ-800 in our work. 2) To preserve the high restoration quality achieved by image-based methods trained on high-quality face images (FFHQ), we freeze the original weights of these methods and train only the newly added TCN to extract temporal information from VFHQ.
}

\begin{figure*}[!t]
\setlength\tabcolsep{1pt}
\scriptsize
\centering
\begin{tabular}{cccccccc}
    \multicolumn{8}{c}{(a) Restored results} \\
    \includegraphics[width=\swexp]{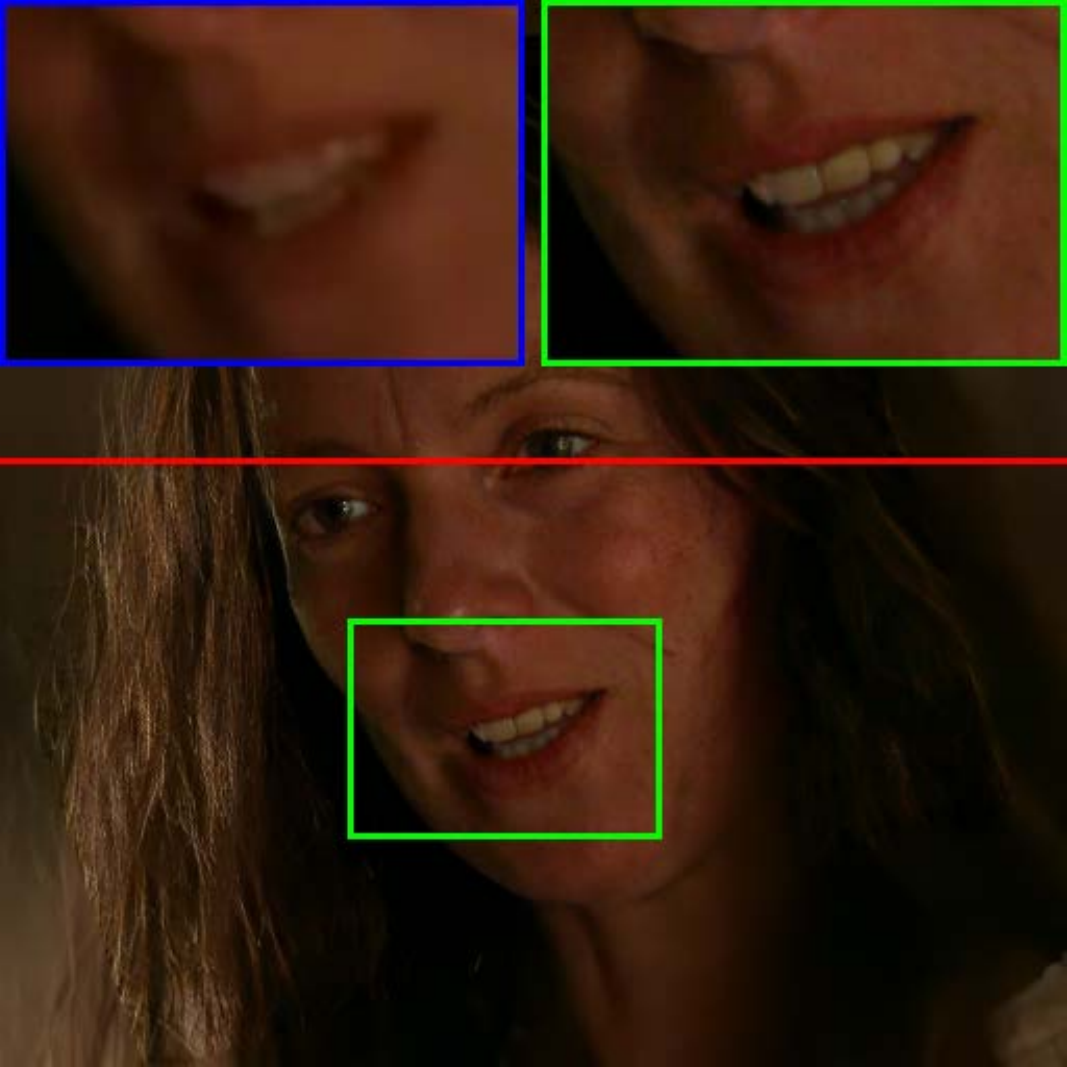} &
    \includegraphics[width=\swexp]{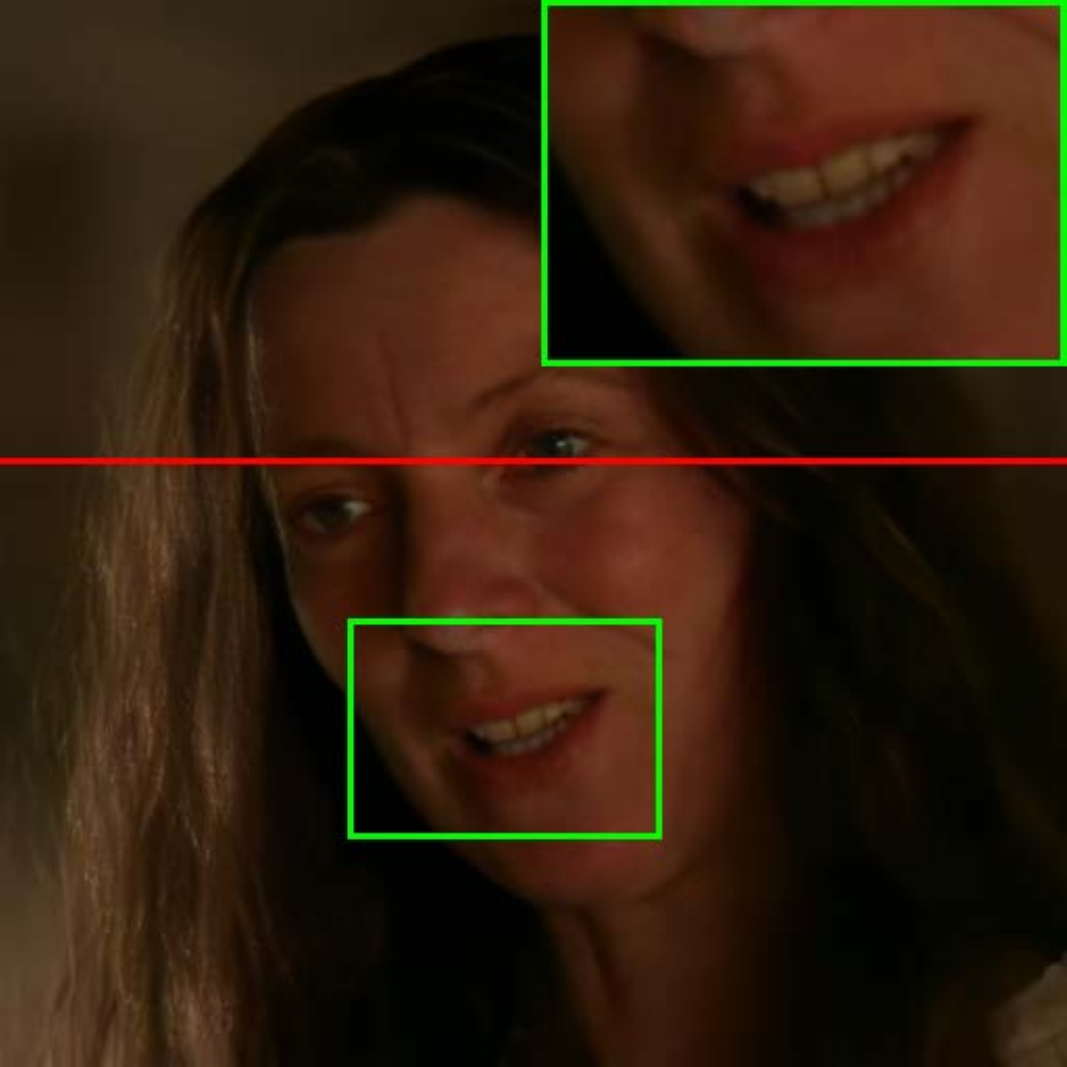} &
    \includegraphics[width=\swexp]{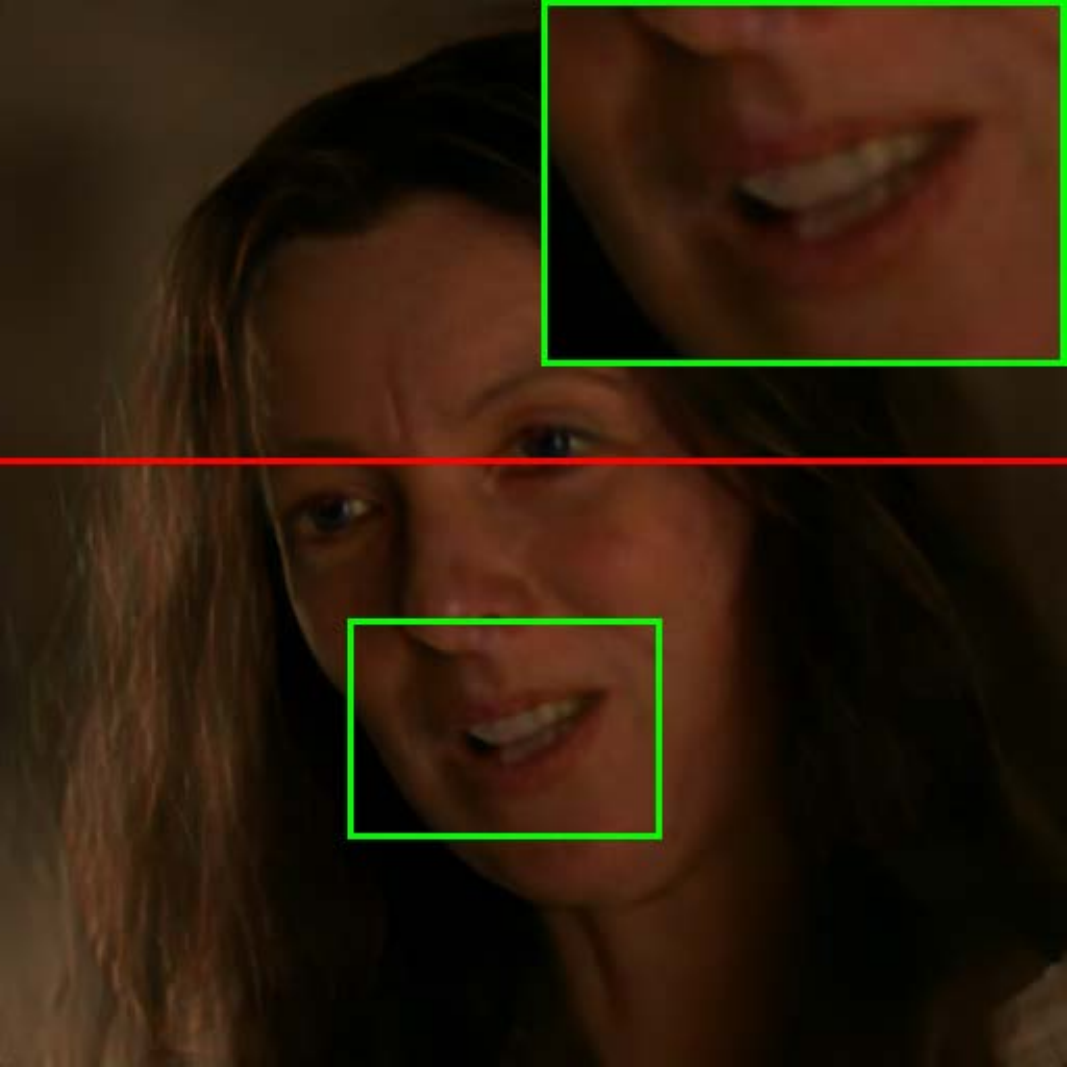} &
    \includegraphics[width=\swexp]{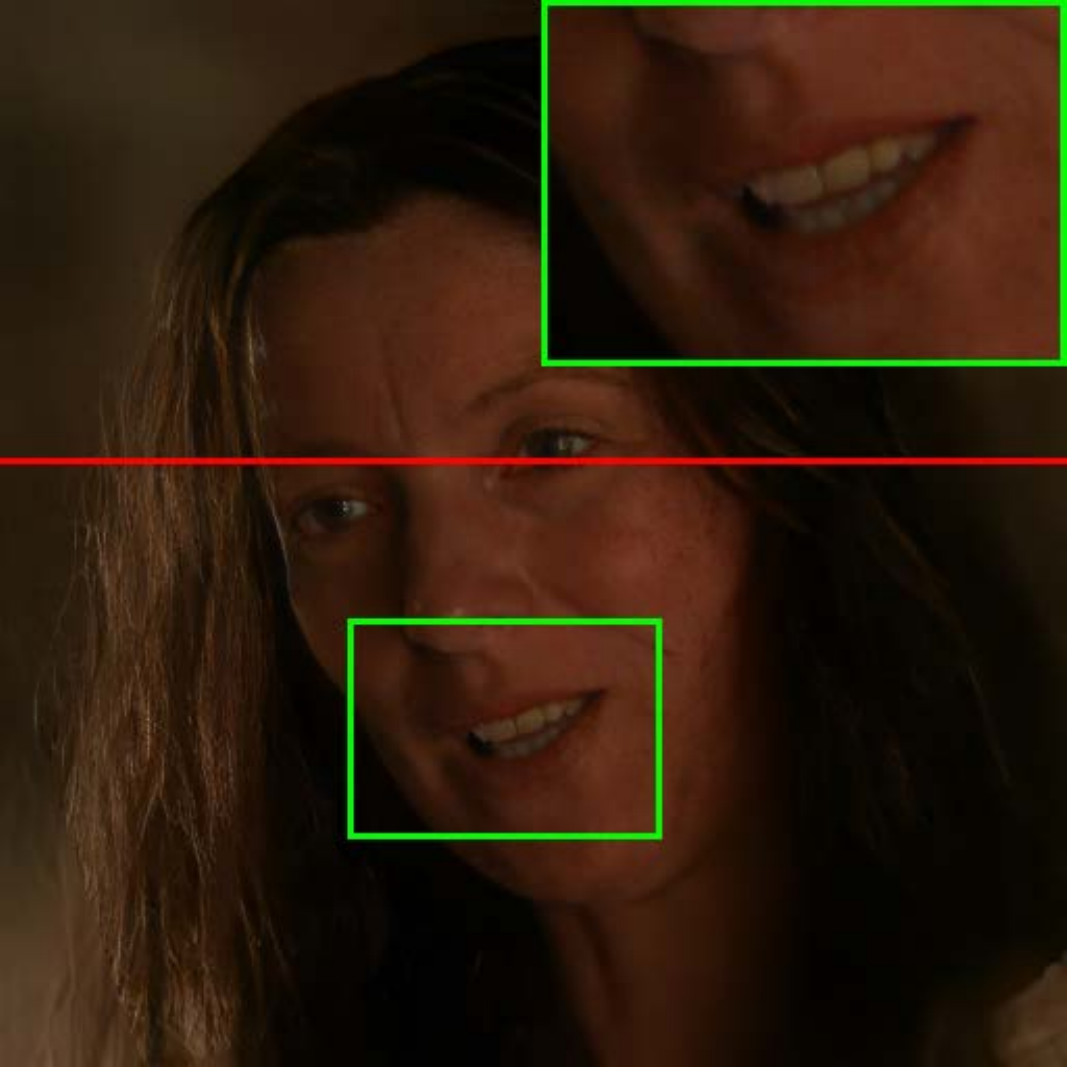} &
    \includegraphics[width=\swexp]{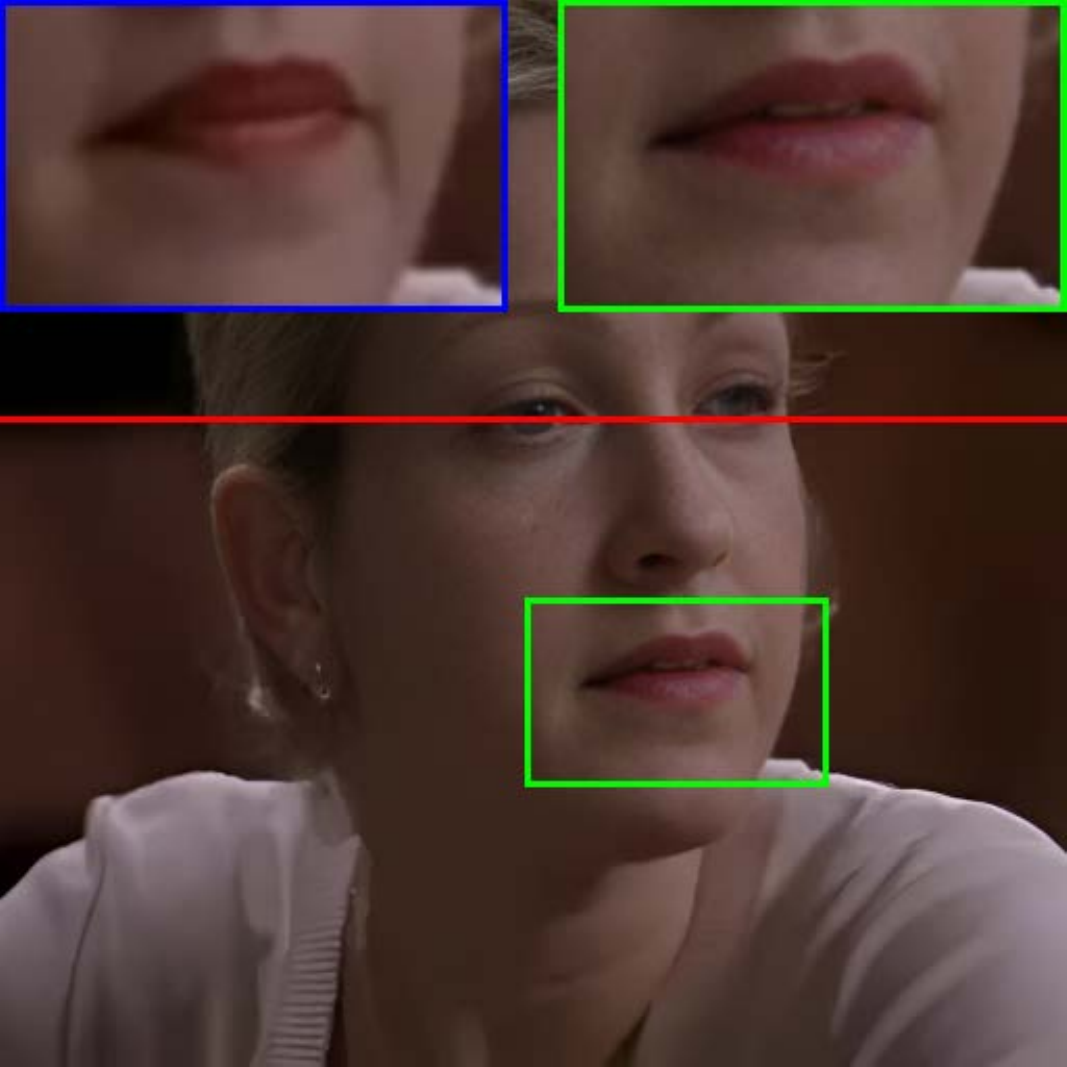} &
    \includegraphics[width=\swexp]{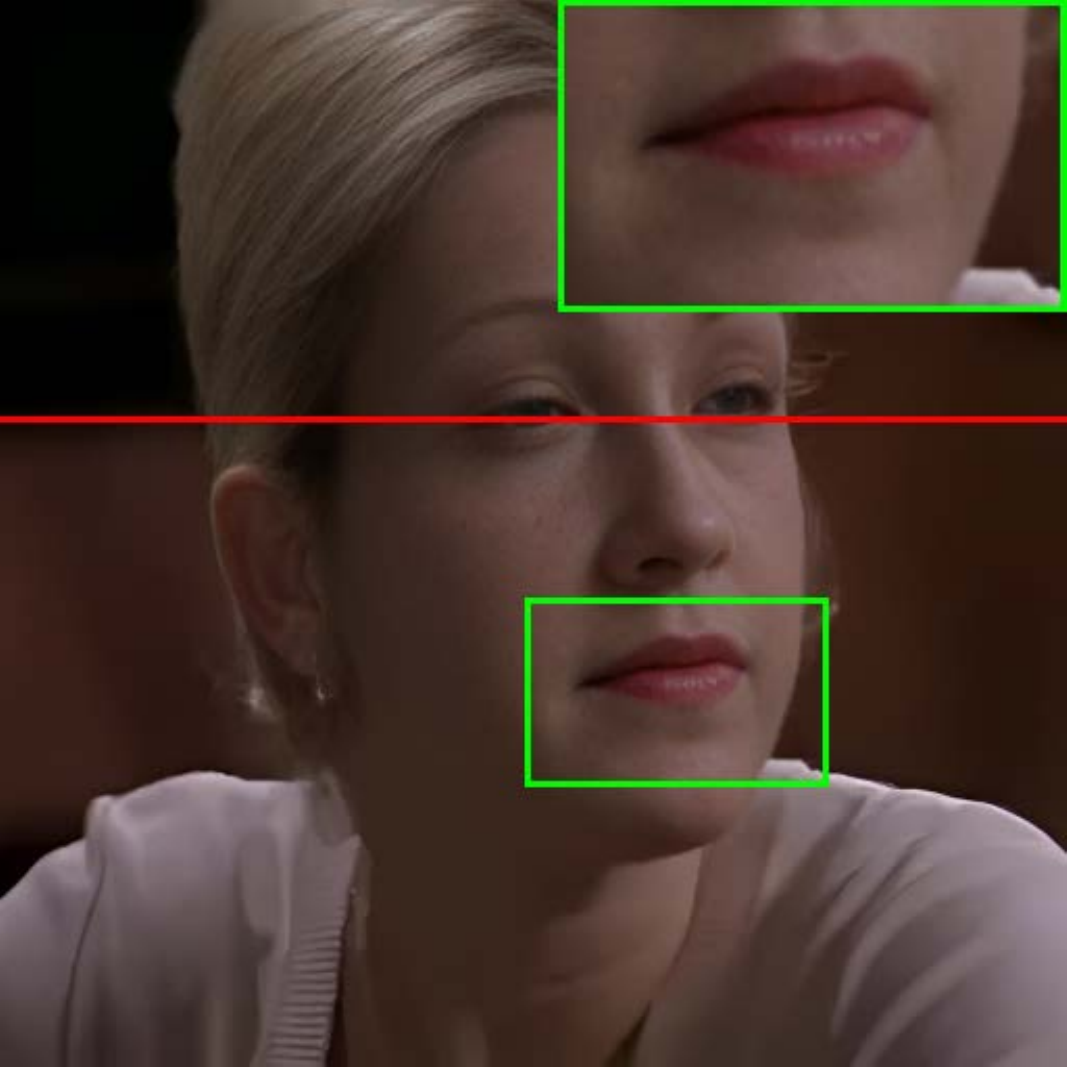} &
    \includegraphics[width=\swexp]{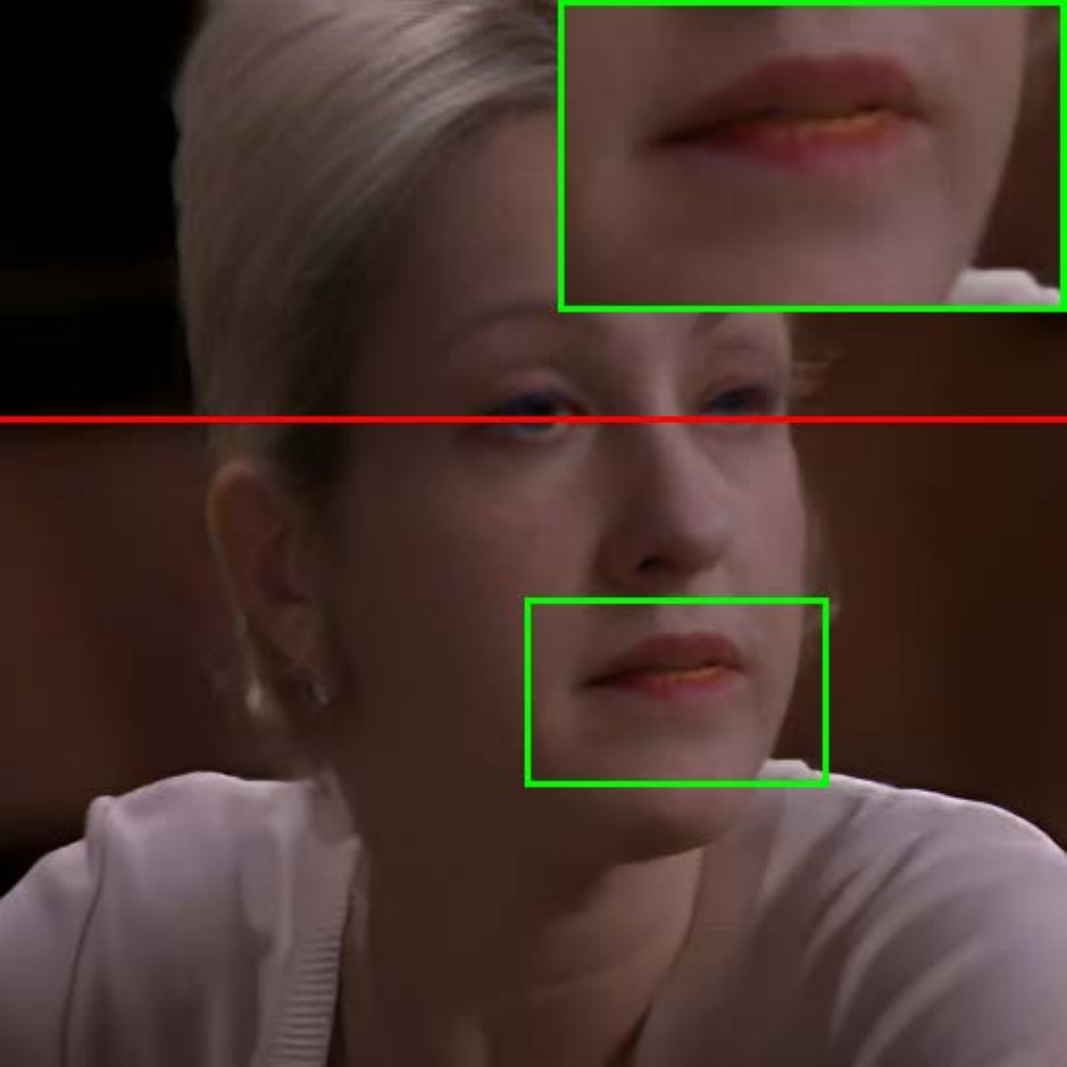} &
    \includegraphics[width=\swexp]{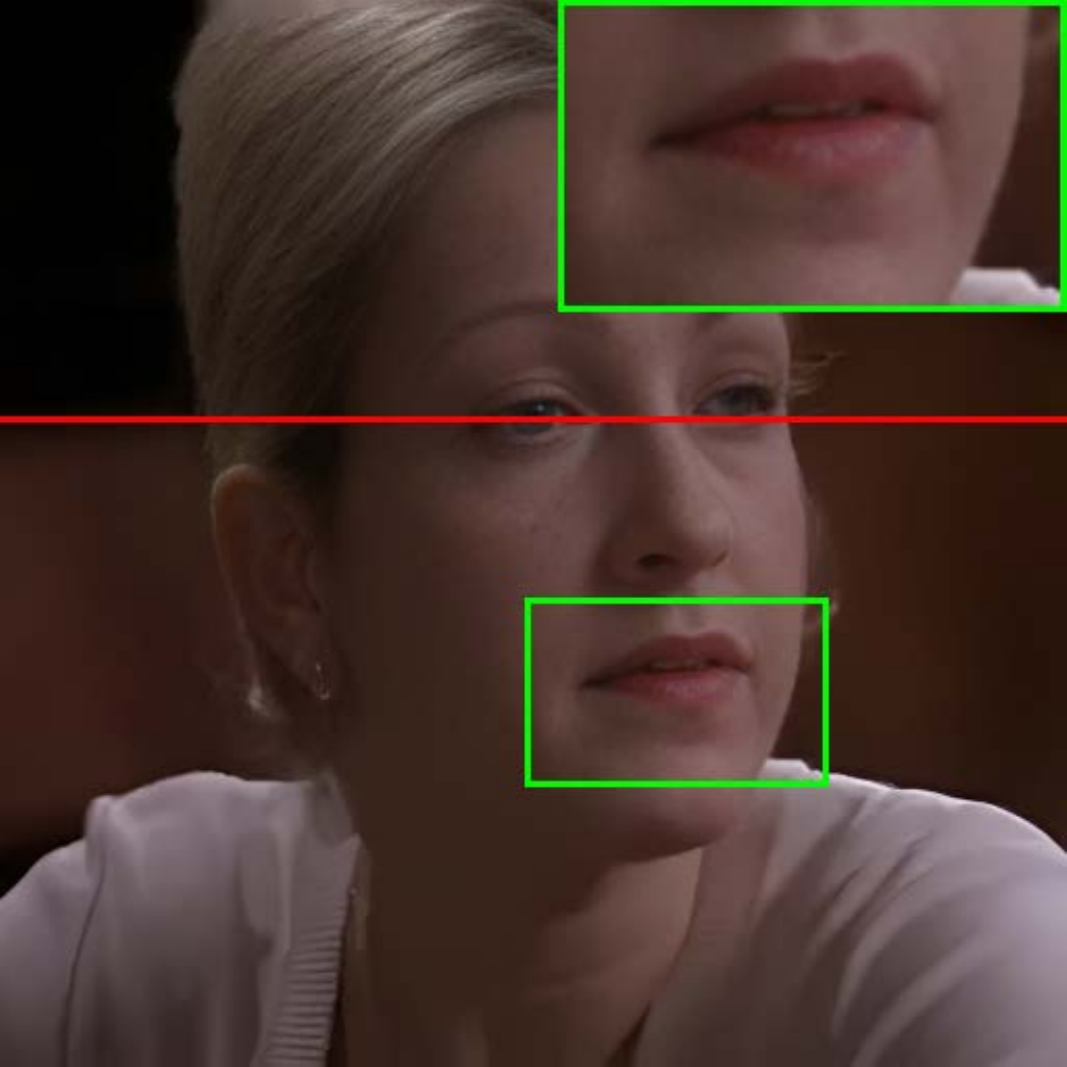} \\
    \multicolumn{8}{c}{(b) Jitters visualization of the red slice in (a)} \\
    \includegraphics[width=\swexp]{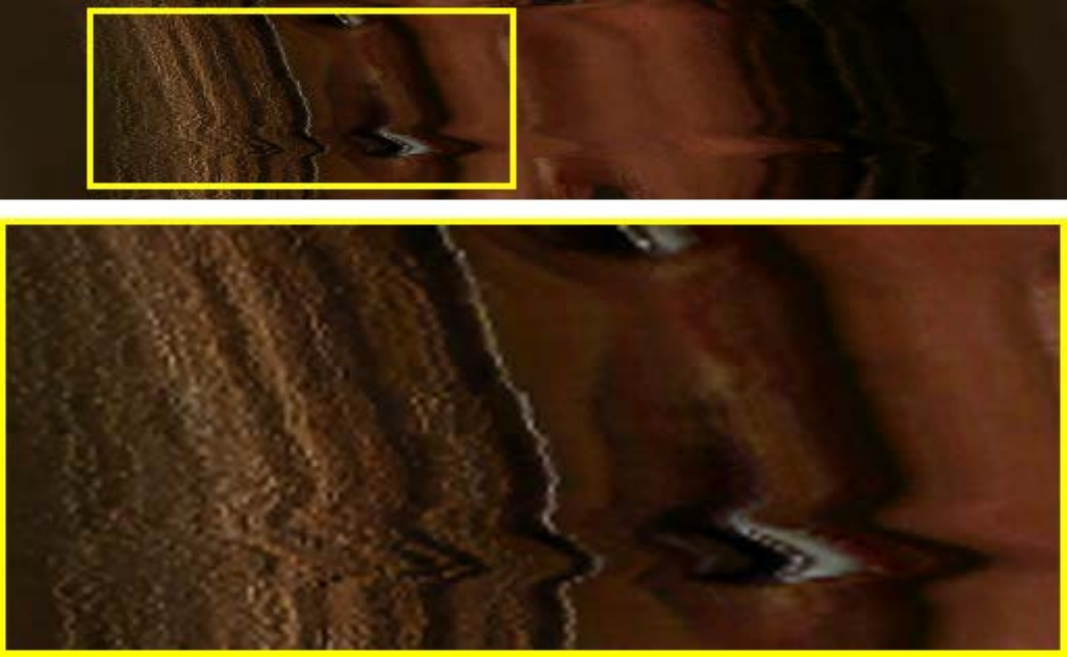} &
    \includegraphics[width=\swexp]{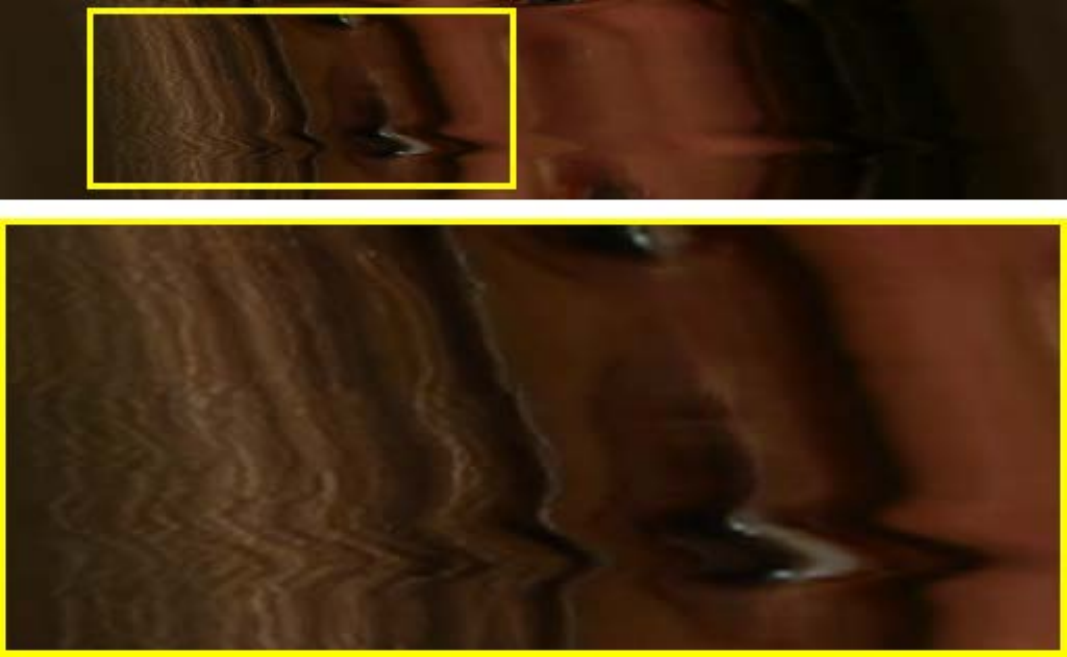} &
    \includegraphics[width=\swexp]{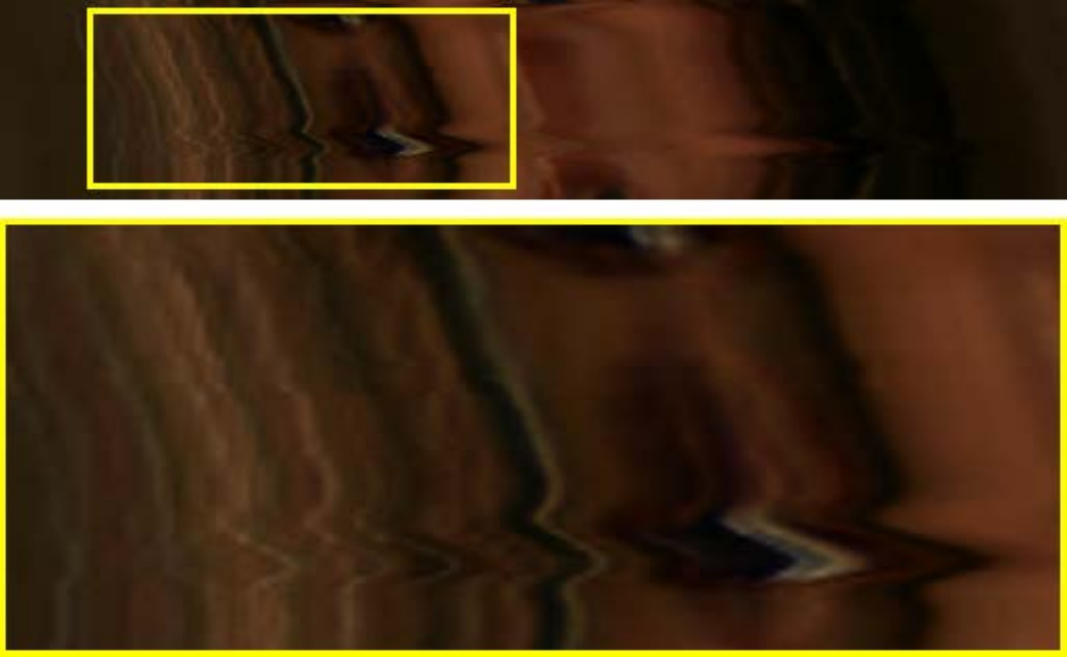} &
    \includegraphics[width=\swexp]{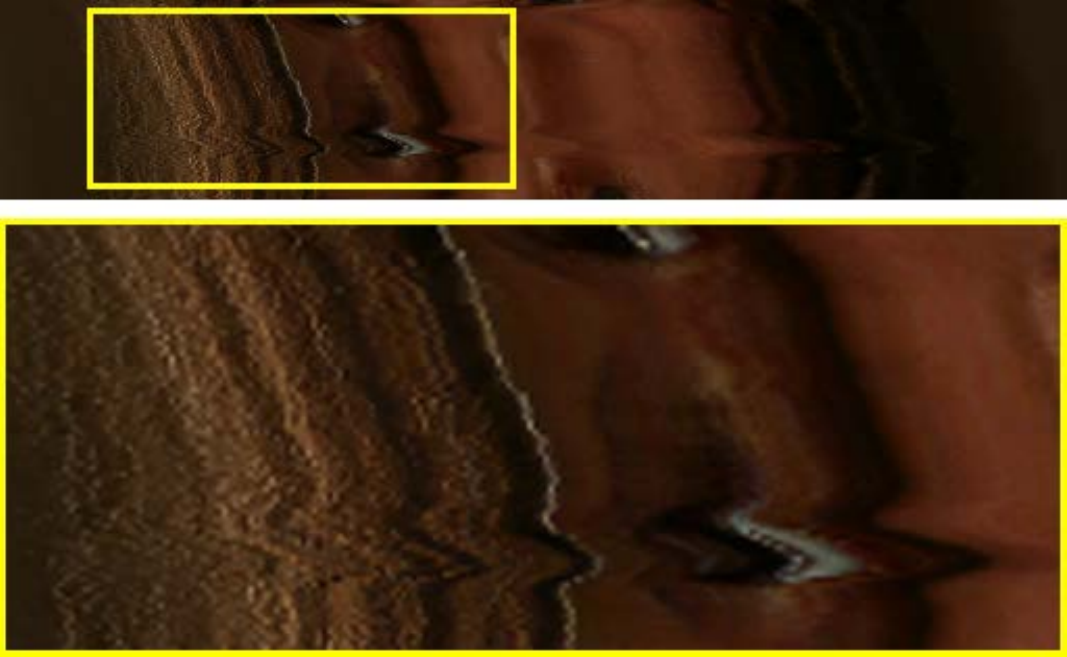} &
    \includegraphics[width=\swexp]{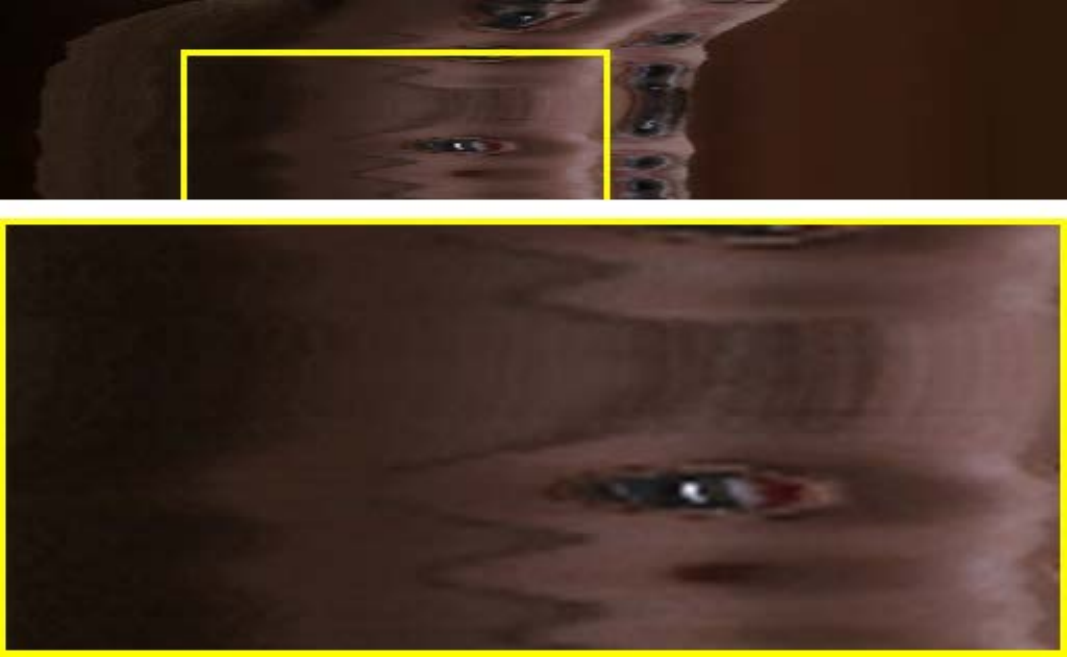} &
    \includegraphics[width=\swexp]{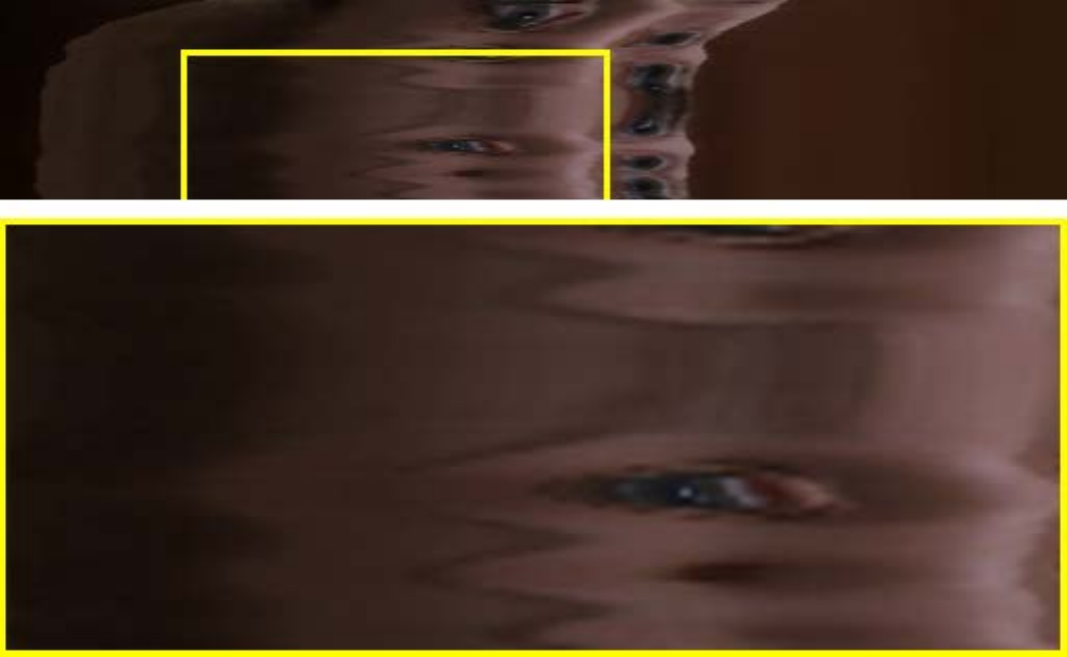} &
    \includegraphics[width=\swexp]{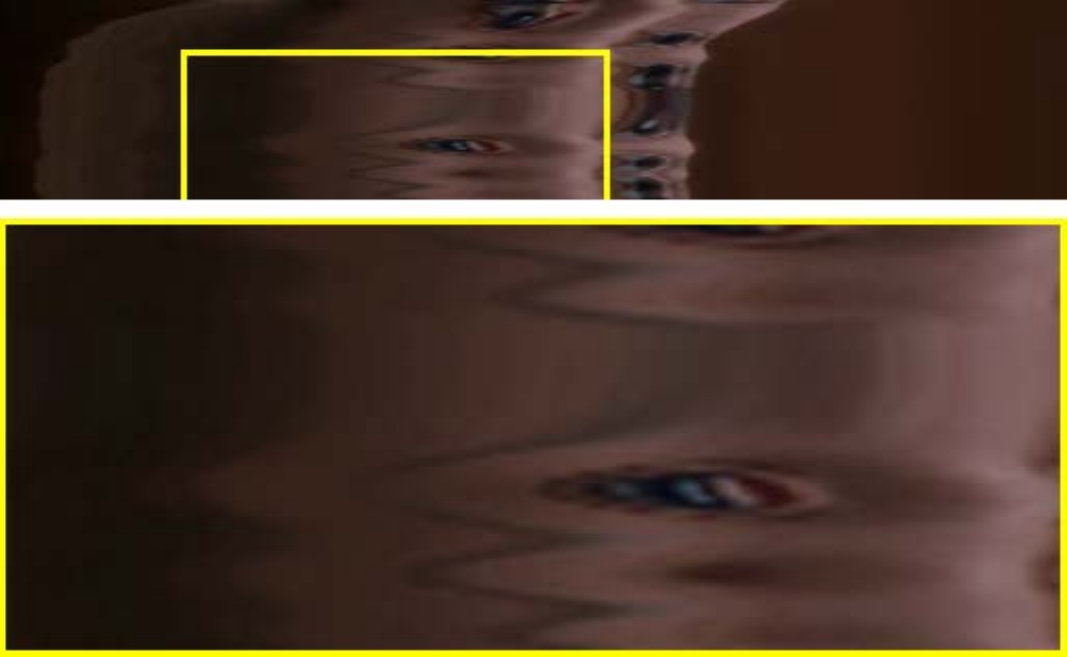} &
    \includegraphics[width=\swexp]{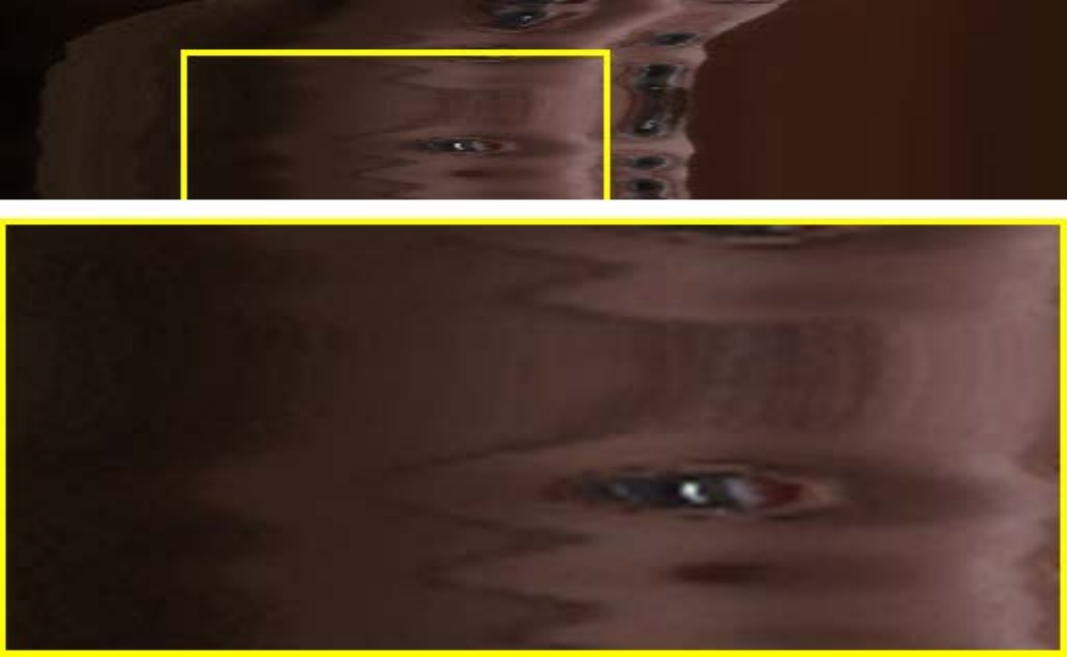} \\
    \multicolumn{8}{c}{(c) Noise-shape flickers visualization} \\
    \includegraphics[width=\swexp]{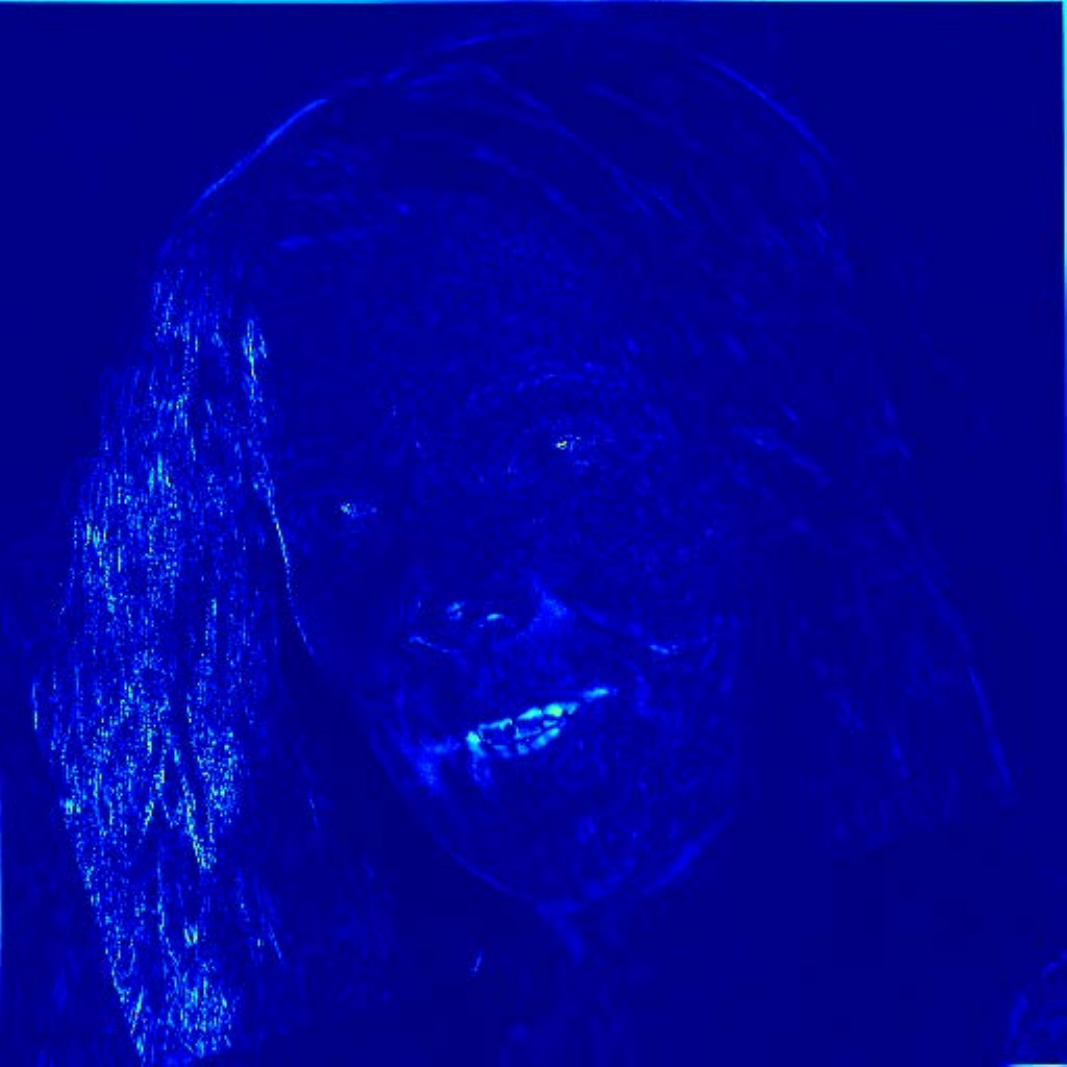} &
    \includegraphics[width=\swexp]{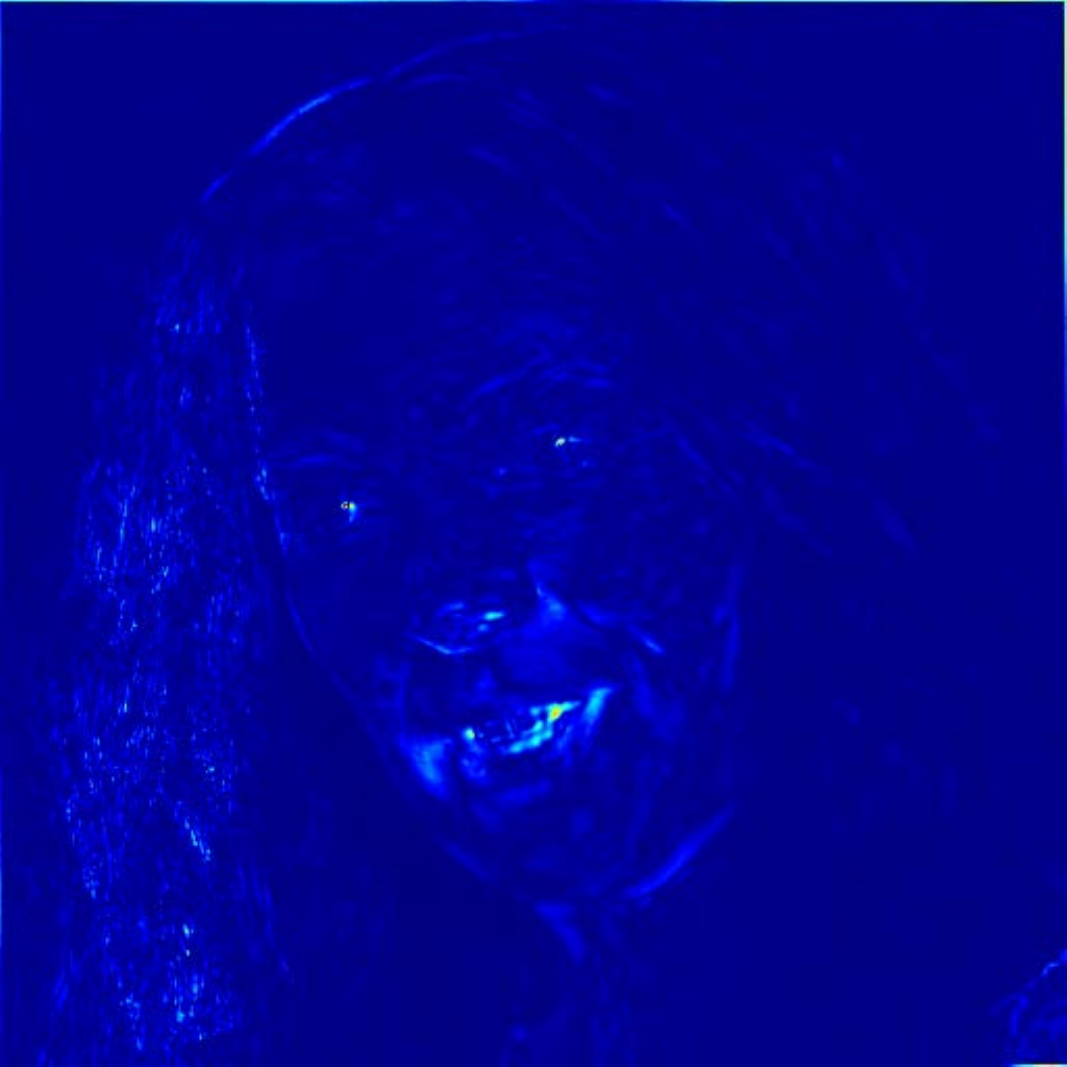} &
    \includegraphics[width=\swexp]{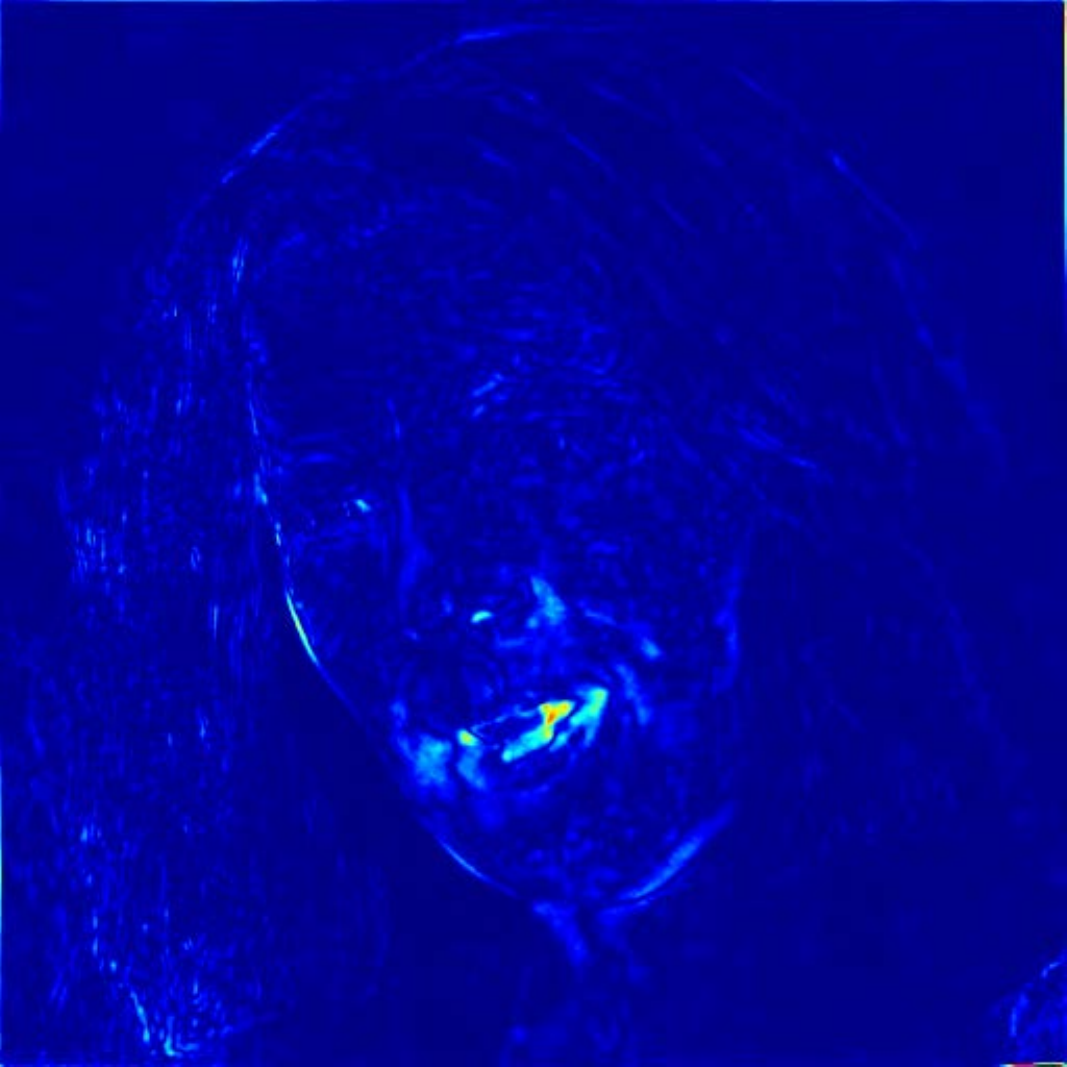} &
    \includegraphics[width=\swexp]{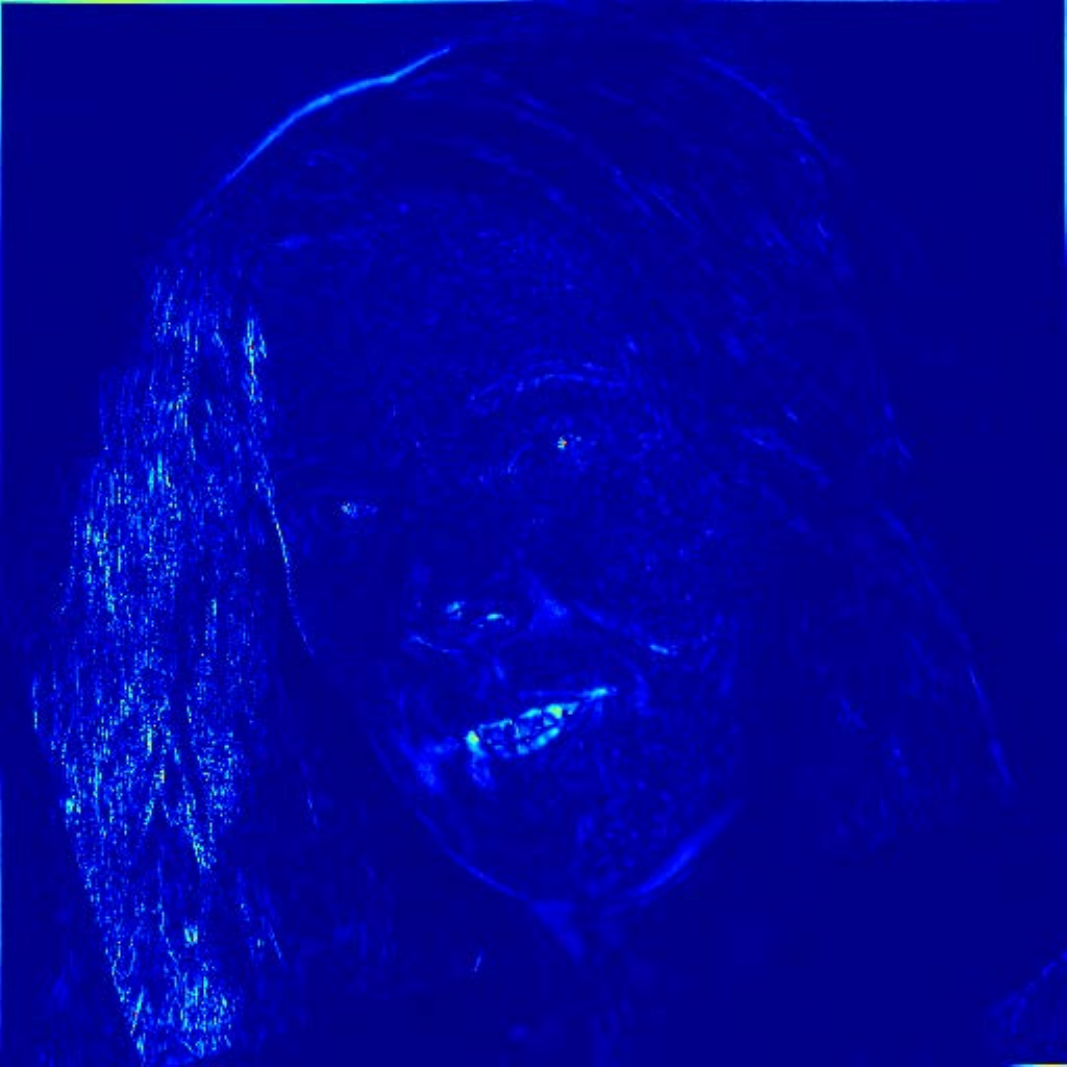} &
    \includegraphics[width=\swexp]{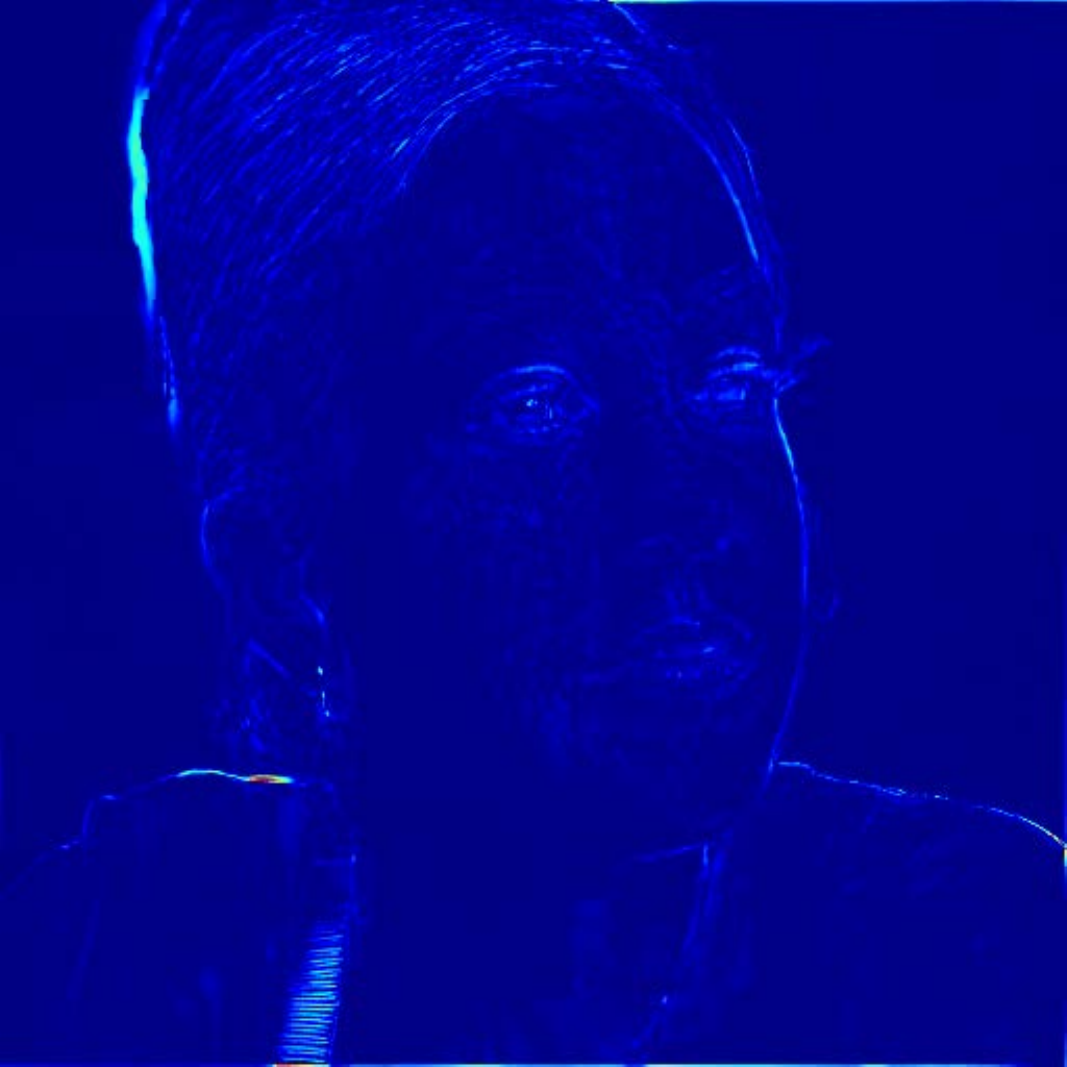} &
    \includegraphics[width=\swexp]{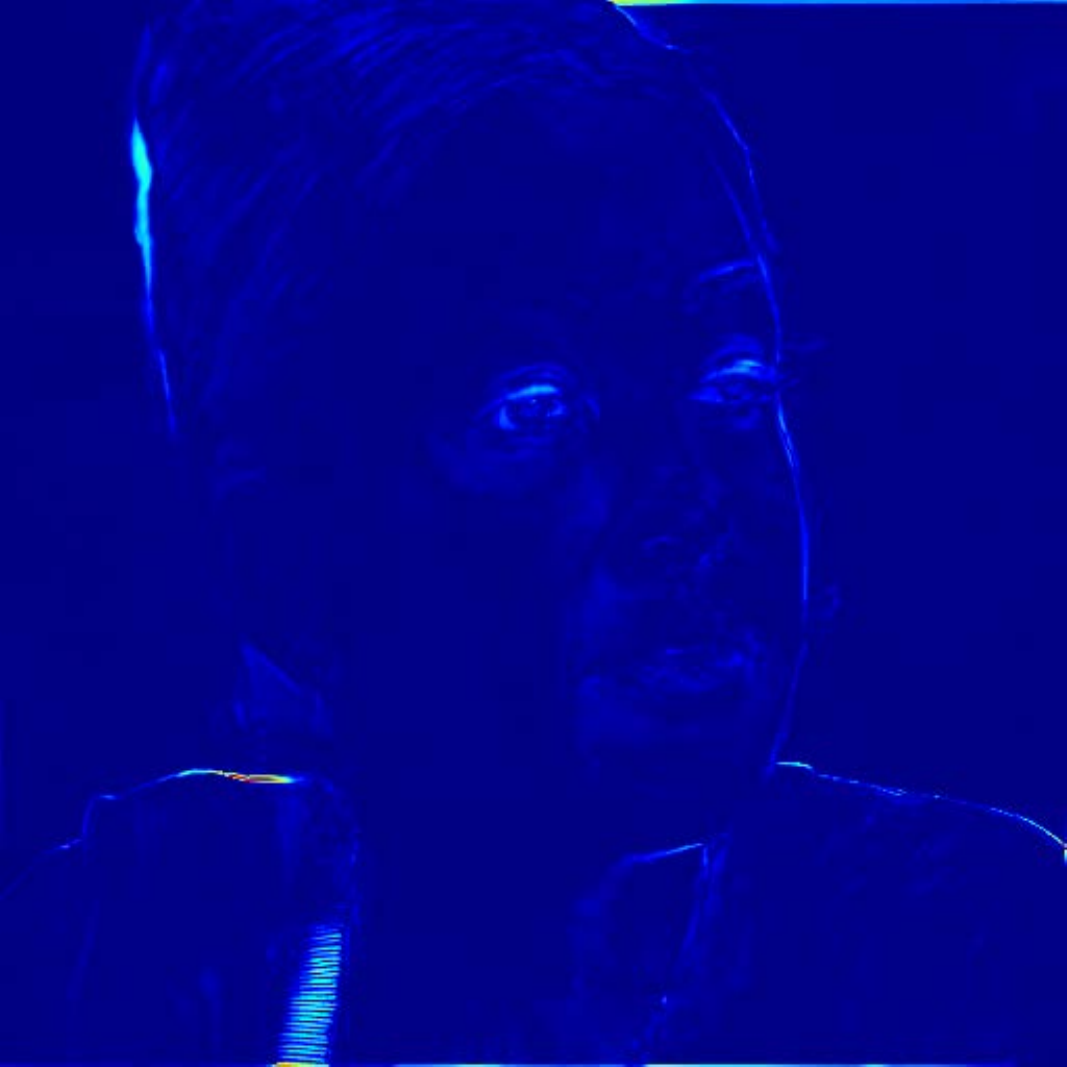} &
    \includegraphics[width=\swexp]{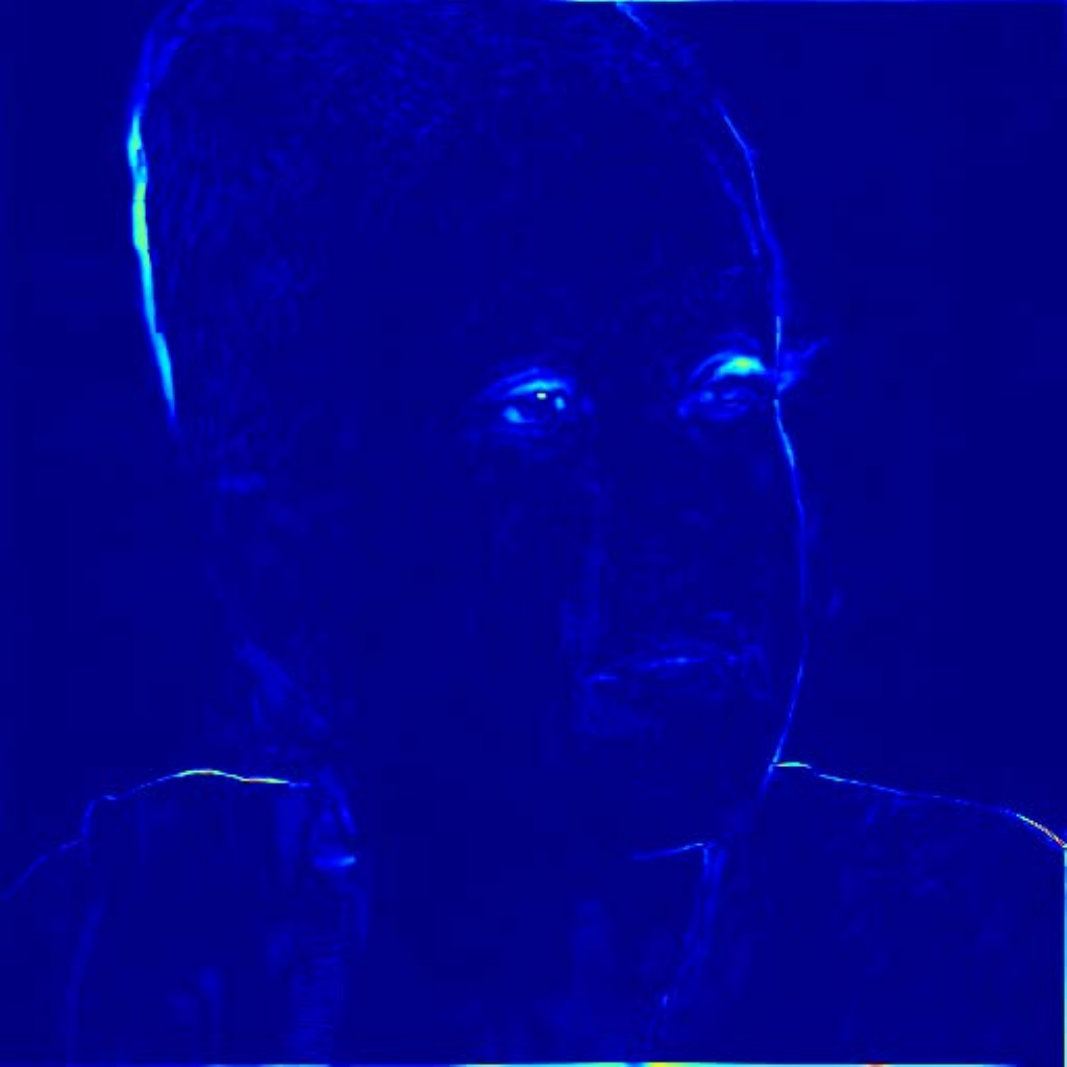} &
    \includegraphics[width=\swexp]{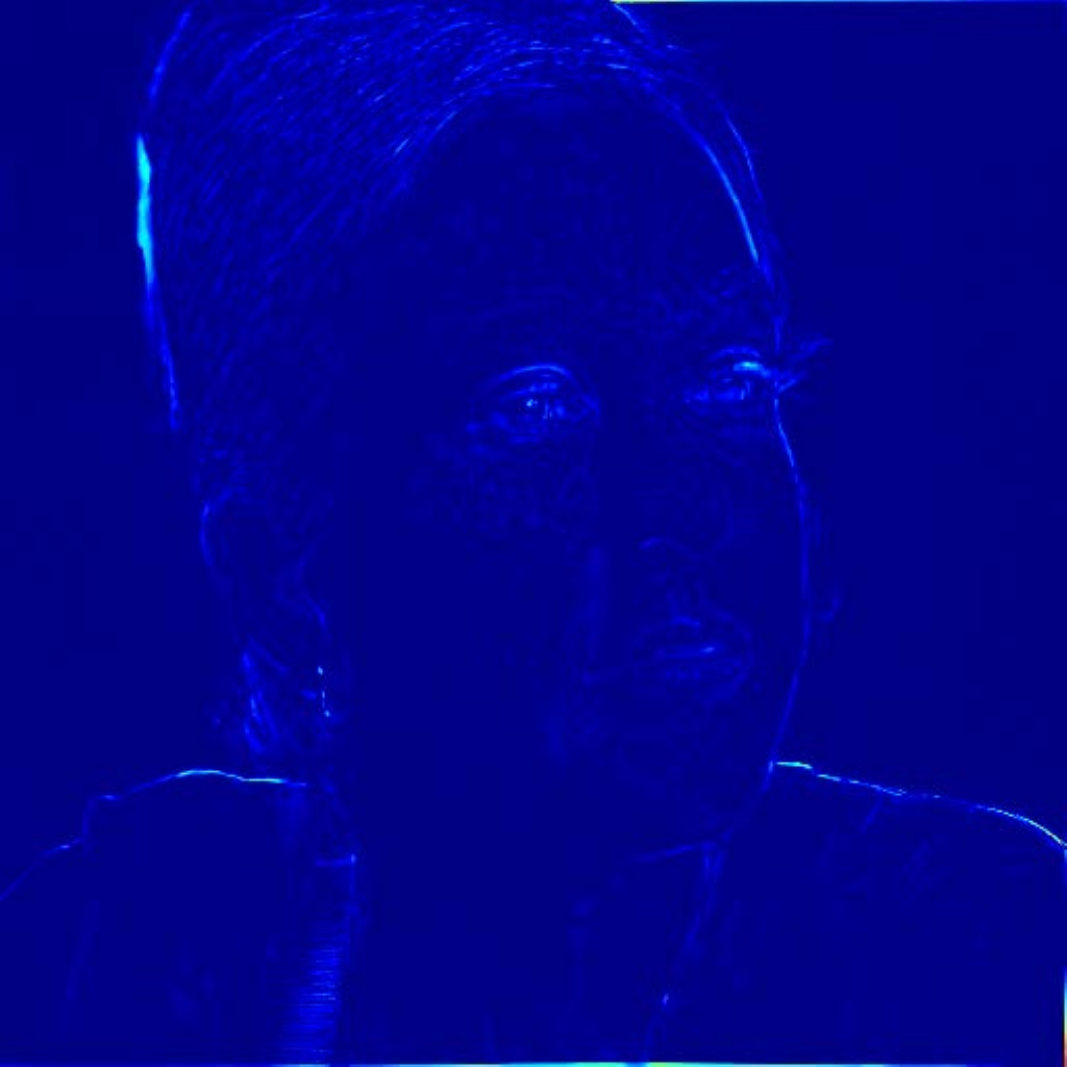}\\
    RestoreFormer++~\cite{wang2023restoreformer++} & + \textbf{Ours} & + DVP~\cite{lei2020blind} & +~\cite{Lei_2023_CVPR} & CodeFormer~\cite{zhou2022towards} & + \textbf{Ours} & + DVP~\cite{lei2020blind} & +~\cite{Lei_2023_CVPR} \\
    \hline
    \multicolumn{8}{c}{(a) Restored results} \\
    \includegraphics[width=\swexp]{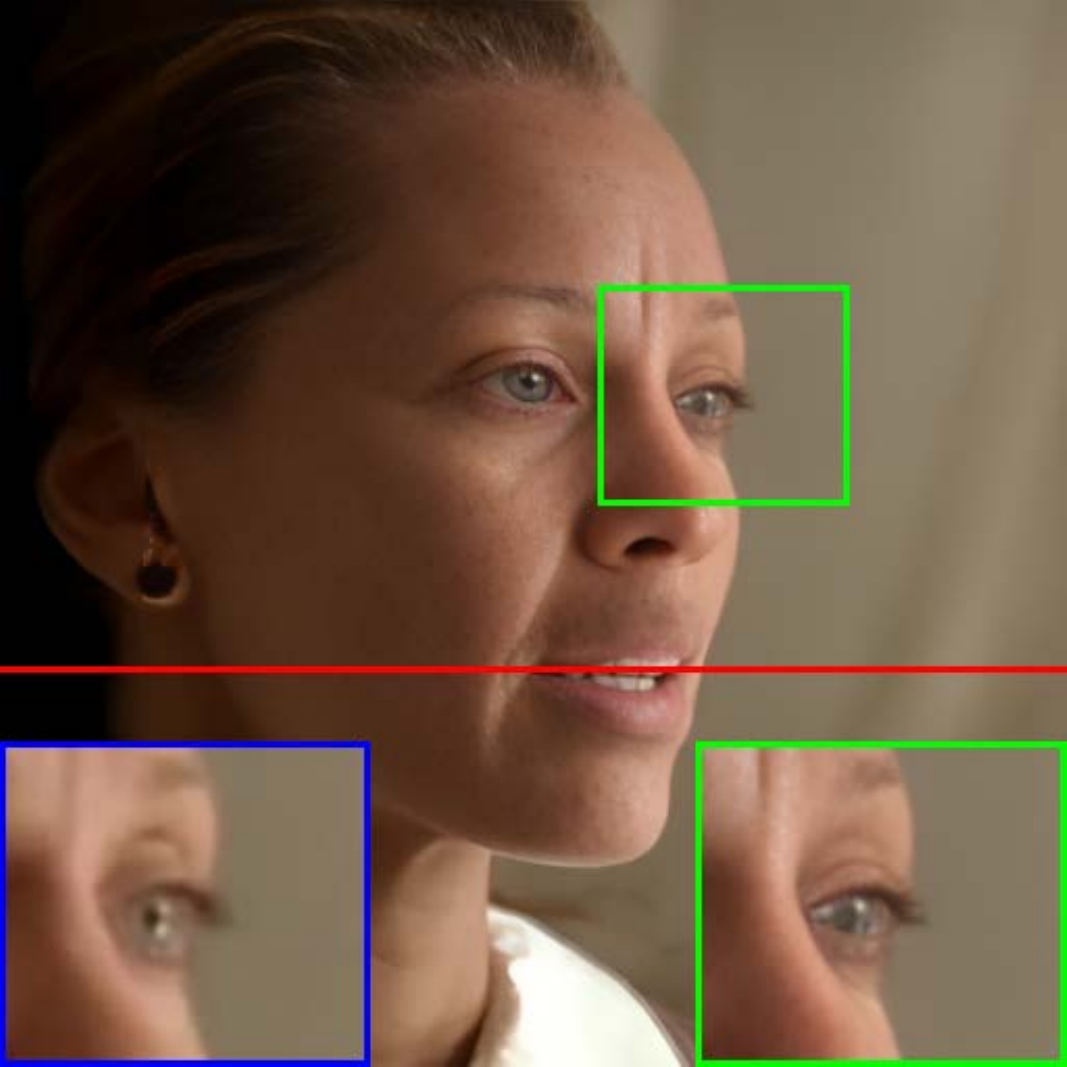} &
    \includegraphics[width=\swexp]{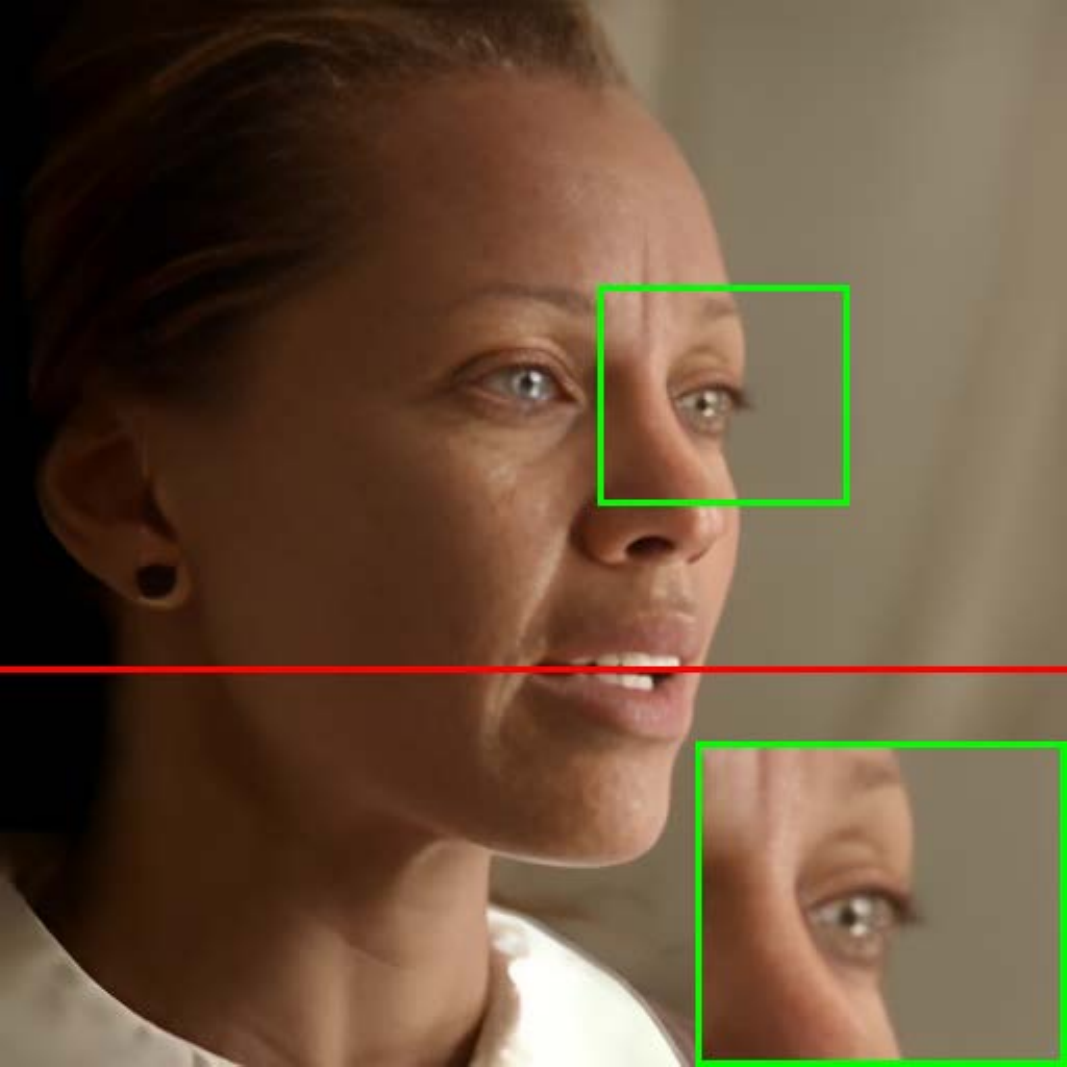} &
    \includegraphics[width=\swexp]{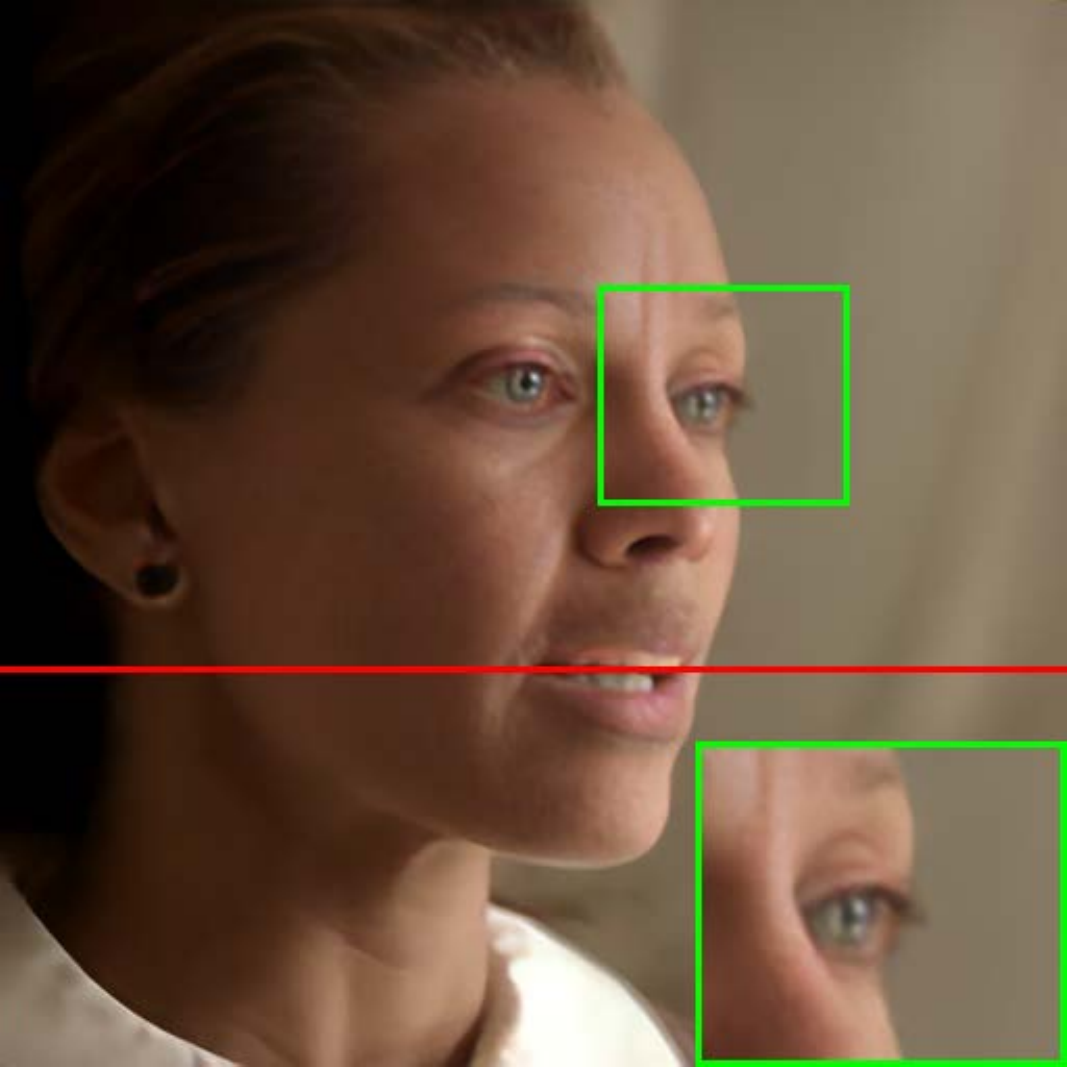} &
    \includegraphics[width=\swexp]{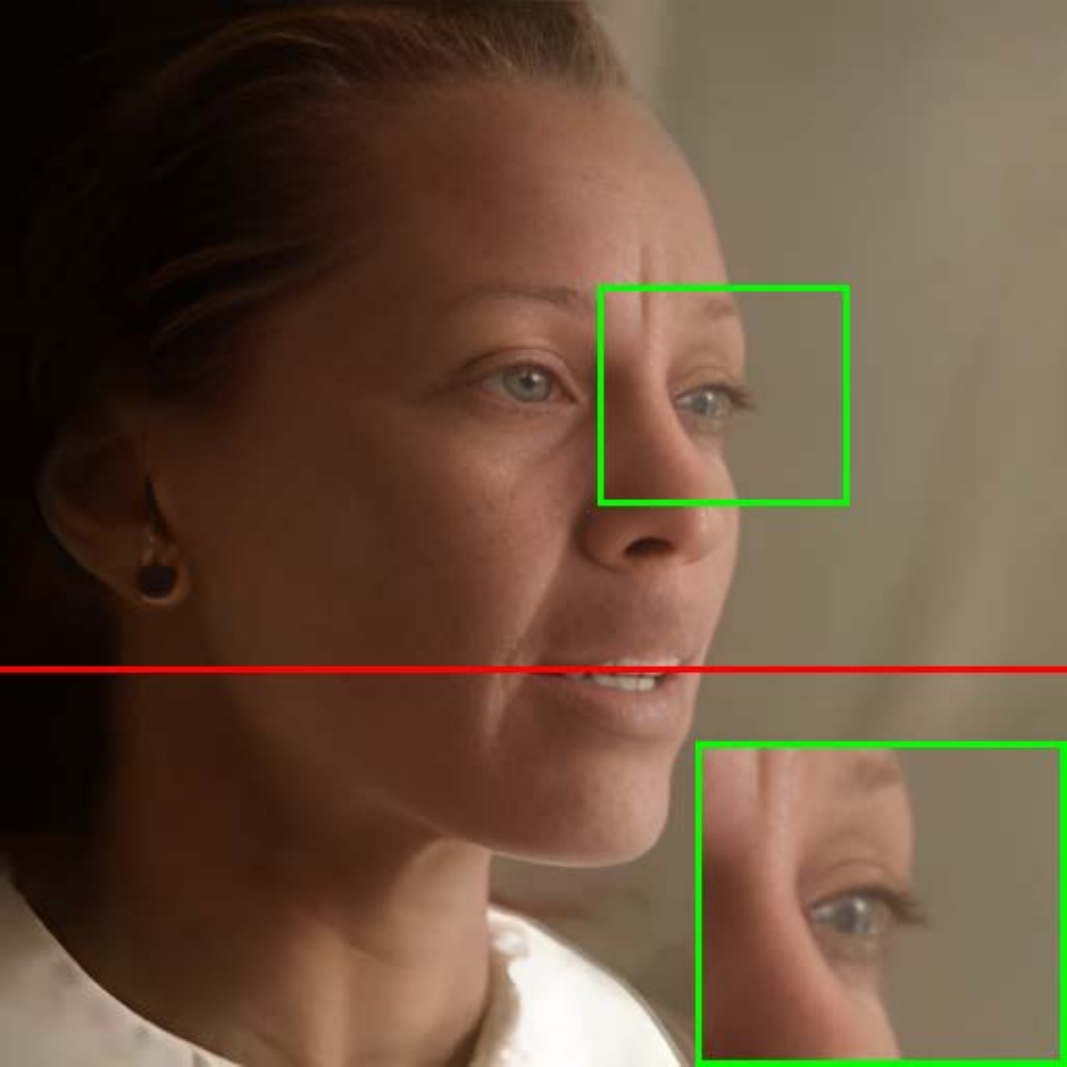} &
    \includegraphics[width=\swexp]{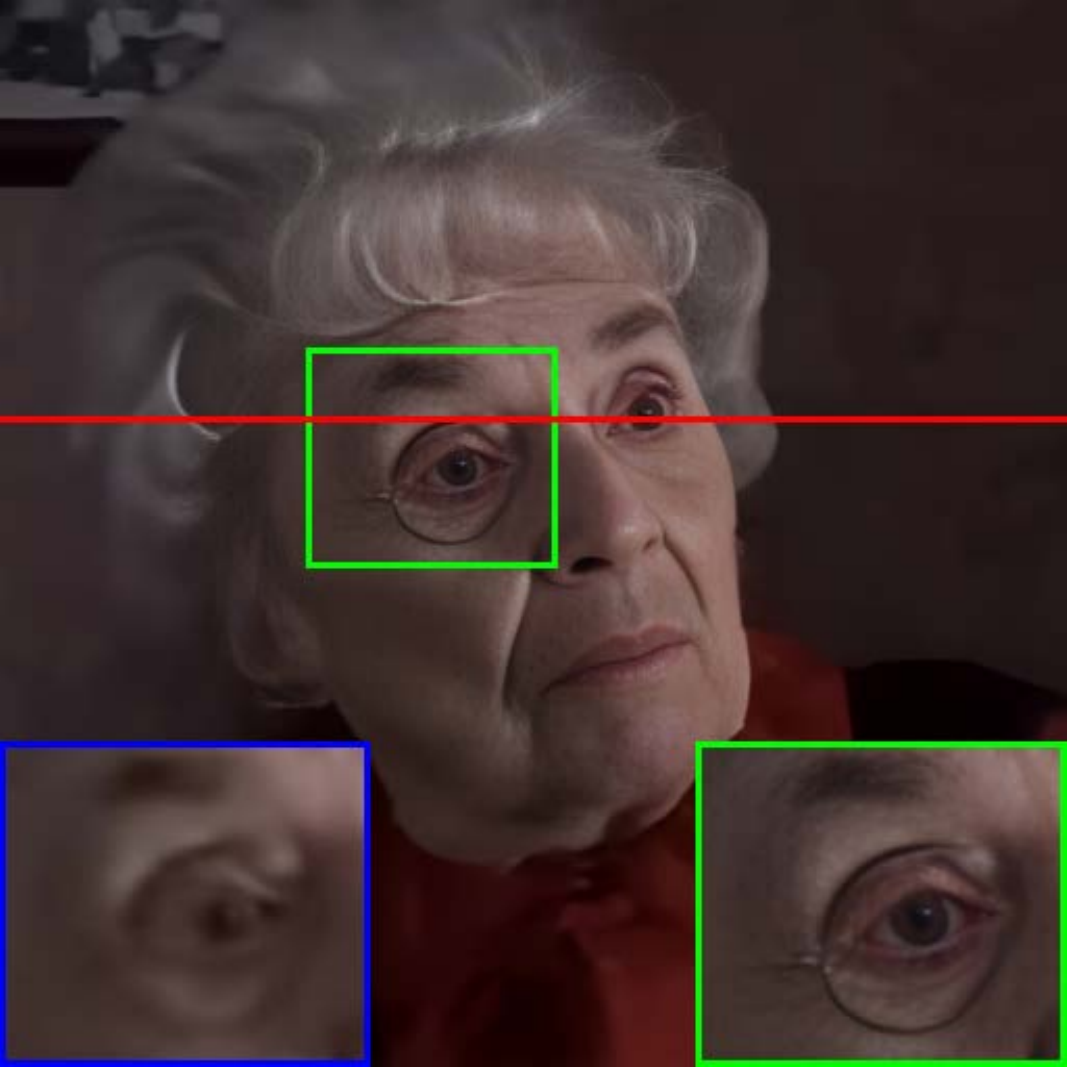} &
    \includegraphics[width=\swexp]{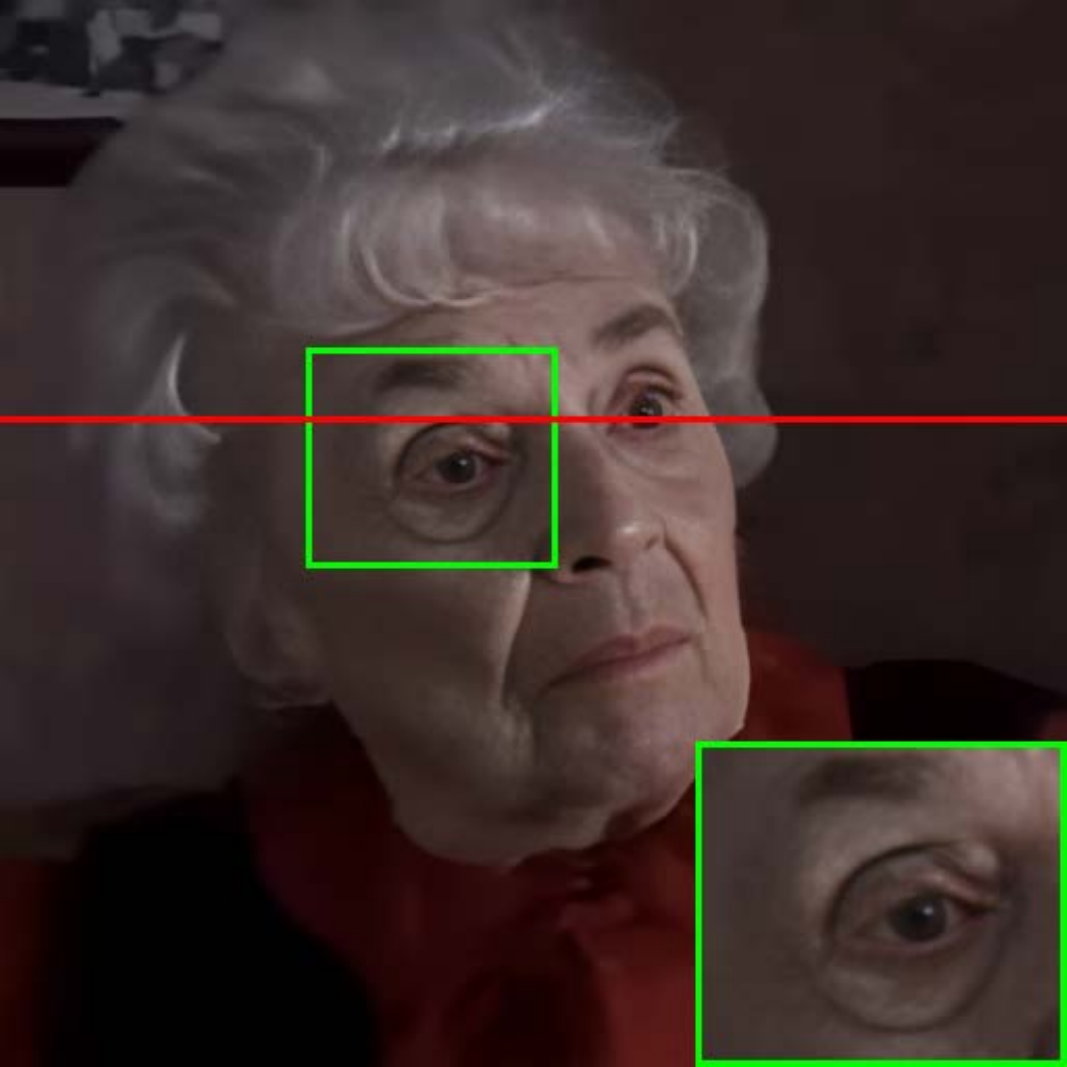} &
    \includegraphics[width=\swexp]{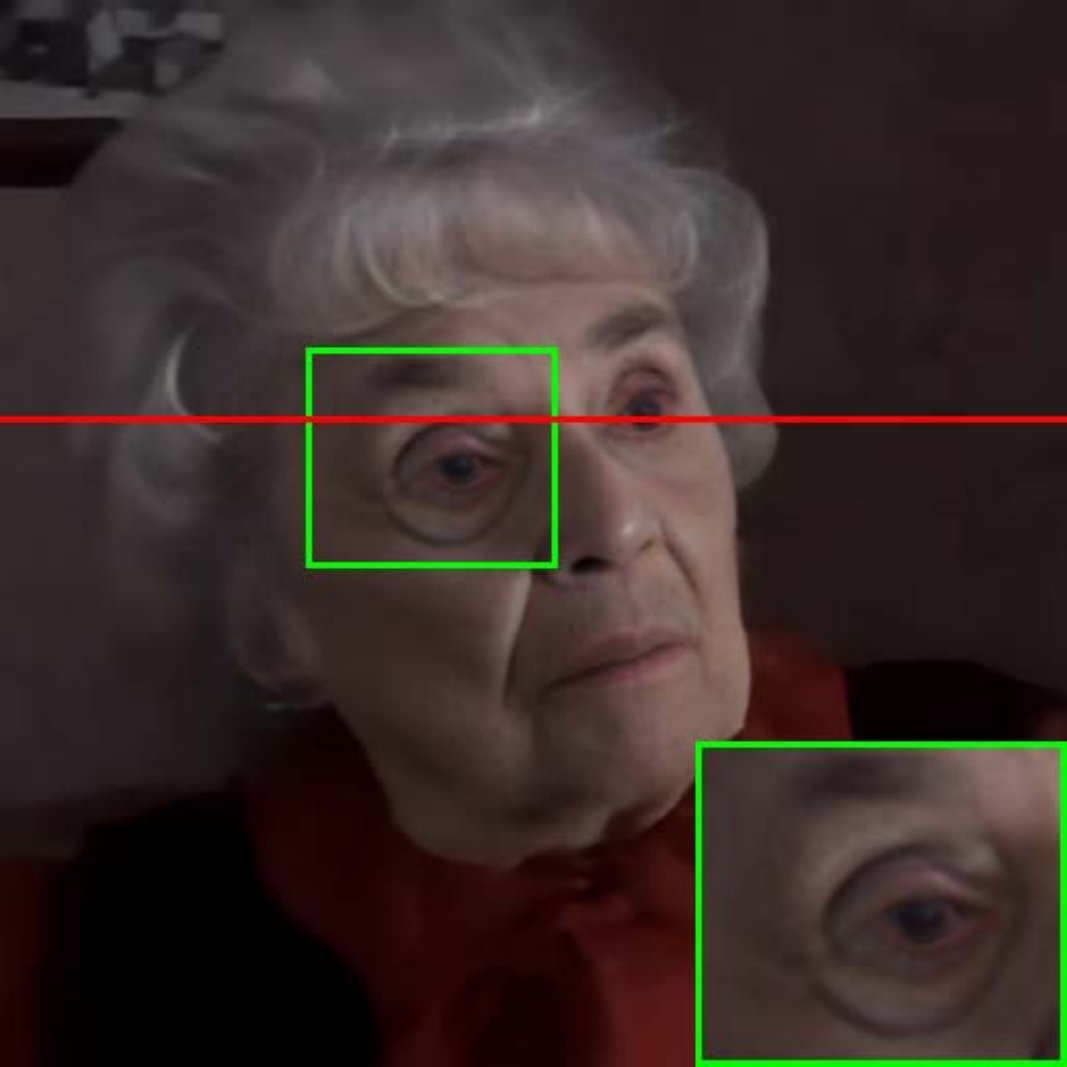} &
    \includegraphics[width=\swexp]{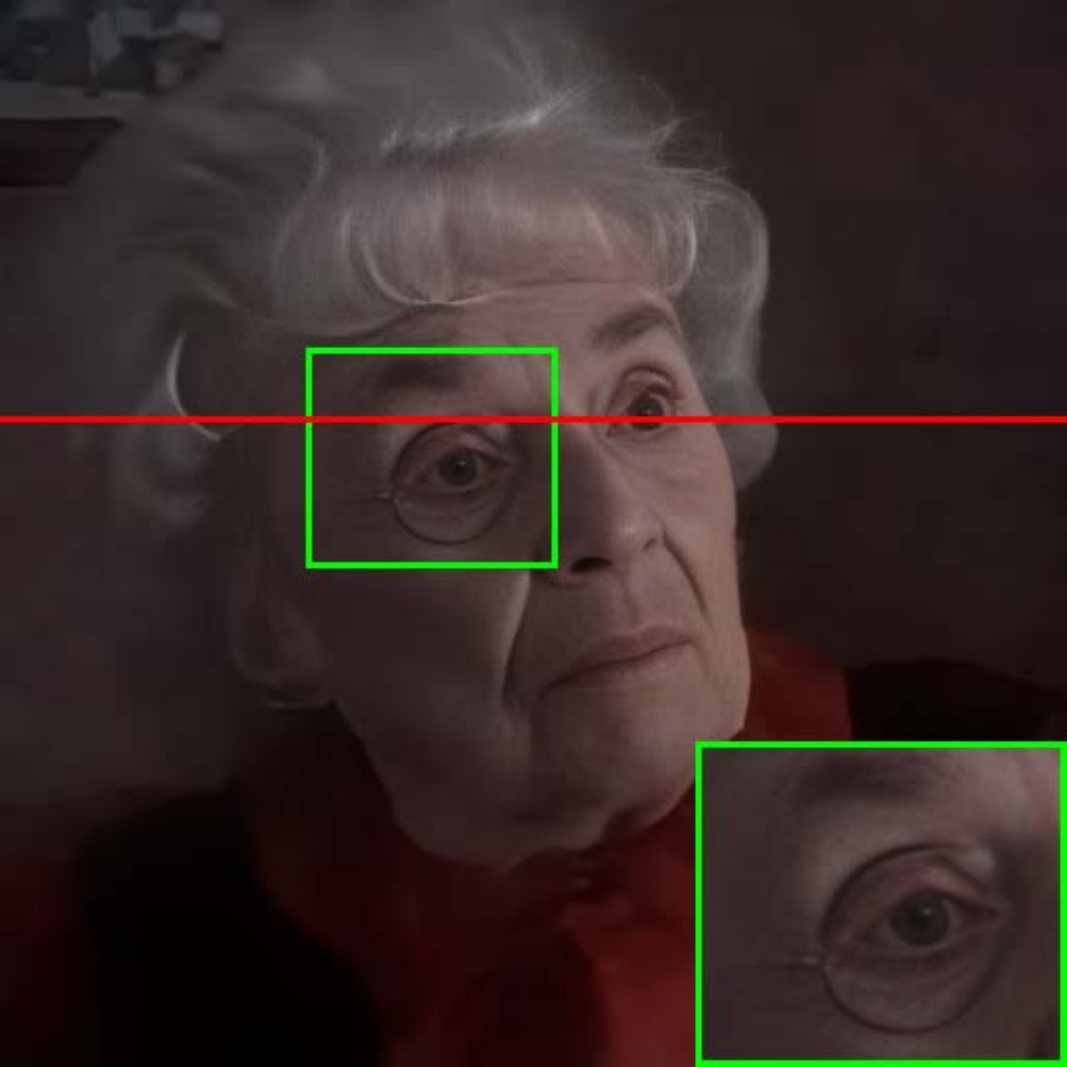} \\
    \multicolumn{8}{c}{(b) Jitters visualization of the red slice in (a)} \\
    \includegraphics[width=\swexp]{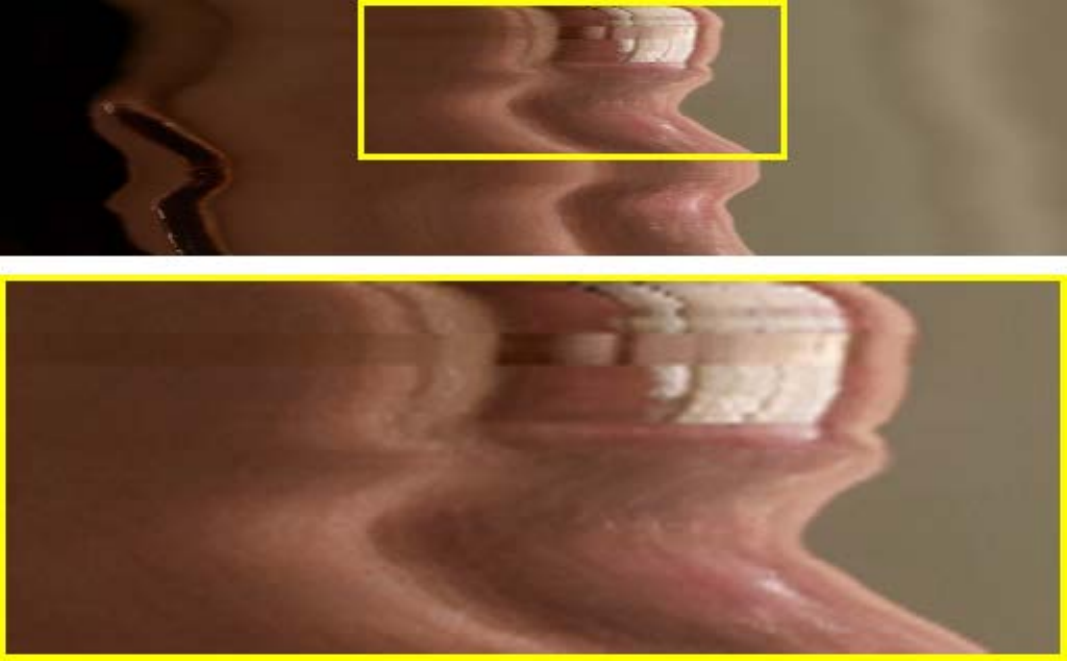} &
    \includegraphics[width=\swexp]{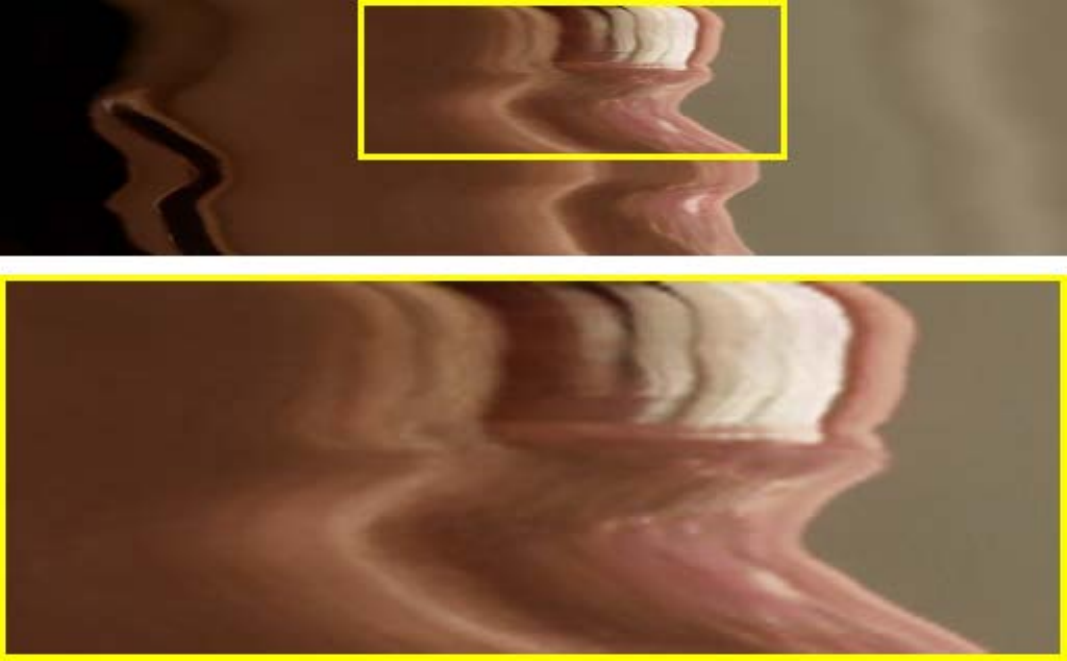} &
    \includegraphics[width=\swexp]{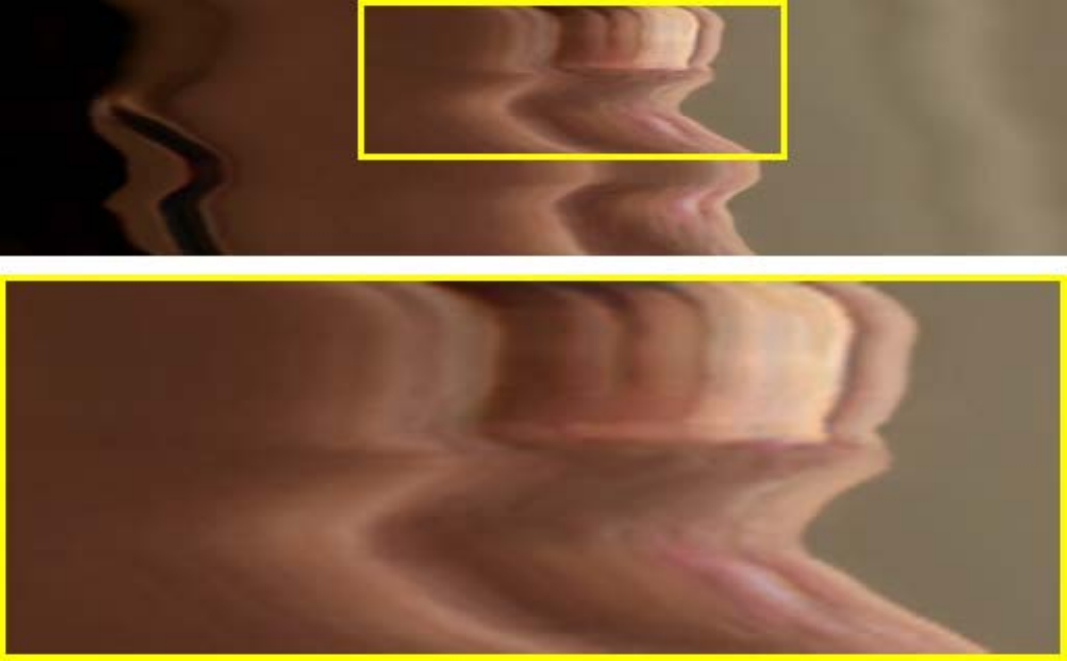} &
    \includegraphics[width=\swexp]{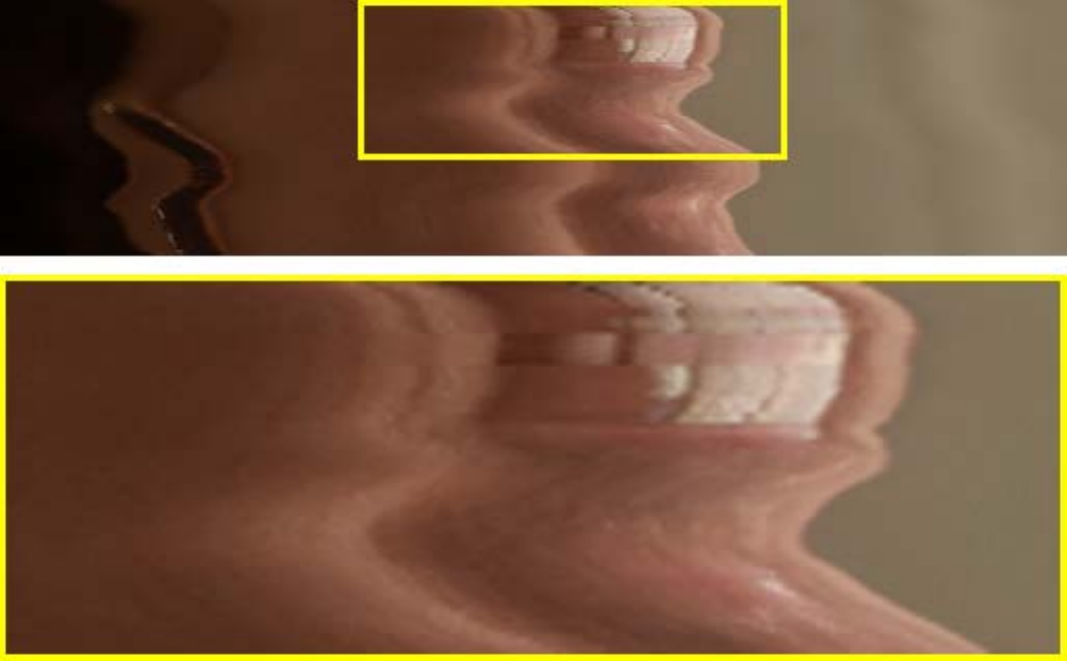} &
    \includegraphics[width=\swexp]{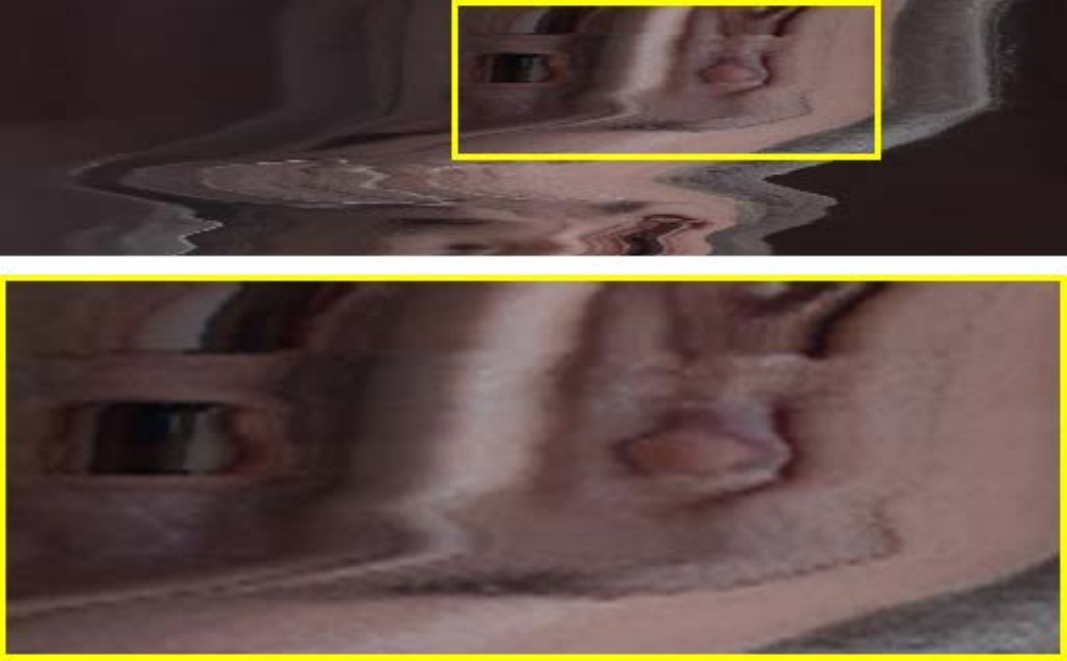} &
    \includegraphics[width=\swexp]{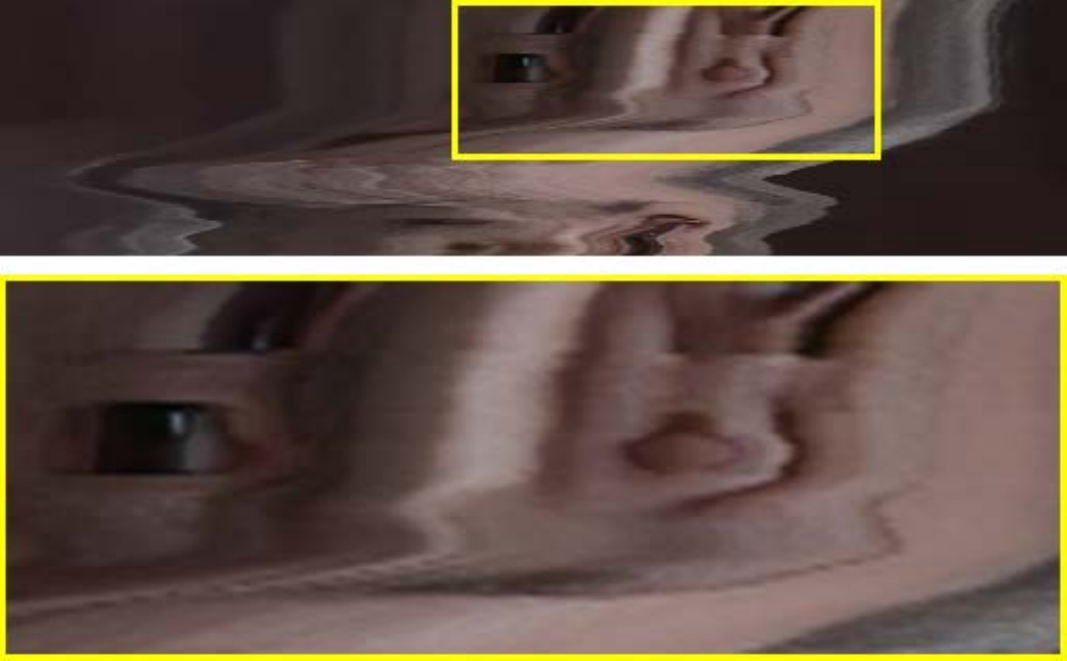} &
    \includegraphics[width=\swexp]{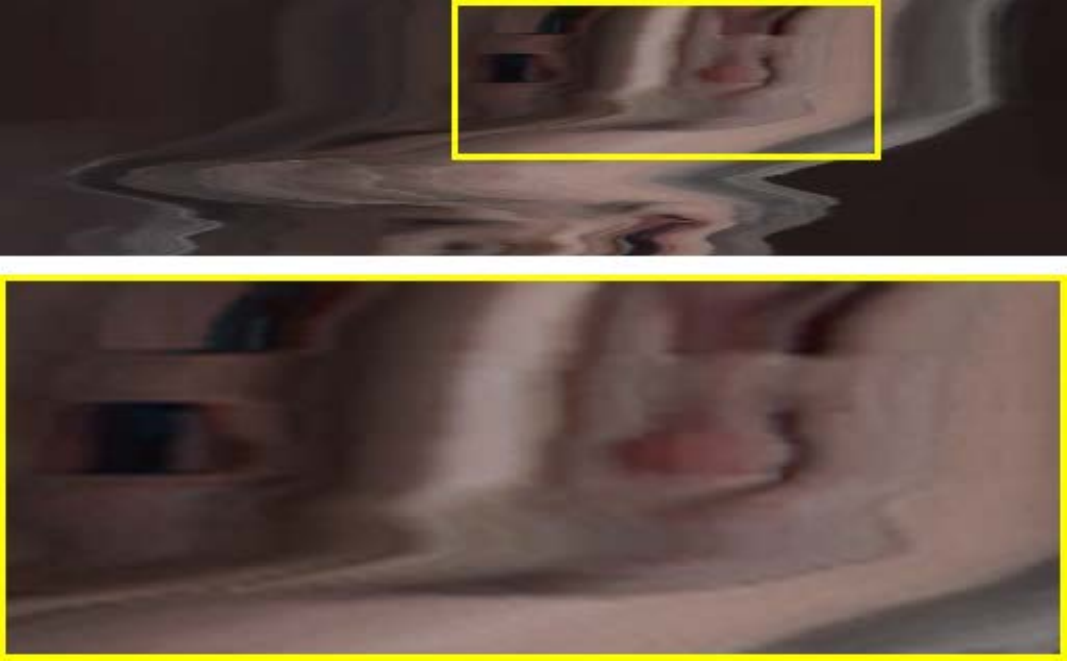} &
    \includegraphics[width=\swexp]{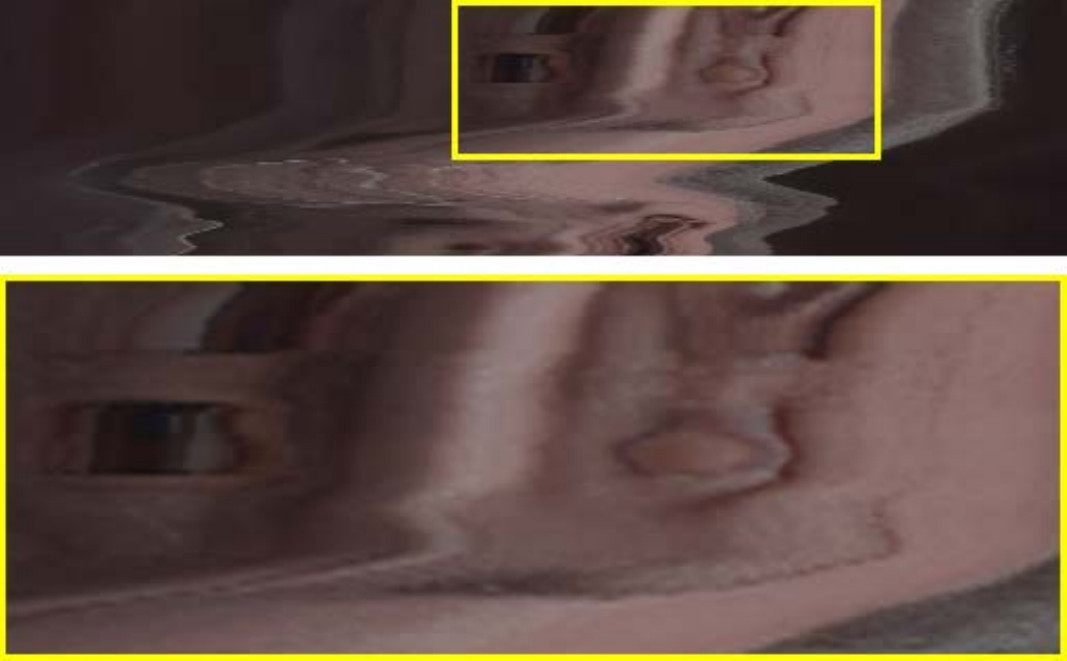} \\
    \multicolumn{8}{c}{(c) Noise-shape flickers visualization} \\
    \includegraphics[width=\swexp]{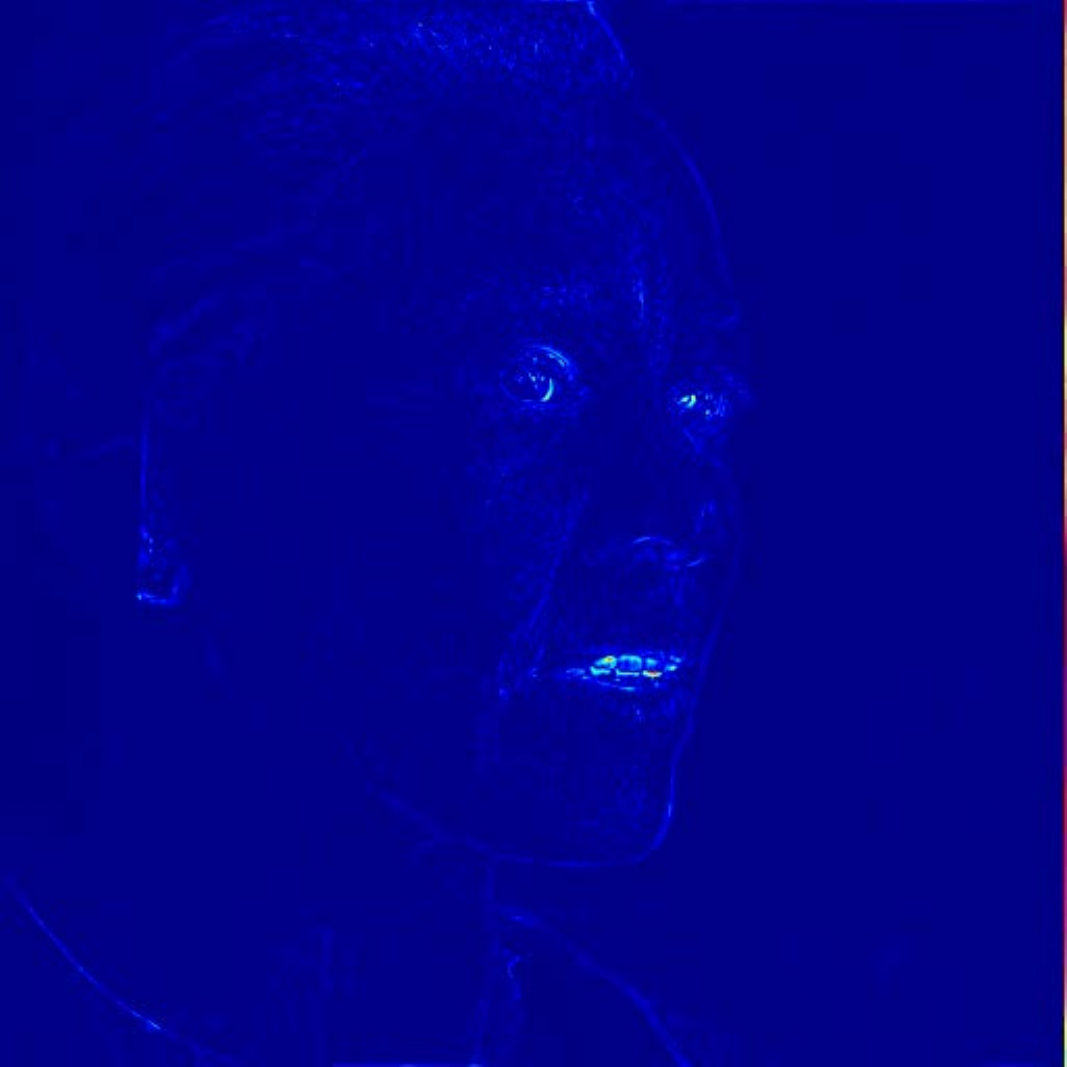} &
    \includegraphics[width=\swexp]{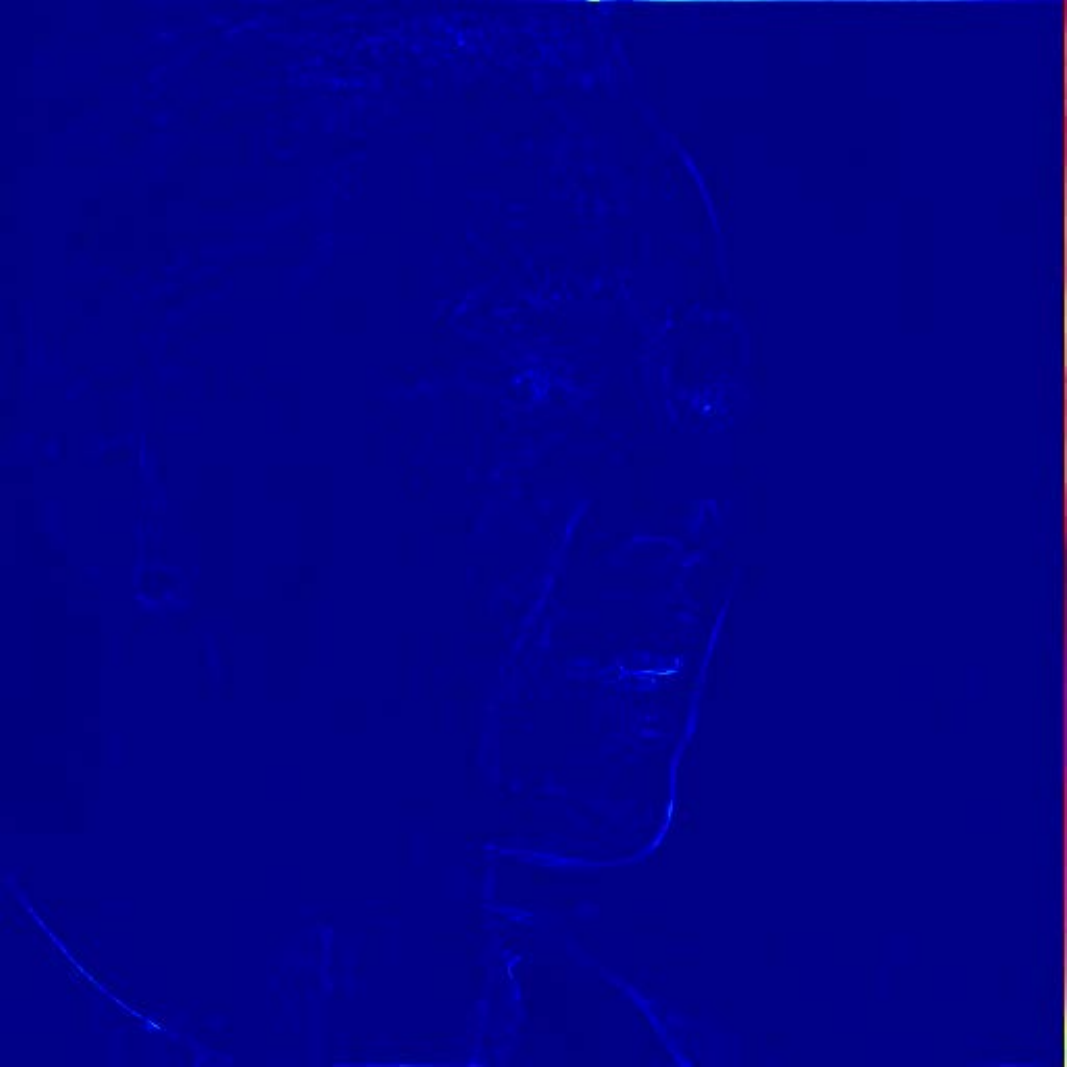} &
    \includegraphics[width=\swexp]{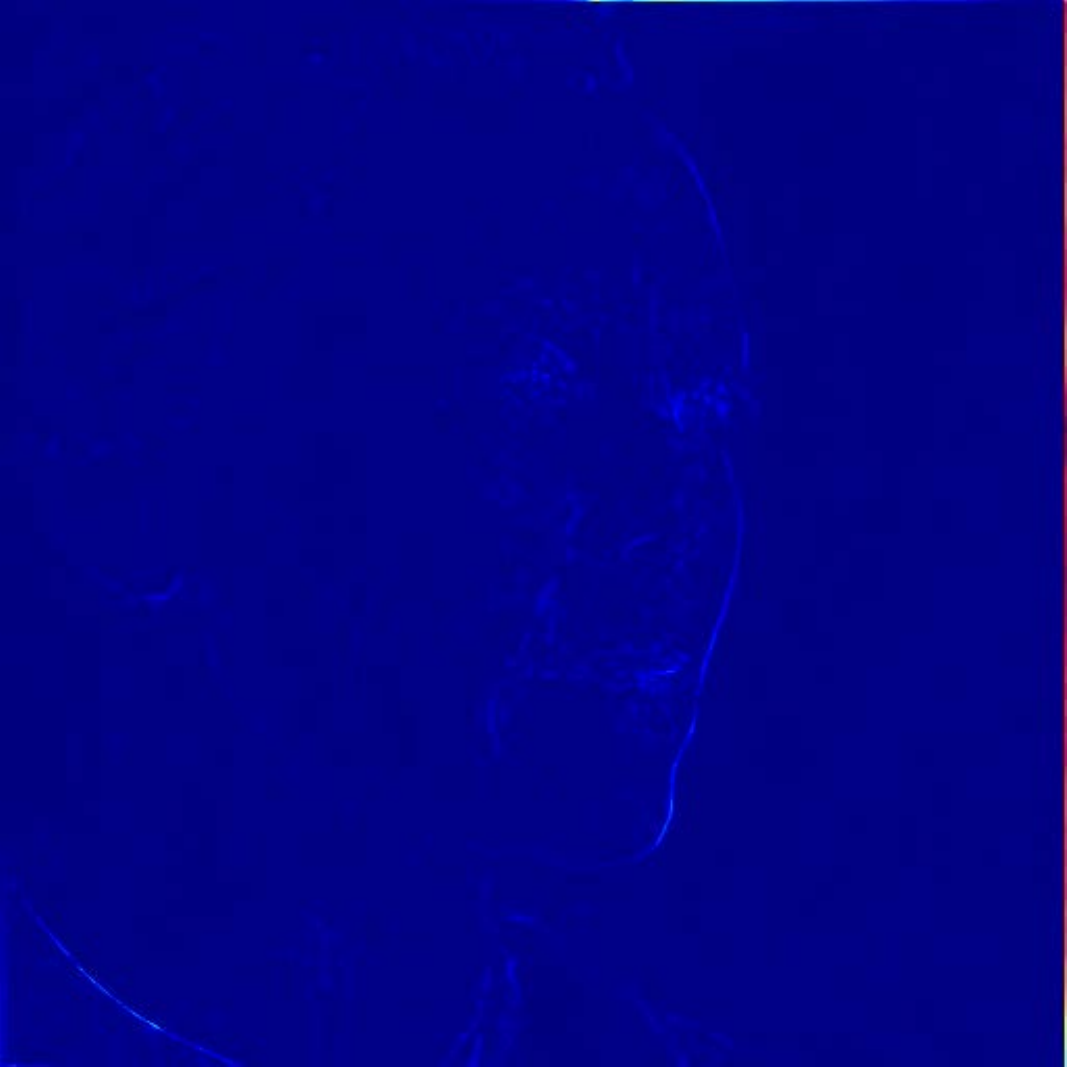} &
    \includegraphics[width=\swexp]{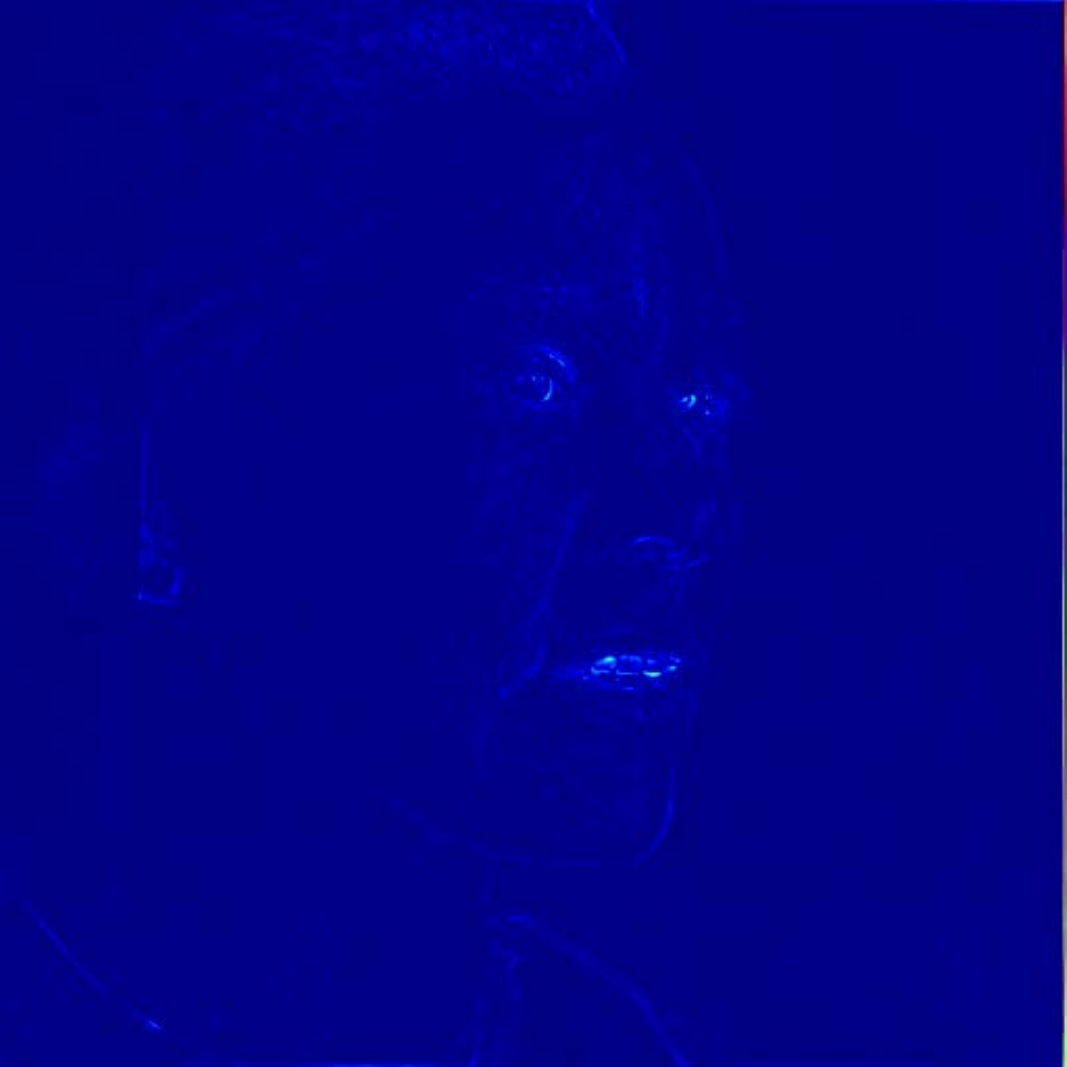} &
    \includegraphics[width=\swexp]{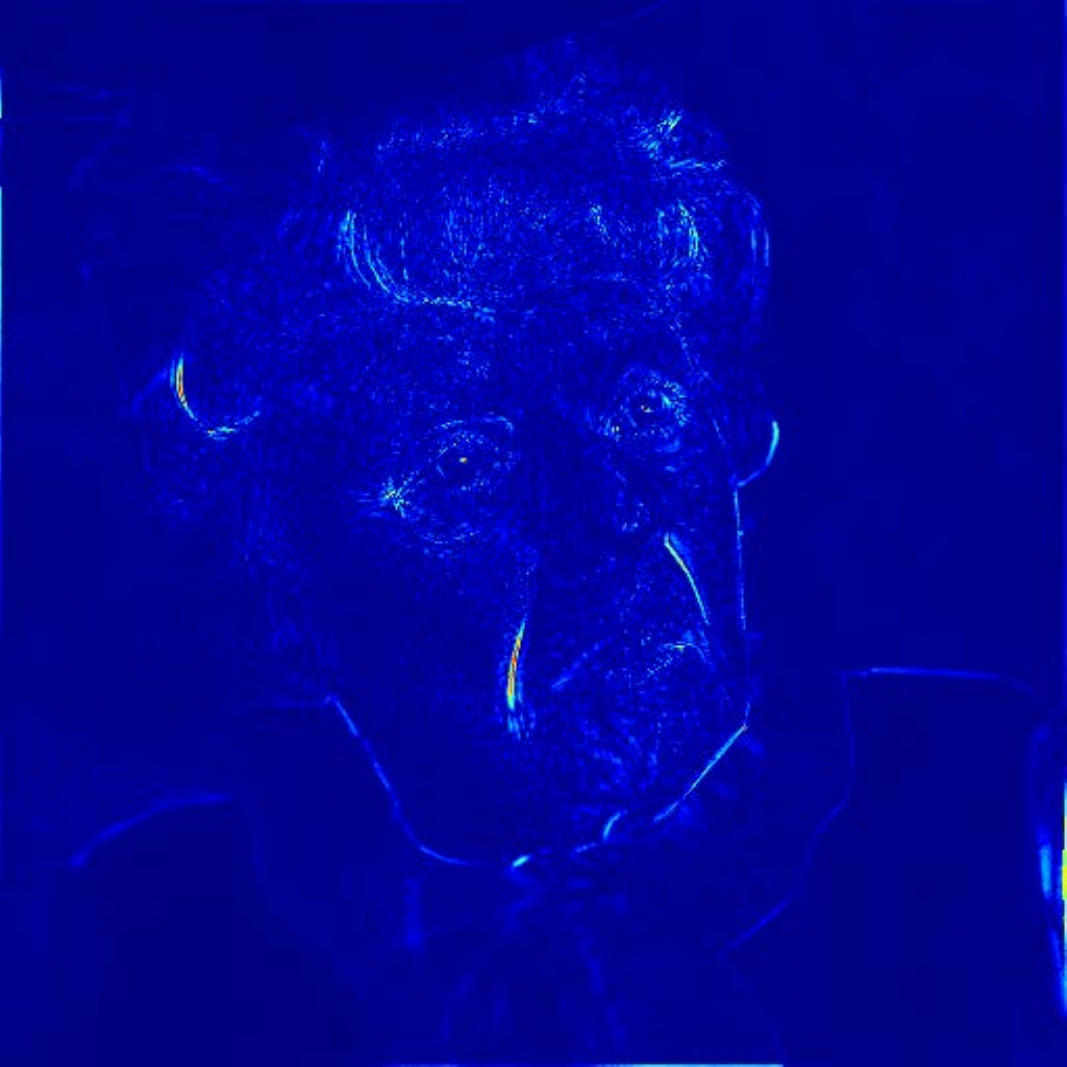} &
    \includegraphics[width=\swexp]{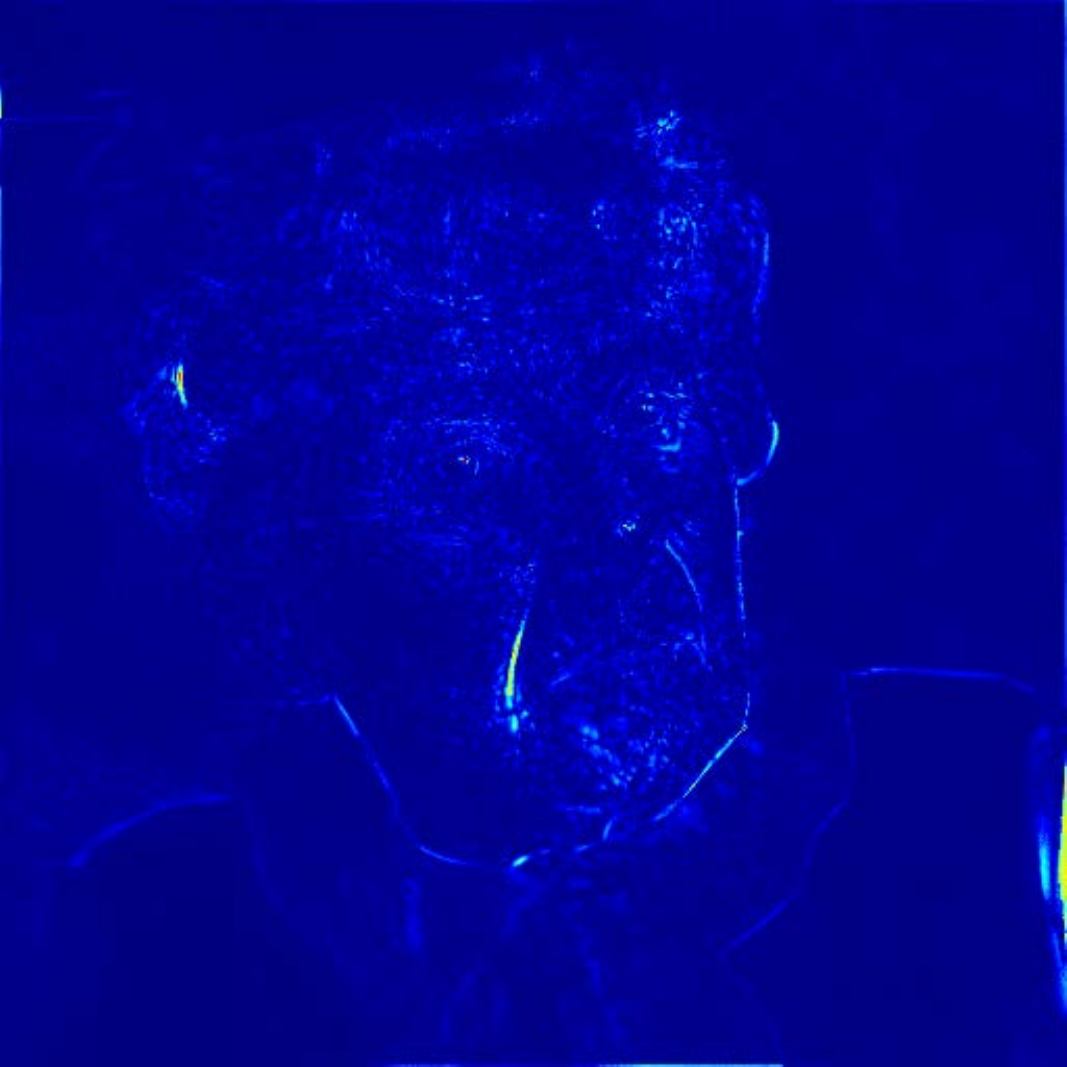} &
    \includegraphics[width=\swexp]{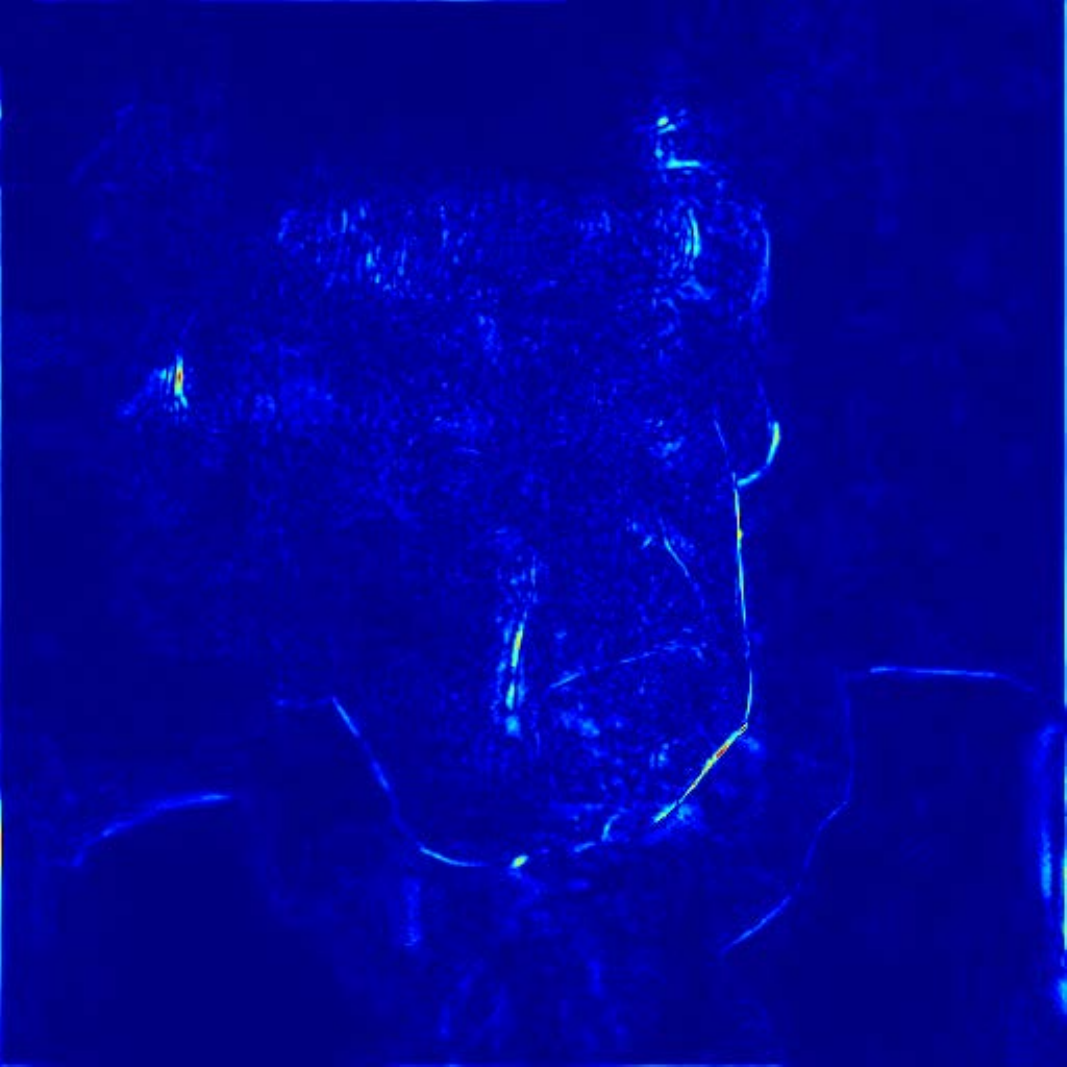} &
    \includegraphics[width=\swexp]{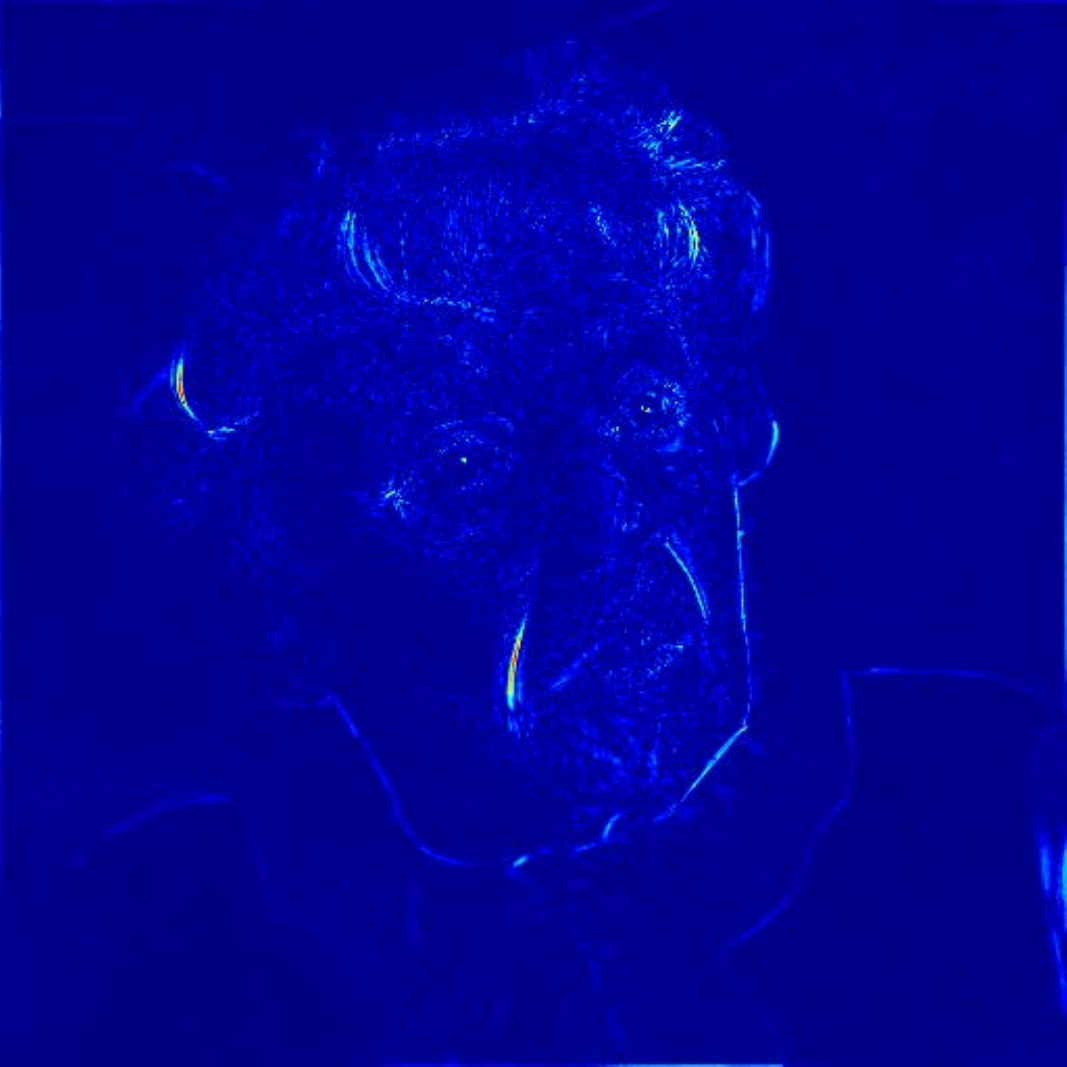}\\
    GFP-GAN~\cite{wang2021towards} & + \textbf{Ours} & + DVP~\cite{lei2020blind} & +~\cite{Lei_2023_CVPR} & GFP-GAN~\cite{wang2021towards} & + \textbf{Ours} & + DVP~\cite{lei2020blind} & +~\cite{Lei_2023_CVPR}
\end{tabular}
\caption{\textbf{Visualized Results of Our Proposed Method and Existing Related Methods.} \redm{The contents in the \textcolor{blue}{blue box} are the parts from degraded face images, while the contents in the \textcolor{green}{green box} are the corresponding restored results. The contents within the \textit{yellow box} depict the temporal stacking of the \textit{red horizontal lines} in the video, and the last row presents the warping error maps. Our proposed method achieves a better balance between restored quality and temporal consistency, and it can correct errors introduced by the original image-based algorithms, such as the teeth in the second sample and the glass in the fourth sample. 
\textbf{Corresponding videos are in the supplementary materials and recommended to zoom in for a better view.}}
}
\label{fig:flicker_update}
\end{figure*}

\section{Experiments and Analysis}
\label{sec:exp}

\subsection{Synthesis of the Degraded Training Dataset}

While training, the degradations on VFHQ-800 are synthesized with a degrading model~\cite{li2020enhanced,li2018learning,wang2021towards,wang2022restoreformer,wang2023restoreformer++,xie2022vfhq}.
Following \cite{xie2022vfhq}, we synthesize the compression artifacts with FFMPEG, which is closer to the real-world degraded videos.
Therefore, the degrading model adopted in this work is 
\begin{equation}
\mathbf{\hat{I}}_t^d = \{[(\mathbf{\hat{I}}_t^h \otimes \mathbf{k}_\sigma) \downarrow_r + \mathbf{n}_\delta]_{{FFMPEG}_{crf}}\}\uparrow_r.
\label{eq:degrading_model}
\end{equation}
In this degrading model, the $t_{th}$ frame $\mathbf{\hat{I}}_t^h$ in a high-quality video is blurred with a Gaussian blur kernel $\mathbf{k}_\sigma$ with sigma $\sigma$.
Its blurry result will be further downsampled by $r\times$ with a bilinear operation and added with white Gaussian noise $\mathbf{n}_\delta$ with sigma $\delta$.
Then FFMPEG compression is applied to it with a constant rate factor $crf$.
Finally, the downsampled degraded frame will be upsampled back to the same size of $\mathbf{\hat{I}}_t^h$ to attain the required degraded frame $\mathbf{\hat{I}}_t^d$.
To ensure the degraded parameters will not vary severely in a video, frames in a video share the same $\sigma$, $r$, and $crf$. The difference of their $\delta$ will not be large than 0.1.

\subsection{Experimental Settings}

Since our proposed method is an extension of several face image restoration algorithms, we follow most of the settings of the original face image restoration models, such as their input size $512 \times 512$, optimizer, and their loss weighting factors, including $\lambda_{1}$, $\lambda_{p}$, and $\lambda_{a}$. However, we also make some adjustments to adapt them to a new block (TCN) and new data. In this paper, we decay the learning rate of the original face image restoration methods by 10. The batch size is 8. In each iteration, our model is trained with 3 frames, while the length of the videos used for inference can vary. $\lambda_{t}$ for temporal loss is 10. Moreover, in the degrading model, we randomly sample $\sigma$, $r$, $\delta$, and $crf$ from $\lbrace1:15\rbrace$, $\lbrace0.8:8\rbrace$, $\lbrace0:5\rbrace$, and $\lbrace10:25\rbrace$, respectively.

\begin{table*}[!t]
\centering
\caption{
\redm{
\textbf{Quantitative Results of Our Proposed Method and Existing Related Methods, conducted on RestoreFormer++~\cite{wang2023restoreformer++}, CodeFormer~\cite{zhou2022towards} and GFP-GAN~\cite{wang2021towards}, and evaluated on RFV-LQ.} Our proposed method outperforms DVP~\cite{lei2020blind} in preserving the restored quality of the original image-based methods, as indicated by lower FID and component-based FID scores. In comparison to ~\cite{Lei_2023_CVPR}, our method achieves superior jitter reduction, as evidenced by lower MSI scores. Additionally, our method demonstrates exceptional efficiency.
}}
\resizebox{\linewidth}{!}{
\begin{tabular}{c|c|ccccc|cc|c}
\hline
FIR & Method & FID~$\downarrow$ & FID-Eyes~$\downarrow$ & FID-Nose~$\downarrow$ & FID-Mouth~$\downarrow$ & FID-Hair~$\downarrow$ & MSI~$\downarrow$ & Warping Error~$\downarrow$ & Time/s~$\downarrow$ \\
\hline
\multirow{4}*{RestoreFormer++~\cite{wang2023restoreformer++}} 
& Original
& 66.1725 
& 53.9941 
& 31.7954
& 43.5653 
& 52.3961 
& 6.873 
& 0.001058 
& \redm{19.9737} \\
& \textbf{+ Ours} 
& 71.0050
& 64.6849
& 47.9572
& 62.6286
& 63.8600
& 6.1823
& 0.000903  
& 38.7133\\
& + DVP~\cite{lei2020blind} 
& 83.9701
& 79.5201 
& 59.070 
& 85.3488
& 79.6759
& 5.3372
& 0.000691 
& 335.8285 \\
& + \cite{Lei_2023_CVPR}
& 72.2167
& 59.6197
& 33.1802
& 67.1150
& 58.2315 
& 6.8641
& 0.000677
& 147.3533 \\
\hline
\multirow{4}*{CodeFormer~\cite{zhou2022towards}} 
& original
& 69.6650 
& 49.1728 
& 32.2229 
& 44.6944 
& 51.6947 
& 6.9899 
& 0.001035 
& \redm{19.9276} \\
& \textbf{+ Ours} 
& 73.9761
& 59.4263 
& 34.250
& 53.6393
& 75.4900
& 6.4907
& 0.000875 
& 36.0371\\
& + DVP~\cite{lei2020blind} 
& 83.8865
& 64.8820
& 48.1960
& 69.9906
& 76.6627
& 5.3324
& 0.000682 
& 335.7824 \\
& + \cite{Lei_2023_CVPR}
& 74.444
& 54.3990
& 32.9940
& 69.4436
& 59.9358
& 7.5328
& 0.000635 
& 147.3072 \\
\hline
\redm{\multirow{4}*{GFP-GAN~\cite{wang2021towards}}}
& \redm{original}
& 70.3217 & 50.4729 & 33.9919 & 47.9370 & 59.3655 & 5.7917 & 0.000920 & 16.6350 \\
& \redm{\textbf{+ Ours} }
& 74.3139 & 56.6721 & 39.0424 & 60.7921 & 68.4474 & 5.5574 & 0.000883 & 33.0217 \\
& \redm{+ DVP~\cite{lei2020blind}} 
& 80.2662 & 60.8234 & 41.2748 & 69.5407 & 76.8258 & 5.0857 & 0.000695 & 332.4898 \\
& \redm{+ \cite{Lei_2023_CVPR}}
& 74.7919 & 58.3820 & 35.9494 & 77.5273 & 66.1199 & 6.1726 & 0.000619 & 144.0146 \\

\hline
\end{tabular}
}
\label{tab:tcn}
\end{table*}

\subsection{Comparison with the State-of-the-art Methods}

\redm{
We assess the effectiveness of our proposed method on RFV-LQ and compare it with two state-of-the-art methods, DVP~\cite{lei2020blind} and the method proposed in~\cite{Lei_2023_CVPR}. Experiments are conducted based on GFP-GAN~\cite{wang2021towards}, RestoreFormer++\cite{wang2023restoreformer++}, and CodeFormer\cite{zhou2022towards}, and the corresponding qualitative and quantitative results are presented in Fig.~\ref{fig:flicker_update} and TABLE~\ref{tab:tcn}, respectively.
}

\redm{
The quantitative scores in Table~\ref{tab:tcn} demonstrate that both our proposed method and existing ones can mitigate jitters and flickers in the results restored from the original GFP-GAN~\cite{wang2021towards}, RestoreFormer++~\cite{wang2023restoreformer++}, and CodeFormer~\cite{zhou2022towards}. However, this improvement comes at the cost of restored quality, as evidenced by lower MSI and Warping Error scores but higher FID and Component FIDs scores.
Specifically, DVP~\cite{lei2020blind}, while performing better in MSI, exhibits a significant drop in restored quality and fails to restore facial details. Its restored results in Fig.~\ref{fig:flicker_update} are relatively blurry compared to the other methods, particularly in the depiction of teeth in the first sample. Moreover, although the method proposed in ~\cite{Lei_2023_CVPR} performs better in Warping Error and maintains a higher restored quality, it shows almost no improvement in MSI, with jitters in its restored results remaining as obvious as those of the original image-based methods.
In contrast, our proposed method achieves improvements in both MSI and Warping Error while better maintaining restored quality. Additionally, unlike DVP~\cite{lei2020blind} and ~\cite{Lei_2023_CVPR}, which function as post-processing operations, our proposed method, a plug-and-play component for image-based algorithms, progressively enhances the temporal consistency of the image-based algorithms while they restore the degraded face video frame by frame. This advantage allows our proposed method to correct the restored results of the original image-based algorithms, such as the teeth (the second sample in Fig.~\ref{fig:flicker_update}) and glasses (the fourth sample in Fig.~\ref{fig:flicker_update}) that are not present in the degraded face (the content in the \textcolor{blue}{blue box}) but appear in the restored results of the image-based algorithms. After processing with our proposed method, the teeth and glasses are revised, and the results are more aligned with the degraded faces, whereas DVP~\cite{lei2020blind} and ~\cite{Lei_2023_CVPR} contain remnants.
}

\redm{
Moreover, our method also excels in efficiency. As shown in Table~\ref{tab:tcn}, our method takes no more than 39 seconds to process a degraded face video with 100 frames using an NVIDIA Tesla T4. 
Only about an additional 17 seconds are needed by our newly added TCN and alignment smoothing compared to the original image-based methods. In contrast, the compared methods, DVP~\cite{lei2020blind} and \cite{Lei_2023_CVPR}, respectively require over 300 seconds and over 100 seconds to process the same degraded video.
Therefore, our proposed method offers a superior trade-off between restoration effectiveness and efficiency compared to previous methods.
}

\begin{table*}[!t]
\centering
\caption{
\textbf{Ablation studies on the setting of $k$ and resolution of the input for TCN based on RestoreFormer++~\cite{wang2023restoreformer++}.} 
}
\begin{tabular}{c|c|ccccc|cc}
\hline
Resolution & k & FID~$\downarrow$ & FID-Eyes~$\downarrow$ & FID-Nose~$\downarrow$ & FID-Mouth~$\downarrow$ & FID-Hair~$\downarrow$ & MSI~$\downarrow$ & Warping Error~$\downarrow$ \\
\hline
\multirow{4}*{256} & 0
& 70.9131
& 64.6652 
& 48.6664
& 62.5615
& 65.9764
& 6.4637
& 0.000922 \\
& 1
& 70.9719
& 64.6719
& 48.7878
& 62.6777
& 65.9858
& 6.2507
& 0.000909 \\
& 3
& 71.0050
& 64.6849
& 47.9572
& 62.6286
& 63.8600
& \textbf{6.1823}
& \textbf{0.000903} \\
& 5
& 71.0627
& 64.9689
& 48.3008
& 62.1979
& 66.2258
& 6.1842
& 0.000904 \\
\hline
128 & 3 
& 73.7823
& 67.2188
& 41.2242
& 66.8583
& 67.3533
& 6.3541
& 0.000906 \\
512 & 3
& 70.3445
& 64.010 
& 58.1716
& 62.3108
& 59.4503
& 6.6841
& 0.000927 \\
\hline
\end{tabular}
\label{tab:ablation}
\end{table*}

\renewcommand{\tabcolsep}{.5pt}
\begin{figure*}
\begin{minipage}{0.18\linewidth}
  \centerline{\includegraphics[width=1\linewidth]{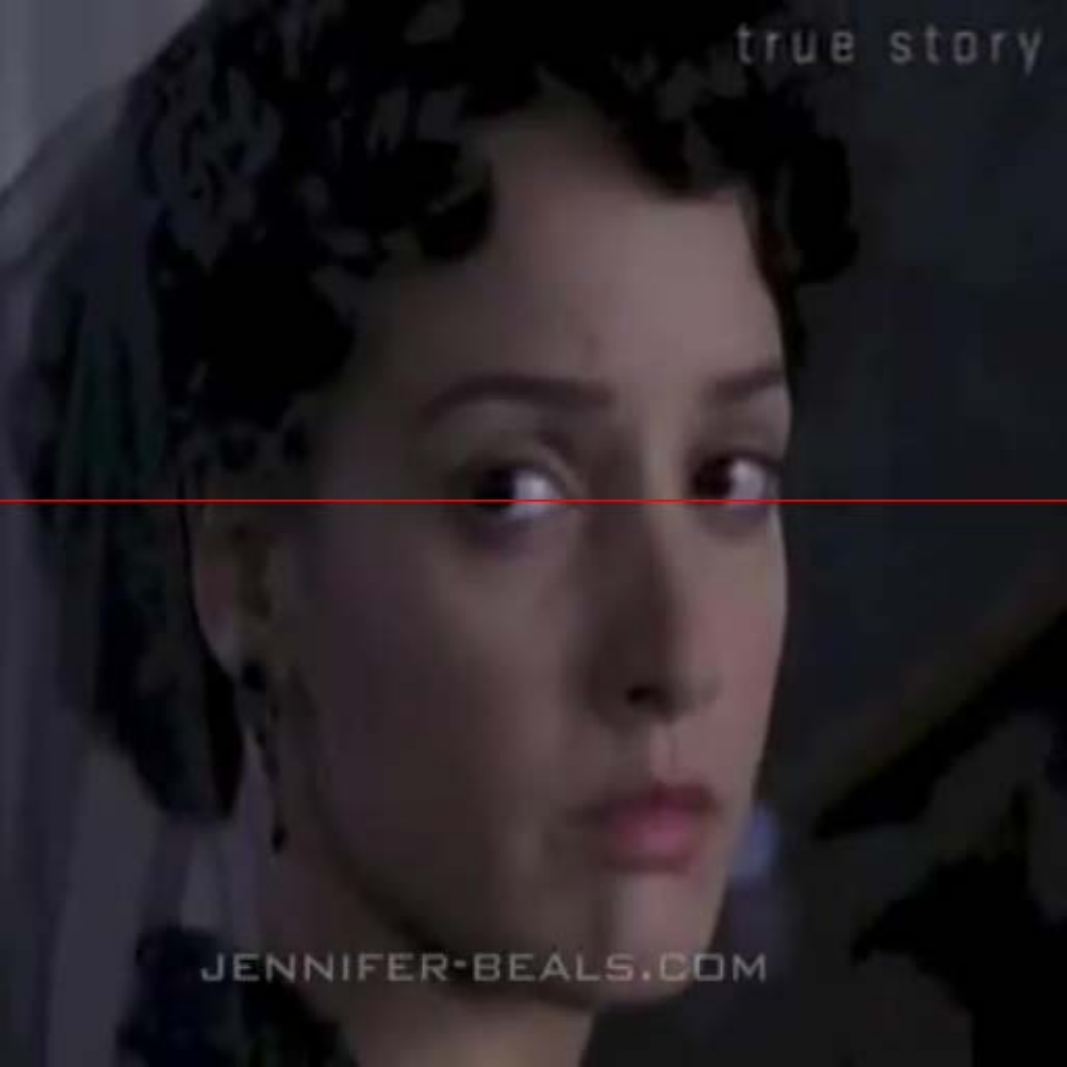}}
  \centerline{(a) Image}
\end{minipage}
\hfill
\begin{minipage}{.90\linewidth}
  \begin{tabular}{ccc}
  \includegraphics[width=\swablscale,angle=0]{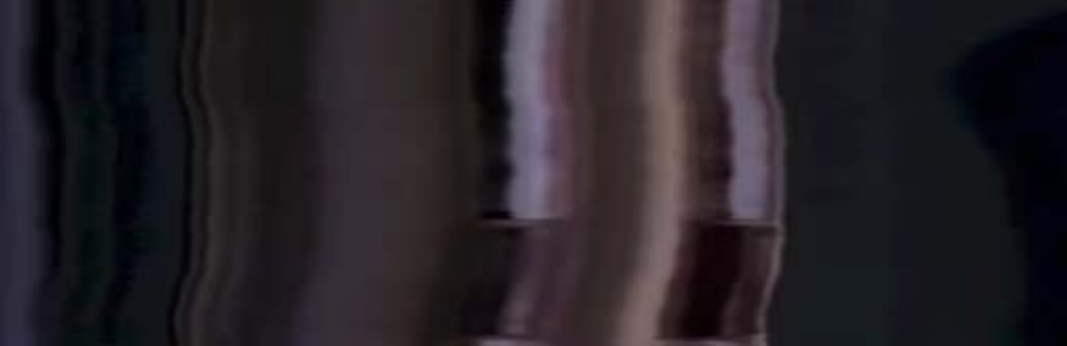} &
  \includegraphics[width=\swablscale,angle=0]{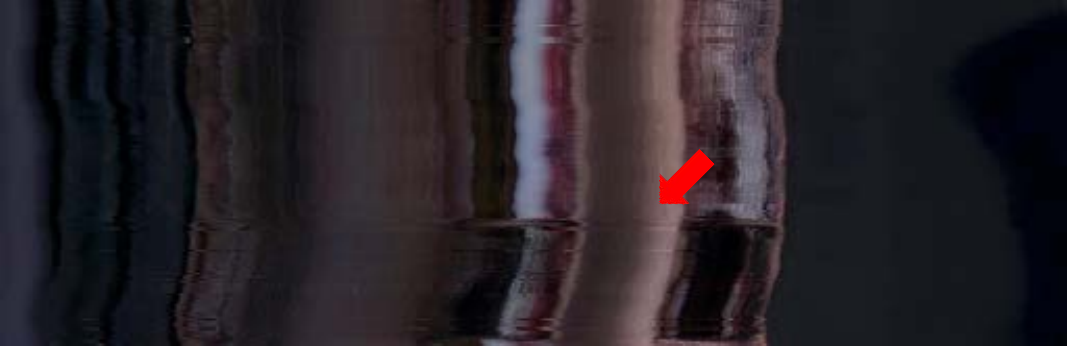} &
  \includegraphics[width=\swablscale,angle=0]{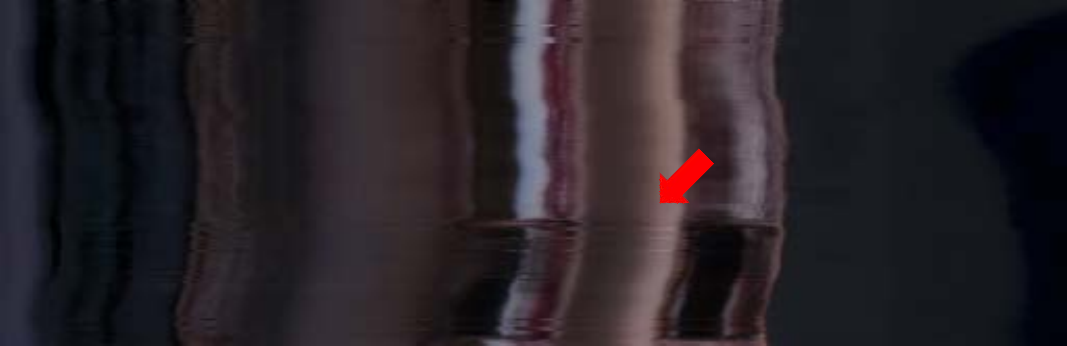} \\
  (b) Input & (c) \scriptsize RestoreFormer++~\cite{wang2023restoreformer++} & (d) $k=0$ \\
  \includegraphics[width=\swablscale,angle=0]{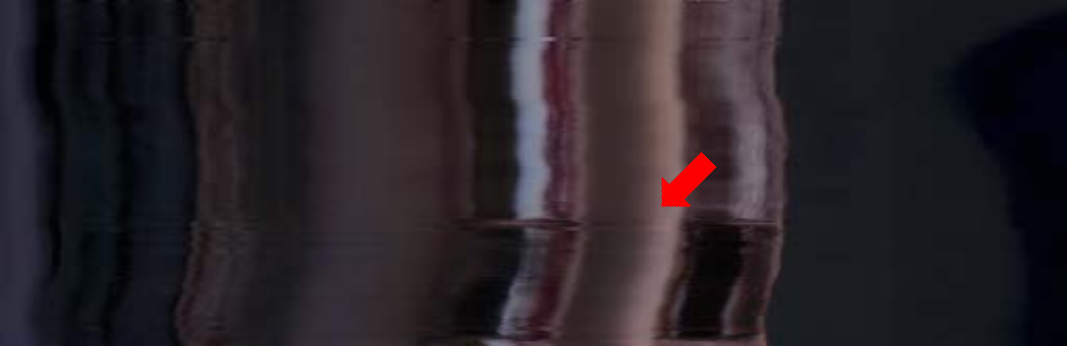} &
  \includegraphics[width=\swablscale,angle=0]{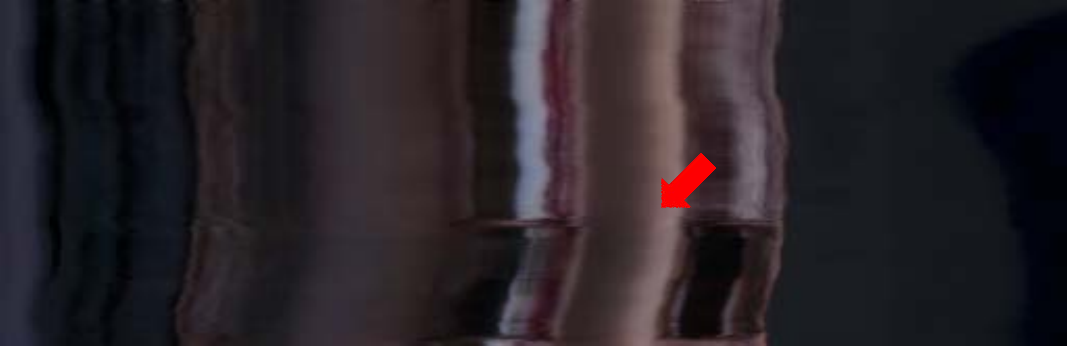} &
  \includegraphics[width=\swablscale,angle=0]{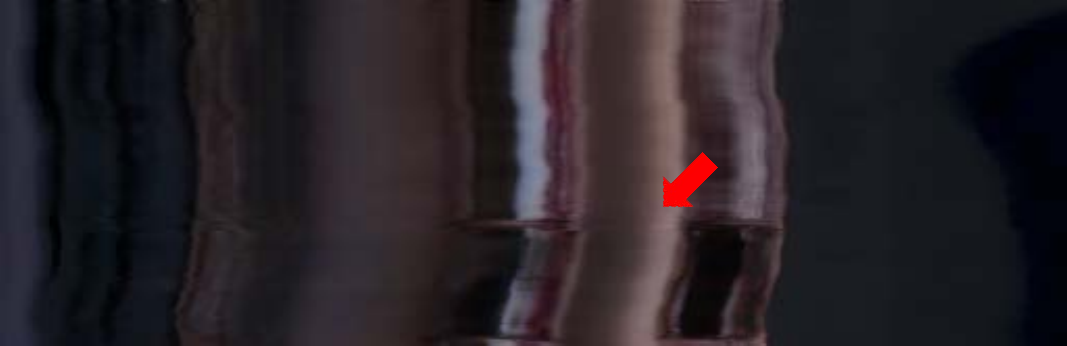} \\
  (f) $k=1$ & (g) $k=3$ & (h) $k=5$ \\
  \end{tabular}

\end{minipage}
\caption{
\textbf{Visualized Results of the Proposed Method with Varying $k$ Settings.} Images in (b)$ \sim $(h) are the temporal stacking of the \textcolor{red}{red line} in (a). Without alignment smoothing (as shown in (d) where $k=0$), our proposed method can effectively reduce the jitters in the restored face videos, compared to the original result of RestoreFormer++~\cite{wang2023restoreformer++} depicted in (c). Nonetheless, employing alignment smoothing with a greater number of neighboring frames can further decrease the jitters effectively. \textbf{Corresponding videos are in the supplementary materials.}
}
\label{fig:k}
\end{figure*}

\subsection{Ablation Studies}
\label{sec:ablation_studies}

\begin{figure}[t]
\scriptsize
\centering
\begin{tabular}{ccccc}
    \includegraphics[width=\swablsm]{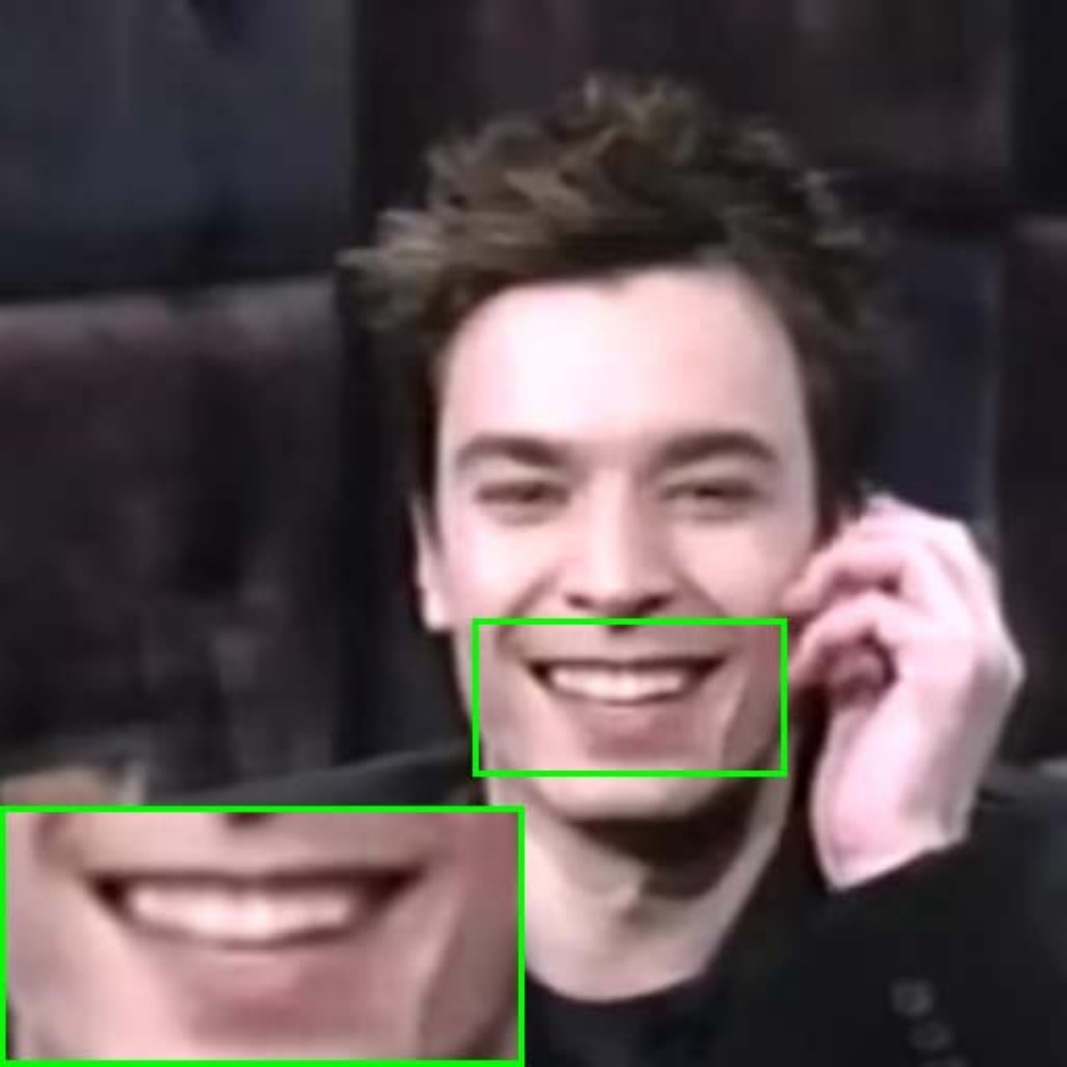} &
    \includegraphics[width=\swablsm]{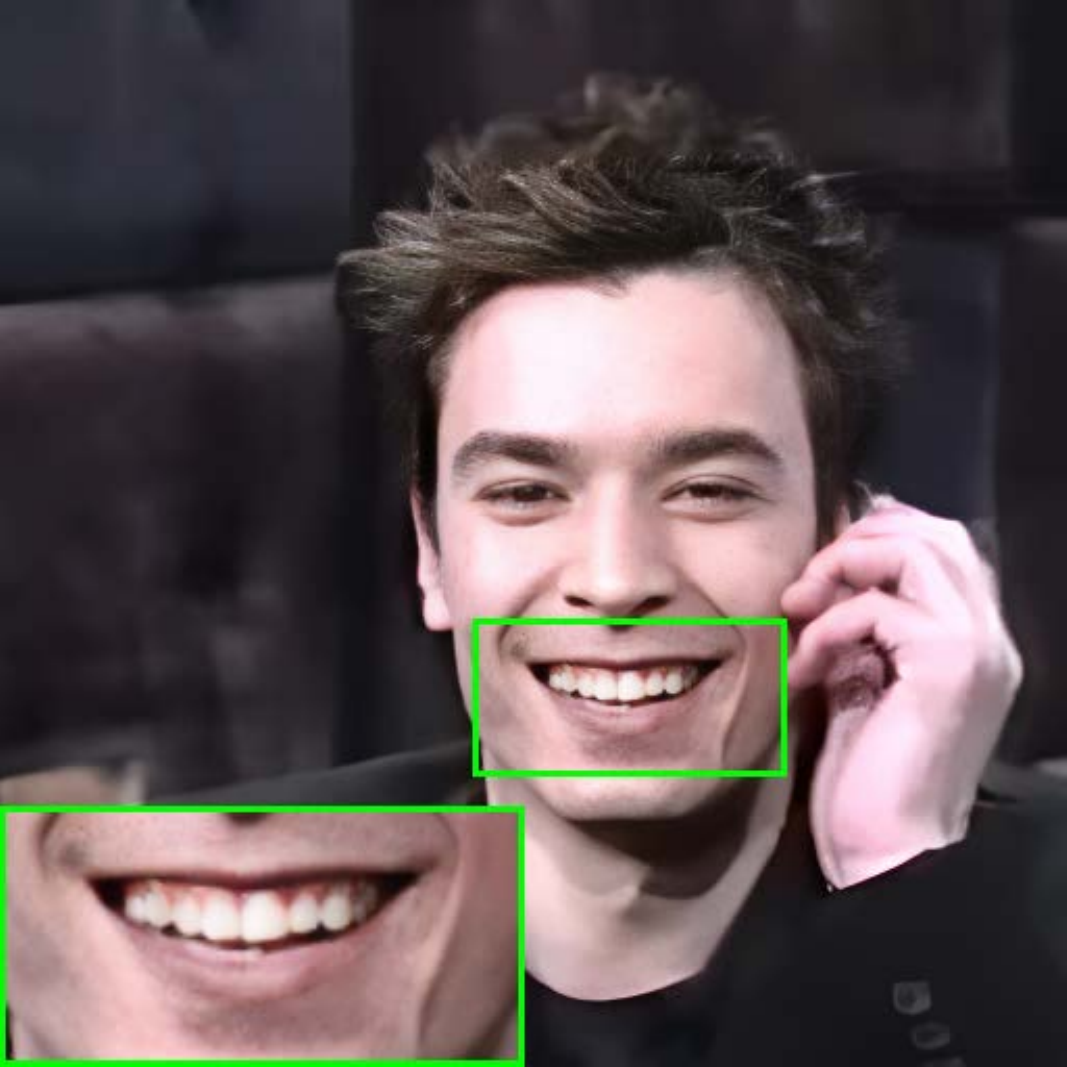} &
    \includegraphics[width=\swablsm]{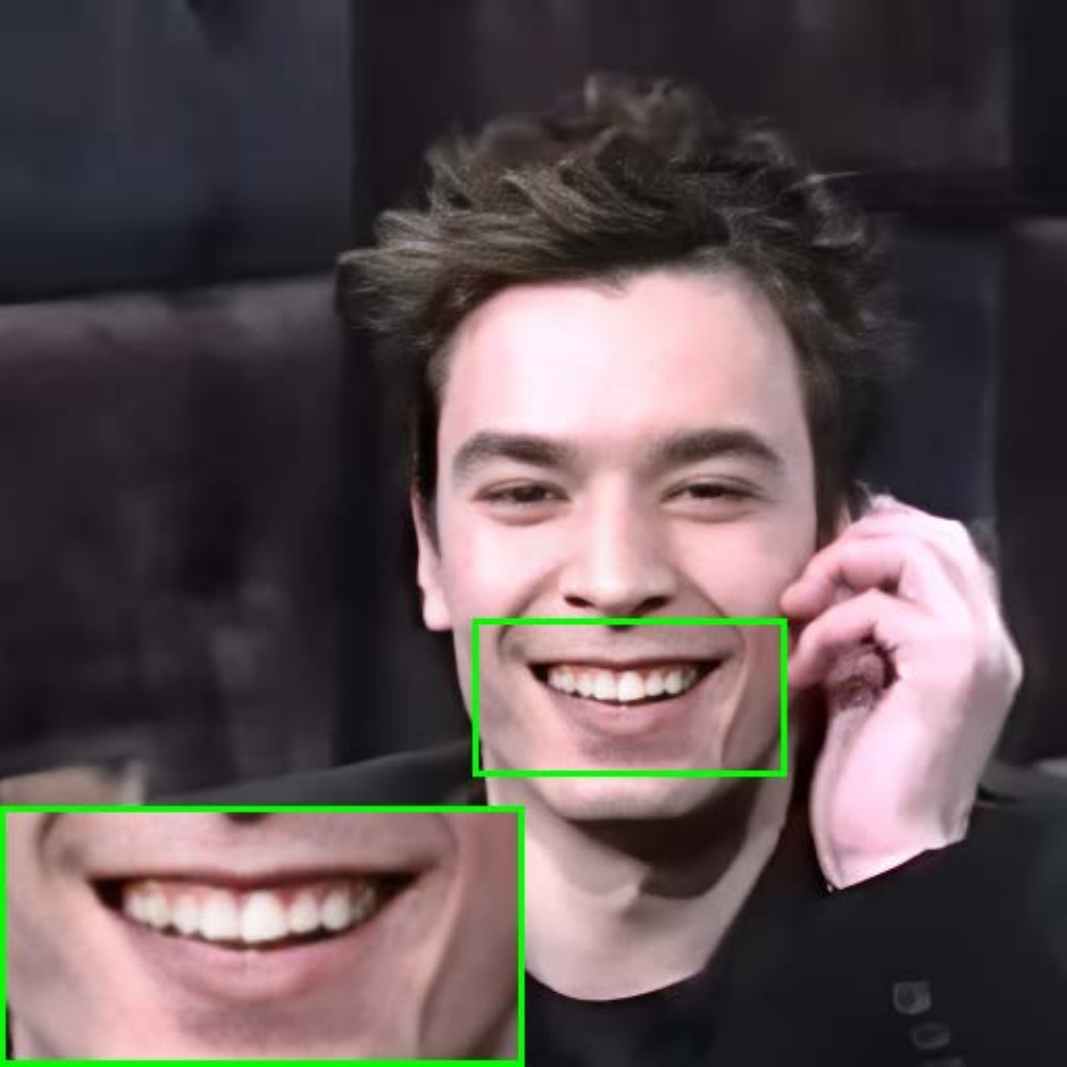} &
    \includegraphics[width=\swablsm]{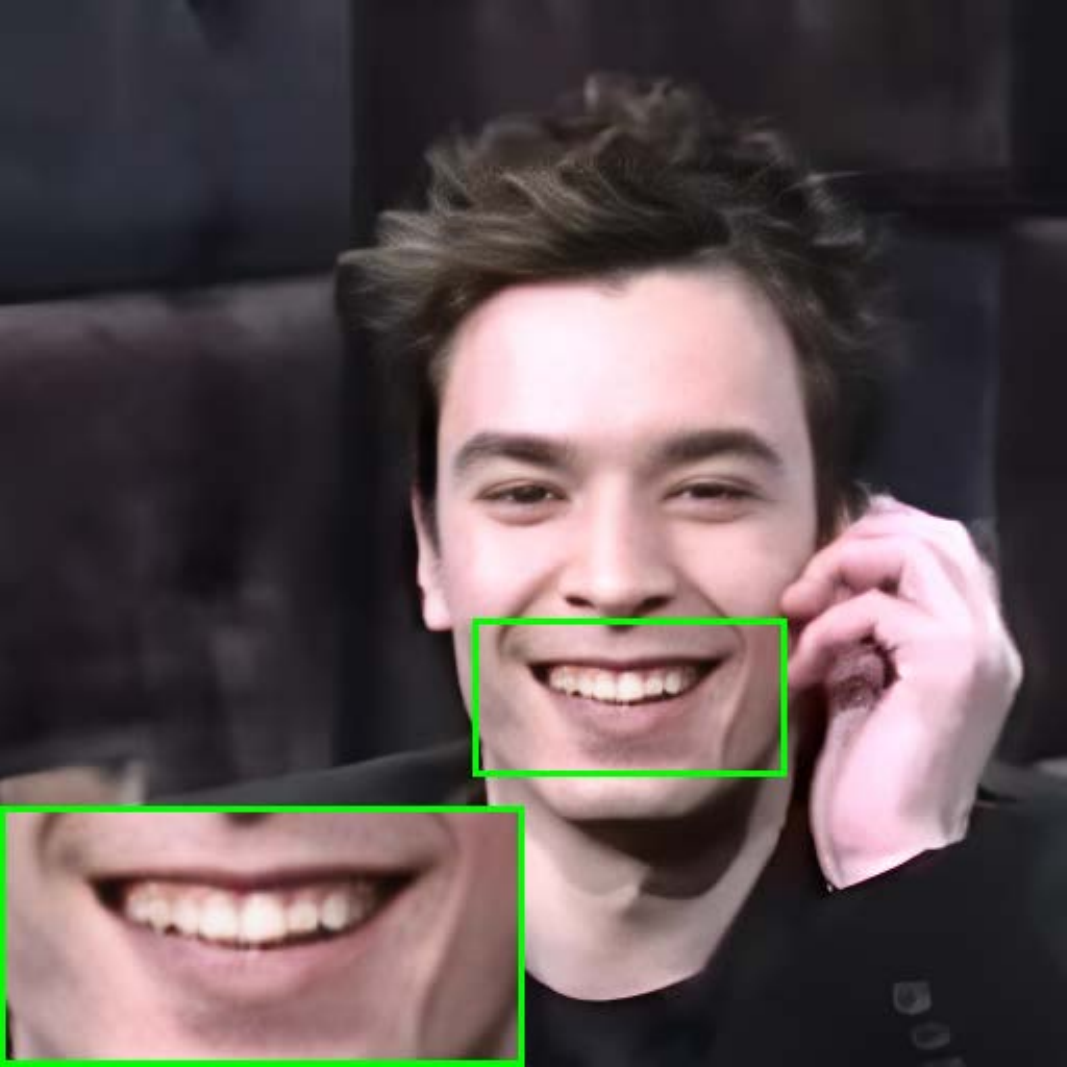} &
    \includegraphics[width=\swablsm]{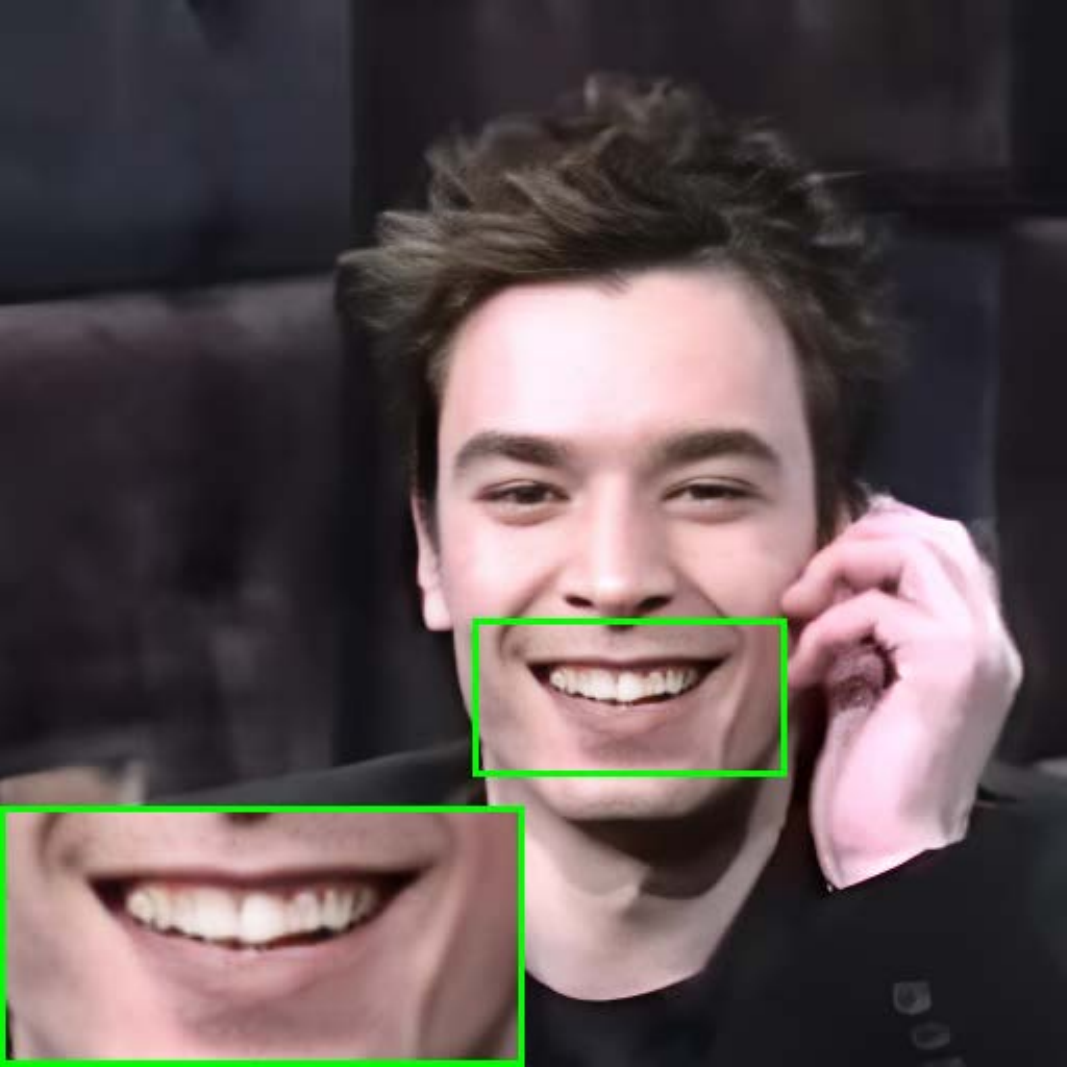} \\
    %
    \includegraphics[width=\swablsm]{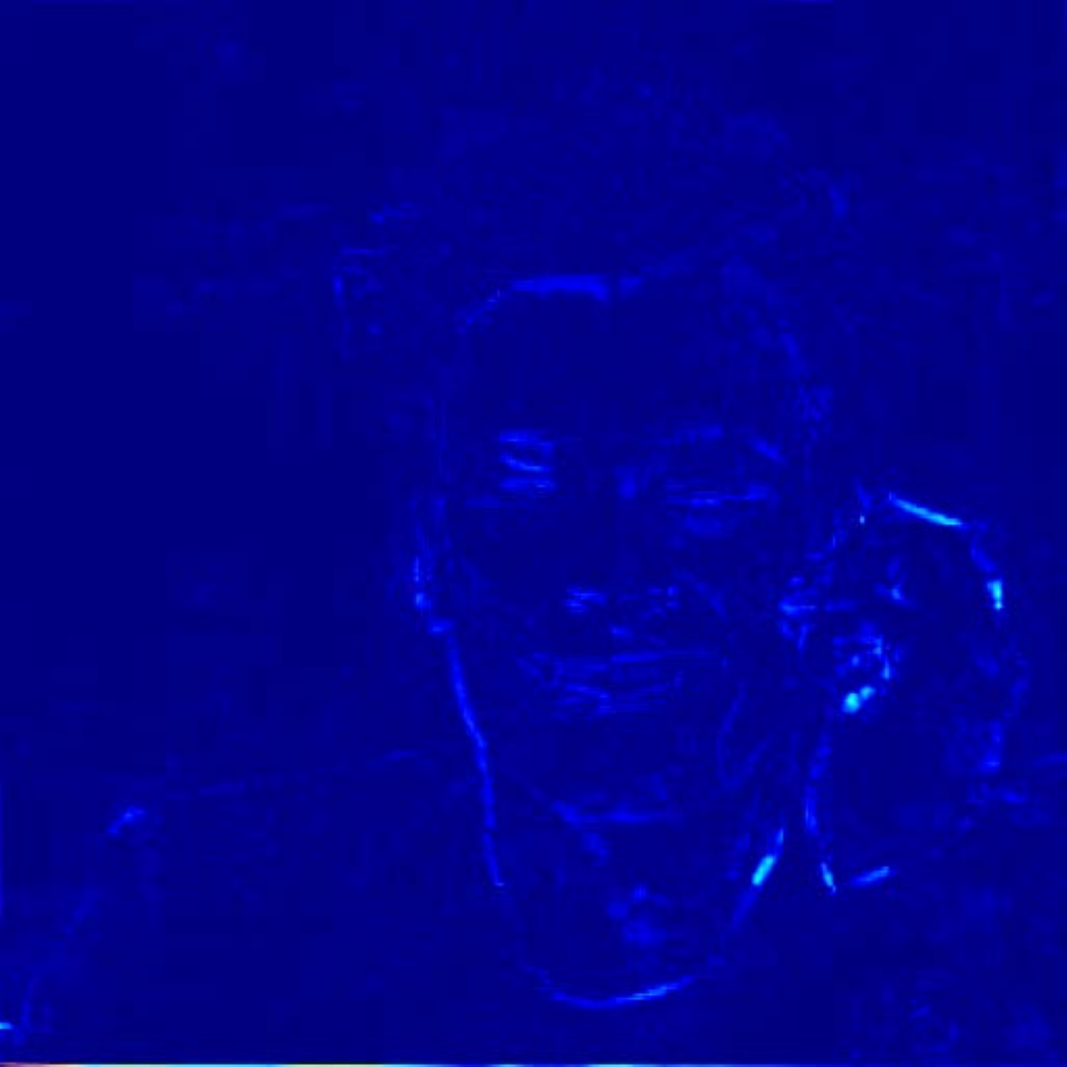} &
    \includegraphics[width=\swablsm]{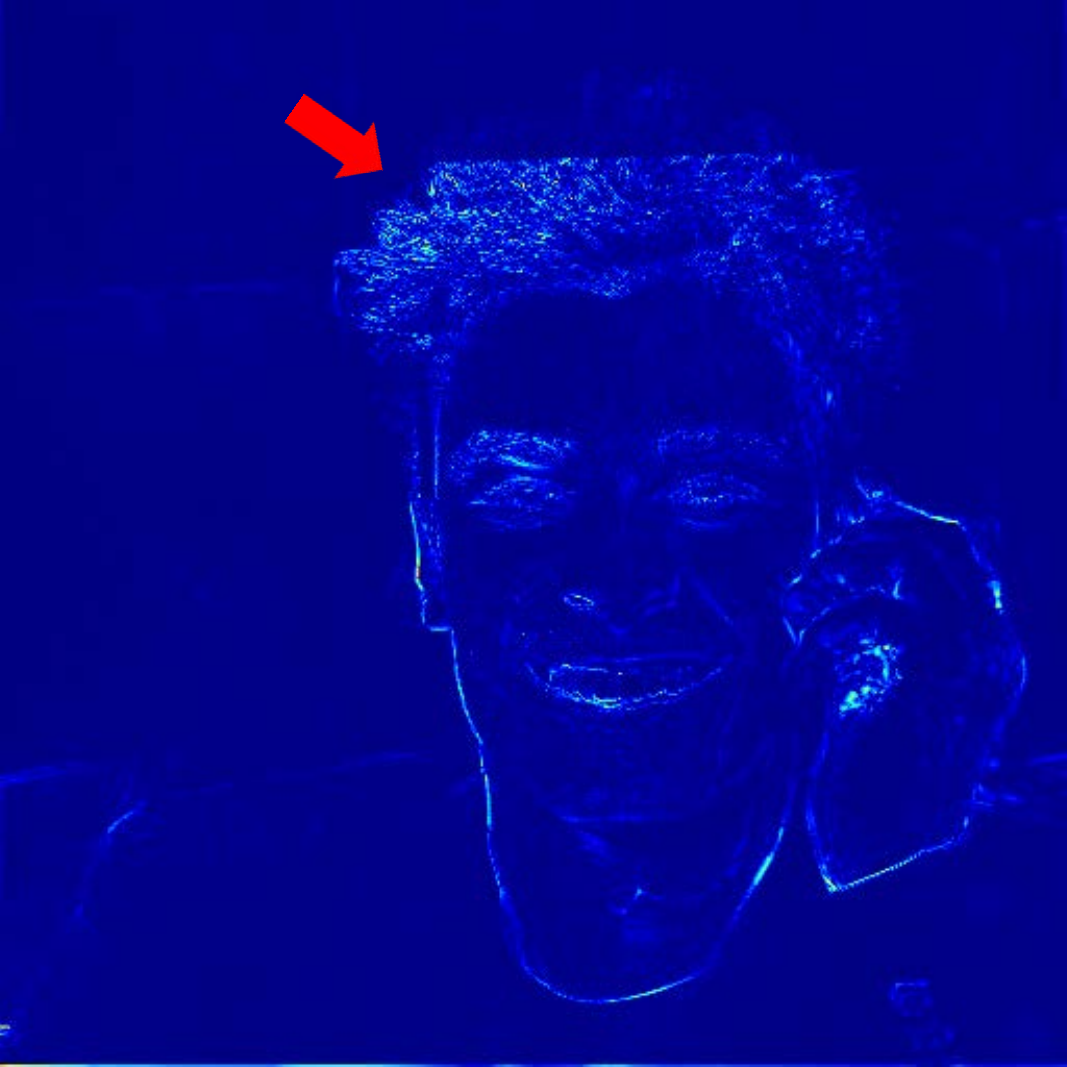} &
    \includegraphics[width=\swablsm]{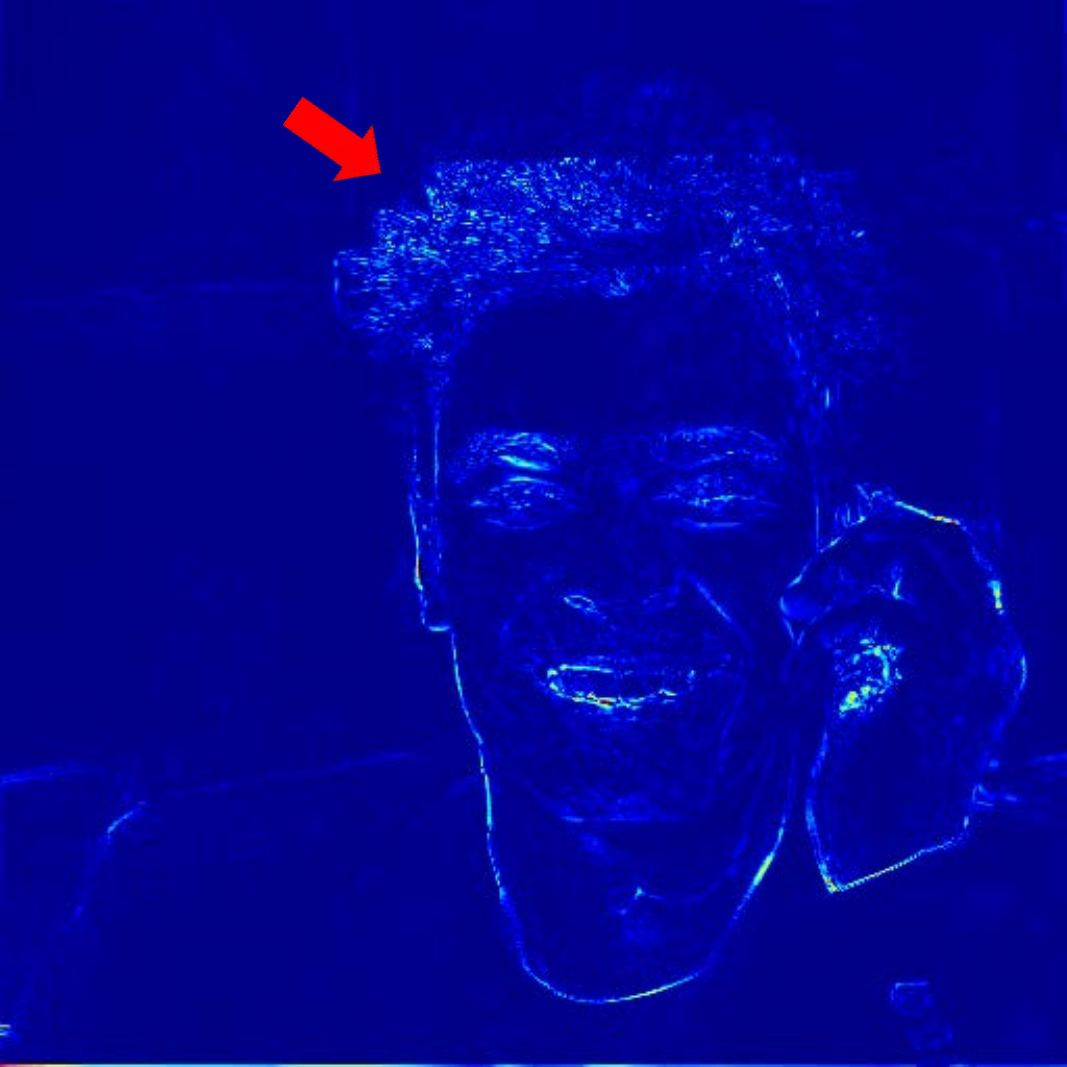} &
    \includegraphics[width=\swablsm]{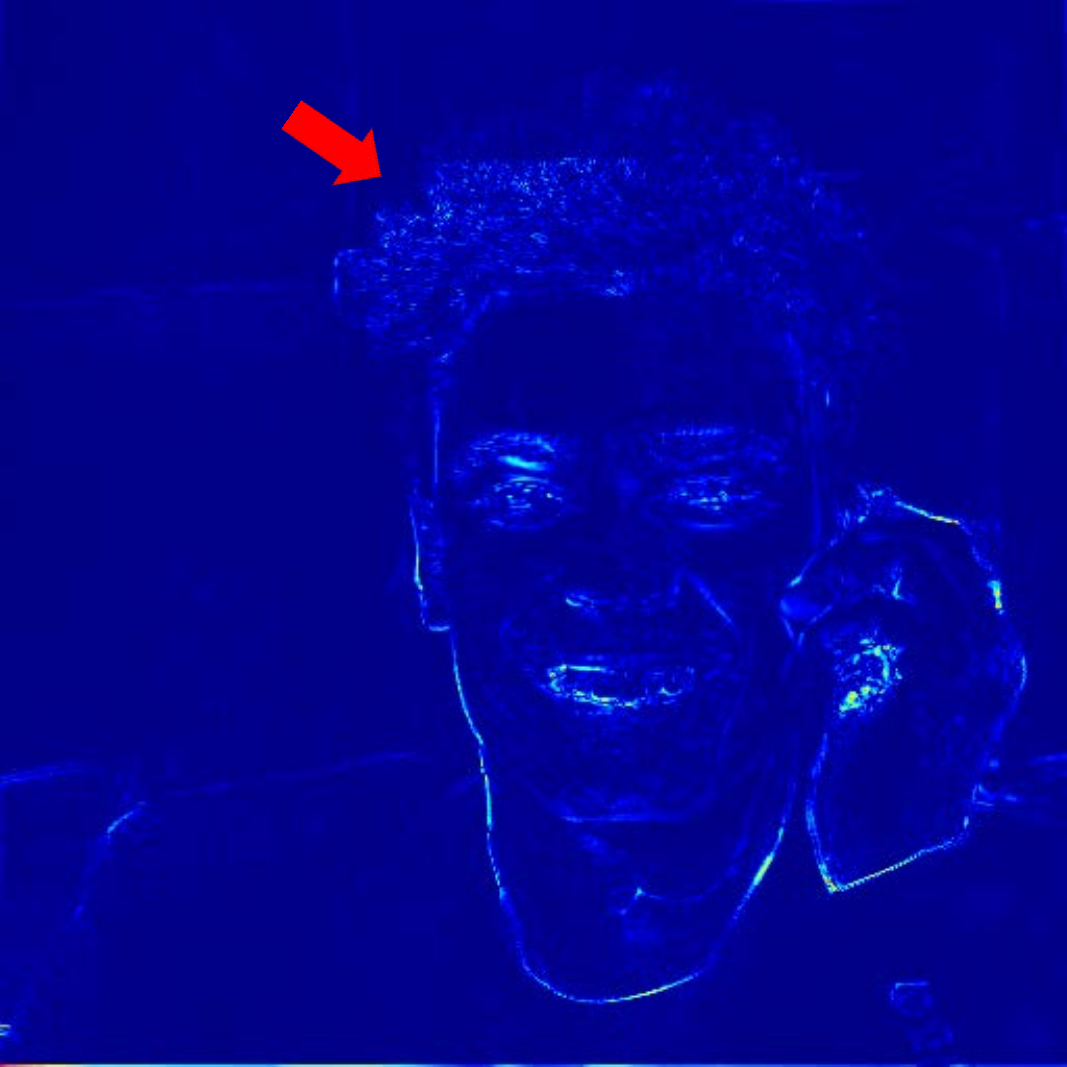} &
    \includegraphics[width=\swablsm]{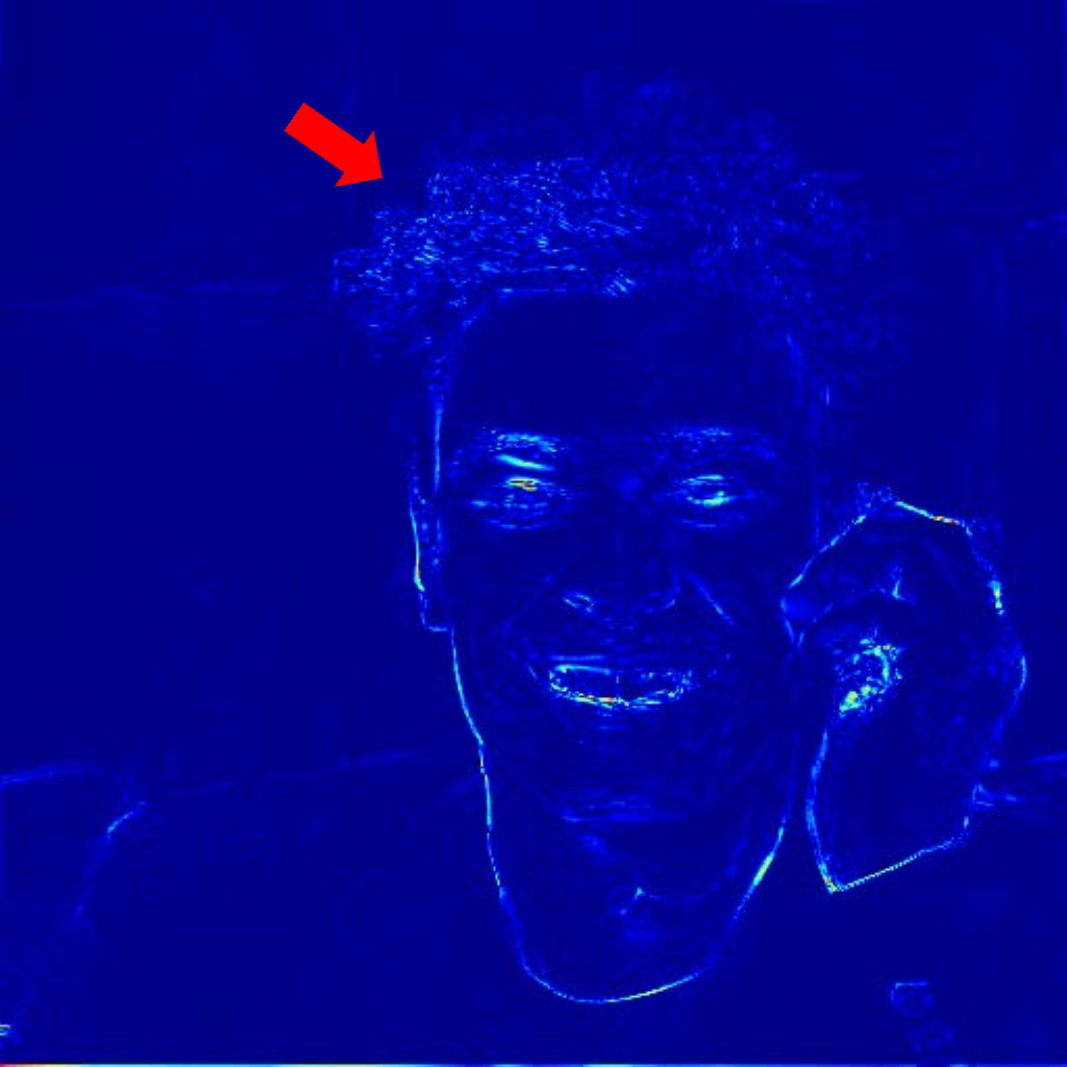} \\
    Input & \tiny RestoreFormer++~\cite{wang2023restoreformer++} & $512 \times 512 $ & $256 \times 256$ & $128 \times 128$ 
\end{tabular}
\caption{\textbf{Visualized Results of the Proposed Method at Different Resolutions.} The application of TCN on the latent feature at various resolutions can reduce the noise-shape flicks in the restored face video. This is evident from the reduced highlight area in the warping error map, particularly in the hair area indicated by the \textcolor{red}{red arrow}. However, using TCN with the latent feature at a resolution of $128 \times 128$ significantly compromises the restored quality, resulting in a lack of detail in the restored teeth. On the other hand, using TCN with the latent feature at a resolution of $256 \times 256$ provides a better balance between restored quality and temporal consistency. \textbf{Corresponding videos are in the supplementary materials.}
}
\label{fig:ablation_scale}
\end{figure}

\noindent\textbf{Integration of TCN.}
\redm{
Since TCN is trained with face videos that inherently contain motion blur (refer Fig.~\ref{fig:ffhq_and_vfhq}), its involvement in the restoration process can degrade the restored quality of the original image-based restoration models. To minimize this degradation while achieving temporal consistency, we conducted experiments to explore the optimal integration of TCN with the image-based restoration models.
The quantitative and qualitative results in TABLE~\ref{tab:ablation} and Fig.~\ref{fig:ablation_scale} indicate that integrating TCN at a later stage (the resolution of $512 \times 512$) of the image-based models can better preserve the quality of the original models. However, the temporal consistency of the restored results is limited due to inadequate interaction with the temporal information from TCN. In contrast, integrating TCN at an earlier stage (the resolution of $128 \times 128$) prematurely disrupts the restoration process of the original image-based model, leading to significant degradation in the restored quality despite improvements in temporal consistency. Based on the experimental results, integrating TCN in the decoder at a resolution of $256 \times 256$ achieves the best balance between restored quality and temporal consistency.
}

\noindent\textbf{Setting of $k$.}
We conducted an exploration of the number of neighboring frames ($k$) used for smoothing the landmarks in alignment smoothing. Our experiments were based on RestoreFormer++~\cite{wang2023restoreformer++}, using its features at a resolution of $256 \times 256$ as inputs for the TCN. As shown in Table~\ref{tab:ablation}, the quantitative results indicate that a setting of $k=3$ yields superior performance on both MSI and Warping Error, resulting in fewer jitters and noise-shaped flicks in the restored face videos. Furthermore, the quality of the results in terms of FID and component-based FID is comparable to those achieved without alignment smoothing ($k=0$). A corresponding visualized result is provided in Fig.~\ref{fig:k}, which corroborates the conclusions drawn from the quantitative results. This suggests that using alignment smoothing with a larger number of neighboring frames can effectively reduce jitters. However, in this work, we adopt $k=3$ as the further reduction of jitters with larger $k$ values was not significantly noticeable.

\section{Conclusion}
\label{sec:conclusion}

This paper systematically analyzes the potential benefits and challenges in extending state-of-the-art image-based face restoration algorithms to blind face video restoration, both qualitatively and quantitatively. It concludes that a key advantage of existing image-based face restoration algorithms lies in their superior restored quality, and proposes the use of component-based FIDs as a more accurate metric for assessing restored quality compared to the whole-face FID calculation. However, the analysis also reveals two significant challenges: 1) the presence of noticeable jitters in the restored face components, and 2) visible noise-shape flickers between frames. These challenges arise from the disregard of temporal information and biases introduced by face alignment while restoring degraded face video with image-based restoration algorithms. To address these issues, this paper introduces a temporal consistency network (TCN) cooperating with an alignment smoothing strategy to alleviate jitters and noise-shape flickers in the restored video produced by image-based face restoration algorithms while preserving their restored quality as much as possible. The effectiveness and efficiency of the proposed method are validated through extensive experiments.


 
\bibliographystyle{IEEEtran}
\bibliography{IEEEabrv,reference}

\end{document}